%% file: main.tex
\documentclass[12pt]{report}

\usepackage{indentfirst} %

\usepackage{setspace}
\usepackage[paper=letterpaper]{geometry}

\usepackage[utf8]{inputenc} 

\usepackage{titlesec, titletoc, etoolbox} %

\usepackage{graphicx} 
\usepackage{tikz} %
\graphicspath{ {figs/} }
\DeclareGraphicsExtensions{.jpg,.png,.pdf} %

\usepackage{multirow}
\usepackage{array} %
\newcolumntype{L}[1]{>{\raggedright\let\newline\\\arraybackslash\hspace{0pt}}m{#1}}
\newcolumntype{C}[1]{>{\centering\let\newline\\\arraybackslash\hspace{0pt}}m{#1}}
\newcolumntype{R}[1]{>{\raggedleft\let\newline\\\arraybackslash\hspace{0pt}}m{#1}}

\usepackage{caption} 
\usepackage{subcaption} 
\DeclareCaptionFormat{continued}{#1 (cont.)\par} %

\usepackage{amsmath}
\usepackage{amssymb}
\DeclareMathOperator*{\minimize}{minimize}

\usepackage[backend=biber, citestyle=authoryear-comp, bibstyle=authoryear-comp, sorting=ynt, maxnames=2, maxbibnames=99, uniquelist=false, uniquename=false]{biblatex} %
\AtEveryBibitem{
\ifentrytype{unpublished}
{	%
}
{	%
	\clearfield{url}
}
}

\usepackage{comment}

\usepackage{hyperref}
\hypersetup{
	colorlinks=true, %
	hidelinks, %
	linktoc=all, %
	linktocpage=false %
}

\usepackage[
	capitalize, %
	noabbrev, %
	nameinlink, %
]{cleveref} %
\crefname{figure}{Figure}{Figures} %
\crefname{table}{Table}{Tables} %
\crefname{equation}{Equation}{Equations} %
\crefname{section}{Section}{Sections}

\newcommand{\MultiLevelTitleFormats}{
	\titleformat{\chapter}[display]
	{\centering\Large\bfseries\singlespacing} %
	{\MakeUppercase{\chaptertitlename}\ \thechapter} %
	{0em} %
	{\MakeUppercase} %
	\titlespacing*{\chapter} %
	{0pt} %
	{*-3} %
	{*3} %
	
	\titleformat{\section}[hang]
	{\large\bfseries\singlespacing} %
	{\thesection} %
	{0.5em} %
	{} 
	\titlespacing*{\section} %
	{0pt} %
	{*2} %
	{*1} %
	
	\titleformat{\subsection}[hang]
	{\normalsize\bfseries\singlespacing} %
	{\thesubsection} %
	{0.5em} %
	{} 
	\titlespacing*{\subsection} %
	{0pt} %
	{*2} %
	{*1} %
	
	\titleformat{\subsubsection}[hang]
	{\normalsize\itshape\singlespacing} %
	{\thesubsection} %
	{0.5em} %
	{} 
	\titlespacing*{\subsubsection} %
	{0pt} %
	{*2} %
	{*1} %
}
\AtBeginDocument{\MultiLevelTitleFormats} 

\newcommand{\RegularTOCEntries}{
	\addtocontents{toc}{\protect\setcounter{tocdepth}{1}}
	  
	\titlecontents{chapter}[0em]{\singlespacing} %
	{\normalfont\normalsize CHAPTER \thecontentslabel: \ \uppercase}%
	{}%
	{\titlerule*[0.4pc]{.}\contentspage}%
	
	\titlecontents{section}[1.5em]{\vspace{-5pt}\singlespacing} %
	{\normalfont\normalsize\thecontentslabel\enspace}%
	{}%
	{\titlerule*[0.4pc]{.}\contentspage}%
}
\AtBeginDocument{\RegularTOCEntries} %

\newcommand\AppendixTOCEntries{
	\addtocontents{toc}{\protect\setcounter{tocdepth}{0}}  
	
	\titlecontents{chapter}[0em]{} %
	{\normalfont\normalsize APPENDIX \thecontentslabel: \enspace\uppercase}%
	{}%
	{\titlerule*[0.4pc]{.}\contentspage}%
}
\pretocmd{\appendix}{\AppendixTOCEntries}{}{} %

\begin{document}
	\newgeometry{margin=1in}

\input{titlepage}

	\pagenumbering{roman}
	\setcounter{page}{2}

\input{abstract}

\input{dedication}

\input{acknowledgments}

	\renewcommand{\contentsname}{\hfill\bfseries\large\uppercase{Table of Contents}\hfill} %
	\tableofcontents

	\cleardoublepage
	\pagenumbering{arabic}
	
	\input{chapter01}
	\input{chapter02}
	\input{chapter03}
	\input{chapter04}

	\input{chapter05}
	\input{chapter06}
	\input{chapter07}
	\input{chapter08}
	\input{chapter09}
	\input{chapter10}
	\input{chapter11}

	\begin{refcontext}[sorting=nyt] %
		\printbibliography[heading=bibintoc, title={\uppercase{References}}] %
	\end{refcontext}

\end{document}

%% file: titlepage.tex
\begin{titlepage}
	\vspace*{\fill}
	\begin{center}
		\copyright\ 2021 Haohang Huang
	\end{center}
	\vspace*{\fill}
\end{titlepage}

\begin{titlepage}
	\begin{singlespace}
	\begin{center}
		\vspace*{1in}
		
		\uppercase{Field Imaging Framework for Morphological Characterization of Aggregates with Computer Vision: \\Algorithms and Applications}
		
		\vspace{1.5in}
		
		\uppercase{BY}
		
		\vspace{\baselineskip}
		
		\uppercase{Haohang Huang}
		
		\vspace{1.2in}
		
		\uppercase{Dissertation}
		
		\vspace{\baselineskip}
		
		Submitted in partial fulfillment of the requirements \\
		for the degree of Doctor of Philosophy in Civil Engineering \\
		in the Graduate College of the \\
		University of Illinois Urbana-Champaign, 2021
		
		\vspace{0.5in}
		
		Urbana, Illinois
		
		\vfill
		
		\begin{flushleft}
			Doctoral Committee: \\
			\vspace{\baselineskip}
			\parindent=0.5in Professor Erol Tutumluer, Chair and Director of Research \\
			\parindent=0.5in Professor Imad Al-Qadi \\
			\parindent=0.5in Professor Jeffery R. Roesler \\
			\parindent=0.5in Associate Professor Mani Golparvar-Fard \\
			\parindent=0.5in Professor Sanjay Patel
		\end{flushleft}
		
	\end{center}
	\end{singlespace}
\end{titlepage}

%% file: abstract.tex
\begin{center}
	\textbf{\uppercase{Abstract}}
\end{center}

\vspace{\baselineskip}

\begin{onehalfspace}
Construction aggregates, including sand and gravel, crushed stone and riprap, are the core building blocks of the construction industry, national economy, and society. In the year 2020, in total 2.42 billion metric tons of aggregates valued at \$27.0 billion were produced by about 5,400 mining companies operating more than 10,000 quarries across all 50 states. Through mining, quarrying, and multi-level crushing and screening processes, aggregates produced in different sizes and forms constitute the main skeleton of civil infrastructure and are extensively used in structural, transportation, geotechnical, and hydraulic engineering applications.

At both quarry production lines and construction sites, the morphological properties of aggregates (such as size, shape, volume/weight, etc.) are some of the most crucial indicators for aggregate Quality Assurance and Quality Control (QA/QC), especially for crushed aggregates and riprap. State-of-the-practice methods mainly use sieving and caliper devices for the size and shape determination of the most commonly used regular sizes of crushed aggregates, and are limited to visual inspection and manual measurement for relatively large-sized aggregates. As a more advanced quantitative approach, state-of-the-art aggregate imaging methods developed to date focus on characterizing aggregate morphology from acquired image data and machine vision analysis, yet with the limitation that most systems are only applicable to regular-sized aggregates under well-controlled laboratory conditions.

The state-of-the-practice and state-of-the-art methods have encountered several major challenges in characterizing aggregate morphology. First, quantitative methods for capturing and analyzing aggregates are required to provide reliable characterization of the material. Second, flexible and effective methods are urgently needed for relatively large-sized aggregates. Furthermore, advanced analyses are necessary to handle the most practical form of aggregate presence, such as densely stacked aggregates in stockpiles and/or in constructed layers. Lastly, three-dimensional (3D) imaging approaches are deemed ideal by providing more comprehensive and realistic aggregate information than two-dimensional (2D) image analyses.

This dissertation presents the research effort to address these major challenges by developing a field imaging framework for the morphological characterization of aggregates as a multi-scenario solution. The framework also has a focus on relatively large-sized aggregates, for which effective and efficient field characterization methods are extremely lacking. 

For individual and non-overlapping aggregates, a field imaging system was designed first, and the associated image segmentation and volume estimation algorithms were developed. The color-based image segmentation algorithm provides robust object extraction under various field lighting conditions such as strong sunlight and shadowing, and the volumetric reconstruction algorithm estimates the particle volume by orthogonal intersection. The approach demonstrated good agreements with ground-truth measurements made at quarry sites and achieved great improvements in the volumetric estimation of individual aggregates when compared with the state-of-the-practice inspection methods.

For 2D image analyses of aggregates in stockpiles, an automated 2D instance segmentation and morphological analysis approach was established based on deep learning. A task-specific stockpile aggregate image dataset was compiled based on images collected from aggregate producers and individual aggregates in the images were manually labeled to provide the ground-truth for learning. A state-of-the-art object detection and segmentation architecture was implemented to train the image segmentation kernel for stockpile segmentation. The segmentation results showed good agreement with ground-truth labeling and provided efficient morphological analyses on images containing densely stacked and overlapping aggregates. 

For 3D point cloud analyses of aggregates in stockpiles, an end-to-end, integrated 3D Reconstruction-Segmentation-Completion (RSC-3D) approach was established by collaborating three developed components, i.e., laboratory and field 3D reconstruction procedures, 3D stockpile instance segmentation, and 3D shape completion. The approach was designed to reconstruct aggregate stockpiles from multi-view images, segment the stockpile into individual instances, and predict the unseen side of each instance based on the partial visible shapes. First, a 3D reconstruction procedure was developed to obtain high-fidelity full 3D models of collected aggregate samples, based on which a 3D aggregate particle library was constructed, and a comparative analysis was conducted regarding the 2D and 3D morphological characteristics. Next, two datasets were prepared based on the 3D particle library for 3D learning purpose: (i) a synthetic dataset of aggregate stockpiles with ground-truth instance labels developed with a synthetic data generation pipeline involving model fabrication, stockpile assembly, and stockpile raycasting; and (ii) a dataset of partial-complete shape pairs, developed with varying-visibility and varying-view raycasting schemes. Based on the two datasets, a state-of-the-art 3D instance segmentation network and a 3D shape completion network were implemented and trained, respectively. The application of the integrated approach was demonstrated on re-engineered stockpiles and field stockpiles, and the validation results against ground-truth measurements showed good performance in capturing and predicting the unseen sides of aggregates, especially in terms of size dimension metrics.

In summary, the developed field imaging framework in this study encompasses three major approaches that characterize various forms and representations of field aggregates with increasing analysis complexity: (i) a volumetric reconstruction approach for individual and non-overlapping aggregates; (ii) a 2D instance segmentation and morphological analysis approach for aggregates in stockpiles based on 2D image analysis; and (iii) a 3D integrated reconstruction-segmentation-completion approach for aggregates in stockpiles based on 3D point cloud analysis. The framework addresses the major challenges of characterizing individual aggregates and aggregate stockpiles in the field, thus provides a multi-scenario solution for efficient 2D and 3D analyses of aggregates. 

\end{onehalfspace}

\clearpage

%% file: dedication.tex
\vspace*{\fill}
\begin{center}
	\textit{To my parents Gang Huang and Xiaoxia Liu, \\
		my sister Shuyan Liu, \\
		and my love Yihui Li.}
\end{center}
\vspace*{\fill}

\clearpage

%% file: acknowledgments.tex
\begin{center}
	\textbf{\uppercase{Acknowledgments}}
\end{center}

\vspace{\baselineskip}

\begin{onehalfspace}
	
First of all, I am deeply indebted to my mentor and advisor, Professor Erol Tutumluer, for his constant support, insightful guidance, and inspiring thoughts throughout this doctoral research. I am fortunate to be nurtured by such a understanding, patient, and humble advisor, without whom any step of this work would not have been possible over the years. He devoted his passion and kindness to the students and bonded the research group as a heart-warming family with his fatherly advice. The methodology and philosophy I learned from him have greatly shaped my mindset of critical thinking and problem solving. His words and trust have given me unlocked powers far beyond knowledge and education, which I know surely, remembered and enshrined, are my constant companions and comforters in life.

I would also like to extend my deepest appreciation to my other doctoral committee members: Professor Imad Al-Qadi, Professor Jeffery R. Roesler, Professor Mani Golparvar-Fard, and Professor Sanjay Patel, for their innovative ideas, insightful comments, and joint contribution towards improving this research. Professor Imad Al-Qadi and Professor Jeffery R. Roesler have shared countless ideas that greatly improve this research to address practical engineering challenges, and the Advanced Transportation Research and Engineering Laboratory (ATREL) facility they manage served as the foundation of every activity conducted in this research. I am also very grateful to Professor Mani Golparvar-Fard and Professor Sanjay Patel, who have been unreservedly offering multi-disciplinary views from computer science and electrical engineering and encouraging me to bring the cutting-edge technology into my research. All doctoral committee members, with their hard-working attitude, well-rounded knowledge, and leadership skills, are the role models for my future professional development.

I would like to thank all research partners and project sponsors who have been involved directly or indirectly in the success of this research. The research was mainly supported by ICT-R27-182 and ICT-R27-214 projects, which were conducted in cooperation with the Illinois Center for Transportation (ICT); the Illinois Department of Transportation (IDOT); and the U.S. Department of Transportation, Federal Highway Administration. The research was conducted as an interdisciplinary collaboration effort with Professor Narendra Ahuja from Electrical and Computer Engineering (ECE) department and John M. Hart, principal research engineer at the Computer Vision and Robotics Laboratory (CVRL). I very much appreciate the unwavering support and effort of them throughout the entire research. I would like to extend my sincere thanks to the help from Andrew Stolba, the project Technical Review Panel (TRP) chair at IDOT, and Sheila Beshears, manager at Riverstone Group and former TRP chair at IDOT. Special thanks go to Chad Nelson, Del Reeves, and Kevin Tressel at IDOT, and Andrew Buck and Dan Barnstable at Vulcan Materials Company, and ICT research engineer Greg Renshaw, for their support and effort in coordinating the quarry field visits. Many thanks also go to Jeb S. Tingle at U.S. Army Engineer Research and Development Center (ERDC) of the United States Army Corps of Engineers (USACE) for the sponsorship through scientific computing related research projects during my doctoral study.

During different stages of this research, I have received selfless help on countless occasions from the students and colleagues at University of Illinois Urbana–Champaign (UIUC). I am specially thankful to my colleague and friend, Dr. Issam I. A. Qamhia, for always providing me with timely help and advice on every aspect in research and life over my entire Ph.D. years. Special thanks also go to Jiayi Luo, who has been working closely with me on numerous research ideas, from conceptualizing, experimenting, implementing, to success/failure. My joy and excitement during the doctoral research, either from a direct success or more commonly after many trials and errors, shall have the resonance with him. My sincere thanks and gratitude go to Professor Gholamreza Mesri, Professor Scott M. Olson, Professor Derek Hoiem, Professor Eric Shaffer, Dr. Yu Qian, Dr. Maziar Moaveni, Dr. Hasan Kazmee, Dr. Yong-Hoon Byun, Dr. Angeli Gamez, Dr. Jianfeng Mao, Dr. Wenting Hou, Dr. Priyanka Sarker, Dr. Huseyin Boler, Dr. Siqi Wang, Dr. Zhoutong Jiang, Zixu Zhao, Scott Schmidt, Sagar Shah, Maximilian Orihuela, Yue Gong, Arturo Espinoza Luque, Punit Singhvi, Guangchao Xing, Linjian Ma, Bin Feng, Jie Shen, Mingu Kang, Wenjing Li, Qingwen Zhou, Jiawei Fan, Zhongyi Liu, Han Wang, Kelin Ding, Taeyun Kong, and Syed Faizan Husain. It had been a great pleasure working with these brilliant minds during my doctoral journey.

Finally, and most importantly, I would like to thank my parents and my sister for their unconditional love and support. You are my anchor, my light and my salvation that I always trust in. I also want to thank my girlfriend, who is soon to become my fianc\'{e}e. Our engagement ring is sitting right next to me as I am on the very last lines of this dissertation, yet you are perfectly unaware at the moment. Your company since our childhood has illumined me at all times and is truly the best gift I can ever have. With all my heart, I dedicate this dissertation to my loved ones.

\end{onehalfspace}

\clearpage

%% file: chapter01.tex
\chapter{Introduction} \label{chapter-1}

\section{Research Statement}

Construction aggregates, including sand and gravel, crushed stone and riprap, are the core building blocks of the construction industry, national economy, and society. In the year 2020, in total 2.42 billion metric tons of aggregates valued at \$27.0 billion were produced by about 5,400 mining companies operating more than 10,000 quarries across all 50 states \parencite{summaries_mineral_2021}. Through mining, quarrying, and multi-level crushing and screening processes, aggregates produced in different sizes and forms serve as essential components in structural, transportation, geotechnical, and hydraulic engineering applications. 

At the production sites, crushed stone aggregate producers first perform quarrying to excavate rocks from the ground and crush them into large-sized aggregates, which can then be screened into specific sizes for immediate use or further processing. These relatively large-sized aggregates, directly as an upstream riprap product or for intermediate temporary storage, are typically categorized by size and stored in separate stockpiles \parencite{greenwell_practical_1913}. For many state Departments of Transportation (DOTs) across the U.S., the characterization of this important engineering material has always been critical for Quality Assurance/Quality Control (QA/QC). For instance, according to Illinois Department of Transportation's Standard Specifications for Road and Bridge Construction \parencite{idot_standard_2016}, Illinois quarries produce construction aggregates in seven categories with increasing size, from RR1 to RR7 (`RR' for `RipRap'). RR1 and RR2 materials are small to medium-sized aggregates with up to 4-in. (10.2 cm) size that are mostly used as building materials, pavement aggregates, and railway ballast. RR3 to RR7 materials are relatively large-sized aggregates that could weigh up to 1,150 lbs. (522 kg). Similarly, Minnesota DOT classifies riprap as Class I to Class V with maximum individual aggregate/rock weight of 2,000 lbs. (907 kg) \parencite{mndot_standard_2018}; Nevada DOT grades riprap from Class 150 to Class 900 with individual rock weighing up to 1,500 lbs. (680 kg) \parencite{ndot_standard_2014}. In the context of this study, the general term ``aggregates'' and several specific terms ``riprap,'' ``riprap rocks'' and ``large-sized aggregates'' are used interchangeably all referring to the aggregate materials in these relatively large-sized categories.

According to \textcite{lagasse_riprap_2006}, uniform specifications or guidelines that ensure reliable and efficient characterization of weight, size, shape, and gradation of riprap categories are critical at both production lines and construction sites. At the current state of the practice, a nationwide American Association of State Highway and Transportation Officials (AASHTO) survey of transportation agencies in the US and Canada has indicated that riprap characterization is mostly based on visual inspection and manual measurements \parencite{sillick_member_2017}. Visual inspection depends greatly on the experience and expertise of practitioners. In this method, certain gauge or keystones and sample stockpiles are usually used as a reference to assist the judgment \parencite{lippert_inspection_2012}. To better estimate the size distribution, the Wolman count method is applied by statistically sampling and measuring rocks within a stockpile \parencite{lagasse_riprap_2006}. For instance, the use of keystones with predefined weight ranges has been adopted recently by IDOT to facilitate the visual inspection process. For manual measurement, transportation agencies either weigh individual particles directly or use size-mass conversion after measuring rock dimensions. U.S. Army Corps of Engineers requires direct weight measurement of individual riprap rocks as specified in \textcite{usace_em_1110-2-2302_engineering_1990} for large stone construction. Alternatively, the size-mass conversion proposed in \textcite{astm_d5519_standard_2015} requires measurement of the midway dimension or circumference from three orthogonal axes and estimates the volume based on a cuboid assumption or averaged sphere-cube assumption. Despite these great efforts, the visual inspection practice is still very subjective and uncertain; and manual size measurement requires heavy machinery to manipulate individual rocks which is time-consuming and labor-intensive. In addition, both methods are qualitative measures in terms of shape characterization lacking the capability to capture the full morphological properties (i.e., size, shape, volume/weight, etc.) of aggregates. As a result, the major challenge for characterizing large-sized aggregates is primarily due to the difficulties associated with its huge size and heavy weight, while an objective and efficient approach for quantitatively characterizing aggregate morphology has yet to be established. In this regard, reliable field imaging techniques are a promising approach to process stockpile images easily and quickly for gradation checks and provide data analytics.

Aggregate imaging techniques have been developed over the past two decades as a promising solution for the quantitative analyses of aggregate morphological properties \parencite{rao_quantification_2002, al-rousan_new_2005, pan_aggregate_2006, wang_evaluation_2013, moaveni_evaluation_2013, hryciw_innovations_2014}. Most of the current aggregate imaging techniques follow a certain pipeline: (i) individual aggregate particles are manually arranged in a laboratory setup under well-controlled background and lighting conditions, (ii) a camera system captures the images of aggregates, and (iii) a computer program analyzes the images to determine the size and shape properties. However, these techniques are only applicable to small and medium-sized aggregates that can be easily manipulated in a laboratory setup and are therefore not scalable for characterizing large-sized aggregates. Moreover, the in-place inspection at the production lines or construction sites poses extra challenge to the image acquisition and analysis steps with natural background and the lighting conditions. 

In summary, the state-of-the-practice and state-of-the-art methods have encountered several major challenges in characterizing aggregate morphology. First, quantitative methods for capturing and analyzing aggregates are required to provide reliable characterization of the material. Second, flexible and effective methods are urgently needed for relatively large-sized aggregates. Furthermore, advanced analyses are necessary to handle the most practical form of aggregate presence, such as densely stacked aggregates in stockpiles and/or in constructed layers. Lastly, three-dimensional (3D) imaging approaches are deemed ideal by providing more comprehensive and realistic aggregate information than two-dimensional (2D) image analyses can provide.

Therefore, there is a pressing need to develop an advanced field imaging framework that can efficiently characterize the morphological properties of large-sized aggregates in field conditions. Better property characterization and optimized material selection can be achieved to improve designs through effective quality control, reduced costs, increased life cycle, and minimum labor and energy consumption. Major cost savings in terms of personnel time, transportation, and laboratory equipment and facility use can be realized.

\section{Research Objectives}

The primary objective of this doctoral research study is to develop a convenient and efficient field imaging framework for aggregates based on computer vision techniques. The framework is supposed to provide an analysis platform for the field collected aggregate data to determine the size, shape, volume/weight, and gradation properties of the large-sized aggregates inspected. The framework will enable the characterization of aggregates at different sophistication levels, i.e. (i) individual and isolated aggregates for volumetric estimation, (ii) in-place aggregates in a stockpile for 2D image analyses, as well as (iii) in-place aggregates in a stockpile or constructed layer for 3D point cloud analyses. The algorithms developed in this framework will be designed as automated and minimally user-dependent and are intended for robust operation under various field and environmental conditions. The applications of this framework should demonstrate the convenience in data acquisition and data analysis with different sophistication levels, together with ground-truth validation confirming the robustness and reliability of the framework. Finally, the framework is envisioned to capture the morphological properties of aggregates for the purpose of fast Quality-Assurance/Quality-Control (QA/QC) inspection as well as advanced morphological analysis based on realistic 3D aggregate data.

\section{Research Methodology and Scope}
To fulfill the above-stated research objectives, this study will consider the following five main research aspects: 

\begin{itemize}
	\item Identifying and acquiring representative aggregate samples and image data. Information will be gathered on the types, geologic origins, and representative sources of riprap and large-sized aggregates materials, as well as the statewide locations of the approved lists of these materials in Illinois. After identifying the approved aggregate sources (from RR3 to RR7 size categories as per IDOT specifications), field visits to aggregate producers in Illinois will be scheduled to collect representative samples and different types of imaging data. 
	
	\item Developing volumetric estimation algorithms for individual aggregates. For individual and isolated aggregates, volumetric estimation algorithms will be developed to quantify the volumetric properties of aggregates inspected from different views. Associated field imaging setup will be designed to provide stable image background that allows accurate extraction of aggregate regions. The algorithms will be compared against ground truth size and weight measurements to validate the potential use and benefits of imaging techniques as compared to the state-of-the-practice methods.
	
	\item Developing automated 2D image segmentation and morphological analyses for aggregate stockpiles. For aggregate stockpiles, automated 2D image segmentation algorithms will be developed based on deep learning to extract the individual aggregates from the stockpile view. An image dataset of aggregate stockpiles will be established and labeled based on the image data collected at quarries in Illinois. State-of-the-art deep learning architecture for object detection and segmentation will be implemented and trained to enable automated segmentation of stockpile images. Next, morphological analysis algorithms will be developed to characterize the size, shape, and gradation properties of the segmented aggregate regions. 
	
	\item Establishing 3D aggregate particle library and generating necessary datasets for deep learning. Based on the image data collected in previous tasks, a 3D aggregate particle library containing riprap and large-sized aggregate models will be established as the database. Next, synthetic aggregate stockpile scenes will be constructed based on the library by simulating the particles with physics and graphics engine. The stockpile scenes in the format of 3D point clouds, associated with ground-truth labels of aggregates in the stockpile and constructed layer, will be used as the training data for 3D detection and segmentation. Moreover, partial and complete aggregate shape pairs will be generated based on the 3D aggregate particle library. This dataset will be used as the training data for 3D shape completion.
	
	\item Developing an integrated framework that implements automated 3D point cloud reconstruction, segmentation, completion, and morphological analyses for aggregate stockpiles and constructed layers. To obtain more comprehensive information of aggregate stockpiles and field constructed layers, 3D point cloud reconstruction approach will be developed based on Structure-from-Motion (SfM) techniques. State-of-the-art deep learning architectures for 3D object detection and instance segmentation will be implemented and trained on the labeled point cloud dataset to enable automated segmentation of stockpile and field constructed aggregate clouds. Next, 3D particle shape completion approach as well as 3D morphological analysis algorithms will be developed to characterize the meaningful 3D size, shape, and volumetric properties of the segmented aggregates. Field application and validation of the developed framework will be conducted on field stockpile data to verify the effectiveness and robustness of the framework.
\end{itemize}

\section{Dissertation Outline}

\begin{figure}[!htb]
	\centering
	\includegraphics[width=\textwidth]{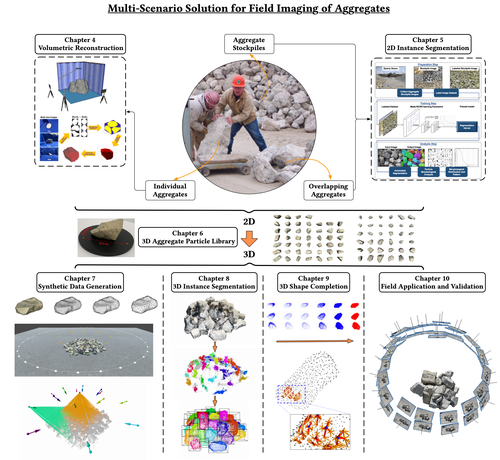}
	\caption{Schematic outline of the dissertation}
	\label{fig: outline}
\end{figure}

This dissertation consists of 11 chapters, including this introduction chapter. A schematic outline of the dissertation is given in \autoref{fig: outline}. The detailed contents of the chapters are as follows:

\begin{itemize}
	\item \textbf{Chapter 2}, titled ``Background Literature Review,'' provides a comprehensive literature review of aggregate production process, aggregate standards and specifications, past aggregate studies and systems that leverage imaging techniques, and key advancements in artificial intelligence and deep learning techniques.
	
	\item \textbf{Chapter 3}, titled ``Field Studies and Sampling of Aggregate Materials at Aggregate Producers,'' provides an overview of field activities undertaken in this research study. This chapter includes aggregate source information from the quarry production sites, material selection and image acquisition criteria, as well as laboratory tests for measuring the ground-truth data of collected samples. 
	
	\item \textbf{Chapter 4}, titled ``Volumetric Reconstruction and Estimation for Individual Aggregates,'' provides the algorithmic details of the segmentation and volumetric reconstruction approach for individual aggregates, and the related ground-truth validation results. This chapter introduces the development of a computer vision–based approach for the volumetric measurement of individual aggregate particles.
	
	\item \textbf{Chapter 5}, titled ``Automated 2D Image Segmentation and Morphological Analyses for Aggregate Stockpiles,'' provides the details of the 2D stockpile segmentation and morphological analysis approaches and the verification results with ground-truth manual labeling. This chapter also includes the established stockpile labeled image dataset, the development of morphological analysis modules, and the completeness and precision analyses of the segmentation results.
	
	\item \textbf{Chapter 6}, titled ``3D Aggregate Particle Library and Comparative Analysis of 2D and 3D Particle Morphologies,'' describes the establishment of a 3D aggregate particle library based on the development of a marker-based 3D reconstruction approach for obtaining full 3D aggregate models. This chapter also includes detailed comparative analyses of 2D and 3D morphological properties and substantiates the advantages of 3D characterization methods for aggregates.
	
	\item \textbf{Chapter 7}, titled ``Synthetic Data Generation of Aggregate Stockpiles for Deep Learning,'' reviews the successful use of synthetic datasets among different tasks in the computer vision domain, as well as the graphics engines that power the synthetic dataset preparation. This chapter introduces a synthetic data generation pipeline designed to simulate densely stacked aggregate stockpiles based on the assembly of instances from the 3D aggregate particle library. The pipeline features the simulation of multi-view cameras and LiDAR sensors and the so-called raycasting techniques to extract 3D dense point clouds with ground-truth labels. 
	
	\item \textbf{Chapter 8}, titled ``Automated 3D Instance Segmentation of Aggregate Stockpiles,'' reviews the state-of-the-art advancements in computer vision regarding the 3D instance segmentation task and analyzes the most suitable strategy for application in the context of dense stockpile segmentation. This chapter discussed the development of a deep learning-based approach for automated stockpile segmentation. Based on the established synthetic dataset, the framework is trained to learn the segmentation of individual aggregate instances from the stockpile.  
	
	\item \textbf{Chapter 9}, titled ``3D Aggregate Shape Completion by Learning Partial-Complete Shape Pairs,'' reviews the current research developments of 3D shape completion in the computer vision domain and implements the state-of-the-art strategy to learn irregular aggregate shapes. This chapter discusses the generation of partial-complete shape pairs based on varying-visibility and varying-view raycasting schemes. A shape completion approach is developed and further evaluated on several unseen aggregate shapes for its robustness and reliability.
	
	\item \textbf{Chapter 10}, titled ``Field Application and Validation of the 3D Reconstruction-Segmentation-Completion Framework,'' presents the collaboration of the developed key components as an end-to-end integrated framework for 3D stockpile analysis. The framework features 3D reconstruction, 3D stockpile segmentation, and 3D shape completion for the morphological characterization of aggregates in dense stockpiles. Field application of the framework is demonstrated and tested on re-engineered stockpiles from collected aggregate samples as well as field stockpiles at the quarry. The robustness and reliability of potential applications using this framework are evaluated by comparing with ground-truth morphological properties and measurement.
	
	\item \textbf{Chapter 11}, titled ``Concluding Remarks and Recommendation,'' provides a summary of research findings as well as recommendations for promising future directions based on this study.
	
\end{itemize}

%% file: chapter02.tex
\chapter{Background Literature Review} \label{chapter-2}

This chapter presents a brief summary of research and practice related to the topics presented in this dissertation. As the background, a detailed literature review is presented on the overall construction aggregates industry, typical aggregate production and manufacturing process, and engineering applications and specifications of aggregates. State-of-the-practice methods and state-of-the-art methods are reviewed to document the currently available techniques for the characterization of aggregates. The essential features of existing machine vision-based aggregate imaging systems are summarized, together with the limitations and knowledge gaps identified in these methods. Next, key concepts and fundamentals in computer vision and deep learning research are presented. Accordingly, the potential for leveraging the advancements in computer vision with deep learning to better characterize aggregates is discussed.

\section{Construction Aggregates Industry}

Against the grand backdrop of the overall mineral and mining industry, construction aggregate materials and their industry have undoubtedly occupied the mainstage in terms of the associated economic volume and value. The use of construction aggregates has even been regarded as an indicator of the economic well-being of the Nation as well as the ``Foundation of America's Future'' \parencite{langer_natural_1988, tepordei_natural_1997, kelly_crushed_1998, wilburn_aggregates_1998}. Construction aggregates are natural mineral and rock materials used in Portland Cement Concrete (PCC), bituminous concrete pavement, road base/subbase, construction fill, railroad ballast, riprap for waterway construction, landscaping, and other construction uses. With aggregates' dual attributes as both engineering materials and commodities, U.S. Department of the Interior (USDOI) and U.S. Geological Survey (USGS) define the construction aggregates industry as the business ecosystem that mine and process crushed stone and/or construction sand and gravel. Domestically, the construction aggregates industry comprised about 5,400 mining companies that manage more than 10,000 operations \parencite{summaries_mineral_2021}. 

\begin{figure}[!htb]
	\centering
	\includegraphics[width=0.5\textwidth]{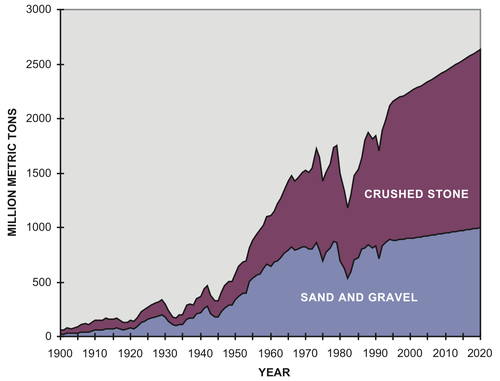}
	\caption{Natural aggregate consumption in the United States (historical and projected) after \textcite{kelly_crushed_1998}}
	\label{fig: 2-1}
\end{figure}

Crushed stone is, by weight, the major raw material used by the construction industry, and sand and gravel are the second mostly used materials. \autoref{fig: 2-1} records the historical consumption and projects the potential consumption of both crushed stone and sand/gravel until the year 2020. The historical trends clearly show the state of the aggregate industry tied closely to the overall economic fluctuations during the growth and recession times. The projections also imply that the crushed stone consumption may increase at a higher rate than that of sand and gravel, which has been observed historically in the U.S. and is very likely to act as the benchmark for developing countries worldwide.

In 2020, the estimated total value of non-fuel mineral production in the United States was \$82.4 billion, in which \$27.0 billion was from construction aggregates production (construction sand and gravel and crushed stone), as shown in \autoref{fig: 2-1}. Among different aggregate types, crushed stone was the leading non-fuel mineral commodity in 2020 with a production value of \$17.8 billion and accounted for 66\% of construction aggregates and 22\% of the total value of U.S. non-fuel mineral production \parencite{summaries_mineral_2021}.
\begin{figure}[!htb]
	\centering
	\includegraphics[width=0.35\textwidth]{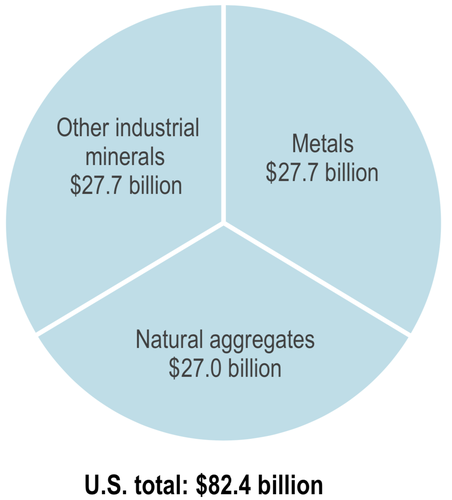}
	\caption{Value of Non-fuel Minerals Produced in 2020 \parencite{summaries_mineral_2021}}
	\label{fig: 2-2}
\end{figure}

\begin{figure}[!htb]
	\centering
	\includegraphics[width=\textwidth]{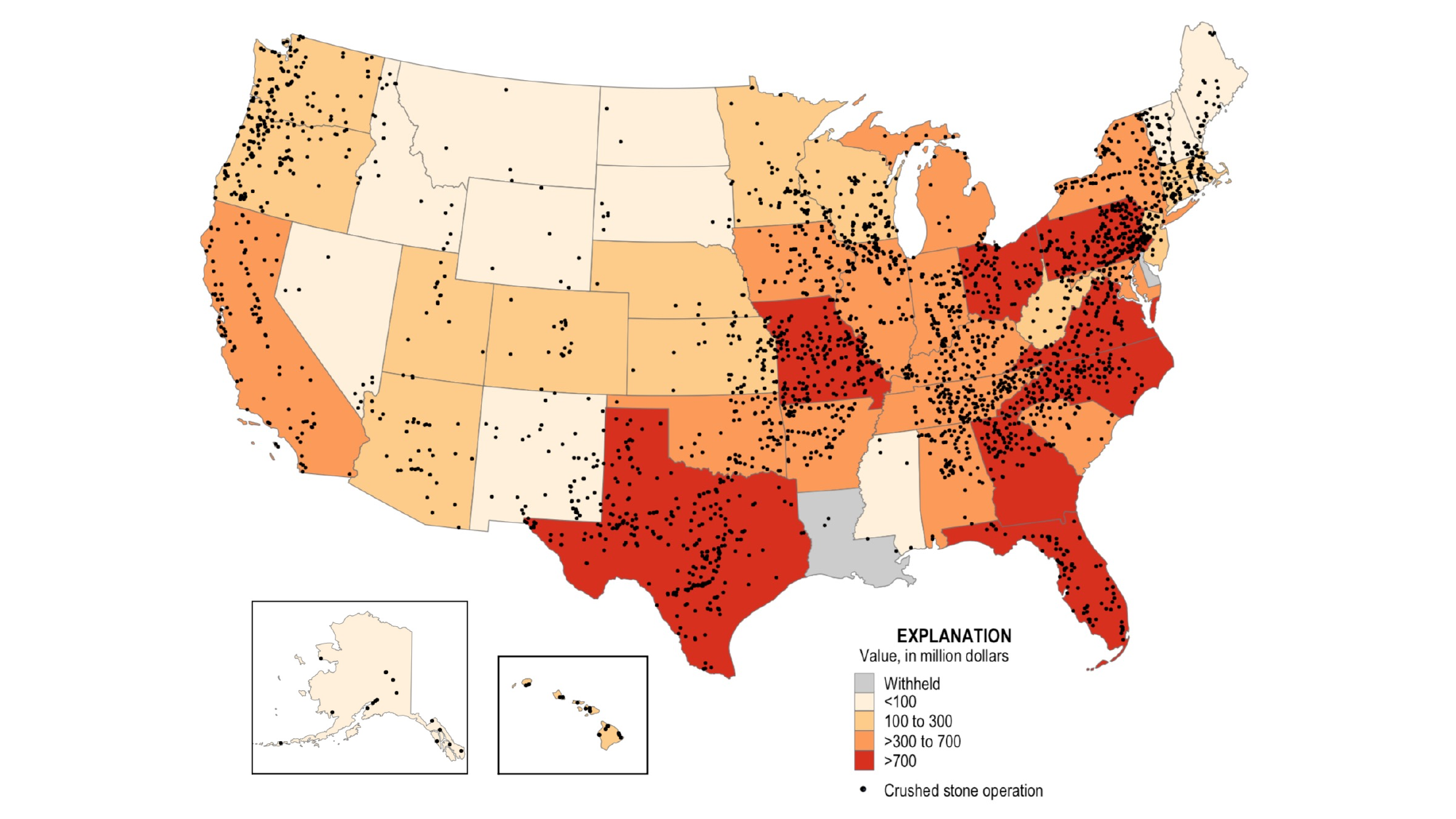}
	\caption{Value of Crushed Stone Produced in 2020 \parencite{summaries_mineral_2021}}
	\label{fig: map-1}
\end{figure}

For crushed stone, 1.46 billion metric tons of crushed stone valued at more than \$17.8 billion was produced by an estimated 1,410 companies operating 3,440 quarries and 180 sales and/or distribution yards across all 50 states. Regarding the mineralogy of the crushed stone, about 70\% was limestone and dolomite; 15\% was granite; 6\% was trap rock; 5\% was miscellaneous stone; 3\% was sandstone and quartzite. At the consumption side, it is estimated that of the 1.5 billion metric tons of crushed stone consumed in 2020, 72\% was used as construction aggregates, mostly for road construction and maintenance; 16\% for cement concrete manufacturing, 8\% for lime manufacturing, 2\% for agricultural uses, and the remainder for other chemical, special, and miscellaneous uses and products. The value and geological sources of crushed stone production in 2020 is illustrated in \autoref{fig: map-1} \parencite{summaries_mineral_2021}.

\begin{figure}[!htb]
	\centering
	\includegraphics[width=\textwidth]{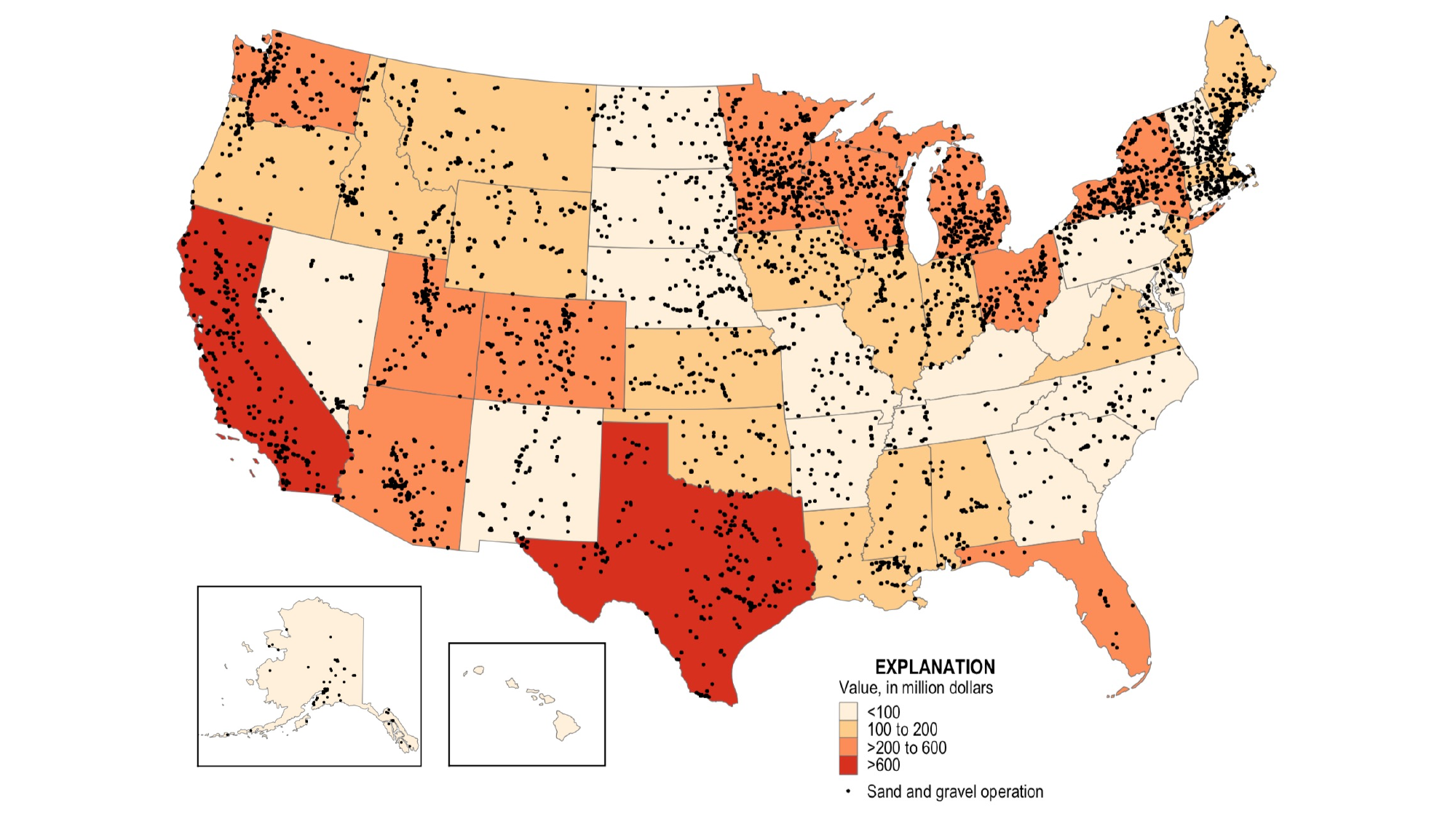}
	\caption{Value of Construction Sand and Gravel Produced in 2020 \parencite{summaries_mineral_2021}}
	\label{fig: map-2}
\end{figure}

As for construction sand and gravel, 960 million metric tons of construction sand and gravel valued at \$9.2 billion was produced by an estimated 3,870 companies operating 6,800 pits and 340 sales and distribution yards in all 50 states. On the consumption side, it is estimated that about 46\% of construction sand and gravel was used as PCC aggregates, 21\% for road base and coverings and stabilization, 13\% for construction fill, 12\% for asphalt concrete aggregate and other bituminous mixtures, and 4\% for other miscellaneous uses. The remaining 4\% was used for applications such as concrete products, railroad ballast, and snow and ice control. The value and geological sources of sand and gravel production in 2020 is illustrated in \autoref{fig: map-2} \parencite{summaries_mineral_2021}.

Overall, the huge volume and value of production and consumption industry have made the construction aggregates to be probably the most fundamental and valuable materials. Therefore, any advancement that may improve the production, storage, transport, quality assurance, quality control, and engineering use of the aggregate materials could lead to game-changing innovations that will have profound impact on the industry.

\section{Aggregate Production and Manufacturing Process}

Aggregate production and related manufacturing processes are usually undertaken in quarries, i.e., open-pit mines where dimension stone, riprap, and construction aggregates are excavated from the ground. The quarrying procedure starts from geological surveying to ascertain the geological conditions at the selected sites. Based on engineering geology, three major types of quarried rocks are sedimentary rocks, igneous rocks, and metamorphic rocks \parencite{gillespie_bgs_1999}. Sedimentary rocks are formed by deposited sediments in water, as precipitations from solution, or as aerial deposits such as volcanic ash. Examples of sedimentary rocks are limestone, dolomite, and sandstone. Igneous rocks are formed from melted and crystallized rocks, with typical examples as granite and basalt. Metamorphic rocks are formed based on the other types of rocks subjected to heat and pressure variations underground, which leads to significant mineralogical changes. Typical metamorphic rocks are slate and marble. As discussed previously, the breakdown of annual aggregate products is about 70\% limestone and dolomite (sedimentary rocks); 15\% granite (igneous rock); 6\% trap rock (igneous rock); 3\% sandstone and quartzite (sedimentary and metamorphic rocks). As a result, most of the aggregate products originate from sedimentary rocks that usually exhibit layered structure that forms beddings in the ground.

\begin{figure}[!htb]
	\centering
	\includegraphics[width=0.8\textwidth]{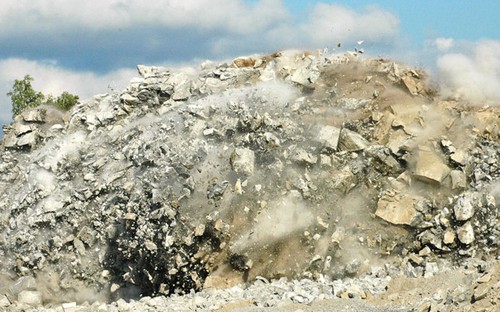}
	\caption{Blasting procedure in the quarrying process. Source: \textcite{quarry_magazine_blasting_2017}}
	\label{fig: blasting}
\end{figure}

After the target bedrock is selected at the quarry sites, a blasting procedure is commonly followed to fragment the rock. Drilling machines are first used to drill vertical boreholes, in which explosives are charged. After further filling the boreholes with clay, ash, fuse and wirings, the blasting holes are fired \parencite{greenwell_practical_1913}. The strong explosives will often break the bottom layers into smaller pieces, while the upper layers will spall to become large rocks, as shown in \autoref{fig: blasting}. Accordingly, large raw rock fragments will fall onto a pile of the small pieces and fines. 

\begin{figure}[!htb]
	\centering
	\includegraphics[width=0.8\textwidth]{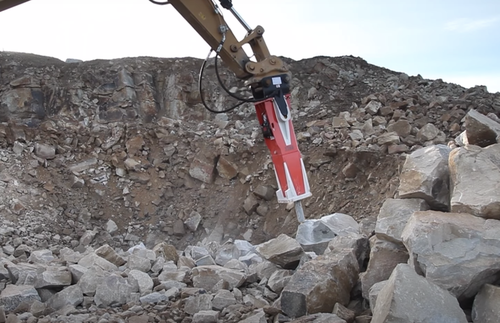}
	\caption{Rock breaking procedure for over-sized rock fragments. Source: \textcite{rammer_hammers_rammer_2013}}
	\label{fig: breaking}
\end{figure}

The large rock fragments usually contain many over-sized rocks (``shot rock'') that either are not cost-effective for on-site transport or cannot fit properly into the next crushing procedure. Therefore, an intermediate rock breaking step is performed at the blasting site, as shown in \autoref{fig: breaking}. These raw fragments are typically beyond the largest riprap size category such as RR7 in IDOT standard \parencite{idot_standard_2016}, Class V in MnDOT standard \parencite{mndot_standard_2018}, and Class 900 in NDOT standard \parencite{ndot_standard_2014}. Hydraulic rock breakers are used to break these rocks into specific size categories. A common practice at the aggregate producers for these large rocks is to directly break them into certain large size categories (e.g., RR5 to RR7 per IDOT standard) by visual judgment and transport them to stockpiles. Therefore, it is noted that the QA/QC is especially lacking in these large size categories, due to the absence of a standard crushing and screening process.

\begin{figure}[!htb]
	\centering
	\begin{subfigure}[b]{0.45\textwidth}
		\centering
		\includegraphics[height=6cm]{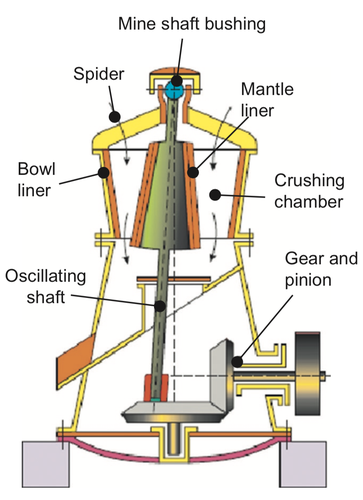}
		\caption{}
	\end{subfigure}
	\hfill
	\begin{subfigure}[b]{0.45\textwidth}
		\centering
		\includegraphics[height=6cm]{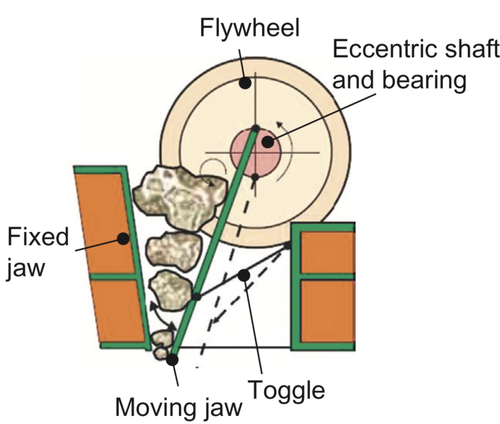}
		\caption{}
	\end{subfigure}
	\newline 
	\begin{subfigure}[b]{0.45\textwidth}
		\centering
		\includegraphics[height=6cm]{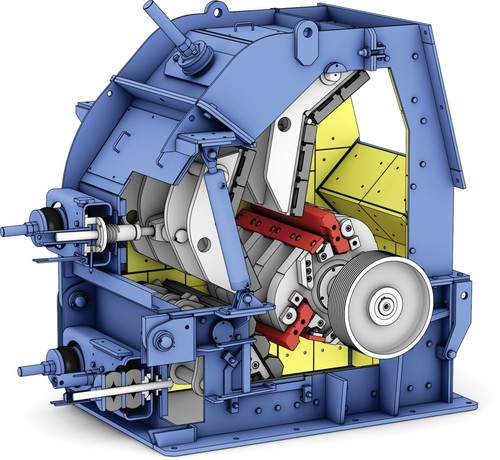}
		\caption{}
	\end{subfigure}
	\hfill
	\begin{subfigure}[b]{0.45\textwidth}
		\centering
		\includegraphics[height=6cm]{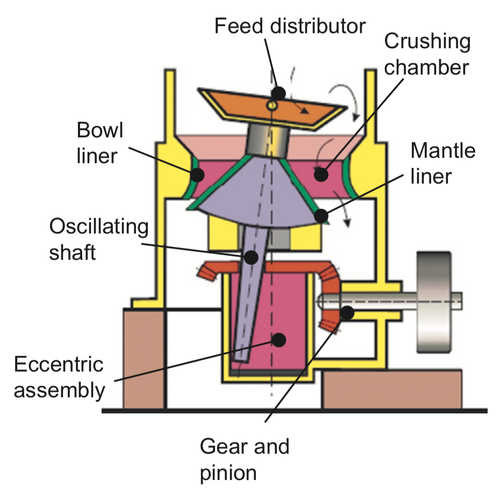}
		\caption{}
	\end{subfigure}
	\caption{(a) Gyratory crusher, (b) jaw crusher, (c) impact crusher, and (d) cone crusher. Sources: \textcite{jankovic_developments_2015, bhs_impact_2021}}
	\label{fig: crusher}
\end{figure}

As the next step, small rocks and the relatively small fragments after rock breaking are transported by loader trucks to the multi-stage crushing and screening system. The crushing system at a quarry usually contains three crushing stages: primary, secondary, and tertiary. The crushers utilized in the primary crushing stage are typically gyratory crushers, jaw crushers, and impact crushers. For secondary and tertiary crushing stages, cone crushers are the most commonly used \parencite{jankovic_developments_2015}. The common crusher types are presented in \autoref{fig: crusher}, and the typical input and output sizes of different crushers are listed in \autoref{tab:crusher}. After each crushing, the material passes through screening and is separated into different stockpiles based on size. 

\begin{table}[!htb]
	\centering
	\caption{Maximum Input and Output Sizes for Common Crusher Types (after \textcite{jankovic_developments_2015})}
	\label{tab:crusher}
	\begin{tabular}{L{3cm}L{3cm}L{3cm}L{3cm}}
		\hline
		\textbf{Crusher Type} & \textbf{Typical Process Stage} & \textbf{Maximum Input Size (mm)} & \textbf{Maximum Output Size (mm)} \\ \hline
		Gyratory Crusher & Primary           & 1,500 & 200-300       \\
		Jaw Crusher      & Primary           & 1,400 & 200-300       \\
		Impact Crusher   & Primary / Secondary & 1,300 & 200-300       \\
		Cone Crusher     & Secondary         & 450   & 60-80         \\
		Cone Crusher     & Tertiary          & 150   & \textless{}30 \\ \hline
		\multicolumn{4}{l}{Note: 1 mm = 0.0393 in.}
	\end{tabular}
\end{table}

\begin{figure}[!htb]
	\centering
	\includegraphics[width=\textwidth]{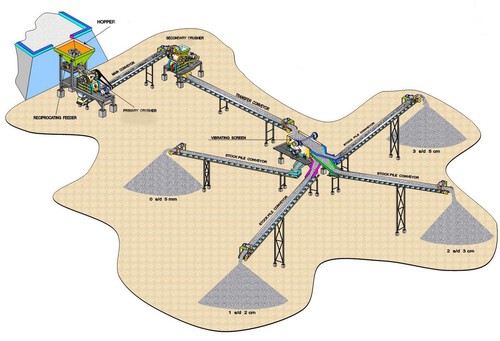}
	\caption{Typical layout of a multi-stage crushing and screening system. Source: \textcite{manufactor_flow_2013}}
	\label{fig: plant}
\end{figure}

\begin{figure}[!htb]
	\centering
	\includegraphics[width=0.5\textwidth]{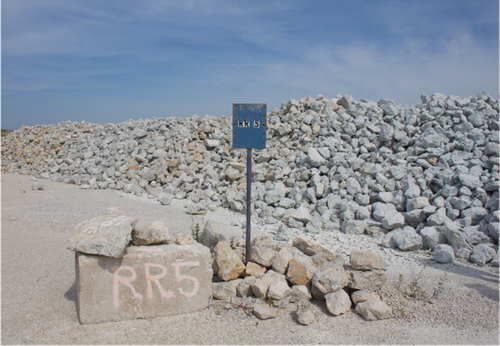}
	\caption{A typical stockpile of large-sized aggregates}
	\label{fig: large-stockpile}
\end{figure}

A typical layout of the multi-stage crushing and screening system is illustrated in \autoref{fig: plant}. After screening, aggregates are separated and stored in different stockpiles. These stockpiles are mostly fine and coarse aggregates that can be conveniently transported over the conveyors. Large-sized aggregates are usually transported by haul trucks (or dump trucks) and stored in more spacious locations away from the crushing-screening system, as shown in \autoref{fig: large-stockpile}. 

By reviewing the general aggregate production and manufacturing process, the bedrock properties and the rock breaking/crushing process can play important roles in the quality and characteristics of aggregate products, especially for large-sized aggregates products. First, the geological beddings usually vary greatly in thickness. They may form thin seams with less than an inch (2.54 centimeters) in thickness, or they may form massive layers with many feet (one foot equals 30.48 centimeters) thick \parencite{greenwell_practical_1913}. Even the same bedding may show variation in thickness from location to location, which mainly depends on the amount and property of the material when forming the sediment. It is readily seen that the size and breakage of raw blasted rocks from the ground will strongly rely on the bedding features of the layer. In addition, thin lamination layers in the bedding may affect the raw size of rocks as well. Lamination is usually a thin clay parting within the bedding that reduces the effective thickness (i.e., vertical distance between the roof and floor of the deposit). With these fundamental geological reasons, the large-sized crushed aggregates may exhibit high randomness in terms of its morphological properties. Additionally, unlike the regular-sized aggregates that undergo multi-stage crushing and screening, the manual rock breaking process for large-sized aggregates described above brings more randomness and has less control on the aggregate quality.

\section{Types of Aggregates and Their Engineering Applications}

Aggregate products are typically categorized as fine aggregates and coarse aggregates based on the standard test procedures established by American Society for Testing and Materials (ASTM) and American Association of State Highway Officials (AASHTO). In the scope of this study, the relatively large-sized aggregates are also considered. Their standards are mostly established by U.S. Army Corps of Engineers (USACE) and Departments of Transportation (DOT) in many states. A brief summary of the typical size range and usage of aggregate types is given in \autoref{tab:aggregate-size}.

\begin{table}[!htb]
	\centering
	\caption{Size and Usage of Typical Aggregate Types}
	\label{tab:aggregate-size}
	\begin{tabular}{L{3cm}L{3cm}L{3cm}L{3cm}}
		\hline
		\textbf{Aggregate Type} & \textbf{Size}    & \textbf{Usage}                                                  & \textbf{Description}                        \\ \hline
		Fine Aggregates         & 0.003 in. (0.075 mm) to 0.187 in. (4.75 mm) & mortar, plaster, concrete, asphalt mixture, pavement filling, etc.               & sand, fly-ash, fine crushed particles, etc. \\
		\hline
		Coarse Aggregates       & 0.187 in. (4.75 mm) to 2.953 in. (75 mm)    & concrete, asphalt mixture, pavement base, railway ballast, etc. & gravel, crushed stone, crushed cement concrete                  \\
		\hline
		Large-sized Aggregates & > 2.953 in. (75 mm) & armor for streambeds, bridge abutments, pilings, and shoreline structures & riprap, jetty stone, cap stone \\ \hline
	\end{tabular}
\end{table}

\subsection{Fine Aggregates and Coarse Aggregates}

Fine aggregates are commonly used in mortar, plaster, and as the filling material in concrete and pavement layers. Specifically, fine aggregates are widely used in PCC as well as for bituminous paving applications. According to the \textcite{astm_c33_standard_2013} specification, fine aggregates are defined as the materials that pass 3/8''-inch (9.525 mm) or No. 4 sieve and are retained on No. 200 sieve. Therefore, the typical size range is denoted as 0.003 in. (0.075 mm) to 0.187 in. (4.75 mm)  in \autoref{tab:aggregate-size}.

Coarse aggregates are often used in PCC, asphalt mixtures, pavement base and subbase, as well as railway ballast. According to the \textcite{astm_c33_standard_2013} specification, coarse aggregates are defined as the material that are retained on the 3/8''-inch (9.525 mm) or No. 4 sieve. Typical size range of coarse aggregates is between 0.187 in. (4.75 mm) to 2.953 in. (75 mm), with the majority of particles sizing between 1.476 in. (37.5 mm) and 1.969 in. (50 mm). For the sake of brevity, fine aggregates and coarse aggregates are referred to herein as ``regular-sized aggregates'' to distinguish from the large-sized aggregates.

\subsection{Riprap Material and Large-Sized Aggregates}

\begin{figure}[!htb]
	\centering
	\begin{subfigure}[b]{0.45\textwidth}
		\centering
		\includegraphics[height=6cm]{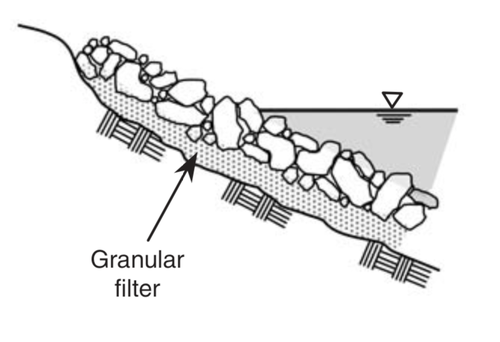}
		\caption{Revetment Riprap for Slope Protection}
	\end{subfigure}
	\hfill
	\begin{subfigure}[b]{0.45\textwidth}
		\centering
		\includegraphics[height=5cm]{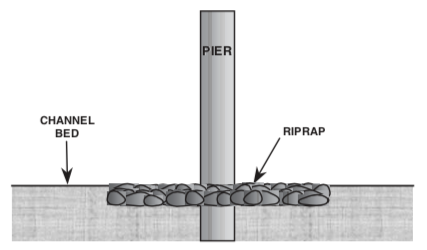}
		\caption{Bridge Pier Protection}
	\end{subfigure}
	\newline 
	\begin{subfigure}[b]{0.45\textwidth}
		\centering
		\includegraphics[height=6cm]{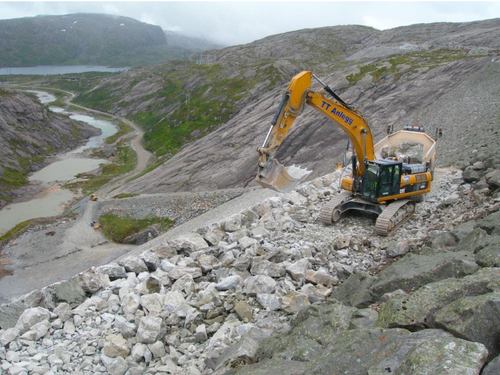}
		\caption{Riprap in Rockfill Dams}
	\end{subfigure}
	\hfill
	\begin{subfigure}[b]{0.45\textwidth}
		\centering
		\includegraphics[height=6cm]{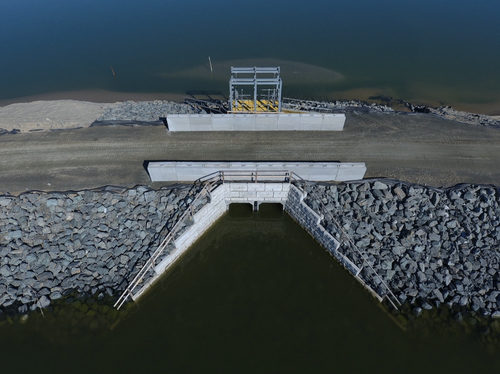}
		\caption{Riprap in Retention Dikes}
	\end{subfigure}
	\newline 
	\begin{subfigure}[b]{0.45\textwidth}
		\centering
		\includegraphics[height=6cm]{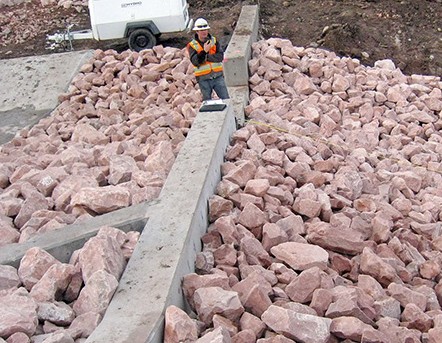}
		\caption{Riprap for Scour/Erosion Control}
	\end{subfigure}
	\hfill
	\begin{subfigure}[b]{0.45\textwidth}
		\centering
		\includegraphics[height=6cm]{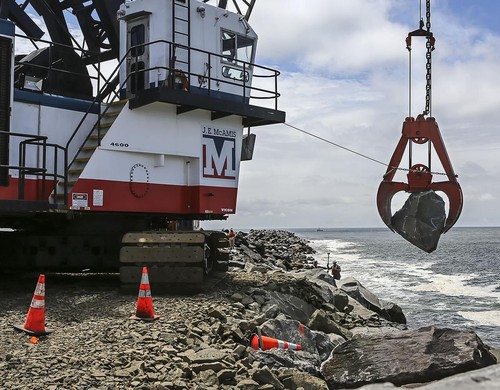}
		\caption{Jetty Stone for Shoreline Protection}
	\end{subfigure}
	\caption{Typical engineering applications of large-sized aggregates. Sources: \textcite{lagasse_riprap_2006, shari_phiel_work_2015, hiller_riprap_2017, magnumstone_eco-friendly_2020, lovas_what_2021}}
	\label{fig: riprap}
\end{figure}

While the applications of fine and coarse aggregates are well known, the importance of riprap material or large-sized aggregates has drawn less attention from the structural and transportation community because they are more commonly used in hydraulic engineering applications. Riprap is large-sized rock used to armor shorelines, streambeds, bridge abutments, pilings, and other coastal structures against scour and water or ice erosion. It is made from a variety of rock types, commonly granite or limestone and occasionally recycled concrete rubble from building and paving demolition. They serve as an important functional component by providing water/ice erosion control, sediment control, rockfill, and scour protection against hydraulic and environmental stresses \parencite{idot_standard_2016}. For a natural material, the reliable and sustainable use of riprap as an integrated system in engineering requires quality control throughout the design, production, transport, installation, inspection, and maintenance stages \parencite{lagasse_riprap_2006}. Apart from the structural geometry, slope stability, and hydraulic analyses of the structures, case studies on riprap failure in stream channels and bridge piers indicate that undersized and open-graded riprap often cause insufficient resistance to hydraulic shear stress \parencite{blodgett_rock_1986, chiew_mechanics_1995, richardson_evaluating_2001, lagasse_bridge_2001}. According to \textcite{usace_em_1110-2-2302_engineering_1990}, large-sized aggregates typically refer to any aggregate with size greater than the regular-sized concrete aggregates. In addition to riprap, jetty stone and cap stone are also types of large-sized aggregates that have even larger dimensions. Several typical engineering applications of riprap and large-sized aggregates are presented in \autoref{fig: riprap}.

\section{State-of-the-Practice Characterization Methods for Regular-Sized and Large-Sized Aggregates}

Despite the vastly different engineering applications of regular-sized aggregates and large-sized aggregates, the characterization methods for these aggregate materials are equally important. Recall that the large-sized aggregates are not only engineering materials used in hydraulic applications, but also they are upstream products during the aggregate production process. Therefore, QA/QC checks are required at both quarry production lines and construction sites by both aggregate producers and state DOTs. 

During the material selection and QA/QC process, characterizing the size and shape properties has become a focal point for aggregate studies. Size and morphological/shape properties of aggregates primarily influence the macroscopic behavior and performance of aggregate skeleton assemblies of constructed layers in transportation infrastructure, e.g., asphalt concrete and Portland cement concrete \parencite{quiroga_effect_2003, polat_correlation_2013}, unbound/bound layers in highway and airfield pavements \parencite{liu_influence_2019, tutumluer_aggregate_2008, bessa_aggregate_2015}, the ballast layer in railway tracks \parencite{huang_discrete_2010, wnek_investigation_2013}, and riprap materials for erosion control and hydraulic applications \parencite{lutton_evaluation_1981, lagasse_riprap_2006}. Across all size ranges, aggregate shape properties in terms of form (e.g., flatness and elongation), angularity, and texture have been used to characterize their morphology \parencite{barrett_shape_1980}. The information on aggregate morphology greatly facilitates the quality control process and the in-depth understanding of aggregate layer behavior linked to its composition and packing. 

For producers and practitioners, the aggregate size and shape are important for QA/QC requirements throughout the production line and mix design \parencite{astm_d448_standard_2017, astm_d2940_standard_2015, astm_d6092_standard_2014}. Different quarrying processes and rock mineralogy introduce randomness to the quality of produced aggregates. Therefore, convenient and continuous monitoring of quarry products is important for the efficient material selection and construction. On the other hand, discrete mechanics that realistically model the inter-particle and assembly behavior of granular materials require properly characterizing the morphological properties of aggregates. Through recently focused research efforts on modeling the aggregate layer behavior using Finite Element Method (FEM) and Discrete Element Method (DEM), aggregate morphological properties have gained increased importance, after the grain size distribution, to capture complex behaviors of granular materials. This is especially challenging for most stone skeleton layers in constructed road pavements, e.g., surface course mixtures such as hot-mix asphalt (HMA) and PCC and unbound aggregate base/subbase, which are subjected to vehicular dynamic loading conditions \parencite{huang_discrete_2010, chen_discrete_2011, ghauch_micromechanical_2014, qian_integrated_2014}.

In summary, the morphological properties of aggregates (such as size, shape, volume/weight, etc.) are some of the most crucial indicators for aggregate QA/QC, especially for crushed stone aggregates and riprap. 

\subsection{Laboratory Methods for Regular-sized Aggregates}
\begin{figure}[!htb]
	\centering
	\includegraphics[width=0.6\textwidth]{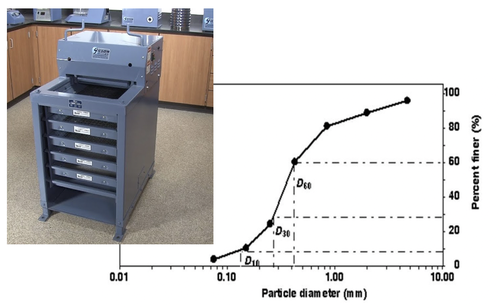}
	\caption{Sieve analysis for regular-sized aggregates}
	\label{fig: sieve}
\end{figure}

\begin{figure}[!htb]
	\centering
	\includegraphics[width=0.4\textwidth]{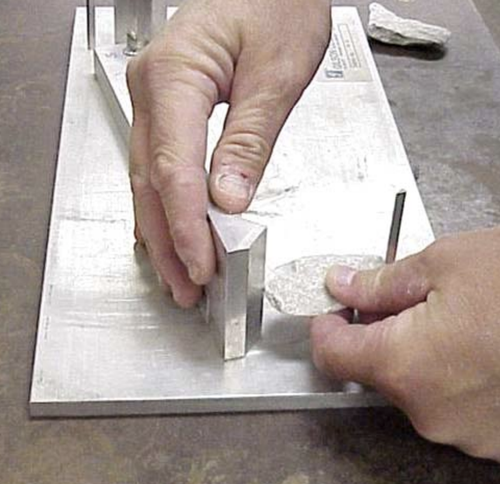}
	\caption{Proportional caliper device for regular-sized aggregates. Source: ASTM D4791}
	\label{fig: caliper}
\end{figure}

For regular-sized crushed aggregates, state-of-the-practice methods include using sieving equipment for size gradation determination (see \autoref{fig: sieve}) and using proportional caliper device (see \autoref{fig: caliper}) for flat and elongated shape determination. 

\subsection{Field Methods for Large-sized Aggregates}
Despite the ongoing development of guidelines for size selection of riprap in design, the practical procedures for characterizing riprap size and shape properties in the field are still subjective and qualitative, primarily because of difficulties associated with measuring sizes of these large rocks. As the gradation requirement, \textcite{idot_standard_2016} specifies that the riprap sizing should be well-graded, with a maximum of $15\%$ of the total test sample by weight may be oversized material, and each oversized piece shall not exceed the maximum permissible particle size by more than $20\%$. 

As compared to coarse aggregates used in transportation engineering, the sizes of which typically range from 0.187 in. (0.475 cm) to 5 in. (12.7 cm) \parencite{astm_d448_standard_2017, astm_d2940_standard_2015}, individual riprap rock can weigh up to 1,150 lbs. (522 kg) with nominal sizes up to 24 in. (61.0 cm) \parencite{idot_standard_2016, astm_d6092_standard_2014}. Laboratory sieve analysis is usually conducted to determine the gradation of small- to medium-sized aggregates, but large-sized riprap material makes this task impractical. Because there is no uniform way to define the sizes or dimensions of individual rocks, standards or guidelines usually specify riprap gradation requirements in terms of weight. The current practice also uses a weight-based metric instead of a size-based one, since weight is easier to measure and quantify for such large-sized aggregates. This approach is based on the assumption that the weight of the riprap correlates with its actual size. However, measuring the weight of individual rock pieces is still a time-consuming and labor-intensive task. 

For relatively large-sized aggregates such as riprap, state-of-the-practice methods have been restricted to subjective visual inspection approaches and/or labor-intensive manual measurement of individual pieces, primarily due to the difficulties in mobilizing these aggregates of large size and heavy weight. At the current state of the practice, a nationwide AASHTO survey of transportation agencies in the US and Canada has indicated that riprap characterization is mostly based on visual inspection and manual measurements \parencite{sillick_member_2017}. Visual inspection depends greatly on the experience and expertise of practitioners. In this method, certain gauge or keystones and sample stockpiles are usually used as a reference to assist the judgment \parencite{lippert_inspection_2012}. To better estimate the size distribution, the Wolman count method is applied by statistically sampling and measuring rocks within a stockpile \parencite{lagasse_riprap_2006}. For instance, the use of keystones with predefined weight ranges has been adopted recently by IDOT to facilitate the visual inspection process. For manual measurement, transportation agencies either weigh individual particles directly or use size-mass conversion after measuring rock dimensions. USACE requires direct weight measurement of individual riprap rocks as specified in \textcite{usace_em_1110-2-2302_engineering_1990} for large stone construction. Alternatively, the size-mass conversion proposed in \textcite{astm_d5519_standard_2015} requires measurement of the midway dimension or circumference from three orthogonal axes and estimates the volume based on a cuboid assumption or averaged sphere-cube assumption. Despite these great efforts, the visual inspection and manual measurement can only provide rough estimations that do not necessarily represent realistic riprap properties, and an objective and efficient approach for quantitatively characterizing the size and shape of riprap has yet to be established. In this regard, establishing reliable field imaging techniques is a promising approach to easily and quickly process stockpile images of riprap for gradation checks and provide data analytics.

Following the overall USACE guidelines in \textcite{usace_em_1110-2-2302_engineering_1990}, state DOTs have implemented customized field approaches to estimate the size and weight of individual rocks. As an example, IDOT's riprap gradation requirements and size/weight categories are introduced as a common standard for aggregate specifications. 

The current IDOT specification for riprap classification into different “RR” categories is based on the grain size distribution, which is determined by the weight distribution of the riprap stones. IDOT published a policy memorandum \parencite{idot_policy_2018} for the classification of riprap based on weight. This memorandum also requires a visual inspection of the riprap stockpiles, including inspections for flat and elongated pieces. A collection of riprap keystones shall be maintained by the producers for all produced riprap gradations, as outlined in \autoref{tab:2-1}, to assist with the visual gradation. IDOT requires that the set of keystones shall be representative of the stockpile gradation and be replaced with a new set if they become non-representative. 

\begin{table}[!htb]
	\centering
	\caption{Keystone Requirements for Different Riprap Size/Weight Categories}
	\label{tab:2-1}
	\begin{tabular}{llll}
		\hline
		Gradation & Keystone \#1 (lbs.) & Keystone \#2 (lbs.) & Keystone \#3 (lbs.) \\ \hline
		RR3       & 50 (±5)             & 10 (±1)             & 1 (±0.1)            \\
		RR4       & 150 (±15)           & 40 (±4)             & 1 (±0.1)            \\
		RR5       & 400 (±40)           & 90 (±13)            & 3 (±0.1)            \\
		RR6       & 600 (±60)           & 170 (±17)           & 6 (±0.5)            \\
		RR7       & 1000 (±100)         & 300 (±30)           & 12 (±1)             \\ \hline
		\multicolumn{4}{l}{Note: 1 lb. = 453.6   grams}                            
	\end{tabular}
\end{table}

If the gradations by visual inspection were disputed by the producer, a second visual inspection is conducted by IDOT Central Bureau of Materials (CBM). If the second visual inspection is again disputed by the producer, a representative sample is excavated from the working face of the stockpile and spread over the length of a marked grid to a one-rock thickness and weighed piece by piece for riprap categories RR3 to RR7. The rock spalls and fines below the minimum specified weight are collected and included in the calculations for each size range. The grid size for each riprap category is outlined in \autoref{tab:2-2}. The grid length is broken into blocks of 5 in. (12.7 cm) long. Note that for riprap categories RR1 and RR2, the grain size distribution is performed by conventional sieve analysis in accordance with Illinois Test Procedure 27, outlined in IDOT's Manual of Test Procedures for Materials \parencite{idot_manual_2019}.

\begin{table}[!htb]
	\centering
	\caption{Grid Size Requirements for Sampling Different Riprap Gradation Categories}
	\label{tab:2-2}
	\begin{tabular}{lll}
		\hline
		Gradation & Grid   Size (ft.) & \begin{tabular}[c]{@{}l@{}}Sample   Size\\    \\ (Min. Number of Tested Blocks)\end{tabular} \\ \hline
		RR3         & 2' by 25'        & 2        \\
		RR4         & 3' by 25'        & 2        \\
		RR5         & 4' by 25'        & 3        \\
		RR6         & 5' by 30'        & 3        \\
		RR7         & 5' by 35'        & 3        \\ \hline
		\multicolumn{3}{l}{Note: 1 ft. = 30.48 cm}
	\end{tabular}
\end{table}

\begin{figure}[!htb]
	\centering
	\includegraphics[width=0.8\textwidth]{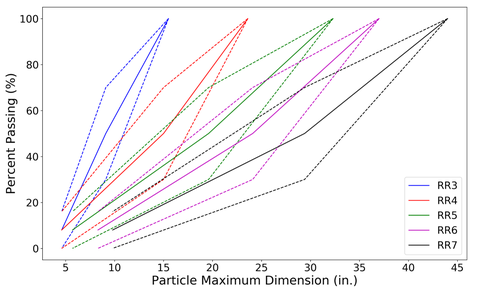}
	\caption{Converted particle-size distribution of IDOT riprap categories RR3 to RR7}
	\label{fig: 3-1}
\end{figure}

Based on IDOT policy memorandum \parencite{idot_policy_2018}, the procedure for riprap size characterization entails using three keystone particles (used as control points) to identify gradations. \autoref{fig: 3-1} shows the upper, lower, and midpoint gradation lines for IDOT's riprap categories RR3 to RR7. This plot assumes a flat and elongated ratio of 2:1, a specific gravity of 2.5, and an ellipsoidal particle shape for a standardized weight to volume/size conversion. Note that the maximum dimension of a particle is used to indicate size on the horizontal axis.

\begin{figure}[!htb]
	\centering
	\begin{subfigure}[b]{0.45\textwidth}
		\centering
		\includegraphics[height=6cm]{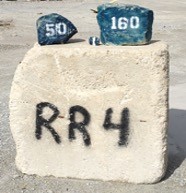}
	\end{subfigure}
	\hfill
	\begin{subfigure}[b]{0.45\textwidth}
		\centering
		\includegraphics[height=6cm]{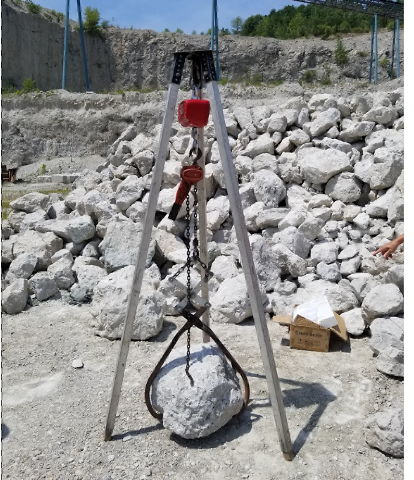}
	\end{subfigure}
	\caption{(a) keystone-based method to facilitate visual inspection and (b) Tripod weighting tool for individual aggregates}
	\label{fig: idot}
\end{figure}

\begin{figure}[!htb]
	\centering
	\includegraphics[width=0.6\textwidth]{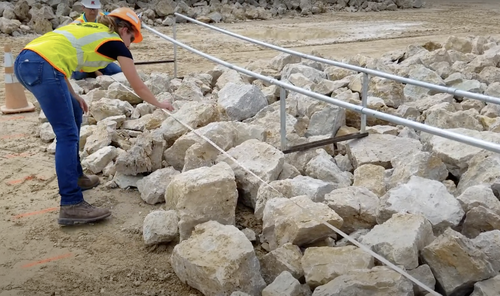}
	\caption{Wolman count method for large-sized aggregates. Source: \textcite{bartelt_wolman_2018}}
	\label{fig: wolman}
\end{figure}

To emulate the sieving process of regular-sized aggregates, the Wolman count method and Galay transect approach are designed to determine a size distribution of large aggregate assemblies based on a random sampling of individual rocks within a matrix. Both methods are widely accepted in practice and rely on samples taken from the surface of the matrix to make the method practical for use in the field. Details of the methods can be found in \textcite{wolman_method_1954, galay_river_1987, bunte_sampling_2001}. The application of Wolman count method is illustrated in \autoref{fig: wolman}. A field approach of the Wolman count method is to stretch a survey tape over the surface and measure each particle located at equal intervals along the tape. The interval recommended for riprap is at least 1 ft. (30.48 cm) for small riprap and increased for larger riprap. The intermediate dimension of each aggregate is then measured for a total of 100 particles. The longer and shorter axes can also be measured to determine particle shape. \textcite{kellerhals_sampling_1971} provide an analysis that supports the conclusion that a surface sample following the Wolman method is equivalent to a bulk sample sieve analysis. In general, the Wolman count method is a combination of visual inspection and manual measurement.

\section{State-of-the-Art Aggregates Characterization Based on Machine Vision}

\subsection{Existing Aggregate Imaging Systems}

Over the past two decades, imaging systems based on machine vision have been widely developed and adopted to characterize size and shape properties from the digital images of aggregates \parencite{rao_quantification_2002, al-rousan_new_2005, pan_aggregate_2006, wang_evaluation_2013, moaveni_evaluation_2013, hryciw_innovations_2014}. Most imaging systems are applicable to aggregates with maximum sizes less than 6 in. (15.2 cm) using a fixed-position camera setup for image acquisition in the laboratory. A brief summary of representative aggregate imaging systems developed is presented as follows.

\begin{figure}[!htb]
	\centering
	\includegraphics[width=\textwidth]{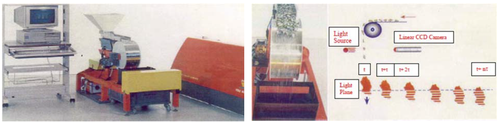}
	\caption{VDG-40 Videograder. Source: \textcite{browne_comparison_2001}}
	\label{fig: vdg-40}
\end{figure}

French public works laboratory (LCPC) developed the VDG‐40 Videograder, which uses
an electromagnetic vibrator to extract the constituents of the sample in a hopper along a feed channel. A separator drum was used to orient the aggregate particles towards the falling plane at the desired speed. A line-scan camera acquires the images of aggregate particles as they fall in front of a back-light \parencite{browne_comparison_2001}. Each particle's third dimension is computed from 2D projected image based on the assumption of elliptical particles. As such, this system is capable of measuring the particle size distribution, flatness and slenderness ratios. Photo of VDG‐40 Videograder and its workflow are presented in \autoref{fig: vdg-40}.

\textcite{masad_development_2003, gates_fhwa-hif-11-030_2011} developed the Aggregate Imaging System (AIMS), which consists of one slide-mounted camera and two lighting sources visualizing aggregates with a maximum size up to 1 in. (2.54 cm). Two modules are incorporated in this system. The first module is for the analysis of fine aggregates; black and white images are captured using a video camera and a microscope. The second module is used for the analyses of coarse aggregates; grayscale images as well as black and white images are captured. Fine aggregates are analyzed for shape and angularity, while coarse aggregates are analyzed for shape, angularity and texture. A video microscope is used to determine the depth of particles, while the images of 2D projections provide the other two dimensions. These three dimensions quantify the shape of particle. Additionally, angularity is determined using gradient method by analyzing the black and white images, while texture is determined by analyzing the grayscale images using wavelet image processing technique.
As an improved version of the prototype AIMS, a second generation of this imaging system, AIMS2 \parencite{gates_fhwa-hif-11-030_2011}, has been developed. The improvements in the new system includes a variable magnification microscopic‐camera system and two different lighting configurations to capture aggregate images for analysis. Additionally, the entire system is placed inside a box with a door to reduce the effect of ambient light on the quality of captured images.
\begin{figure}[!htb]
	\centering
	\begin{subfigure}[b]{0.45\textwidth}
		\centering
		\includegraphics[height=5cm]{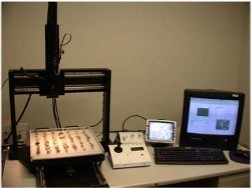}
		\caption{AIMS}
	\end{subfigure}
	\hfill
	\begin{subfigure}[b]{0.45\textwidth}
		\centering
		\includegraphics[height=5cm]{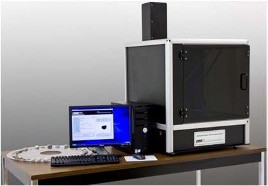}
		\caption{AIMS2}
	\end{subfigure}
	\caption{Aggregate Imaging System (AIMS). Source: \textcite{gates_fhwa-hif-11-030_2011}}
	\label{fig: aims}
\end{figure}

\begin{figure}[!htb]
	\centering
	\begin{subfigure}[b]{0.45\textwidth}
		\centering
		\includegraphics[height=6cm]{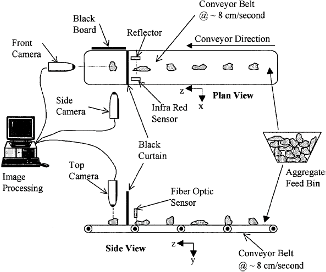}
		\caption{Schematic drawing}
	\end{subfigure}
	\hfill
	\begin{subfigure}[b]{0.45\textwidth}
		\centering
		\includegraphics[height=6cm]{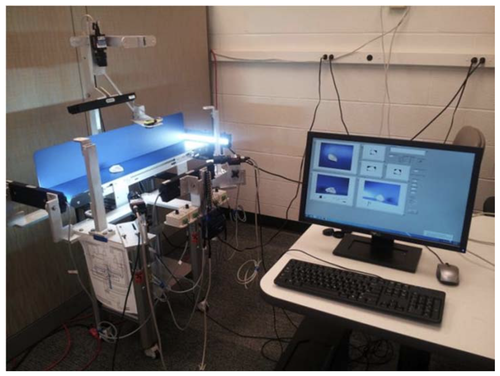}
		\caption{Final assembly of the E-UIAIA system}
	\end{subfigure}
	\caption{Enhanced-University of Illinois Aggregate Image Analyzer (E-UIAIA). Sources: \textcite{tutumluer_video_2000, moaveni_evaluation_2013}}
	\label{fig: euiaia}
\end{figure}

\textcite{tutumluer_video_2000} originally developed and later \textcite{moaveni_evaluation_2013} improved the Enhanced-University of Illinois Aggregate Image Analyzer (E-UIAIA) to automate the process of measuring the shape and size properties of coarse aggregates.  E-UIAIA uses three Charge-Coupled Device (CCD) cameras with sensor resolution of $1292\times 964$ pixels to capture images of aggregate particles from top, side and front views. Using these three orthogonal views, the volume, surface area, surface texture, angularity and size of each aggregate particle are evaluated. Infrared and fiber optic sensors detect the location of the particles on the conveyor and they send a signal to trigger three cameras. A delay of 1/30 of a second is set between detecting a particle and camera triggering to let the particles move into the field of view of three cameras. This system can process aggregates with size no greater than 3 in. (7.6 cm). \autoref{fig: euiaia} demonstrates the schematic drawing and final assembly of E-UIAIA. 

\begin{figure}[!htb]
	\centering
	\includegraphics[width=0.5\textwidth]{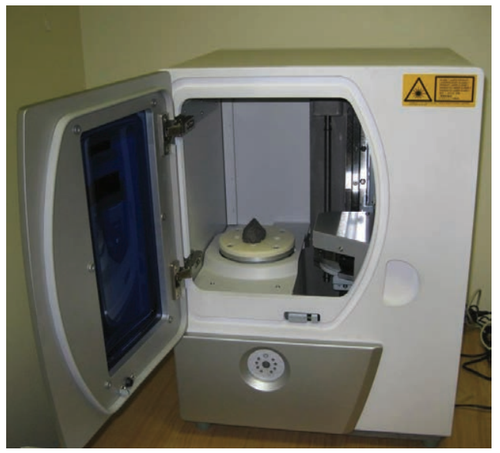}
	\caption{3D laser scanning system of aggregates. Sources: \textcite{komba_analytical_2013, anochie-boateng_three-dimensional_2013}}
	\label{fig: csir}
\end{figure}

\textcite{komba_analytical_2013, anochie-boateng_three-dimensional_2013} used a 3D laser scanning type of aggregate system that scans individual aggregates and analyzes the generated 3D mesh model with a maximum size of 0.75 in. (1.9 cm), as illustrated in \autoref{fig: csir}. The 3-D laser scanning device used is an LPX-1200, originally designed by Roland DGA Corporation for solid-shape modeling in medical and manufacturing applications. The device uses a laser beam moving horizontally and vertically to scan objects at a predefined resolution. The maximum scanning resolution is 0.1 mm (100 µm).

\textcite{ohm_translucent_2013, hryciw_innovations_2014} developed a Translucent Segregation Table (TST) to evaluate the effect of aggregate size and morphology on the shear strength properties of the material. It is a 36 in. $\times$ 36 in. (91 cm $\times$ 91 cm)
translucent backlit plate which tilts upwards 35 degrees for specimen preparation. The
soil is introduced at the top of the incline and the particles slide or roll downward
passing beneath a series of “bridges” having progressively smaller underpass heights.
Particle blockages behind the bridges can be disrupted by mild brushing of the grains
with horizontal strokes. Following segregation, the TST is lowered, the bridges are
removed and the backlit specimen is photographed by a ceiling-mounted camera. The
TST backlighting enhances the contrast between the particles and the background. \textcite{zheng_soil_2014} later used stereo photography in the TST setup to determine the size as well as surface information of aggregates in the size range of sand and gravel (maximum size of 1.2 in. [3.0 cm]). 
\begin{figure}[!htb]
	\centering
	\includegraphics[width=0.5\textwidth]{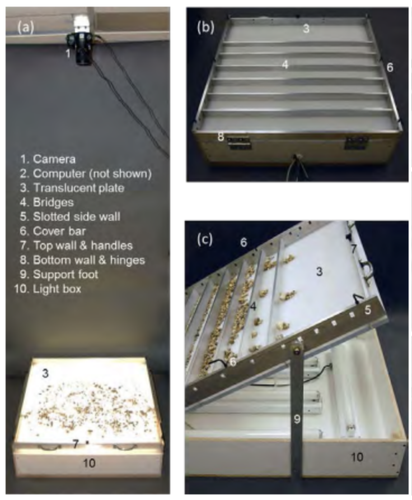}
	\caption{Translucent Segregation Table (TST) for stereo photography of aggregates. Sources: \textcite{hryciw_innovations_2014, zheng_soil_2014}}
	\label{fig: tst}
\end{figure}

\begin{figure}[!htb]
	\centering
	\includegraphics[width=\textwidth]{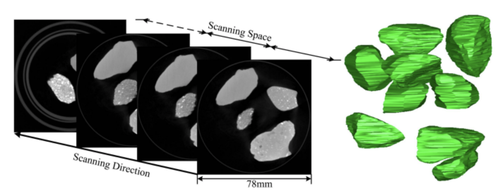}
	\caption{X-ray CT images of the cylindrical container and the developed 3D models of aggregates. Source: \textcite{jin_aggregate_2018}}
	\label{fig: ct}
\end{figure}

\textcite{jin_aggregate_2018} performed aggregate shape characterization and volume estimation based on a 3D solid model constructed from X-ray CT images. The pack of aggregates was subjected to scanning with Compact225 (YXLON, Hamburg, Germany) X-ray CT equipment. With a 0.02 in. (0.5 mm) scanning spacing, several images were acquired and 3D solid models were constructed from a route searching algorithm. The X-ray CT scans and the constructed aggregate models are illustrated in \autoref{fig: ct}.

Although these imaging systems were developed and validated with ground-truth measurements under laboratory conditions, their capabilities for field application have not been verified. First, these systems are designed with a laboratory-scale setup for small to medium-sized aggregates. Thus, they may not be easily transported, assembled, and deployed for field applications, especially for those involving advanced devices such as a 3D-laser scanner or an X-ray CT scanner. Moreover, most of the systems have a maximum aggregate size restriction, limiting their application for handling large-sized aggregates. Further, the lighting conditions for these systems are controlled using backlighting or multiple light sources to minimize the shadow and surface reflection effects. Consequently, these laboratory-scale imaging systems are not directly applicable or adaptable for field inspection of large-sized aggregates. Furthermore, most of the laboratory imaging systems (except the ones using laser-scanner or X-ray CT scanner) focus on 2D aggregate size and shape analysis instead of 3D volumetric information. Since the weights of individual rocks are needed for determining the size distribution of riprap material, volumetric information is preferred. 

\begin{figure}[!htb]
	\centering
	\includegraphics[width=0.6\textwidth]{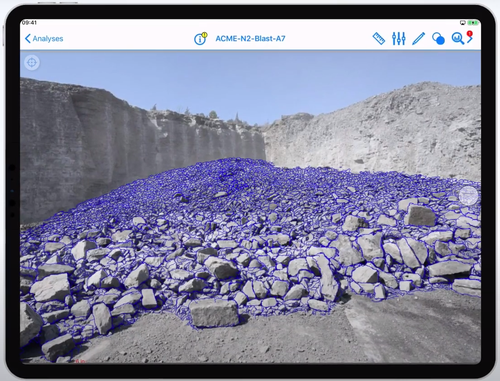}
	\caption{WipFrag software for rock fragmentation analysis. Source: \textcite{wipware_wipfrag_2020}}
	\label{fig: wipfrag}
\end{figure}

For field inspection of large-sized aggregates, the WipFrag software \parencite{wipware_wipfrag_2020} developed by \textcite{maerz_wipfrag_1996, maerz_wipfrag_1999} and commercialized by WipWare, Inc. is the only imaging system found in the literature that was used to provide riprap characterization. The example segmentation analysis is presented in \autoref{fig: wipfrag}. It was integrated with mobile devices for on-site use to roughly estimate the aggregate size gradation in a stockpile. Nevertheless, the image segmentation procedure used in this software is highly user-dependent and its gradation property estimation is based on a single-view of riprap stockpile. Also, it is based on 2D analysis and does not characterize 3D aggregate shape or volumetric information. Specifically, the WipFrag software was tried by IDOT in the field for riprap imaging with limited success. It was found that the analysis quality and accuracy were affected by shadows and aggregate overlaps in the image. 

In summary, among these emerging machine vision-based techniques, a field imaging system with robust and efficient algorithms for obtaining comprehensive 2D and 3D information of aggregates—especially riprap and large-sized aggregates—has not yet been developed. 

\subsection{Image Processing Procedure for Aggregate Evaluation}

Traditional methods for aggregate evaluation include visual inspection, geometry measurements, and sieve analysis, while the accurate characterization of aggregate shape is challenging and labor-intensive for humans to visually or manually determine. In this regard, machine vision techniques have been widely adopted to characterize aggregate size and shape properties in the above-mentioned aggregate imaging systems developed to date.

Analysis of an aggregate image typically consists of an image segmentation module followed by a morphological analysis module based on computational geometry \parencite{al-rousan_evaluation_2007}. Image segmentation extracts the interested region of target aggregate(s) from the image background, which is a key step for filtering the noisy and useless information from the raw image. The morphological analysis step is relatively consistent across different imaging systems since it usually processes the binary silhouette images after the segmentation step. Flat and Elongated Ratio (FER), Angularity Index (AI), and Surface Texture Index (STI) are developed as key indices in aggregate shape characterization \parencite{al-rousan_evaluation_2007}. 

To achieve robust image segmentation, the setups of aggregate imaging systems are usually configured to provide a clean background and ensure spacing among aggregates such that the effort required to separate overlapping or touching aggregates during the image segmentation step is minimized. The AIMS system \parencite{masad_development_2003, gates_fhwa-hif-11-030_2011} can capture multiple aggregates that are spread onto a tray and manually separated. Further post-processing is required by conducting the convex hull test to select valid aggregate regions. The E-UIAIA system \parencite{tutumluer_video_2000, moaveni_evaluation_2013} acquires aggregate photos from orthogonal views of individual aggregates placed in front of a blue background. The E-UIAIA system deals with individual-aggregate imaging with no touching or overlapping involved. Other imaging systems such as 3D laser-based system \parencite{komba_analytical_2013, anochie-boateng_three-dimensional_2013} and stereo-photography based system \parencite{zheng_soil_2017} also mainly operate on aggregates with minimal contact or overlap.

The above aggregate imaging systems manually control the arrangement of aggregates and achieve high-precision measurements of separated or non-overlapping aggregates. However, when aggregates are in a densely-stacked stockpile form or in a constructed layer, which are the most practical scenarios, their capability to simultaneously characterize a large quantity of aggregates may not be sufficient for several reasons. First, these systems manually separate the aggregates and provide a clean background to simplify the image segmentation task. This condition can no longer be satisfied when aggregates are in a stockpile background or other field scenes. In addition, the image segmentation algorithms originally intended for laboratory conditions may also have accuracy and robustness issues under field lighting conditions. Second, imaging many aggregates using these systems is inefficient since they are designed for inspecting aggregates one by one or in a manually arranged pattern. Moreover, the application of these imaging systems is further limited when only in-place evaluation is available at production or construction sites or when the size of aggregates is beyond the system capability. 

\begin{figure}[!htb]
	\centering
	\includegraphics[width=\textwidth]{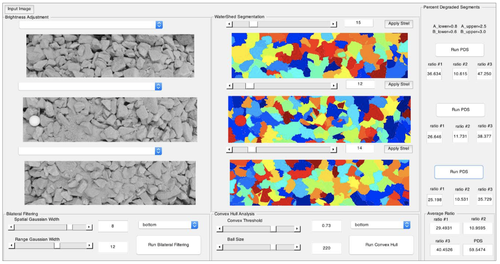}
	\caption{Watershed segmentation used for railway ballast analysis. Source: \textcite{tutumluer_field_2017} }
	\label{fig: watershed}
\end{figure}

To overcome the challenges of analyzing stockpiled and densely stacked aggregate images, more advanced image segmentation techniques are required. Traditional 2D image segmentation methods include three major types, region-based, edge detection-based, and watershed, among which the variations of edge-based and watershed segmentation algorithms have been shown to perform better in the presence of mutually touching particles in dense images such as stockpile aggregate views \parencite{wani_edge-region-based_1994, muthukrishnan_edge_2011, vincent_watersheds_1991}. In this connection, several research and industrial applications have been developed. For example, \textcite{tutumluer_field_2017, huang_evaluation_2018} applied watershed segmentation to characterize the degradation level in trench-view images of railway ballast by classifying the size distributions of image segments, as shown in \autoref{fig: watershed}. Similarly, the commercial software WipFrag, developed by \textcite{maerz_wipfrag_1996, maerz_wipfrag_1999}, uses edge-based segmentation to partition rock fragments and estimate the aggregate size distribution from stockpile images. Nevertheless, both image segmentation algorithms are very user dependent. Considerable user interaction by either parameter fine-tuning or interactive editing is required to achieve an acceptable segmented image.

\section{Computer Vision Techniques with Deep Learning}

Over the last decade, machine learning-based methods have enabled significant advances in many challenging vision tasks benefiting from the development of artificial intelligence and computer vision techniques \parencite{prince_computer_2012}. Dense image-segmentation tasks, along with many object classification and detection algorithms, are difficult in the sense that the features in the image are usually implicit and thus cannot be easily extracted and represented by human intuition. While traditional image segmentation methods are not effectively applicable to identifying these features, machine learning methods may better handle such tasks by capturing the underlying features based on data-driven mechanisms. During recent developments in the deep learning domain, deep neural networks proposed by \textcite{lecun_deep_2015}—architecture that has many layers of different types—exhibit advantages over conventional machine learning techniques because of the better capability to discover intricate features in large datasets with minimal human-guided interaction. With multiple levels of abstraction in the neural network, deep neural networks have dramatically improved the state-of-the-art in many complicated tasks in computer vision, such as image classification, object detection, semantic segmentation, etc. In the context of aggregate studies, deep learning techniques have not been fully utilized to solve the challenging task of stockpile aggregate imaging. Considering this fact, applying deep learning techniques to aggregate imaging research has the great potential to provide more robust and user independent 2D and 3D analyses. A brief introduction of the fundamentals in deep learning is presented as follows.

\subsection{General Categories of Machine Learning and Deep Learning Problems}
\begin{figure}[!htb]
	\centering
	\includegraphics[width=\textwidth]{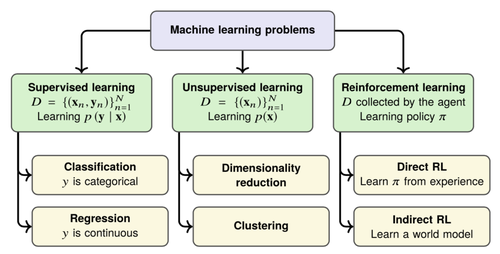}
	\caption{Main categories of machine learning and deep learning problems. Source: \textcite{nikolenko_synthetic_2021}}
	\label{fig: taxonomy}
\end{figure}

As a more general context of machine learning and deep learning research, learning problems are usually classified into three main categories: supervised learning, unsupervised learning, and reinforcement learning, as illustrated in \autoref{fig: taxonomy}.

Supervised learning problems are for data given in the form of pairs $D =\{(x_n,y_n)\}_{n=1}^N$, with $x_n$ being the $n^{th}$ data point (input of the model) and $y_n$ being the target variable (i.e., label). In classification problems, the target variable $y_n$ is categorical and discrete, while in regression problems, the target variable $y_n$ is continuous. For more complex problems, the data and label may take different forms.

Unsupervised learning problems is for learning a distribution of input data, where no corresponding label $y_n$ is given. Typical tasks for unsupervised learning are dimensionality reduction that captures key information from a high-dimensional dataset; and clustering that reduces the dimensionality into a discrete set of clusters. The difference between the two tasks in continuous and discrete space is similar to the difference between the regression and classification tasks in supervised learning.

Furthermore, reinforcement learning problems are more abstract in that data input does not even exist before learning begins, and an ``agent'' is supposed to collect its own dataset by interacting with the ``environment''. Agents are trying to learn a policy $\pi$ based on the actions they take along with the feedback from the environment. Generally, a reinforcement learning agent is supposed to perceive and observe its environment, take actions and learn through trial and error.

\subsection{Fundamental Deep Learning Designs in Computer Vision}

\subsubsection{Convolution Mechanism for Visual Feature Learning}
The emerging deep learning techniques have many successful applications, such as Natural Language Processing (NLP) and speech recognition. Among all the success, Computer Vision (CV) is an essential application of deep learning. It is an interdisciplinary field that focuses on how computers can gain high-level understanding from digital image or video data. Vision sensors replicate the human eye perception of the world, and human vision skills and/or more complicated skills are learned on these sensor data, such as object recognition, object tracking, visual measurement, and segmentation.

Convolutional Neural Network (CNN) is a popular and widely used algorithm in deep learning, which has been extensively applied in different applications such as NLP, speech processing, and computer vision \parencite{pouyanfar_survey_2018}. The design of CNN structure is inspired by the neurons in animal and human brains. Specifically, it simulates the visual cortex in a cat's brain containing a complex sequence of cells \parencite{hubel_receptive_1962}. As described in \textcite{goodfellow_deep_2016}, CNN has three main advantages: parameter sharing, sparse interactions, and equivalent representations. To fully utilize the two- dimensional structure of an input data (e.g., image signal), local connections and shared weights in the network are utilized. This results in fewer but essential parameters, which makes the network faster and easier to train. This operation is similar to the one in the visual cortex cells, that are more responsive to local observations than to the entire scene.

\begin{figure}[!htb]
	\centering
	\includegraphics[width=\textwidth]{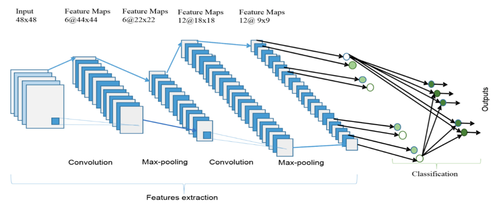}
	\caption{Typical architecture of CNNs. Source: \textcite{alom_state---art_2019}}
	\label{fig: cnn}
\end{figure}

In typical CNNs, there are a number of convolutional layers followed by pooling (subsampling) layers, and in the final stage layers, fully connected layers are used to generate the output. \autoref{fig: cnn} shows an example CNN architecture. The layers in CNNs have the inputs $x$ with three dimensions, $H\times W\times C$, where $H$ and $W$ refer to the height and width of the input, respectively, and $C$ refers to the depth or the channel numbers (e.g., $C=3$ for an RGB image). In each convolutional layer, there are several filters (kernels) of size $n\times n\times k$, where $n$ is usually a small number indicating the size of the kernel, and $k$ is the number of output feature maps that can be generated from the convolution. The filters share the same parameters (weight $W_k$ and bias $b_k$) and are convoluted with the input to generate $k$ feature maps ($h_k$). The convolutional layer computes a dot product between the weights and its inputs, and adds the bias to the product. Then, a nonlinear activation function $f$ is usually applied to the output of the convolutional layers:

\begin{equation*} \label{eqn:cnn}
	h_k =f(W_k\cdot x+b_k)
\end{equation*}

After that, maxpooling layers (or subsampling layers) are commonly applied to condense the feature space, by downsampling each feature map with less parameters. The pooling operation (e.g., average-pooling or max-pooling) is performed to extract the representative feature values over a receptive field. Finally, the last layers in CNNs are usually fully connected layers that establish point-to-point connection between each input feature and output feature. Overall, CNN designs learns from the low-level features of the input and are capable of obtaining the high-level abstraction and understanding from the data. 

\subsubsection{Attention Mechanism and the Transformer Design}
\begin{figure}[t]
	\centering
	\includegraphics[width=0.8\textwidth]{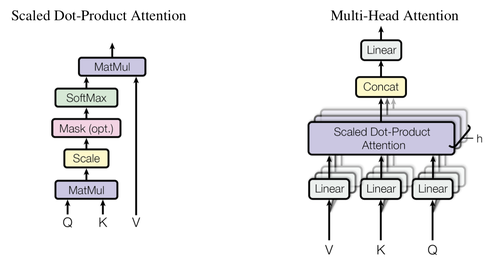}
	\caption{(left) Scaled Dot-Product Attention. (right) Multi-Head Attention consists of several attention layers running in parallel. Source: \textcite{vaswani_attention_2017}}
	\label{fig:attention}
\end{figure}

The powerful transformer design was originated in the NLP domain and was recently adopted in CV tasks. The key idea behind the transformer design is the attention mechanism. Encoding the semantics of a long sequence into a single hidden state is usually difficult in practice. This is because the original feature embedding design does not encode the actual context information. Therefore, the attention mechanism was proposed by \textcite{vaswani_attention_2017} to let decoder utilize each of the encoder's hidden states and better capture the contextual semantics of the sequence. The implementation of attention mechanism is essentially a function that maps a query and a set of key-value pairs to an output attention vector. Different forms of attention are hard attention, soft attention, global attention, local attention, and self attention. Self attention is the most interesting design. By assigning the key, value, and query as the same feature vector, its goal is to learn the feature dependencies and capture the internal structure of the context. As shown in \autoref{fig:attention}, the scaled dot-product attention normalizes the dot product by the key dimension such that the numerical stability is improved during training. And the multi-head attention is a stack of multiple self-attention units that can learn the feature in different representation spaces.

\begin{figure}[t]
	\centering
	\includegraphics[width=0.5\textwidth]{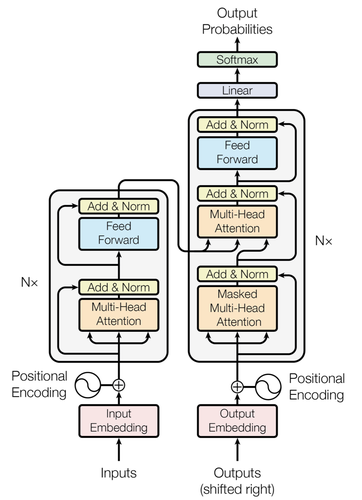}
	\caption{The transformer model architecture. Source: \textcite{vaswani_attention_2017}}
	\label{fig:transformer}
\end{figure}

The design of transform architecture is a famous breakthrough that boosts the advancements in NLP. Transformer is basically an encoder-decoder framework. As shown in \autoref{fig:transformer}, the decoder has one extra multi-head attention layer than the encoder. This extra layer takes the encoder outputs as the key and value, and uses the output of the first decoder layer (masked multi-head attention) as the query. The ``Add" represents the residual connection between layers to mitigate the gradient vanishing problem. The ``Norm" represents layer normalization that improves the numerical stability and accelerates the training process. Another important design is the feed forward layer after the multi-head attention layer that increases the non-linearity in both encoder and decoder blocks. Recently, the key transformer concept has been further applied to computer vision domain in the development by \textcite{dosovitskiy_image_2020}, known as the Vision Transformer (ViT).

\clearpage 

\section{Summary}

This chapter provided a review of aggregate standards and specifications, findings from previous aggregate studies, relevant equipment that leveraged imaging techniques, and the applications of deep learning–based technology. Traditional methods for assessing riprap geometric properties involve subjective visual inspection and time-consuming hand measurements. As such, achieving the comprehensive in-situ characterization of riprap materials remains challenging for practitioners and engineers. In this regard, several advanced aggregate imaging systems developed over the years utilized machine vision techniques to approach this task in a quantitative, objective, and efficient manner. These aggregate imaging systems developed to date for size and shape characterization, however, have primarily focused on measurement of separated or non-overlapping aggregate particles. The development of efficient computer vision algorithms based on traditional computer vision techniques and/or emerging deep-learning techniques is urgently needed for image-based evaluations of densely stacked (or stockpile) aggregates, which require image segmentation of a stockpile for the size and morphological properties of individual particles.

%% file: chapter03.tex
\chapter{Field Studies and Sampling of Aggregate Materials at Aggregate Producers} \label{chapter-3}

This chapter presents details of field data collection and sampling of aggregate materials used in this study, with a focus on riprap and large-sized aggregates. Geographical and geological information of the aggregate quarry sites visited during a series of the field investigation are summarized. Moreover, details are presented for the imaging data collection of both individual aggregates and aggregate stockpiles. Explanation of imaging setups, data acquisition procedures, and required ground-truth measurements are also discussed in this chapter.

\section{Selection of Aggregate Sources and Aggregate Producers}

A series of field site visits of aggregate quarries in Illinois were made in this study to sample and collect data for riprap rocks and aggregate stockpiles. The main purpose of the field visits was to collect data for a comprehensive database of large-sized aggregates together with ground-truth measurements. The database aims to provide high quality aggregate imaging data with studied morphological properties, and thus serves as the benchmark for any validation needed during the development of the field imaging framework. 

\begin{figure}[!htb]
	\centering
	\includegraphics[width=0.5\textwidth]{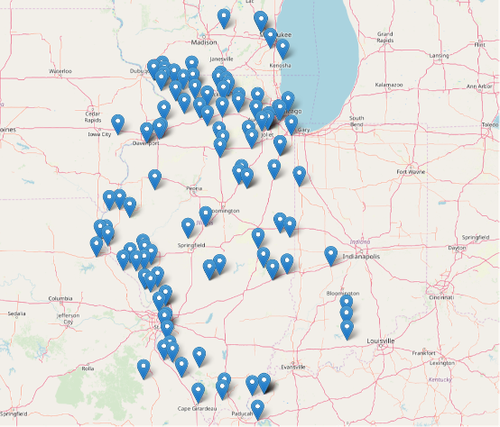}
	\caption{Geographical information for selected IDOT-approved aggregate producers. Source: \textcite{idot_approvedqualified_2019}}
	\label{fig: 3-2}
\end{figure}

During the field site visits, different types of aggregate imaging data had to be collected for the development of multiple imaging algorithms, requiring a careful selection of riprap aggregate producers for these visits at the beginning. Information was first gathered regarding the locations, gradations, and product quality of the IDOT-approved list of Illinois aggregate producers. The map in \autoref{fig: 3-2} provides the geographical information of aggregate producers collected during the site selection phase.

\begin{figure}[!htb]
	\centering
	\includegraphics[width=0.6\textwidth]{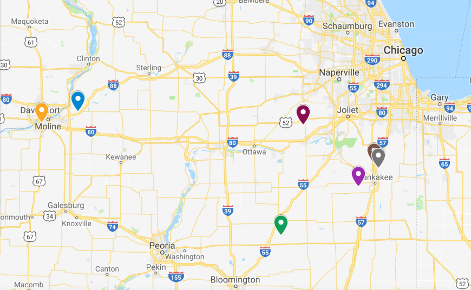}
	\caption{Geographical information for the aggregate producers that participated in this study}
	\label{fig: 3-3}
\end{figure}

Upon further communication with the aggregate producers, several aggregate producers were selected from the list as the collaborated producers involved in this research study. With the approval from the quarry managers of these collaborated aggregate producers, aggregate imaging data were collected on-site from multiple scheduled field visits. Several aggregate producers were not scheduled for field visits, but they also participated in this study by either shipping aggregate material to Advanced Transportation Research Engineering Laboratory (ATREL) for further study or providing aggregate image data from their routine inspection process. All aggregate producers in Illinois that participated in this study are identified in \autoref{fig: 3-3}, with detailed information of the field site visits presented in \autoref{tab:3-1}.  In addition, existing photo libraries of aggregates maintained by aggregate producers and IDOT office were also collected. These photo libraries mainly contain a limited amount of inspection photos during their previous QA/QC activities.

\begin{table}[!htb]
	\centering
	\caption{Information of Collaborated Aggregate Producers in Illinois}
	\label{tab:3-1}
	\begin{tabular}{L{4cm}L{3cm}L{3cm}L{4cm}}
		\hline
		\textbf{Aggregate Producer}  & \textbf{Aggregate Size Categories} & \textbf{Mineralogy Description}  & \textbf{Field Study}   \\ \hline
		Vulcan Materials Company - Lisbon, Illinois             & RR3                        & Dolomite, yellowish & Material Shipped, June 2018 (I) \& August 2021 (III) \\\hline
		Prairie Material – Ocoya, Illinois               & RR4, RR5, and RR6                        & Limestone, bluish gray & Inspection Photos Provided (II) \\\hline
		RiverStone Group,   Midway Stone – Hillsdale, Illinois   & RR6  & Dolomite, yellowish to   bluish gray, fossiliferous & July 2018 (II) \\\hline
		RiverStone Group,   Allied Stone – Milan, Illinois & RR4, RR5, and RR6                        & Dolomite, white   & July 2018 (I, II) \& March 2019 (II, III)  \\\hline
		Vulcan Materials   Company – Kankakee, Illinois     & RR4, RR5, RR6, and RR7                   & Dolomite, white to yellowish & May 2019 (II) \& August 2021 (III) \\ \hline
	\end{tabular}
\end{table}

\section{Multi-Phase Field Studies for Aggregate Imaging}
To develop robust 2D and 3D imaging algorithms progressively, the field studies were carried out in three main phases:
\begin{itemize}
	\item Phase I: For field study of individual aggregates, a field inspection system was designed to collect three orthogonal views of aggregates. The aggregate sources involved in this phase is labeled as `I' in \autoref{tab:3-1}. The details of the aggregate source information and the individual-aggregate imaging procedure are discussed in \cref{sec-individual}.
	
	\item Phase II: For field study of 2D stockpile aggregates, stockpile images of different size categories and geological origins were collected. All aggregate sources involved in this phase are labeled as `II' in \autoref{tab:3-1}. The details of the aggregate source information and the 2D stockpile imaging procedure are discussed in \cref{sec-stockpile2d}.
	
	\item Phase III: For field study of 3D stockpile aggregates, multi-view stockpile images of different size categories and geological origins were collected. All aggregate sources involved in this phase are labeled as `III' in \autoref{tab:3-1}. The details of the aggregate source information and the 3D stockpile imaging procedure are discussed in \cref{sec-stockpile3d}.
\end{itemize}

\section{Aggregate Sources and Field Imaging Procedure for Individual-Aggregate Study} \label{sec-individual}

\subsection{Aggregate Sources for the Individual-Aggregate Study}

For the Phase I field study, i.e., the individual-aggregate study, images of individual large rocks were taken. In total, 85 particles were collected and analyzed from two aggregate sources. All particles were analyzed using a field imaging system (\cref{sec-3-2-1}), and ground-truth volume/weight information was measured. The purpose of this Phase I study was to collect data for individual aggregates using the field inspection system and to later validate the robustness of the associated volumetric reconstruction algorithms (\cref{chapter-4}). 

The first source was aggregates from a pile of IDOT's CS02 material obtained from Vulcan Materials Company in Lisbon, IL. Only aggregate particles larger than 3 in. (76.2 mm) were selected for this study. Thus, the collected 40 rocks complied to IDOT's RR3 riprap size category. The second source was from RiverStone Group in Moline, IL. A field visit was arranged to this quarry site to sample and image the individual rocks. For this source, 40 rocks complying to IDOT's RR3 size category and five rocks complying to the RR5 category were selected for imaging and manual size/weight measurements. The material size and source information for the riprap rocks collected for this individual-aggregate study is summarized in \autoref{tab:3-3}. 

\begin{table}[!htb]
	\centering
	\caption{Material Size and Source Information for the Individual-Aggregate Study}
	\label{tab:3-3}
	\begin{tabular}{llll}
		\hline
		Aggregate Producer                     & Source Name    & Number of Rocks & Size Range (in.) \\ \hline
		Vulcan Materials   Company, Lisbon, IL & Source 1       & 40              & {[}3, 6{]}       \\
		RiverStone Group,   Milan, IL          & Source 2       & 40              & {[}5, 16{]}      \\
		RiverStone Group,   Milan, IL          & Source 2–Large & 5               & {[}16, 26{]}     \\ \hline
		\multicolumn{4}{l}{Note: 1 in. = 2.54 cm}                                                   
	\end{tabular}
\end{table}

\subsection{Individual-Aggregate Image Data Acquisition Procedure} \label{sec-3-2-1}

\begin{figure}[!htb]
	\centering
	\begin{subfigure}[b]{0.45\textwidth}
		\centering
		\includegraphics[height=6cm]{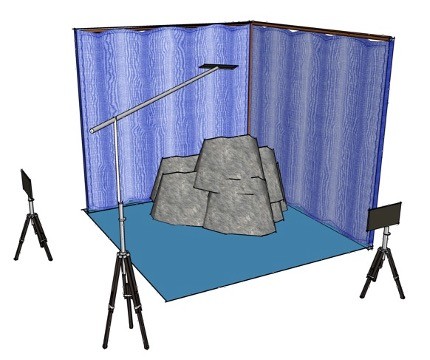}
		\caption{}
		\label{fig: 3-4a}
	\end{subfigure}
	\hfill
	\begin{subfigure}[b]{0.45\textwidth}
		\centering
		\includegraphics[height=6cm]{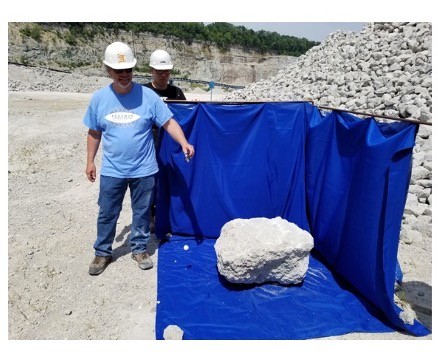}
		\caption{}
		\label{fig: 3-4b}
	\end{subfigure}
	\caption{(a) A conceptual illustration, and (b) the actual constructed setup of the 
		field-imaging system for individual aggregates}
	\label{fig: 3-4}
\end{figure}

An imaging-based riprap inspection system was designed and built to acquire field images of individual aggregates. The schematic drawing and actual photo of the field inspection system are shown in \autoref{fig: 3-4}. The system consists of five major parts: 
\begin{itemize}
	\item Three smartphones with high-resolution cameras and remote shutter control.
	\item Three pieces of 5 ft. (1.52 m) copper tubing. The copper-tubing framework was connected by plumbing joints and can be easily assembled and disassembled for mobility.
	\item A 22 lb. (10 kg) patio umbrella base as an anchorage for the copper tubing framework. 
	\item Three 5 ft. by 5 ft. (1.52 m by 1.52 m) blue curtains as the background and bottom surfaces.
	\item Three camera tripods for fixing smartphones to the top/front/side. (One tripod was designed with a cantilever arm for holding the phone from a top view.)
\end{itemize}

The inspection system was developed to image the selected 85 riprap rocks. The selected rock samples were intended for a comprehensive database—which covers both medium- and large-sized particles—for a reasonable validation of the algorithms. Note that only a small number of very large particles were inspected because of limited access to operational machinery at the quarry site for hauling rocks to the inspection system and rotating them between trials. Therefore, most riprap samples for field validation were medium-sized particles rather than very large riprap rocks, such as the one shown in \autoref{fig: 3-4}b. 

In the beginning of the field imaging procedure, three smartphones were aligned to achieve approximately orthogonal views. A white-colored calibration ball having a 1.5 in. (38.1 mm) diameter was first captured as the standard reference object. Riprap particles were then placed at the same location as the calibration ball and captured in sequence by triggering the shutter. For each riprap sample in Source 1 and Source 2 (see \autoref{tab:3-3}), the top-front-side image triplet was repeated three times, each time rotating the particle to a random angle. The purpose of this rotate-repeat testing is to check the reproducibility of the imaging procedure and to further investigate the variations resulting from viewing the single rocks from different angles.

\begin{figure}[!htb]
	\centering
	\begin{subfigure}[b]{0.45\textwidth}
		\centering
		\includegraphics[height=6cm]{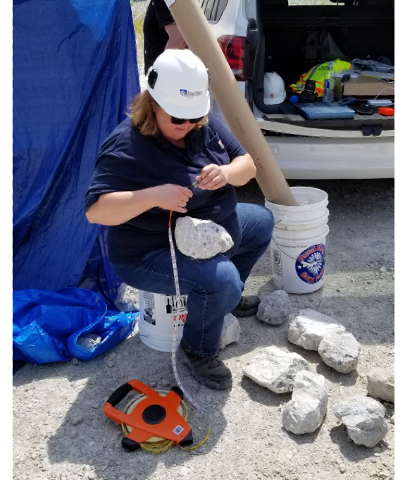}
		\caption{}
		\label{fig: 3-5a}
	\end{subfigure}
	\hfill
	\begin{subfigure}[b]{0.45\textwidth}
		\centering
		\includegraphics[height=6cm]{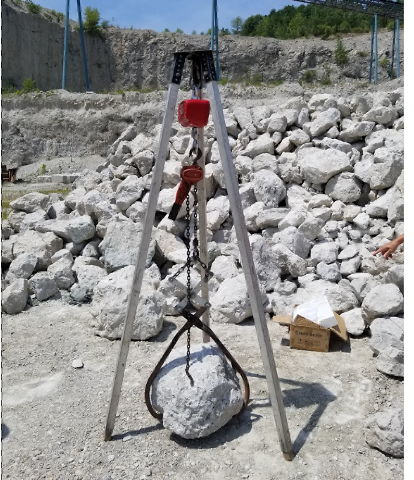}
		\caption{}
		\label{fig: 3-5b}
	\end{subfigure}
	\caption{(a) Manual measurements of riprap dimensions and (b) tripod scale system used by IDOT for measuring the weight of large riprap rocks}
	\label{fig: 3-5}
\end{figure}

The accuracy of size (and shape) measurements were checked by manual measurements of individual riprap rock dimensions and weights. As the ground-truth data, all particles in Source 1 were collected from the quarry site and specific gravity tests as per \textcite{astm_c127_standard_2015} were conducted to obtain weight, volume, and specific gravity information. For Source 2 and Source 2–Large riprap rocks, only the weight of individual rocks was measured on-site. Manual measurements for rock dimensions were also conducted by experienced IDOT engineers on 20 randomly selected rocks from Source 2 and on all Source 2–Large rocks as the state-of-the-practice result. The dimension measurements were taken at three midway locations as determined by rough estimates and by the judgment of the experienced IDOT engineers and practitioners. \autoref{fig: 3-5} shows the current practice of measuring riprap size and weight in the field, which was used to establish the ground truth for Source 2 and Source 2–Large riprap rocks. Note that a tripod scale system shown in \autoref{fig: 3-5}b was specially designed as the best practice of IDOT for a heavy-duty weight measurement setup.

\section{Aggregate Sources and Field Imaging Procedure for the 2D Aggregate Stockpile Study} \label{sec-stockpile2d}

\subsection{Aggregate Sources for the 2D Aggregate Stockpile Study}

For the Phase II field study, i.e., the 2D aggregate stockpile study, images of riprap stockpiles were taken at four Illinois quarries. Sources were selected that covered a wide variety of geological origins, riprap size categories, rock sizes, texture (shape properties), and rock colors. For each aggregate producer, several images were taken for riprap stockpiles of different sizes and from different viewing angles. The details of aggregate producers, riprap size categories, and descriptions of riprap rocks are given in \autoref{tab:3-4}. The number of images reported in this table include multiple images taken at the same stockpile from different viewing angles. Note that different rock types and colors typically quarried in Illinois were chosen so that the stockpile images can later be used to train a neural network to detect different rock shapes, colors, and sizes in riprap stockpile images (\cref{chapter-5}). Also note that two rock sizes were generally considered: medium-sized riprap rocks, ranging in weight between 1 lb. and 40 lbs. (0.45 kg and 18.1 kg), and large-sized rocks, ranging in weight between 40–600 lbs. (18.1–272.2 kg). 

In addition to the stockpile images taken from the field visits, a small amount of stockpile images conforming to different gradation categories were selected from the QA/QC photo libraries from multiple quarries. These quarries include Prairie Materials, Ocoya; Prairie Materials, Manteno; Vulcan Materials, Manteno; and Vulcan Materials, Kankakee. Since the number of images is very limited in this source, they are all denoted as the Prairie Material - Ocoya, IL in \autoref{tab:3-4}. These images were also used to train and/or validate the robustness of the associated segmentation and reconstruction algorithms.

\begin{table}[!htb]
	\centering
	\caption{Source Information and Description of Stockpile Aggregate Image Dataset}
	\label{tab:3-4}
	\begin{tabular}{L{5cm}L{4cm}L{2cm}L{4cm}}
		\hline
		\textbf{Aggregate Producer}                        & \textbf{Aggregate Size Categories} & \textbf{Number of Images} & \textbf{Rock Description}               \\ \hline
		Prairie Material –   Ocoya, Illinois               & RR4, RR5, and RR6      & 6                   & Limestone, bluish gray,   medium-sized* \\
		RiverStone Group,   Allied Stone – Milan, Illinois & RR4, RR5, and RR6  & 14                      & Dolomite, white,   medium-sized*        \\
		RiverStone Group,   Midway Stone – Hillsdale, Illinois   & RR6+  & 100 & Dolomite, yellowish to   bluish gray, fossiliferous, large-sized*  \\
		Vulcan Materials   Company –Kankakee, Illinois     & RR4, RR5, RR6, and RR7   & 44                & Dolomite, white to yellowish,   large-sized*     \\ \hline
		\multicolumn{4}{l}{* Note: “Medium sized” refers to aggregates weighing between 1 lb. (0.45 kg) and 40 lbs. (18.1 kg);} \\
		\multicolumn{4}{l}{“Large sized” refers to aggregates weighing between 40 lbs. (18.1 kg) and 600 lbs. (272.2 kg);}      \\
		\multicolumn{4}{l}{+ Images were taken from separate, adjacent, and stockpile views.}      
	\end{tabular}
\end{table}

\subsection{2D Aggregate Stockpile Image Data Acquisition Procedure} \label{sec-2d-procedure}

\begin{figure}[!htb]
	\centering
	\begin{subfigure}[b]{\textwidth}
		\centering
		\includegraphics[width=0.5\linewidth]{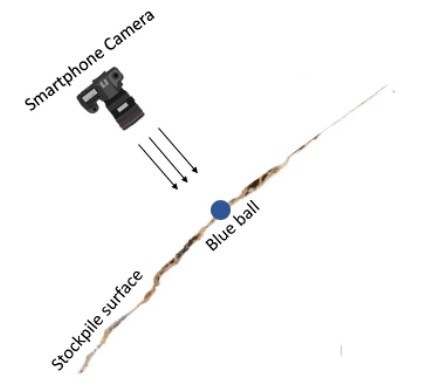}
		\caption{}
	\end{subfigure}
	\newline
	\begin{subfigure}[b]{0.31\textwidth}
		\centering
		\includegraphics[width=\linewidth]{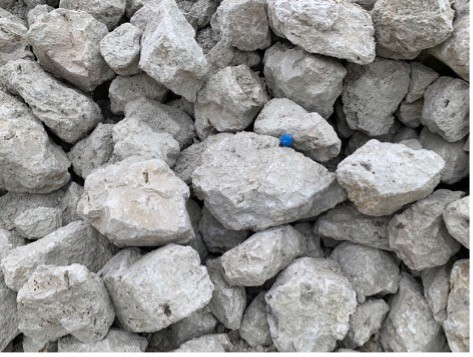}
		\caption{}
	\end{subfigure}
	\begin{subfigure}[b]{0.31\textwidth}
		\centering
		\includegraphics[width=\linewidth]{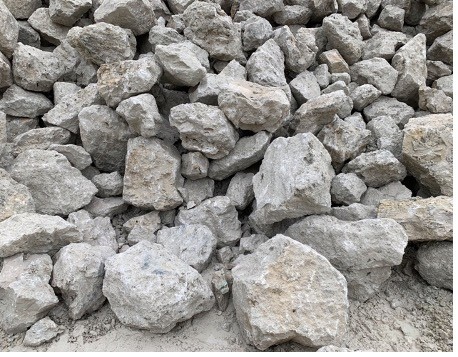}
		\caption{}
	\end{subfigure}
	\begin{subfigure}[b]{0.31\textwidth}
		\centering
		\includegraphics[width=\linewidth]{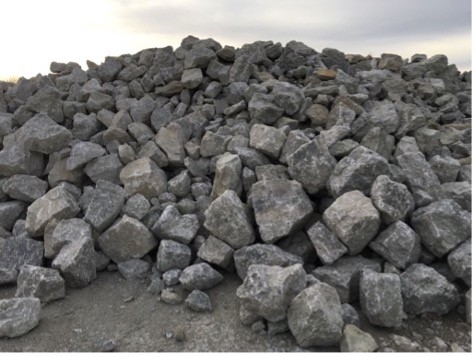}
		\caption{}
	\end{subfigure}
	\caption{(a) Proper positioning of the camera relative to the stockpile, (b) satisfactory image with a calibration ball, (c) satisfactory image without a calibration ball, and (d) unsatisfactory image of a stockpile}
	\label{fig: 3-6}
\end{figure}

For the imaging of riprap stockpiles, conventional smartphone cameras were used to obtain the images. Ideally, multiple views of the same stockpiles were taken with a calibration ball and with the camera positioned parallel to the slope of the stockpile. The following guidelines/procedures were closely followed for imaging riprap stockpiles:
\begin{itemize}
	\item The camera requirement for acquiring an image with a sufficient resolution is 2400 x 3000 or higher. Most smartphone cameras will pass this requirement. 
	\item 	Stockpile images were taken from a nearly perpendicular direction against the stockpile surface, as illustrated in \autoref{fig: 3-6}a.
	\item The stockpile images were taken from a location close to the stockpile so that most of the images were filled with useful rock pixels, and the calibration ball (if present) would not appear too small in the image. Images with and without a calibration ball were taken for the purposes of developing a robust algorithm, particularly for training and segmentation purposes. In \autoref{fig: 3-6}, images (b) and (c) are satisfactory images because they are taken at a close distance, the ball is at the center of the image (c), all image pixels in the image contain a rock, and the view is perpendicular to the stockpile face. Image (d) is less satisfactory because it was taken from a far distance and the camera was not perpendicular to the sloped surface.
	\item For all images with a calibration ball, the location of the ball was approximately at the center of the image so that the distortion effect is minimized.
	\item Riprap size groups from RR3 to RR7 categories were all imaged. Additionally, rocks with special colors/textures were selected, as they contributed to the robustness of the dataset.
\end{itemize}

\section{Aggregate Sources and Field Imaging Procedure for the 3D Aggregate Stockpile Study}  \label{sec-stockpile3d}

\subsection{Aggregate Sources for the 3D Aggregate Stockpile Study}
For the Phase III field study, i.e., the 3D aggregate stockpile study, multi-view images of aggregate stockpiles were taken at two Illinois quarries. Aggregate sources were selected that covered various geological origins and size categories. In preparation for the 3D aggregate stockpile study, a 3D aggregate particle library was first established as the essential database (see later in \cref{chapter-6}). The library contains aggregate samples collected throughout the field studies and is summarized in \autoref{tab:3-5}. A total of 46 RR3 rocks and 36 RR4 rocks were collected in this library across four Illinois quarries.

\begin{table}[!htb]
	\centering
	\caption{Source Information of 3D Aggregate Particle Library}
	\label{tab:3-5}
	\begin{tabular}{L{5cm}L{3cm}L{4cm}L{2cm}}
		\hline
		\textbf{Aggregate Producer}  & \textbf{Aggregate Size Categories} & \textbf{Mineralogy Description}  & \textbf{Number of Collected Samples}   \\ \hline
		Vulcan Materials Company - Lisbon, Illinois             & RR3                        & Dolomite, yellowish & 46 \\\hline
		RiverStone Group,   Allied Stone – Milan, Illinois & RR4   & Dolomite, white   & 20  \\\hline
		RiverStone Group,   Midway Stone – Hillsdale, Illinois & RR4   & Dolomite, white   & 6  \\\hline
		Vulcan Materials   Company – Kankakee, Illinois     & RR4                  & Dolomite, white to yellowish & 10 \\\hline
	\end{tabular}
\end{table}

Next, for the 3D stockpile study, a sequence of multi-view images were taken for riprap stockpiles from different viewing angles. The details of aggregate producers, aggregate size categories, and descriptions of riprap rocks are given in \autoref{tab:3-6}. First, stockpiles of different size categories were re-engineered at ATREL based on studied rocks in the 3D aggregate particle library. For example, all 46 RR3 aggregate samples were used to build a re-engineered RR3 stockpile, and multi-view images were taken for the stockpile. After the image acquisition step was completed, the aggregate samples were randomly permuted (e.g., rocks buried inside the current stockpile were placed preferably on the surface for the next stockpile) to vary the stockpile configuration. As a result, six RR3 and six RR4 stockpiles were built. Moreover, field stockpile data were collected during quarry visits to Vulcan Materials Company at Kankakee, IL. RR4 and RR5 stockpiles were prepared at the quarry site by front loader trucks. A similar process was followed to prepare three RR4 and three RR5 stockpiles in the field. The information of the stockpiles used in the 3D aggregate stockpile study is summarized in \autoref{tab:3-6}.

\begin{table}[!htb]
	\centering
	\caption{Source Information of Aggregate 3D Stockpile Study}
	\label{tab:3-6}
	\begin{tabular}{L{5cm}L{3cm}L{4cm}L{2cm}}
		\hline
		\textbf{Aggregate Source}  & \textbf{Aggregate Size Categories} & \textbf{Mineralogy Description}  & \textbf{Number of Stockpiles}   \\ \hline
		RR3 Source in 3D Aggregate Particle Library            & RR3                        & Dolomite, yellowish & 6 \\\hline
		RR4 Sources in 3D Aggregate Particle Library  & RR4   & Dolomite, white to yellowish   & 6  \\\hline
		Vulcan Materials   Company – Kankakee, Illinois     & RR4                  & Dolomite, white to yellowish & 3 \\\hline
		Vulcan Materials   Company – Kankakee, Illinois     & RR5                   & Dolomite, white to yellowish & 3 \\ \hline
	\end{tabular}
\end{table}

\subsection{3D Aggregate Stockpile Image Data Acquisition Procedure}
The camera configuration and data acquisition procedure for 3D aggregate stockpile study is very similar to the 2D stockpile imaging procedure described in \cref{sec-2d-procedure}, with repeated details omitted herein. The major differences between the 2D and 3D stockpile imaging procedures are: (i) multiple views around the stockpiles were taken with a marker system (discussed in detail in \cref{chapter-10}) rather than the calibration ball in 2D procedure, (ii) the camera was moving and positioned at different viewing heights and angles without the restriction to be facing parallel to the slope of stockpile, and (iii) an all-around inspection is required in 3D procedure to obtain full representation of the stockpile, which usually requires a sequence of around 40 multi-view images to be taken.

\section{Summary}

This chapter presented an overview of the field studies and sampling procedures of aggregate materials in quarries. First, representative size categories for riprap and lists of approved riprap sources were identified. Aggregate source information for the individual-aggregate study, the 2D aggregate stockpile study, and the 3D aggregate stockpile study were presented. The procedure for collecting images for individual aggregates using a field inspection system was described, along with the laboratory testing and field size/weight measurements. The detailed procedure for collecting proper images of aggregate stockpiles was presented. Lastly, procedures for collecting 2D images and multi-view inspections of aggregate stockpiles were demonstrated. The following chapters will present details regarding the development of the field imaging framework for the individual aggregate study and the 2D and 3D aggregate stockpile analyses.

%% file: chapter04.tex
\chapter{Volumetric Reconstruction and Estimation for Individual Aggregates} \label{chapter-4}

This chapter describes a volumetric estimation approach developed for the single-particle study. A computer vision–based image-segmentation algorithm was developed for extracting object information while also reducing effects of sunlight and shadowing. Based on multi-view information from the image-segmentation algorithm, a 3D reconstruction algorithm was then integrated for quantifying the volumetric properties of objects. Both algorithms are designed with minimal user input required during image-processing stages for ease in implementation and practical use. The scope of this chapter establishes the relevant algorithms needed for field imaging and volumetric reconstruction of individual riprap and large-sized aggregates. The image analysis results are validated against ground-truth measurements. The full development of this field-inspection system is intended to be portable, affordable, and convenient for data acquisition, with reliable and efficient image-processing algorithms.

The full form of this chapter is published in \textcite{huang2019field, huang2020size}.

\section{Color-Based Image Segmentation Algorithm for Object Detection}

Given an image of individual or multiple rocks under uncontrolled field lighting conditions, the first and foremost task is to accurately recognize and extract the region that comprises the objects. From the perspective of digital image processing in computer vision, this includes partitioning and registering the image pixels into either a foreground object or a background scene, which is often referred to as image segmentation \parencite{gonzalez_digital_2002}. Since color can provide valuable information for the human eye and machine vision systems, color-based image-segmentation techniques are widely used in object detection and recognition. Hence, a color-based image-segmentation algorithm is developed herein for the reliable and accurate extraction of rock particles from field images. The developed image-segmentation scheme involves color-space representation, foreground-background contrast enhancement, adaptive thresholding, and morphological de-noising. The field-inspection system described in \cref{chapter-3} was used to capture images of single large-sized rocks to develop and test the accuracy of the segmentation algorithm. 

\subsection{Color Representation Using CIE L*a*b* Space}

\begin{figure}[!htb]
	\centering
	\begin{subfigure}[b]{0.31\textwidth}
		\centering
		\includegraphics[width=\linewidth]{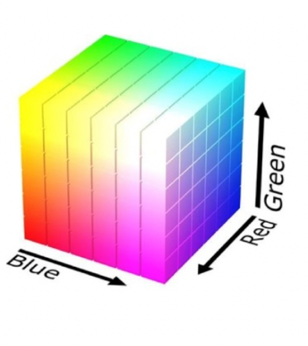}
		\caption{}
	\end{subfigure}
	\begin{subfigure}[b]{0.31\textwidth}
		\centering
		\includegraphics[width=\linewidth]{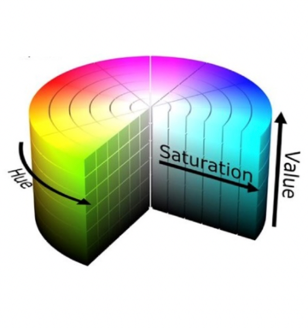}
		\caption{}
	\end{subfigure}
	\begin{subfigure}[b]{0.31\textwidth}
		\centering
		\includegraphics[width=\linewidth]{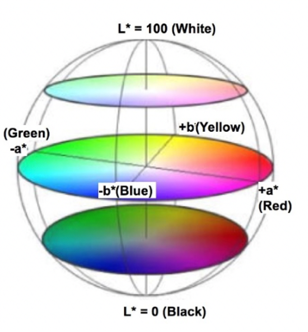}
		\caption{}
	\end{subfigure}
	\caption{Schema of (a) RGB color space, (b) HSV color space, and (c) CIE L*a*b* color space \parencite{kothari_what_2018}}
	\label{fig: 4-1}
\end{figure}

In trichromatic theory, color is perceived by humans as a combination of red, green, and blue, i.e., three primary colors. Along with the development of digital image technology, several 3D color spaces are proposed to represent the colors. The most popular color space is Red-Green-Blue (RGB), which is used in nearly all digital camera devices and display screens, where each pixel is represented as coordinate $(R,G,B)$ with $R,G,B \in [0,255]$ and $R,G,B\in \mathbb{N}$. However, for the purpose of color image segmentation, RGB space is not recommended. It fails to satisfy the perceptual uniformity principle of color, namely, two colors that are perceptually similar to the human eye are not closely located in RGB space in terms of Euclidean distance \parencite{cheng_color_2001, busin_color_2008}. Instead, approximately uniform Hue-Saturation-Value (HSV) space and, further, uniform International Commission of Illumination (CIE) L*a*b* space are formulated via nonlinear transformations from RGB space and can provide better performance in color image representation \parencite{alata_is_2009, fernandez-maloigne_advanced_2012, wang_comparison_2014}. The conceptual schema of different color representation spaces is shown in \autoref{fig: 4-1}. 

\begin{figure}[!htb]
	\centering
	\includegraphics[width=\textwidth]{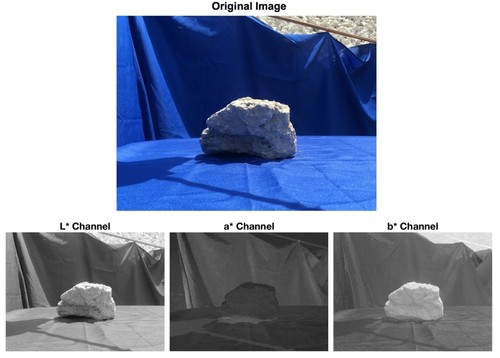}
	\caption{Channel images of a large riprap rock studied using CIE L*a*b* color space}
	\label{fig: 4-2}
\end{figure}

In CIE L*a*b* space, the L* channel represents luminance or intensity value, and a* and b* channels track the green-to-red and blue-to-yellow transition, respectively. Particularly, it makes color chroma information less luminance-dependent, which enables effective measurement of small color differences in the shadow and highlights regions of the scene. For the original rock image presented in \autoref{fig: 4-2}, although an artificial blue background has been used for the convenience of segmentation, the background shadow and the large shading area on the rock surface are inevitable under field lighting conditions. Other color spaces become insufficient for object detection in these shadow regions, while CIE L*a*b* space can properly eliminate the shadow effect by separating the useful information into its a* and b* channels. Note that the useful object information can be accumulated in a* channel, b* channel, or both, depending on what the object color and background color are. For example, when a bright-colored rock on a blue background is used (see \autoref{fig: 4-2}), there are few green-to-red color components in the image. Therefore, minimal object-background contrast is available in a* channel while most of the useful object information accumulates in b* channel. 

The features of CIE L*a*b* space help systematically improve the robustness of the image-segmentation algorithm under various lighting conditions. Therefore, CIE L*a*b* is selected in this paper as the appropriate color space for the following image-segmentation process. Additionally, based on the observation that normal rock colors are rarely blue or green, selection of a background color in the blue-green zone will yield better performance in color segmentation.

\subsection{Foreground and Background Representation Using Pixel Statistics}
\begin{figure}[!htb]
	\centering
	\begin{subfigure}[b]{0.45\textwidth}
		\centering
		\includegraphics[width=\linewidth]{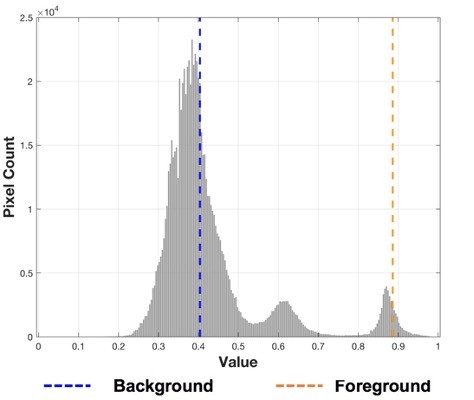}
		\caption{}
	\end{subfigure}
	\begin{subfigure}[b]{0.45\textwidth}
		\centering
		\includegraphics[width=\linewidth]{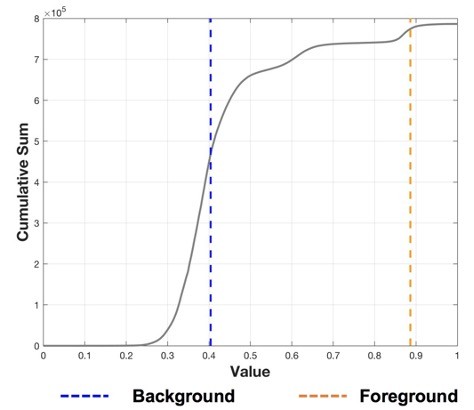}
		\caption{}
	\end{subfigure}
	\caption{(a) Pixel histogram of color channel and (b) pixel cumulative distribution function of color channel}
	\label{fig: 4-3}
\end{figure}

To further differentiate the rock from its surroundings, the representative colors of foreground and background are calculated based on pixel statistics. As an example, the b* channel image in \autoref{fig: 4-2} (bottom-right image) is analyzed. For the reader's convenience, a* and b* channels are denoted as the “color channel” in the following context, as compared to the L* channel, which is the “intensity channel.” 
A pixel-wise histogram is first obtained for the color channel, as shown in \autoref{fig: 4-3}a. However, the histogram is comprised of discrete pixel counts, where the representative color values of foreground and background pixels can be visually defined rather than consistently quantified. Therefore, a Cumulative Distribution Function (CDF) that allows numerically characterizing pixel statistics from a continuous curve is calculated based on the pixel histogram, as shown in \autoref{fig: 4-3}b. Note that the horizontal axis is the pixel value of color channel after scaling into a range $[0,1]$. 

\autoref{fig: 4-3} shows that when a significant number of pixels clusters around the peaks in the histogram, an increasing slope is associated in the CDF. A turning-point detection algorithm proposed in signal processing \parencite{killick_optimal_2012} is then used to capture abrupt changes in the CDF. As illustrated in \autoref{fig: 4-3}b, representative background and foreground colors $b_{\text{background}}^*=0.40,b_{\text{foreground}}^*=0.89$ are detected based on the pixel statistics of the b* channel image in \autoref{fig: 4-2}.

\subsection{Contrast Enhancement Based on Color Distance}

With the representative colors of the background scene and foreground object, a color difference measure can be designed based on Minkowski distance in color space. Suppose two pixel-color vectors in the 2D $a^*-b^*$ space
\begin{equation} \label{eqn:4-1}
	\mathbf{p_1}=(a_1^*,b_1^* ), \mathbf{p_2}=(a_2^*,b_2^* )  
\end{equation}
       
Then the Minkowski distance of order p (i.e., p-norm distance) between the two colors is
\begin{equation} \label{eqn:4-2}
	\lVert \mathbf{p_1}-\mathbf{p_2} \rVert_p = (|a_1^*-a_2^* |^p+|b_1^*-b_2^* |^p )^{\frac{1}{p}} 
\end{equation}

Considering the principle of gamma correction that can bring more contrast between the object and background, the proposed color distance to be calculated at each pixel location is revised from \autoref{eqn:4-2} and presented in the following generalized form:
\begin{equation} \label{eqn:4-3}
	d(\mathbf{p},\mathbf{p_0})=|a^*-a_0^* |^\gamma+|b^*-b_0^* |^\gamma 
\end{equation}
where
$\mathbf{p}$ is the color $(a^*,b^* )$at the current pixel location, $\mathbf{p_0}$ is the reference color $(a_0^*,b_0^*)$ obtained from the pixel statistics (either background or foreground representative color value can be selected as reference), and $\gamma$ is the gamma correction coefficient. $\gamma>1$ is used to contrast the object with background, typically 2.0 (squared-Euclidean distance).

\begin{figure}[!htb]
	\centering
	\begin{subfigure}[b]{0.45\textwidth}
		\centering
		\includegraphics[width=\linewidth]{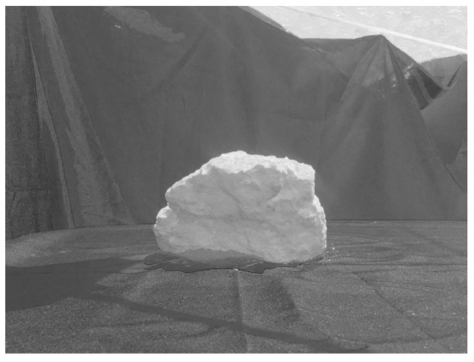}
		\caption{}
	\end{subfigure}
	\begin{subfigure}[b]{0.45\textwidth}
		\centering
		\includegraphics[width=\linewidth]{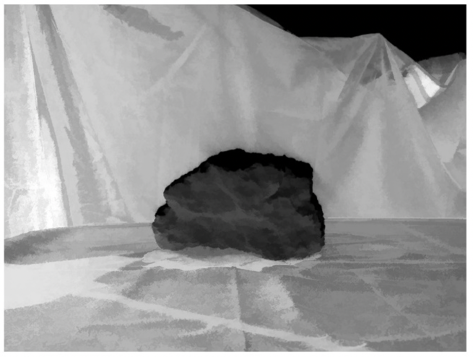}
		\caption{}
	\end{subfigure}
	\caption{(a) Original b* channel image and (b) distance map with foreground representative color as the reference}
	\label{fig: 4-4}
\end{figure}

\autoref{fig: 4-4} illustrates the effectiveness of background and foreground contrast enhancement using the proposed color distance approach. In \autoref{fig: 4-4}b, the pixel grayscale intensity represents the magnitude of the color distance with the foreground's representative color as the reference. Therefore, the closer the pixel color to the foreground representative color, the smaller the color distance and the darker the intensity in the distance map, and vice versa. Note that the measure of color distance helps better contrast the background and foreground and further eliminate the shadow effect.

\subsection{Adaptive Thresholding and Morphological De-noising}

Based on the enhanced distance map, image thresholding (or binarization) is applied and a binary image can be obtained as follows:
\begin{equation} \label{eqn:4-4}
	\text{Pixel value} = 
		\begin{cases}
			1\ \text{(white)} & \text{if}\ v\leq v_{\text{threshold}}\\
			0\ \text{(black)} & \text{if}\ v >  v_{\text{threshold}}
		\end{cases}
\end{equation}

\begin{figure}[!htb]
	\centering
	\includegraphics[width=\textwidth]{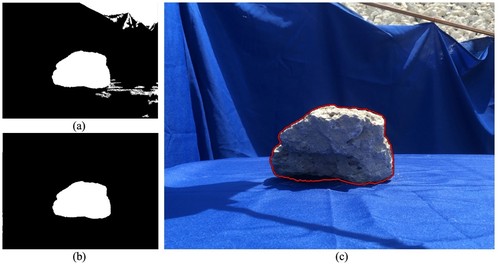}
	\caption{(a) Adaptive thresholding applied image, (b) morphological de-noising applied image, and (c) image-segmentation result}
	\label{fig: 4-5}
\end{figure}

The thresholding algorithm can either follow a fixed threshold value (user-defined or computed based on Otsu's bimodal method \parencite{otsu_threshold_1979}) or a flexible threshold value known as the adaptive thresholding method \parencite{bradley_adaptive_2007}. Since digital images are discretized by pixel, it is common that a binary image can include a significant amount of noise pixels, as shown in \autoref{fig: 4-5}a. De-noising is then required to clean the noise pixels as well as complement discontinuities along the object's boundary. A series of image morphological operations are applied on the binary image, including image erosion, dilation, hole filling, etc. In addition, regions that are closer to the image border are removed, because they are often unidentified objects such as equipment or field surroundings. \autoref{fig: 4-5}b shows the binary image after morphological de-noising, and \autoref{fig: 4-5}c illustrates the object boundary detected by the image-segmentation algorithm. The detected and visualized object boundary is accurate under the intervention of strong shadows, surface reflection, unidentified objects, and so on. Furthermore, based on experiments throughout the development of this algorithm under both laboratory and field conditions, the robustness and versatility of the proposed image-segmentation algorithm are verified.

\section{Volumetric Reconstruction Algorithm for Individual Aggregates}

For the volumetric reconstruction of a 3D object from 2D images, three orthogonal and equal-distance views are required. The orthogonality of views can be well controlled in laboratory via precise camera positioning and distance measurement. However, in the field, neither ideal criteria can be easily achieved without time-consuming and lengthy efforts. To efficiently estimate the volume with sufficient precision, a 3D volumetric reconstruction algorithm using orthogonality calibration and volume correction is proposed herein. 

\subsection{Image Resizing Based on Calibration Ball Reference}

A calibration ball is commonly used as a standard reference object to facilitate the volume/size estimation. Two options are available for using the calibration ball: (i) if the location of camera/smartphone is fixed during image acquisition, a calibration ball can be first captured before any object; and (ii) if the location of camera/smartphone keeps changing or only a limited number of devices are available, the calibration ball should be captured together with the object in every image for a consistent reference. The first option usually provides more efficiency, but the latter is more versatile if the ideal condition cannot be achieved. In each case, the user is expected to take three approximately orthogonal views of the object, i.e., from the top, front, and side.

\begin{figure}[!htb]
	\centering
	\includegraphics[width=\textwidth]{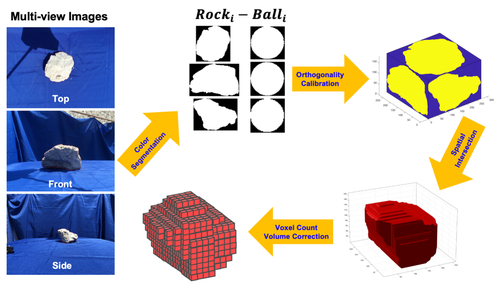}
	\caption{Flowchart of 3D volumetric reconstruction algorithm}
	\label{fig: 4-6}
\end{figure}

After a successful image segmentation, three silhouettes of an individual rock, $Rock_i$,  are cropped from the corresponding binary images. Accordingly, three silhouettes of the calibration ball, $Ball_i$, are also cropped, and their equivalent diameter $r_i$ are measured. The information can be paired as $Rock_i-Ball_i-r_i, i=1 (top),2 (front),3 (side)$, as shown in \autoref{fig: 4-6}. The three rock silhouette images $Rock_i$ are then resized based on the equivalent diameter $r_i$ of its calibration ball correspondence. The purpose of this step is to counteract the effect of different lens zoom and unequal camera-object distances. 

\subsection{Orthogonality Calibration Using Least Squares Solution}

Although the rock silhouettes have been resized with respect to the calibration ball, the dimension of these silhouettes can rarely achieve an exact match primarily because both lack perfect orthogonality and photogrammetry error. Therefore, a linear system of equations is formed as follows to obtain a standardized dimension for orthogonality correction:
Suppose the target orthogonal dimension is $[x_0\times y_0\times z_0]$, and each silhouette has an image height-width dimension of $[h_{top} \times w_{top}],[h_{front}\times w_{front} ],[h_{side}\times w_{side}]$. By aligning each silhouette dimension with the orthogonal dimension (see \autoref{fig: 4-6} above), the following linear system of equations should be satisfied:
\begin{equation} \label{eqn:4-5}
	A\ \mathbf{x}=\mathbf{b}
\end{equation}
where
\begin{equation*}
	A=\begin{bmatrix}
		0 & 0 & 1\\
		1 & 0 & 0\\
		1 & 0 & 0\\
		0 & 1 & 0\\
		0 & 0 & 1\\
		0 & 1 & 0
	\end{bmatrix},
	\mathbf{x}=\begin{bmatrix}
		x_0\\
		y_0\\
		z_0
	\end{bmatrix},
	\mathbf{b}=\begin{bmatrix}
		w_{top}\\
		h_{top}\\
		w_{front}\\
		h_{front}\\
		w_{side}\\
		h_{side}\\
	\end{bmatrix}
\end{equation*}

The linear system in \autoref{eqn:4-5} can be solved as a least squares problem to minimize the residual error term, i.e.:
\begin{equation}
	\begin{aligned} \label{eqn:4-6}
		\mathbf{x_*}=\arg\min_{x} &\lVert A\mathbf{x}-\mathbf{b}\rVert_2^2\\
		&\big\Updownarrow\\
		A^TA\mathbf{x_*}&=A^T\mathbf{b}
	\end{aligned}
\end{equation}
where the target orthogonal dimension $x_*$ is obtained by solving the normal \autoref{eqn:4-6}.

\subsection{Spatial Intersection of Multi-View Silhouettes}

Three silhouettes are then calibrated to the orthogonal dimension $x_*=[x_0\times y_0\times z_0]$, and the object solid can now be reconstructed by replicating each binary silhouette along its orthogonal dimension and determining their intersection set, as shown in \autoref{fig: 4-6} above. The intersection set is represented as a binary matrix where the object solid has a value of 1. The volume of the intersected body can then be calculated in terms of “voxels,” namely a 3D cuboid version of pixel. The reconstructed volume can be obtained from the voxel ratio between the rock and calibration ball.

\subsection{Volume Correction }

Note that the reconstruction algorithm based on orthogonal views will always overestimate the volume of the object \parencite{rao_development_2001}. The overestimation mainly results from the following two aspects: systematic overestimation, which is related to the algorithm methodology, and image resolution–based overestimation, which is limited by the image precision. A detailed analysis is conducted on both aspects, and corresponding corrections are applied to the reconstructed volume result.

\subsubsection{Systematic Correction}

Denoting the actual voxel set of the object as $S$ and the reconstructed one as $S'$, then the volume (total number of voxels) of the reconstructed object, $V(S')$ should always be equal to or greater than the volume of the actual object, $V(S)$:
\begin{equation} \label{eqn:4-7}
	V(S)\leq V(S')  
\end{equation}

\autoref{eqn:4-7} can be proven by contradiction as follows. The reconstructed object $S'$ must share identical silhouette with the actual object $S$ from three orthogonal views, i.e., the following statement must hold during the reconstruction process:
\begin{equation} \label{eqn:4-8}
	\pi_i(S)= \pi_i(S' )=s_i\ (i=1,2,3)  
\end{equation}

\noindent where
$\pi_i$ is the silhouette-projection operation along direction $i$ and $s_i$ is the projected silhouette along direction $i$.

Suppose the proposition in \autoref{eqn:4-7} is false, then $V(S)>V(S' )$. Accordingly, there must be a voxel $M$ in $S$ but not in $S'$, i.e., $
\exists\ M(x,y,z),M\in S,M\notin S'$. This implies that at least one of the three silhouette projections of $M$ does not lie within the silhouettes of both $S$ and $S'$, i.e.
\begin{equation} \label{eqn:4-9}
	\exists\ i\in\{1,2,3\}, \pi_i(M)\in \pi_i(S),\pi_i(M)\notin \pi_i(S') \Rightarrow pi_i(S)\neq \pi_i (S') 
\end{equation}

\noindent which contradicts the statement in \autoref{eqn:4-8}. By contradiction, \autoref{eqn:4-7} is proven.

However, this systematic overestimation is hard to measure quantitatively for two reasons. The first is the high randomness of the riprap shape, which is a natural property of riprap based on the productive process. The second is the insufficient surficial information on dents, hollow portions, or cavities that are not in sight of cameras \parencite{rao_development_2001}. Therefore, a correction factor $c_1$ for eliminating the systematic overestimation can only be selected empirically. Based on preliminary laboratory data, $c_1=0.954$ is reckoned and used in the volume correction process. 

\subsubsection{Resolution-based Correction}

\begin{figure}[!htb]
	\centering
	\includegraphics[width=0.5\textwidth]{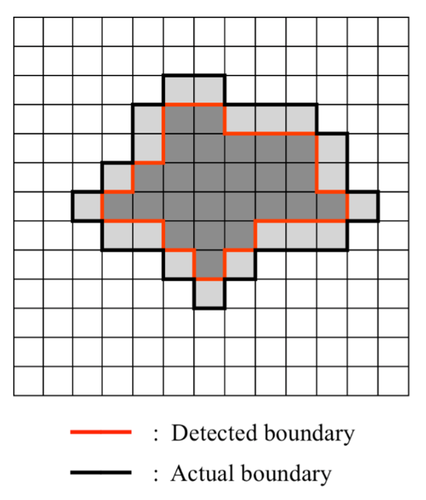}
	\caption{Pixel-scope difference between the detected boundary and actual boundary}
	\label{fig: 4-7}
\end{figure}

As the first step in the image-segmentation algorithm, images are compressed into $w\times h$ ($w\geq h$, typically $1024\times 768$) resolution. In a digital image, there is usually a pixel-wise transition from an object to the background. Since the segmentation algorithm is based on foreground-background contrast, the detected boundary will slightly shrink from the actual object boundary. Typically, a one-pixel difference can be observed between the detected boundary and actual boundary, as illustrated in \autoref{fig: 4-7}. 

This effect can cause resolution-based overestimation controlled by two factors. The first one is the relative size ratio between the rock and calibration ball, and the second is the absolute pixel occupancy of the calibration ball. The influence of these two factors is presented as follows.

For the convenience of analysis and explanation, sphere-shaped objects are assumed for both the calibration ball and riprap rock. Suppose the actual radii of the calibration ball and rock are $r_{ball}$ and $r_{rock}$, respectively. Based on the observation in \autoref{fig: 4-7}, their detected radii in image will be $r_{ball}-1$ and $r_{rock}-1$, both in terms of pixel. Denote the volume of reconstructed ball as $V_{ball}$ and the volume of reconstructed rock (based on detected boundary before applying any volume correction) as $V_{rock_ detected}$, both in terms of voxels. Then, the ratio between the reconstructed rock's volume and the reconstructed ball's volume is given as follows:
\begin{equation} \label{eqn: 4-10}
	\frac{V_{rock_detected}}{V_{ball}}=\frac{(r_{rock}-1)^3}{(r_{ball}-1)^3}
\end{equation}

The actual volume of rock is denoted as $V_{rock}$ in terms of voxels. The ratio of the actual rock's volume and the actual ball's volume is given as follows:
\begin{equation} \label{eqn: 4-11}
	\frac{V_{rock}}{V_{ball}}=\frac{r_{rock}^3}{r_{ball}^3}
\end{equation}

Then, the resolution-based correction factor $c_2$ is calculated by:
\begin{equation} \label{eqn: 4-12}
	c_2=\frac{V_{rock}}{V_{rock_detected}}=\frac{(r_{ball}-1)^3 r_{rock}^3}{(r_{rock}-1)^3 r_{ball}^3}
\end{equation}

Let $t=\frac{r_{rock}}{r_{ball}}$, then \autoref{eqn: 4-12} can be simplified as:
\begin{equation} \label{eqn: 4-13}
	c_2=(1-\frac{t-1}{t\cdot r_{ball}-1})^3
\end{equation}

\begin{figure}[!htb]
	\centering
	\includegraphics[width=0.8\textwidth]{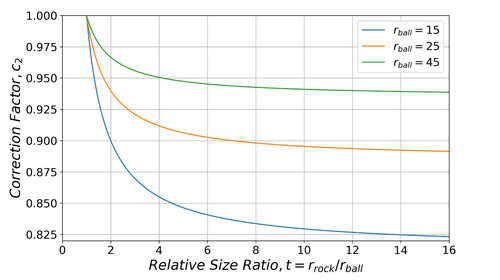}
	\caption{Effect of rock/ball ratio and calibration ball size on volume overestimation}
	\label{fig: 4-8}
\end{figure}

The correction factor $c_2$ is a function of the relative size ratio $t$ and the absolute pixel occupancy $r_{ball}$, as shown in \autoref{fig: 4-8}. Note that with the increase of $t$, or decrease of $r_{ball}$, the value of resolution-based correction factor $c_2$ will decrease. To better illustrate the effect, typical values of $r_{ball}={45,25,15}$ and $t\in [1,15]$ are selected for a parametric analysis. For example, in a $1024\times 768$ image, when the calibration ball has a radius of 25 pixels and the relative size ratio between rock and ball equals 7, the correction factor to be applied will be $c_2$=0.90. 

As a result, correction factors $c_1=0.95$ for systematic correction and $c_2$ from \autoref{eqn: 4-13} for resolution-based correction will be applied to the reconstructed rock volume at the end of the reconstruction algorithm. 

\section{Comparison with Ground-Truth Measurement and Manual Method}

After applying the image segmentation and 3D reconstruction algorithms to all image triplets, the reconstructed volume of each riprap rock was obtained. As described in \cref{sec-3-2-1}, the reconstructed results were validated with ground-truth volume/weight measurements and compared with results from IDOT's manual measurement practice. 

\begin{figure}[!htb]
	\centering
	\includegraphics[width=0.85\textwidth]{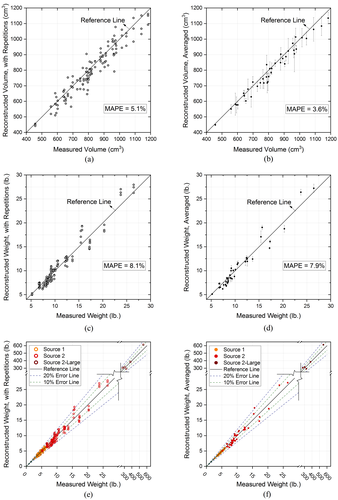}
	\caption{Comparisons between image analysis results and ground-truth measurements for: (a) Source 1 material with rotate repetitions, (b) Source 1 material after averaging, (c) Source 2 material with rotate repetitions, (d) Source 2 material after averaging, (e) all sources with rotate-repetitions, and (f) all sources after averaging (1 lb. = 0.454 kg, 1 $in.^3$= 16.4 $cm^3$)}
	\label{fig: 4-9}
\end{figure}

\begin{figure}[!htb]
	\centering
	\includegraphics[width=0.9\textwidth]{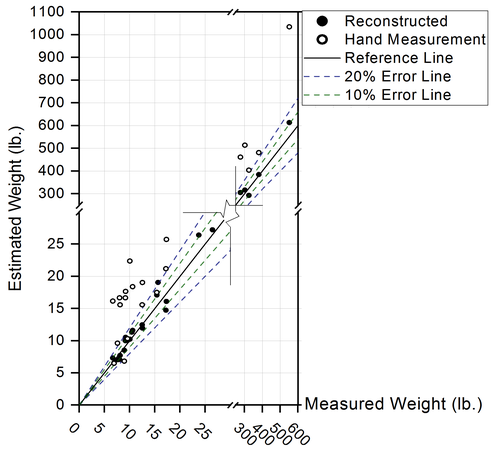}
	\caption{Comparisons between weights estimated from image analyses and weights estimated from hand measurements on Source 2 and Source 2–Large particles (1 lb. = 0.454 kg)}
	\label{fig: 4-10}
\end{figure}

For Source 1 particles (see \autoref{tab:3-3}), since the volume was directly measured during the specific gravity test, comparison was made between the reconstructed volume and measured volume. For Source 2 and Source 2–Large particles, since only the weight information was available on-site, the reconstructed volume and the volume calculated from hand measurement were first converted to weight based on a typical specific gravity value $G_s=2.66$ and then compared with the measured weight. The results are presented in \autoref{fig: 4-9}, \autoref{fig: 4-10}, and \autoref{tab: 4-1}.

\autoref{fig: 4-9}a compares all reconstructed volume results (i.e., with three rotate repetitions) with the ground-truth measurements. \autoref{fig: 4-9}b compares the averaged volume results with the ground-truth measurements for Source 1 particles. Similarly, \autoref{fig: 4-9}c and \autoref{fig: 4-9}d compare the reconstructed results with repetitions and after averaging with the ground-truth measurements for Source 2 particles, but in terms of weight. A 45-degree line is plotted as the reference for the ground-truth comparisons. Error bars are used in the averaged results plot to present the standard deviation among three rotate repetitions for individual particles. For consistently quantifying the error for each source, the following statistical indicator Mean-Absolute-Percentage-Error (MAPE) was calculated as follows:
\begin{equation} \label{eqn: 4-14}
	MAPE(\%)=\frac{\sum_{i=1}^N |\frac{E_i-M_i}{M_i}|}{N}
\end{equation}

\noindent where
$E_i$ is the estimated result from image analysis or hand measurement of $i^{th}$ particle, $M_i$ is the ground-truth measurement of $i^{th}$ particle, and $N$ is the total number of particles.

Note that the average results have less deviations from the ground-truth measurements in terms of MAPE, i.e., $3.6\%$ and $7.9\%$, as compared to the $5.1\%$ and $8.1\%$ for Source 1 and Source 2 particles, respectively. This indicates that increasing the number of image viewing angles and averaging the results can help reduce the random sampling error by obtaining more comprehensive stereo-photography information of the object. Moreover, by comparing Source 1 and Source 2 particles, it is observed that the assumption made on material specific gravity could introduce an additional source of error. Source 1 was compared by accurately measured rock volumes and thereby smaller deviation from ground truth $(3.6\%)$ was achieved, while the assumption of $G_s=2.66$ made on Source 2 rocks can bring additional error to the weight comparison ($7.9\%$). Additionally, the main drawback of the silhouette-based 3D reconstruction approach is its inability to reconstruct concavities \parencite{cremers_multiview_2010}. This is an inherent property of the reconstruction approach and introduces an inevitable source of error. 

\autoref{fig: 4-9}e and \autoref{fig: 4-9}f show the complete database of the image analysis results with Source 1, Source 2, and Source 2-Large particles. Axis breaks are made to better visualize the results that span a large range. Note that most data points lie within $\pm 20\%$ error band from the reference line, and more than half of them locate within the $\pm 10\%$ band, for all the 85 particles. Good agreement is achieved between the image analysis results and the ground-truth measurements, in terms of either accurately measured volume or converted weight with given specific gravity.

\autoref{tab: 4-1} presents the hand measurement data and image analysis results. Volume estimation procedure of the hand measurements is based on a cuboid assumption where the volume of riprap is the multiplication of three estimated midway dimensions from roughly orthogonal axes, i.e., $V=a\times b\times c$. The volume results from two approaches are then converted to weight using $G_s=2.66$. 

\begin{table}[!htb]
	\centering
	\caption{Comparisons between Image Analysis Results and Manual Measurements on Source 2 and Source 2-Large Particles (1 lb. = 0.454 kg)}
	\label{tab: 4-1}
	\begin{tabular}{L{2cm}L{2cm}L{2cm}L{2cm}L{2cm}L{2cm}}
		\hline
		\textbf{ID} &
		\textbf{Measured Weight} &
		\textbf{Weight Estimated from Hand Measurement} &
		\textbf{Weight Estimated from Image Analysis} &
		\textbf{Error Associated with Hand Measurement} &
		\textbf{Error Associated with Image Analysis} \\ 
		(-)       & (lb.) & (lb.)  & (lb.) & (\%)  & (\%) \\\hline
		2-1       & 9.6   & 10.3   & 10.0  & 7.4   & 4.7  \\
		2-2       & 9.2   & 17.7   & 10.5  & 91.9  & 14.4 \\
		2-3       & 15.6  & 50.2   & 19.0  & 220.8 & 21.7 \\
		2-4       & 10.6  & 18.4   & 11.6  & 73.4  & 9.5  \\
		2-5       & 9.1   & 16.6   & 10.0  & 82.0  & 9.2  \\
		2-6       & 15.4  & 17.5   & 17.1  & 13.6  & 11.0 \\
		2-7       & 17.2  & 21.2   & 14.7  & 23.2  & 14.4 \\
		2-8       & 10.4  & 33.2   & 11.3  & 217.3 & 8.4  \\
		2-9       & 6.6   & 16.1   & 7.3   & 142.8 & 9.8  \\
		2-10      & 12.5  & 19.0   & 11.9  & 51.6  & 4.8  \\
		2-11      & 8.0   & 16.6   & 7.1   & 108.1 & 11.5 \\
		2-12      & 17.3  & 25.7   & 16.1  & 48.7  & 6.9  \\
		2-13      & 8.1   & 15.6   & 7.7   & 92.2  & 4.9  \\
		2-14      & 10.0  & 22.4   & 10.2  & 122.6 & 1.7  \\
		2-15      & 7.6   & 9.6    & 7.0   & 26.4  & 7.4  \\
		2-16      & 8.9   & 6.8    & 8.5   & 23.9  & 5.0  \\
		2-17      & 6.9   & 6.5    & 6.6   & 6.0   & 5.0  \\
		2-18      & 12.5  & 15.6   & 12.5  & 24.5  & 0.3  \\
		2-19      & 26.4  & 32.8   & 27.2  & 24.0  & 2.9  \\
		2-20      & 23.7  & 31.1   & 27.4  & 31.1  & 15.4 \\
		2L-1      & 380.0 & 481.1  & 384.7 & 26.6  & 1.2  \\
		2L-2      & 324.5 & 403.6  & 292.9 & 24.4  & 9.7  \\
		2L-3      & 302.0 & 513.7  & 315.9 & 70.1  & 4.6  \\
		2L-4      & 552.0 & 1034.4 & 613.3 & 87.4  & 11.1 \\
		2L-5      & 277.0 & 461.3  & 305.3 & 66.5  & 10.2 \\
		MAPE (\%) & -    & -     & -    & 68.3  & 8.2  \\ \hline
	\end{tabular}
\end{table}

\autoref{fig: 4-10} shows the comparisons between hand measurements and imaging-based reconstructed results. Note that the hand measurement results have much larger errors than the imaging-based results, with MAPE=$68.3\%$ when compared against the ground truth. Furthermore, a consistent overestimation has been observed that most of the particles fall outside the $\pm 20\%$ error band. This can be explained since the midway dimensions are based on visual estimation and hand measurement, which have more subjectivity and higher variability. In contrast, imaging-based results have a much better accuracy with a MAPE=$8.2\%$. 

Overall, the consistent improvements in estimating volume/weight of medium- to large-sized particles validates the robustness and accuracy of the algorithm and shows great potential of the algorithm for further development and implementation. Moreover, the similarity between the results within three rotate repetitions implies the reproducibility of the algorithm.
\clearpage 

\section{Summary}

This chapter presented an innovative approach for characterizing the volumetric properties of riprap by establishing a field-imaging system associated with newly developed color image-segmentation and 3D reconstruction algorithms. The field-imaging system described in this chapter with its algorithms and field application examples was designed to be portable, deployable, and affordable for efficient image acquisition. 

The robustness and accuracy of the image segmentation and 3D reconstruction algorithms were validated against ground-truth measurements collected in stone quarry sites and compared with state-of-the-practice inspection methods. The imaging-based results showed good agreement with the ground truth and provided improved volumetric estimation when compared to currently adopted inspection methods. Based on the results and findings, the innovative imaging-based system is envisioned for full development to provide convenient, reliable, and sustainable solutions for the on-site QA/QC tasks relating to individual riprap rocks and large-sized aggregates.

%% file: chapter05.tex
\chapter{Automated 2D Image Segmentation and Morphological Analyses for Aggregate Stockpiles} \label{chapter-5}

As compared to the individual-aggregate imaging approach developed in \cref{chapter-4}, this chapter presents an innovative approach to provide aggregate stockpile image segmentation and morphological analysis. Aggregate imaging systems developed to date for size and shape characterization have primarily focused on measurement of separated or slightly non-overlapping aggregate particles. Development of efficient computer vision algorithms is urgently needed for image-based evaluations of densely stacked (or stockpile) aggregates, which require image segmentation of a stockpile for the size and morphological properties of individual particles. Deep-learning-based techniques are utilized to achieve effective, automated, and user-independent segmentation and morphological analyses. 

The full form of this chapter is published in \textcite{huang2020automated, huang2020size, huang2021riprap, luo2021riprap, luo2023deep}.

\section{Deep Learning Based Workflow}

\begin{figure}[!htb]
	\centering
	\includegraphics[width=\textwidth]{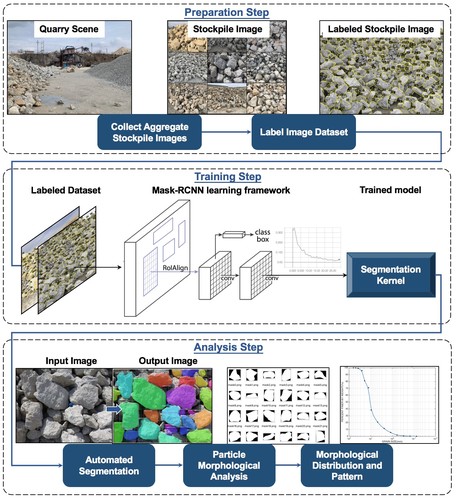}
	\caption{Flowchart of deep learning-based image segmentation and morphological analysis approach}
	\label{fig: 5-1}
\end{figure}

To analyze stockpile aggregate images, the objective is to establish an innovative approach consisting of an image-segmentation kernel based on deep learning framework and a morphological analysis module for particle shape characterization. The flowchart of the research approach follows a preparation-training-analysis pipeline (see \autoref{fig: 5-1}).

The deep learning-based image-segmentation process is data-driven and thus requires a high-quality labeled training dataset from which to extract and learn the intricate image features needed. In the preparation step, stockpile aggregate images are collected, and individual riprap rock particles in the images were manually labeled. These manual labels, or annotations, commonly serve as the ground-truth data for both training and validation purposes in image-segmentation problems. Furthermore, transfer learning is usually used as a time-saving and cost-effective solution in deep-learning research by utilizing generalized models that are already pretrained on a large dataset and fine-tuned to task-specific data \parencite{pratt_machine_1997}. Therefore, a pretrained object recognition model on the Microsoft COCO (Common Objects in Context) image dataset \parencite{lin_microsoft_2014} is used in the training process together with 164 manually labeled stockpile aggregate images. Twenty additional manually labeled images are used as the validation set to measure the performance of the resulting trained model.

To train the image-segmentation kernel, a state-of-the-art deep-learning framework for object detection and segmentation, Mask R-CNN \parencite{he_mask_2017}, is selected as the candidate architecture. Training parameters are tested and tuned to achieve optimal performance of the final image-segmentation kernel. Upon input of a stockpile aggregate image, the segmentation kernel performs object detection and semantic segmentation and outputs the regions of each segmented aggregate. In the analysis step, morphological analysis is conducted on each aggregate with reference to a calibration object, then collective statistics of the particle properties in the stockpile image are illustrated in the form of histogram and cumulative distribution. Completeness and precision on the validation dataset are analyzed to investigate the accuracy and robustness of the approach. 

\section{Labeled Dataset of Aggregate Stockpile Images}

As the task-specific data for the training, stockpile aggregate images in the dataset are collected based on the following criteria: (i) the dataset should include aggregates from various geological origins and aggregate producers, and (ii) the dataset should generally cover aggregates with varying size, color, texture, and from different viewing angles. Based on the source information specified in \cref{chapter-3} (\autoref{tab:3-4}), the details of aggregate producers, number of images taken, and number of labeled aggregates in all images for establishing the stockpile image dataset are detailed in \autoref{tab:5-1}. 

\begin{table}[!htb]
	\centering
	\caption{Source Information and Description of Stockpile Aggregate Image Dataset}
	\label{tab:5-1}
	\begin{tabular}{L{5cm}L{3cm}L{3cm}}
		\hline
		\textbf{Aggregate   Producer} & \textbf{Number of Images} & \textbf{Number of Aggregates} \\ \hline
		Prairie Material – Ocoya, Illinois                   & 6            & 520             \\
		RiverStone Group, Allied Stone – Milan, Illinois     & 14           & 982             \\
		RiverStone Group, Midway Stone – Hillsdale, Illinois & 100          & 6,766           \\
		Vulcan Materials Company – Kankakee, Illinois        & 44           & 3,527           \\ \hline
		\textbf{Total}                                       & \textbf{164} & \textbf{11,795}
	\end{tabular}
\end{table}

To provide the neural network with ground-truth data for learning, it is necessary to manually identify the locations and regions of all aggregate particles present in each stockpile aggregate image. This manual segmentation process is called “labeling,” or “annotation.” The VGG Image Annotator (VIA, \cite{dutta_vgg_2016})  is selected as the tool to ease the labor-intensive manual labeling process. Each aggregate region is marked by a polygon with all vertex coordinates recorded in pixel dimension. These regions are given a label named “rock” so that when processing this image from the dataset, the neural network will search for this label and locate every aggregate region in the image. 

\begin{figure}[!htb]
	\centering
	\begin{subfigure}[b]{0.45\textwidth}
		\centering
		\includegraphics[height=6cm]{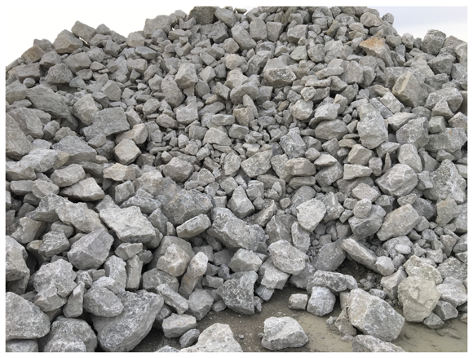}
		\caption{}
	\end{subfigure}
	\hfill
	\begin{subfigure}[b]{0.45\textwidth}
		\centering
		\includegraphics[height=6cm]{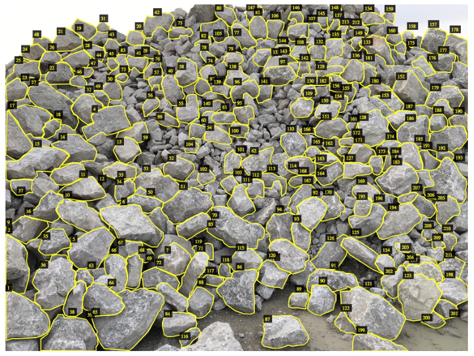}
		\caption{}
	\end{subfigure}
	\caption{Stockpile aggregate image (a) before labeling and (b) after labeling}
	\label{fig: 5-2}
\end{figure}

The main idea is to label as many particles as possible in a stockpile image according to the following criteria: (i) the polygonal line should carefully approximate the particle boundary with no large deviation from the real shape; (ii) one should try to label all human-identifiable particles, except very tiny ones that the naked eye cannot clearly recognize and those that are indistinguishable in dark areas; and (iii) incomplete particles at the image boundary should also be labeled so that the segmentation model can show consistent performance at different locations in an image. Example raw and labeled images are illustrated in \autoref{fig: 5-2}. In this example, a total of 213 aggregate particles were manually labeled in the stockpile image.

Following the above procedure and criteria established, 164 stockpile images containing 11,795 labeled aggregate particles constituted the stockpile image dataset for training. This labeled dataset serves as the human vision ground truth for the deep-learning framework described in detail in the next section.

\section{Deep Learning Framework for Automated Image Segmentation}

\begin{figure}[!htb]
	\centering
	\includegraphics[width=\textwidth]{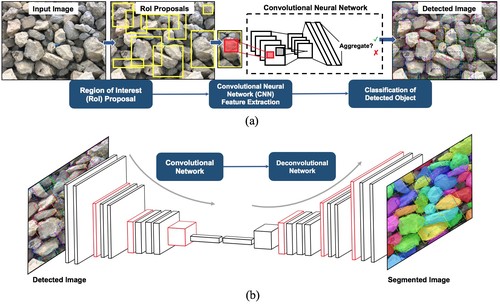}
	\caption{Model architecture of the Mask R-CNN framework composed of (a) Region-based Convolutional Neural Network (R-CNN) and (b) Fully Convolutional Network (FCN)}
	\label{fig: 5-3}
\end{figure}

Automated segmentation of stockpile aggregate images aims to identify and extract each aggregate particle, which is essentially an “instance segmentation” problem in the research domain of computer vision. Instance segmentation refers to the general task of detecting and delineating each object of interest appearing in an image, which is a popular area of interest in computer vision \parencite{romera-paredes_recurrent_2016, zhao_object_2019}. Research developments targeting algorithms related to this task have been applied in many real-life scenarios, such as urban surveillance, autonomous driving, and scene reconstruction. 
In the context of a single stockpile aggregate image, each aggregate particle becomes the target instance to be segmented. Inspired by this similarity in concept, this research adopts a recently developed and successful instance segmentation architecture named Mask Region-based Convolutional Neural Network or Mask R-CNN \parencite{he_mask_2017}, which has proven to be a breakthrough solution for general-purpose instance segmentation tasks. Mask R-CNN is a flexible and efficient framework for instance-level recognition, which can be applied to other general tasks with minimal modification. For this project's purpose, it was selected for the rock segmentation task. It is an efficient and flexible framework with convenient extensibility for task-specific applications. The instance segmentation task in Mask R-CNN is divided into an object detection step followed by a semantic segmentation step, and accordingly, Mask R-CNN is composed of two neural networks: Region-based Convolutional Neural Network (R-CNN) for object detection and Fully Convolutional Network (FCN) for semantic segmentation. The model architecture of the neural network is illustrated in \autoref{fig: 5-3}.

\subsection{R-CNN Framework for Object Detection}
Object detection is employed to estimate the contents and locations of the objects contained in an image. As one of the fundamental problems in computer vision, object detection provides comprehensive information for semantic understanding of the target image. For the stockpile aggregate image segmentation, aggregate particles should be generalized as a target category of the object to be detected. Traditionally, this task involves three stages: region selection, feature extraction, and object classification \parencite{zhao_object_2019}. Following this pipeline, this research adopts the R-CNN architecture that consists of a region proposal scheme and an object classification scheme.

The region proposal scheme simulates the attentional mechanism of the human brain during the object recognition process. The model first generates a large set of Regions of Interest (RoI), or region proposals, using a Region Proposal Network (RPN). The yellow boxes in \autoref{fig: 5-3}a are several example region proposals generated during this step. Each region proposal is then condensed into a feature map via the traditional CNN-based feature extraction network. As the next step, the object classification model feeds the feature map into a linear Support Vector Machine (SVM) and reports the object classification and confidence level of each region using non-maximum suppression. At locations with a high confidence level, overlapping bounding boxes are merged into one final bounding box marked as a detected object \parencite{he_mask_2017}. As a result, R-CNN can efficiently extract high-level features and significantly improve the quality and accuracy of the detected objects. In a general stockpile image, the algorithm is expected to detect, recognize, and locate only valid aggregate particles and distinguish them from other elements such as sky, ground, workers, etc. Detected aggregates are marked with colored bounding boxes associated with confidence levels, as illustrated in the detected image in \autoref{fig: 5-3}a. 

\subsection{FCN Framework for Semantic Segmentation}
After object detection, semantic segmentation is needed to further extract the valid aggregate pixels inside each bounding box to obtain the particle shape and boundary. During the past few years, significant research effort has been made to accomplish this task accurately and rapidly and has achieved substantial progress. FCN is one of the most powerful models for semantic segmentation; it associates each pixel with an object class description \parencite{arnab_pixelwise_2017, long_fully_2015}. Fully convolutional, as shown by its name, is a network with pure convolutional and pooling layers, and thereby requires fewer hyper-parameters while preserving high accuracy. The network is composed of a convolutional network followed by a symmetric deconvolutional network. Through the forward inference and backward propagation mechanism, the trained network can take an input image of any arbitrary size and output localized object regions for the designated class. At the pixel level, the network will screen out the invalid non-aggregate pixels and extract the aggregate surface inside the detected bounding box. This semantic segmentation process is illustrated in \autoref{fig: 5-3}b. 

The proposed neural network in \autoref{fig: 5-3} was trained based on the pretrained COCO model using the labeled stockpile image dataset. After the training, a Mask R-CNN model, referred to as the segmentation kernel in the following context, was established for the stockpile image analysis. Following the machine learning concepts, the training of a neural network follows a forward-pass and back-propagation scheme. The forward pass will feed input image(s) to the neural network, and output is generated in the form of segmented image(s). However, since the neural network parameters are randomly initialized at the beginning, these segmentation results can deviate significantly from the ground-truth labeling. This deviation between output and ground truth is calculated by a loss function that quantifies the error. Accordingly, in the back-propagation step, the model parameters of the neural network will update based on the forward pass error. Therefore, the neural network obtains the ability to self-adjust, or “learn,” to tackle the segmentation task.

\section{Morphological Analyses of Segmented Aggregates}
After the successful segmentation, each region that belongs to a different aggregate particle is then fed into the morphological analysis module. The equivalent sizes and Flat and Elongated Ratios (FER) are calculated for these segmented particles and are presented as histogram and cumulative distribution. The unit of length used in the morphological analyses is determined with reference to a 2.25-in. (5.7-cm) blue calibration ball in the stockpile image.
The equivalent size of a particle used in this research study follows the definition of the Equivalent Spherical Diameter (ESD), which is commonly used to characterize the size of an irregularly shaped object as follows:
\begin{equation} \label{eqn: 5-1}
	\mathbf{ESD}=2\cdot \sqrt{\frac{A}{\pi}}
\end{equation}
\noindent where $A$ is the measured area of the irregularly shaped object. Users can use other size metrics such as longest, shortest, or intermediate dimension at their discretion. 
For the FER calculation, Feret dimensions \parencite{feret_grosseur_1930} are used to measure the particle shape along specified directions. Generally, the Feret dimension, also called the caliper diameter, is defined as the distance between two parallel planes restricting the particle perpendicular to the direction of the planes. The calculation of FER needs to find a maximum and a minimum Feret dimension. The maximum, or longest Feret dimension, $L_{max}$, is first determined by searching for the longest intercept with the particle region in all possible directions. Next, by searching the intercepts along the orthogonal directions against the $L_{max}$, the minimum or shortest Feret dimension,  $L_{min}$, is obtained. The FER is then defined as the ratio between maximum and minimum dimensions:
\begin{equation} \label{eqn: 5-2}
	\mathbf{FER}=\frac{L_{max}}{L_{min}}
\end{equation}

Note that an individual particle shape is typically characterized using three morphological factors at three different scales. These are global form (large scale), related to the flatness and elongation or sphericity of a particle; angularity (intermediate scale), linked to crushed faces, corners, and edges of a particle; and, finally, surface texture (small scale), related to the smoothness and roughness of aggregate particles. Other previously developed 2D shape descriptors, i.e., Angularity Index (AI) and Surface Texture Index (STI), quantify the shapes at each scale that are widely considered to be accurate indices in the construction aggregate community. Although characterization of all shape indices is important, this research study limits the focus on the global form using FER, not on the surface texture characterization. This is because (i) such small-scale characterization typically requires an ultra-high resolution that may not be practical; (ii) surface texture has been often mechanically characterized in terms of surface roughness (i.e., effect of surface texture) such as friction coefficient (or inter-particle friction angle); and (iii) surface texture is closely linked to mineralogy and crushed faces. 

\begin{figure}[!htb]
	\centering
	\includegraphics[width=\textwidth]{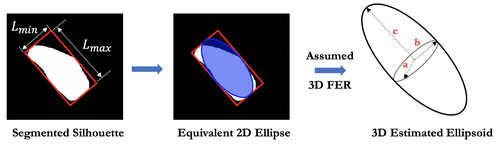}
	\caption{Volume/Weight estimation based on 2D segmented silhouette}
	\label{fig: 5-7}
\end{figure}

The size and shape indicators above define quantitatively the general 2D silhouette information. However, the current riprap classification methods adopted by most DOTs use volume or weight as the references, which are 3D size properties. Therefore, a conversion is needed to bridge between the 2D silhouette information and the 3D volume/weight estimation for practical use. For this purpose,  a volume/weight estimation module was developed. Since there is always a hidden dimension (i.e. the depth dimension) that is not directly available in 2D images, the module takes into account a typical 3D FER value (true definition of FER which is computed by dividing the longest dimension of a 3D object by its shortest dimension) as a user input and estimates the particle volume/weight based on an ellipsoidal shape assumption. The detailed procedure is illustrated in \autoref{fig: 5-7} and summarized below. The recommendation for selecting the typical 3D FER value is based on Table 1 and will be discussed after describing the procedure.

\begin{itemize}
	\item Step 1: Convert the 2D segmented silhouette to an equivalent 2D ellipse based on the Feret dimensions, $L_{max}$ and $L_{min}$, computed previously in the shape analysis. 
	
	\item Step 2: Assign the longest dimension identified in the 2D silhouette, $L_{max}$, to the longest dimension $2\cdot c$ of the 3D ellipsoid (see \autoref{fig: 5-7}). This assumes the longest dimension will be visible in 2D images. Otherwise, if the longest dimension of the aggregate is in the hidden dimension, a valid volume estimation would not be possible because the magnitude of the hidden dimension cannot be bounded. Hence, the volume estimation assumes the longest dimension is visible on the stockpile surface.
	
	\item Step 3: Use the user-provided 3D FER to determine the shortest and intermediate dimensions. Note that 3D FER is defined as the ratio between the longest and shortest dimension, as shown in \autoref{eqn: fer3}:
	\begin{equation} \label{eqn: fer3}
		\mathbf{FER_{3D}}=\frac{c}{a}, \ (c\geq b\geq a)
	\end{equation}

	Therefore, if the FER of 2D ellipse is greater than the assumed 3D FER, it indicates that the shortest dimension is present in the 2D ellipse and the given 3D FER value is rejected. In this case, set the shortest dimension $2\cdot a$ and the intermediate dimension $2\cdot b$ both equal to the shorter dimension of the 2D ellipse, $L_{min}$. Otherwise, the hidden dimension is the 3D shortest dimension and the 2D dimension $L_{min}$ is in fact the intermediate dimension. In this case, set the intermediate dimension $2\cdot b$ equal to the 2D shorter dimension, $L_{min}$, and infer the shortest dimension $2\cdot a$  based on the longest dimension $2\cdot c$ and the assumed 3D FER.
	
	\item Step 4: Calculate particle volume based on the ellipsoid volume equation \autoref{eqn: vol}:
	\begin{equation} \label{eqn: vol}
		\mathbf{V}=\frac{4}{3}\cdot \pi \cdot a \cdot b \cdot c
	\end{equation}
	
\end{itemize}

As a reference to the typical 3D FER values to be used, all riprap particles collected in the previous individual-aggregate study (see \cref{chapter-4}) have been analyzed for their 3D FER statistics. Three orthogonal views of each particle were analyzed, and the shortest/longest dimensions were calculated based on the minimum/maximum measurement among the multiple views. The information of the riprap particles studied, and the 3D FER statistics are presented in \autoref{tab:fer3}. Note that the studied particles conform to the RR3 to RR5 categories in IDOT classification \parencite{idot_standard_2016}, and the results only suggest the typical 3D FER values for these materials produced in Illinois quarries.

\begin{table}[]
	\centering
	\caption{Flat and Elongated Ratios (3D FER) for Different Riprap Categories in Individual Aggregate Study}
	\label{tab:fer3}
	\begin{tabular}{L{2.5cm}L{2cm}L{2cm}L{2cm}L{2cm}L{1.5cm}L{1.5cm}}
		\hline
		\textbf{Source Name} & \textbf{Number of Particles} & \textbf{Size Range (in.)} & \textbf{Minimum} & \textbf{Maximum} & \textbf{Median} & \textbf{Average} \\ \hline
		Source 1-RR3 & 40 & 3 to 6   & 1.55 & 3.26 & 1.97 & 2.16 \\
		Source 2-RR4 & 40 & 5 to 16  & 1.24 & 2.94 & 1.89 & 1.94 \\
		Source 2-RR5 & 5  & 16 to 26 & 1.27 & 1.9  & 1.55 & 1.56 \\ \hline
		\multicolumn{7}{l}{Note: 1 in. = 2.54 cm}  
	\end{tabular}
\end{table}

\section{Evaluation of Instance Segmentation Performance}

To validate and visualize the performance of the segmentation kernel, 20 labeled images were randomly selected as the validation set. The validation set typically serves as a benchmark for measuring the performance of trained models, since the images in this set have never been used in the training process. Model performance on the validation set indicates the generality and robustness of the model when processing unseen images. 

\subsection{Comparison of Image Segmentation Results}
\begin{figure}[!htb]
	\centering
	\includegraphics[width=\textwidth]{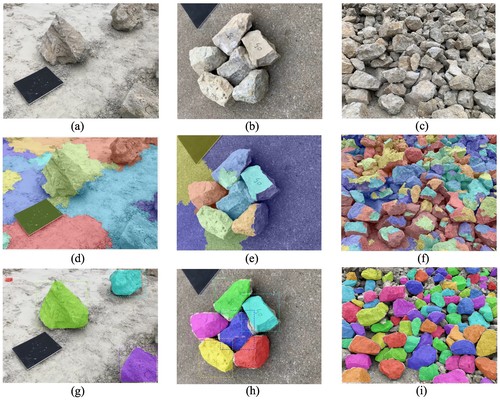}
	\caption{Raw images of (a) separated particles, (b) contacting or overlapping particles, and (c) densely stacked particles. Watershed segmented images of (d) separated particles, (e) contacting or overlapping particles, and (f) densely stacked particles. Mask R-CNN segmented images of (g) separated particles, (h) contacting or overlapping particles, and (i) densely stacked particles}
	\label{fig: 5-4}
\end{figure}

The kernel takes the images in the validation set as input and outputs the segmentation results with each aggregate particle marked by a colored mask. \autoref{fig: 5-4} illustrates the segmentation results on various types of sample images in the validation set as well as comparison with traditional watershed segmentation results. As shown in \autoref{fig: 5-4}g to \autoref{fig: 5-4}i, the segmentation kernel successfully completes the image-segmentation task and achieves robust performance on different types of aggregate images, such as separated particles, non-overlapping particles, and densely stacked particles. An interesting phenomenon to note is that although the training dataset contains only stockpile aggregate images, the trained model has gained a more general and consistent skill of segmenting aggregates in different types of backgrounds. This indicates that this segmentation kernel may have learned certain intrinsic morphological features of aggregate particles, and thereby has the potential to process general aggregate images other than the stockpile form.

When compared with the traditional watershed segmentation results, the Mask R-CNN segmentation results have much better partitioning along the aggregate boundary (see \autoref{fig: 5-4}). This can be explained by the different mechanisms behind the watershed and CNN-based methods. The watershed method tries to separate all pixels in the image based on the regional intensity with no clues on the semantic meaning of an object's existence. The CNN-based method, on the other hand, first conducts object detection and locates all potential particle regions and then segments the aggregate pixels in detail. This mechanism, along with the confidence value reported for each detected region, ensures that the model would rarely recognize irrelevant pixels as aggregates. Additionally, without the object detection mechanism, the watershed algorithm tends to categorize every pixel in the entire image into one of the regions, which is counter-intuitive in the context of aggregate image segmentation. For example, in \autoref{fig: 5-4}d to \autoref{fig: 5-4}f, watershed segmentation results include many non-aggregate fragments such as the ground or blackboard, which are very difficult to eliminate by fine-tuning the algorithm parameters. Additional post-processing steps may be needed to select the valid aggregate regions before conducting the morphological analyses. Conversely, CNN-based segmentation identifies the greatest number of individual particles in visually reasonable shapes and thus requires little or no post-processing, as illustrated in \autoref{fig: 5-4}g to \autoref{fig: 5-4}i.

Another important observation is that the problematic shadow issue is well handled in the CNN based segmentation results. The shadow effect has always been a challenge during digital image processing, since computer vision algorithms have difficulty distinguishing between an on-surface shadow and a cast shadow, especially when the algorithms rely heavily on human-defined features. In \autoref{fig: 5-4}d to \autoref{fig: 5-4}f, the watershed method is also misled by the shadow such that several aggregate particles are segmented into two adjacent regions along the light-shadow divide. The Mask-RNN based method—which better emulates the perception of the human vision system—unambiguously extracts the whole particle. This is because the convolutional scheme of this neural network recognizes implicit features among multiple levels of abstraction instead of focusing on local features such as texture or pixel intensity. Such advantage enhances the reliability and precision of each segmented aggregate particle as compared to the watershed method, as clearly illustrated in \autoref{fig: 5-4}e and \autoref{fig: 5-4}h.

Note that in \autoref{fig: 5-4}i, not all human-identifiable particles are detected and segmented. These non-segmented regions generally include two types. First, particles that are highly occluded are not detected, since unrecognizable, highly incomplete, or extremely tiny particles were not labeled during the manual labeling process. They are deliberately screened off because such particles may become outliers during the morphological analysis as they affect the accuracy of the total particle statistics. The trained model follows this feature of the labeled dataset and is therefore selective as well. On the other hand, there are valid aggregate particles that are not detected by the segmentation kernel. They are usually particles with special shape, orientation, color, or texture that are quite different from labeled ones in the dataset. This indicates that the dataset should be further enlarged to account for robustness.

\subsection{Morphological Analysis Results}
\begin{figure}[!htb]
	\centering
	\includegraphics[width=0.8\textwidth]{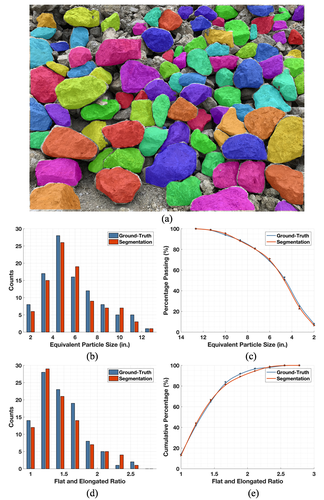}
	\caption{(a) Mask R-CNN segmented image (enlarged from Figure 4[i]); (b) histogram distributions and (c) cumulative distribution curves for equivalent particle size; and (d) histogram distributions and (e) cumulative distribution curves for flat and elongated ratio}
	\label{fig: 5-5}
\end{figure}

After the successful segmentation, each region that belongs to a different aggregate particle is then fed into the morphological analysis module. The equivalent sizes and FERs are calculated for these segmented particles and are presented as histogram and cumulative distribution. The size and shape metrics are also calculated for the corresponding labeled image and are plotted as the ground-truth comparison. The morphological analysis results for an example stockpile image in the validation set (shown in \autoref{fig: 5-4}i) are presented in \autoref{fig: 5-5}. The unit of length in the following analyses is determined with reference to a 2.25-in. (5.7-cm) blue calibration ball in the image.

For the example stockpile image in \autoref{fig: 5-5}a, 93 particles are segmented by the image-segmentation kernel, and a total of 100 particles are identified during the ground-truth labeling process. From the particle-size analysis results in \autoref{fig: 5-5}b and \autoref{fig: 5-5}c, the sizes of the aggregate particles are between 2 in. and 13 in., with about $70\%$ of the particle sizes ranging from 3 in. to 8 in. The segmentation results demonstrate good agreement with the ground truth in histogram counts and cumulative distribution. From the FER analysis results in \autoref{fig: 5-5}d and \autoref{fig: 5-5}e, the FERs range from 1.0 to 3.0, and more than $90\%$ of the particles have FERs less than 2.0. The segmentation results again capture the trends in the ground-truth histogram and cumulative distribution. Both analyses show reasonable statistical distributions for the morphological properties in an aggregate stockpile and achieve good agreement with the ground-truth labeling. The particle-size distribution curve indicates a uniform gradation of the stockpile. The FER distribution, influenced by the crushing process for this batch of aggregates, implies that more cubical particles were produced instead of long and slender ones.

Note that the morphological analysis presented herein is an example analysis with simplified analytical components. Users should be attentive to the following aspects during a formal and comprehensive morphological analysis step. First, it is highly recommended for users to take images from a perpendicular direction against the stockpile slope. The images in the training dataset have no restrictions on viewing angle, since they are meant for the development of the segmentation kernel. But the images for morphological analysis should be normal facing in order to minimize the perspective distortion (or foreshortening) effect of images. Images taken in this way can ensure the accuracy and reliability of the morphological analysis results. In addition, incomplete particles segmented at the image boundary may be removed from the morphological analysis because such shapes are caused by artifacts at the image boundary. Secondly, advanced morphological analysis modules—such as the ones in existing aggregate imaging systems—can be used for a more comprehensive characterization of particle shape regarding the form, angularity, and texture. Finally, and as a limitation, since only the stockpile surface is visible to the users, note that the morphological analysis results only represent the aggregate statistics for the surface particles in a stockpile. 

\subsection{Statistical Analyses of Segmentation Results}
\begin{figure}[!htb]
	\centering
	\includegraphics[width=\textwidth]{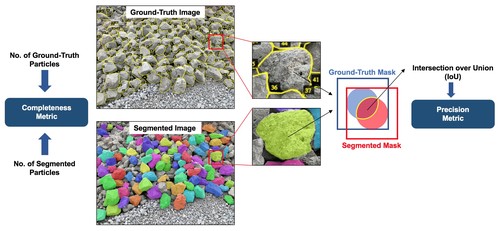}
	\caption{Completeness and precision metrics used to compare the segmentation results with the ground-truth labeling}
	\label{fig: 5-6}
\end{figure}

To evaluate the performance of the segmentation results, two important statistical indices, completeness and precision, are selected as the performance indicators. They are widely used for model evaluation in image-segmentation problems. The illustration of these two metrics is given in \autoref{fig: 5-6}. To assess the completeness of the segmentation results, the ratio between the number of segmented particles and the number of ground-truth labeled particles is calculated. This defined ratio describes the percentage of particle regions correctly detected as compared to the ground-truth labeling, which measures the overall performance of the object detection step. As for the precision metric, the Intersection over Union (IoU) score calculates the percent overlap between the segmented particle mask and the corresponding ground-truth mask. This metric measures the number of pixels in common between the segmented and ground-truth masks divided by the total number of pixels present across both masks, as given in \autoref{eqn: 5-3}.
\begin{equation} \label{eqn: 5-3}
	\mathbf{IoU}(\%)=\frac{\text{Segmented}\cap\text{Ground-Truth}}{\text{Segmented}\cup\text{Ground-Truth}}
\end{equation}
\noindent where “Segmented” denotes the region of segmented mask and “Ground-Truth” denotes the ground-truth labeled mask. The average IoU score of all segmented particles in an image measures the overall accuracy of the semantic segmentation step. 

Following the completeness and precision metrics, the model performance is evaluated on 15 stockpile images from the validation set. Note that there were in total 20 riprap images in the validation set, including five non-stockpile images of separated or non-overlapping particles (see \autoref{fig: 5-4}a and \autoref{fig: 5-5}b). To better measure the model performance on stockpile images, those five images were excluded from the completeness and precision validation and only stockpile images were selected. As listed in \autoref{tab:5-2}, the average completeness and precision values are $88.0\%$ and $86.7\%$, respectively, which are both considerably high for dense image-segmentation and analysis tasks. They indicate that the model has been well trained to detect and segment only those “true” aggregate regions instead of reporting ambiguous aggregates with a large error, which is ideal for stockpile image segmentation. 

\begin{table}[!htb]
	\centering
	\caption{Completeness and Precision Results of Randomly Selected Validation Set Images}
	\label{tab:5-2}
	\begin{tabular}{L{2cm}L{2cm}L{2cm}L{2cm}L{2cm}}
		\hline
		\textbf{ID} & \textbf{Number of Labeled   Particles} & \textbf{Number of Segmented   Particles} & \textbf{Completeness} & \textbf{Precision} \\
		(-) & (-) & (-) & (\%)  & (\%) \\ \hline
		1   & 70  & 67  & 95.7  & 87.7 \\
		2   & 56  & 53  & 94.6  & 88.3 \\
		3   & 131 & 102 & 77.9  & 87.4 \\
		4   & 73  & 65  & 89.0  & 86.1 \\
		5   & 111 & 92  & 82.9  & 83.2 \\
		6   & 99  & 84  & 84.8  & 86.9 \\
		7   & 114 & 91  & 79.8  & 83.7 \\
		8   & 106 & 88  & 83.0  & 87.9 \\
		9   & 115 & 96  & 83.5  & 86.8 \\
		10  & 117 & 95  & 81.2  & 87.5 \\
		11  & 60  & 56  & 93.3  & 88.0 \\
		12  & 149 & 127 & 85.2  & 87.5 \\
		13  & 56  & 51  & 91.1  & 87.2 \\
		14  & 62  & 62  & 100.0 & 87.8 \\
		15  & 116 & 114 & 98.3  & 85.3 \\\hline
		\multicolumn{3}{l}{\textbf{Average}}                                                            & \textbf{88.0}         & \textbf{86.7}      \\
		\multicolumn{3}{l}{\textbf{Standard Deviation}}                                                 & \textbf{7.1}          & \textbf{1.5}       \\ \hline
	\end{tabular}
\end{table}

The average completeness value shows that more than $85\%$ of aggregate particles can be identified as compared to the ground truth, and those particles can be segmented with over $85\%$ accuracy. The model misses about $10\%$ to $20\%$ of the ground-truth labeled particles, which can be explained by its conservative behavior during the detection step. By setting the confidence threshold at 0.7, the model only reports the aggregates with a relatively high precision. This often leads to a lower completeness rate, since non-aggregate regions are screened off. For morphological analysis, the target is to process reliable aggregate regions rather than poorly segmented ones. Hence, it would be adequate to maintain this conservative behavior and further improve the completeness of model performance by retraining the model with a more comprehensive dataset. In addition, the standard deviation values for completeness and precision are $7.1\%$ and $1.5\%$, respectively. This implies good generality and robustness of the model performance on different unseen input images.

Furthermore, as compared to existing aggregate imaging systems, the trade-off between the number of analyzed particles and the precision of segmented particle shape is noteworthy. These aggregate imaging systems are not efficient for massive evaluation because of the additional setup and human effort required to separate aggregate particles. But, the characterized shapes are of high precision, since the particles are all analyzed separately under controlled conditions. Namely, the morphological analyses in those systems are considered high-precision on a small sample portion of aggregates. 

In contrast, this research study has developed an efficient massive analysis based on stockpile images, but the particle shapes are of medium precision because of the inevitable occlusion and overlapping effect occurring in stockpile aggregate images. This intrinsic difference is recognized, and this approach has great practical merits. The reasons are twofold. First, this new approach may better serve the tasks when quick and massive analyses of aggregate stockpiles are demanded, e.g., in a quarry or a construction site, especially during the time-sensitive quality control process. Second, the stockpile image analysis does not require additional setup and can handle in-place evaluation of small- to large-sized aggregates, while the existing systems are limited to small-sized aggregates under laboratory conditions. In addition, as for realistic representation of an entire stockpile of aggregate material, more statistical analysis is needed to determine whether a high-precision result from a small sample group or medium-precision result of the whole stockpile surface will be more representative and informative.
\clearpage

\section{Summary}

This chapter presented an innovative approach for automated segmentation and morphological analyses of stockpile aggregate images based on deep-learning techniques. A task-specific stockpile aggregate image dataset was established from images collected from quarries in Illinois. Individual particles from the stockpile images were manually labeled on each image associated with particle locations and regions. A state-of-the-art object detection and segmentation framework called Mask R-CNN was then implemented to train the image-segmentation kernel, which enables user-independent segmentation of stockpile aggregate images. The segmentation results showed good agreement with ground-truth labeling and improved the efficiency of size and morphological analyses conducted on densely stacked and overlapping particle images. Based on the presented approach, stockpile aggregate image analysis can become an efficient and innovative application for field-scale and in-place evaluations of aggregate materials.

%% file: chapter06.tex
\chapter{3D Aggregate Particle Library and Comparative Analyses of 2D and 3D Particle Morphologies} \label{chapter-6}

The 2D instance segmentation approach developed in \cref{chapter-5} provides a convenient way to analyze stockpile images of aggregates. However, there are certain limitations of 2D imaging approaches since a significant amount of useful spatial information is lost when projecting a 3D scene onto a 2D image plane. 3D size and shape information on the other hand offers more comprehensive geometric features as well as more accurate shape characterization of aggregate material. Reliable and efficient 3D imaging techniques that can facilitate convenient QA/QC checks are still in great demand for accurately evaluating aggregate stockpiles. This first requires a 3D segmentation approach to be developed. Additionally, particles observed from a stockpile surface do not exhibit their full shapes. Thus, predicting shape information in the unseen part of the particles is also important for stockpile analysis. With all these challenges, a 3D aggregate particle database/library would serve as the cornerstone for any development relating to 3D aggregate research. This chapter presents the development of a photogrammetry approach for obtaining full 3D aggregate models, based on which an in-depth investigation is conducted regarding the 2D and 3D morphological properties.

\section{Marker-Based 3D Reconstruction Approach for the Construction of 3D Aggregate Particle Library}

\subsection{Review of Existing 3D Reconstruction Approaches}

To fully reconstruct the aggregates as 3D models, many 3D scanning-based approaches have been developed in the past decade. \textcite{anochie-boateng_three-dimensional_2013, komba_analytical_2013} used a 3D laser scanning device to obtain 3D aggregate models by a spot-beam triangulation scanning method, and similarly, \textcite{miao_feasibility_2019} used a handheld 3D laser scanner to obtain one-side 3D surface of aggregate models. Although the bottom supporting surface could usually be approximated by a flat plane, it should be noted that these 3D laser scanning results are not strictly full 3D models of the aggregates. \textcite{jin_aggregate_2018} constructed 3D solid models of nine aggregates by merging X-Ray Computed Tomography (CT) slices from the cross-sections of the specimens. Complicated searching and merging algorithms were developed to orient the CT slices to form valid 3D shapes. \textcite{thilakarathna_aggregate_2021} used a structured light 3D scanner to reconstruct 3D models by projecting preset light patterns onto the aggregate surface. Overall, these 3D scanning-based approaches usually utilize expensive scanning devices and require external lighting sources. Alternatively, more convenient and cost-effective photogrammetry approaches were investigated and demonstrated a comparable reconstruction quality when compared to the approaches requiring expensive imaging devices. \textcite{paixao_photogrammetry_2018} reconstructed 18 ballast particles by fixing the aggregate with a support pedestal and obtaining all-around views at three elevations. The particle sizes were below 3 in. (7.6 cm) to ensure stable support from the pedestal. The photogrammetry results were compared with the results from 3D laser scanning, and both methods demonstrated very close results. \textcite{ozturk_photogrammetry_2020} followed a similar photogrammetry procedure that captures all-around views from different viewing angles when the aggregate particle is glued to a stick and elevated in the air. The particle sizes were around 0.5 in. (1.3 cm) to be stably fixed using glue. Both researchers used a support system to elevate the aggregate in the air such that all-around views are accessible. However, the size range of aggregates that can be reconstructed by the procedure is greatly limited by the design of the support system.

Based on the literature review of available techniques, the major limitations of existing aggregate reconstruction systems are as follows:
\begin{itemize}
	\item Devices are costly. Most of the aggregate imaging systems that can obtain high-fidelity 3D aggregate models involve expensive devices such as 3D structured light scanner, 3D laser scanner, or X-Ray CT scanner. Commercial software tools usually come with the expensive devices. Photogrammetry-based methods using digital cameras have a much lower cost but may lack a well-established pipeline for the pre-processing and post-processing of the data. 
	
	\item Limited range of aggregate sizes that can be scanned. Unlike X-Ray CT devices, 3D laser scanners and 3D structured light scanners can generally scan a larger size range of aggregates. For existing photogrammetry-based approaches, however, the feasible size ranges are greatly limited because the procedure uses a support system to elevate the aggregate in the air for all-around inspection.

	\item Operating condition. The available 3D systems require sophisticated light control, especially for the 3D structured light devices. Photogrammetry-based approaches have more relaxed restrictions on the operating condition since digital cameras can work under various lighting conditions. However, the existing photogrammetry approaches are not designed and are less suited for field conditions. 

\end{itemize}

To address these limitations, a convenient and cost-effective procedure for the 3D reconstruction of individual aggregate particles from multi-view images was developed. The proposed photogrammetry approach follows a marker-based design that enables background suppression, point cloud stitching, and scale referencing to obtain high-quality aggregate models. The approach allows reconstruction across flexible size ranges (especially for relatively large-sized aggregates) and is potentially extensible to work under field conditions as well. The equipment setup, reconstruction mechanism, and the key designs of the reconstruction approach are detailed as follows.

\subsection{Equipment Setup}
The equipment of the reconstruction system includes a digital camera, a camera tripod, a 12-in. (30.5-cm) diameter turntable, and white cardboard background, as shown in Figure 1. The digital camera used in this study was a smartphone camera (Model: iPhone XR) with 4032-pixel by 3024-pixel resolution, but other types of digital cameras can also be used if the collected images are of sufficient quality and resolution. The camera was mounted on the tripod at a viewing angle of 30 degrees to 45 degrees with respect to the horizontal plane. A proper viewing angle ensures the top and side surfaces of the inspected aggregate particle are visible to the camera. During reconstruction, the camera was at a fixed position, and the multi-view images of the aggregate were obtained by manually rotating the turntable. The smartphone camera was programmed with an automatic shutter (with a beeping sound) every two seconds. In between two shutters, the operator rotates the turntable around 30 degrees and switches to the next view. Note that the use of a turntable and a white background with a fixed-position camera is one proposed setup to collect multiple views. The approach is flexible and designed to accommodate different configurations. For example, when applying this approach to larger aggregates that cannot easily fit onto a turntable, or a turntable is not available for field inspection, it is recommended to acquire multi-view images by moving the camera around the static object.

\begin{figure}[!htb]
	\centering
	\includegraphics[width=0.6\textwidth]{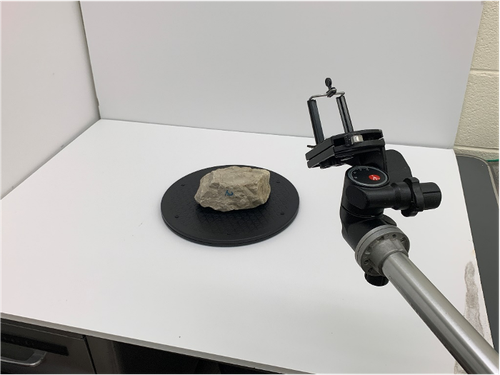}
	\caption{Equipment setup for 3D reconstruction of aggregates}
	\label{fig: 6-1}
\end{figure}

\subsection{Reconstruction by Structure-from-Motion }
In the computer vision domain, Structure-from-Motion (SfM) technique is a powerful photogrammetry method for 3D reconstruction of static scenes. The previous photogrammetry-based methods used by aggregate researchers \parencite{ozturk_photogrammetry_2020, paixao_photogrammetry_2018} also belong to the SfM category. SfM solves the problem of recovering 3D stationary structure from a collection of multi-view 2D images. A typical SfM pipeline involves three main stages: (i) extracting local features from 2D views and matching the common features across views, (ii) estimating the motion of cameras and obtaining relative camera positions and orientations, and (iii) recovering the 3D structure by jointly minimizing the total re-projection error \parencite{andrew_multiple_2001, longuet-higgins_computer_1981}. The fundamentals and implementation of SfM are omitted from this discussion, but the key steps, (ii) and (iii), are discussed herein with necessary details. The process of simultaneously estimating the camera parameters and the 3D structure is also called bundle adjustment, which is essentially an optimization problem as shown in \autoref{eqn:6-1}:
\begin{equation} \label{eqn:6-1}
	\minimize_{\{\mathbf{P},\mathbf{X}\}}\sum_{i=1}^m \sum_{j=1}^n \lVert \mathbf{x}_{ij}-\mathbf{P}_i\mathbf{X}_j\rVert^2
\end{equation}
\noindent where $\mathbf{P}_i$ is the projection matrix of the $i^{th}$ camera, $\mathbf{X}_j$ is the coordinates of the $j^{th}$ feature point in the 3D structure, and $x_{ij}$ is the projected pixel location of $\mathbf{X}_j$ in the $i^{th}$ camera view. The total re-projection error, the objective function in \autoref{eqn:6-1}, is the squared pixel distance of all feature points across all camera views. Bundle adjustment process then iteratively finds the best estimates of the camera parameters and the point coordinates by minimizing the objective. After convergence, the reconstructed structure is available as a sparse 3D point cloud and can be further processed to generate a dense point cloud.

\subsection{Background Suppression by Masking for Noise Reduction}
\begin{figure}[!htb]
	\centering
	\begin{subfigure}[b]{0.45\textwidth}
		\centering
		\includegraphics[width=\linewidth]{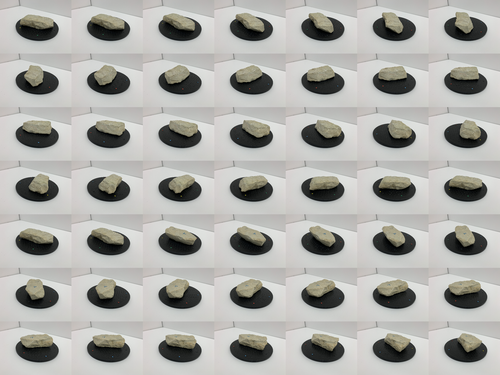}
		\caption{}
	\end{subfigure}
	\hfill
	\begin{subfigure}[b]{0.45\textwidth}
		\centering
		\includegraphics[width=\linewidth]{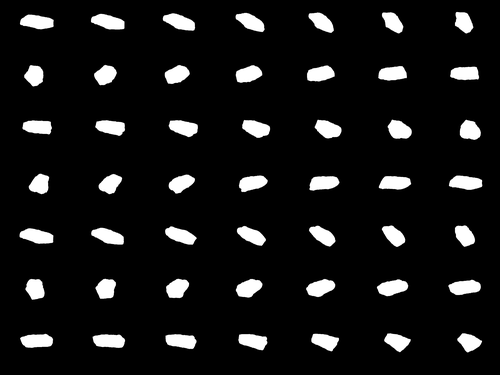}
		\caption{}
	\end{subfigure}
	\caption{(a) Multiple view images of an example aggregate particle, (b) salient object masks for each view}
	\label{fig: 6-2}
\end{figure}

\begin{figure}[!htb]
	\centering
	\begin{subfigure}[b]{0.45\textwidth}
		\centering
		\includegraphics[width=\linewidth]{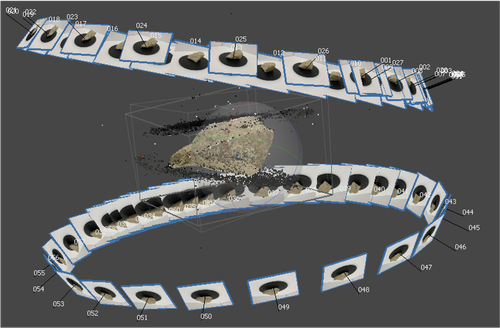}
		\caption{}
	\end{subfigure}
	\hfill
	\begin{subfigure}[b]{0.45\textwidth}
		\centering
		\includegraphics[width=\linewidth]{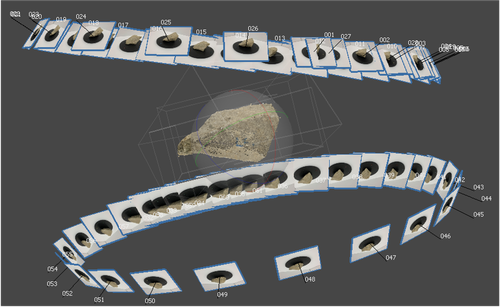}
		\caption{}
	\end{subfigure}
	\caption{Reconstructed sparse point cloud (a) without background suppression and (b) with background suppression}
	\label{fig: 6-3}
\end{figure}

The standard SfM procedure extracts features from the whole 2D images and attempts to reconstruct the entire scene, as shown in \autoref{fig: 6-2}a. This usually results in a 3D model that requires manual cleaning to remove unrelated background information (noise) and obtain a clean model of the aggregate sample only. Depending on how much of the background is reconstructed, the manual cleaning process could become considerably time-consuming, especially in regions where the aggregate is touching the background surface, as illustrated in \autoref{fig: 6-3}a. It is noteworthy that this manual cleaning requirement is not only limited to the SfM procedure. During the 3D reconstruction with costly devices (i.e., laser scanner, structured light scanner, etc.), manual cleaning is also a necessary step. This is because the scanning mechanism does not distinguish the foreground from background since their relative definition will vary from one application to another.
To reduce the various noise from unrelated background regions, the proposed approach improves the standard SfM approach by generating a foreground object mask M for each image. During bundle adjustment, the object mask is applied as additional constraints in the original objective function, as shown in \autoref{eqn:6-2}:
\begin{equation} \label{eqn:6-2}
	\minimize_{\{\mathbf{P},\mathbf{X}\}}\sum_{i=1}^m \sum_{j=1}^n M_{ij}\cdot \lVert \mathbf{x}_{ij}-\mathbf{P}_i\mathbf{X}_j\rVert^2
\end{equation}
\noindent where $M_{ij}$ is the object mask indicating the inclusion or suppression of feature $\mathbf{X}_j$ in the $i^{th}$ camera view.

The generation of this type of foreground object mask is an image segmentation problem. Although traditional segmentation methods can be applied using the color and edge information, the proposed approach adopts a deep learning-based segmentation method. The neural network architecture used is called $U^2$-Net, which is a successful design for the salient object detection task \parencite{qin_u2-net_2020}. Salient object detection is utilized to detect and extract the potential Region of Interest (RoI) of objects that may be salient in the image. The network uses deep nested U-shape convolutional-deconvolutional blocks to capture multi-scale contextual information without significantly increasing the computation cost. The training dataset was image-mask pairs prepared by both manual labeling and 3D to 2D projection of several manually cleaned 3D models. Based on experiments, around 100 image-mask pairs yield very robust and accurate foreground extraction for a given background environment. Note that for a given background environment, the network is trained only once, and no further training is involved in the reconstruction workflow. The raw images and the generated foreground masks of an example aggregate are illustrated in \autoref{fig: 6-2}.

The reason of adopting a deep learning-based method is to improve the flexibility of the proposed approach. Although the experiments conducted in this study were set up with a fixed background, the approach is designed to work in different environments, such as using different colors of the turntable and background, or under field conditions with natural lighting conditions. In such cases, a traditional segmentation method may not generate masks robustly; while the deep learning-based method only requires a few image-mask pairs to tune its behavior. The robustness of detection in natural background has been validated in the original $U^2$-Net development.

By applying the foreground masks, the unrelated background is suppressed, and the reconstructed model is noise-free and does not require any further manual cleaning. The resulting background suppression effect is illustrated in \autoref{fig: 6-3}.

\subsection{Object Markers for Robust Point Cloud Stitching }
Unlike small-sized aggregates that can be easily elevated by a support pedestal, medium- and large-sized aggregates usually need to sit on a flat surface during reconstruction or scanning. This limits the possibility of obtaining all-around views of the aggregate and reconstructing with one run of SfM. Two or more rounds of reconstruction are required on different parts of the aggregate by adjusting its pose in between, and the partial point clouds must be stitched into a complete 3D model. The most common way to stitch multiple point clouds is to use point set registration algorithms \parencite{choi_robust_2015}. However, based on experiments, automatic registration algorithms are not always robust and may fail for certain aggregate samples with less distinct surface features. 

\begin{figure}[!htb]
	\centering
	\begin{subfigure}[b]{0.31\textwidth}
		\centering
		\includegraphics[width=\linewidth]{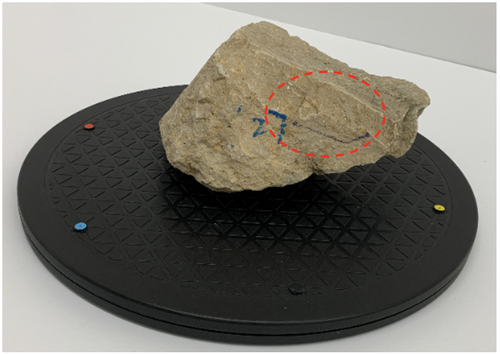}
		\caption{}
	\end{subfigure}
	\hfill
	\begin{subfigure}[b]{0.31\textwidth}
		\centering
		\includegraphics[width=\linewidth]{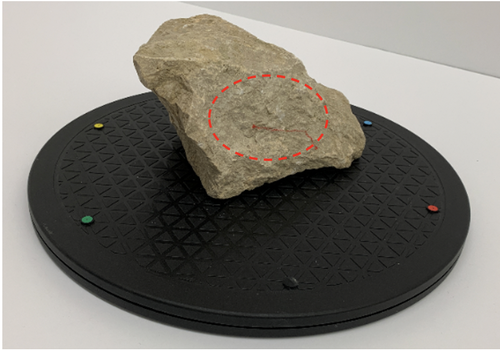}
		\caption{}
	\end{subfigure}
	\hfill
	\begin{subfigure}[b]{0.31\textwidth}
		\centering
		\includegraphics[width=\linewidth]{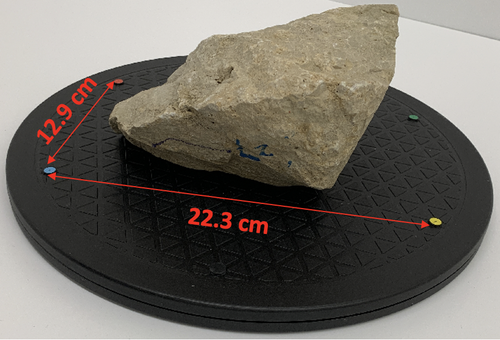}
		\caption{}
	\end{subfigure}
	\caption{(a) Purple-colored and (b) red-colored object markers for robust point cloud stitching, and (c) background markers for scale reference}
	\label{fig: 6-4}
\end{figure}

In this regard, a set of object markers was designed to provide robust feature matching during point cloud stitching. Two markers were drawn with colored pencils on the side of each aggregate. The markers were designed to have a head-tail pattern with purple and red colors, as shown in \autoref{fig: 6-4}a and \autoref{fig: 6-4}b. Note that the selected colors are not fixed and can be adjusted based on the color of the aggregate for better contrast. The head and tail of each marker are the ends of a short and long line segments, respectively. Such pattern is invariant to different viewing angles and can thus be identified robustly. After the sparse reconstruction is completed, manual labeling of the markers is required on few views (typically three views) to obtain a consistent localization of the markers in 3D coordinates. When the marker localization is completed for each partial point cloud, the stitching process can be conducted successfully, and a complete 3D model is obtained for the aggregate.

\subsection{Background Markers for Scale Reference}
The reconstructed 3D model from previous steps is in a local coordinate system. To bring the model into a true physical scale and a global coordinate system, a set of background markers was designed to provide a scale reference. The design follows the same concept of Ground Control Points (GCP) in land surveying \parencite{bernhardsen_geographic_2002}. Color-coding labels with red, green, blue, and yellow colors were placed at four corners of the turntable, as illustrated in \autoref{fig: 6-4}c. The distances between the markers were measured in advance and given as the scale factor. As discussed previously, when the proposed approach is applied without a turntable, the background markers could take other forms such as GCPs.

\subsection{Reconstruction Workflow}
The reconstruction workflow can be summarized by the following steps:
\begin{itemize}
	\item Step 1: Preparation step (executed only once for each environment). This involves setting up the equipment, tuning the foreground detection network, and placing the background markers.

	\item Step 2: Placing the aggregate sample. The sample is placed in the camera view, and object markers are labeled on the side surface.

	\item Step 3: Capturing visible sides (two or more) of the sample. By rotating the turntable (or moving the camera), multiple view images are taken. The same procedure is repeated for each side. In our experiments, 30 views were taken for each side with a two-second shutter interval, resulting in two minutes per sample for a two-side inspection.

	\item Step 4: Reconstruction. First, foreground masks are generated from the foreground detection network. Second, SfM is executed using the raw multi-view images and the associated foreground masks. Next, object markers and background markers are labeled on a subset of images (usually three images from each side). Finally, a complete 3D point cloud model is obtained by stitching the partial point clouds together, and an associated 3D mesh model is reconstructed from the complete dense cloud using the screened Poisson surface reconstruction method \parencite{kazhdan_screened_2013}.
	
	\item Steps 2 to 4 are repeated for each aggregate sample.
\end{itemize}

The reconstructed results presented in this study were generated by extending the Agisoft Metashape \parencite{agisoft_agisoft_2021} software program. Note that the implementation of the reconstruction step is not limited to certain software tools. Commercial software programs such as Agisoft Metashape \parencite{agisoft_agisoft_2021}, free software available such as VisualSFM \parencite{wu_visualsfm_2011}, or open-sourced software available such as Meshroom \parencite{griwodz_alicevision_2021}, can all be extended to implement the proposed approach. Also note that even though this research study focused on relatively large-sized aggregates, the setup previously shown in \autoref{fig: 6-1} is expected to work for smaller sizes such as base course aggregates or ballast without further adjustments.

\section{Material Information and Properties of the 3D Aggregate Library}
The outlined reconstruction procedure was used to inspect a set of 46 RR3 aggregate particles and 36 RR4 aggregate particles collected from field site visits to aggregate producers in Illinois. The samples conform to the `RR3' and `RR4' categories based on IDOT specification, which typically refers to aggregates that have weights above 10 lbs. (4.54 kg). In the specification, RR1 and RR2 categories refer to small-sized riprap aggregates having the same size ranges as aggregate subgrade material in pavement engineering and ballast material in railway engineering, and RR3 to RR7 categories are medium to large-sized aggregates or rocks that are more common in riprap applications. 

\begin{figure}[!htb]
	\centering
	\begin{subfigure}[b]{0.31\textwidth}
		\centering
		\includegraphics[width=\linewidth]{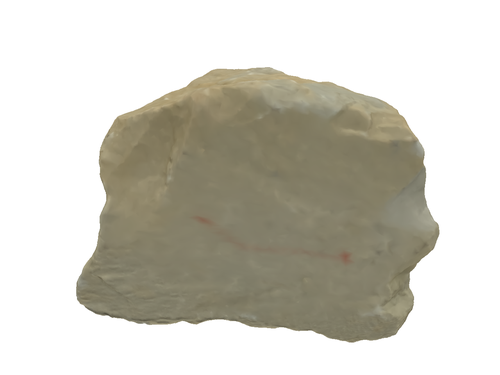}
		\caption{}
	\end{subfigure}
	\hfill
	\begin{subfigure}[b]{0.31\textwidth}
		\centering
		\includegraphics[width=\linewidth]{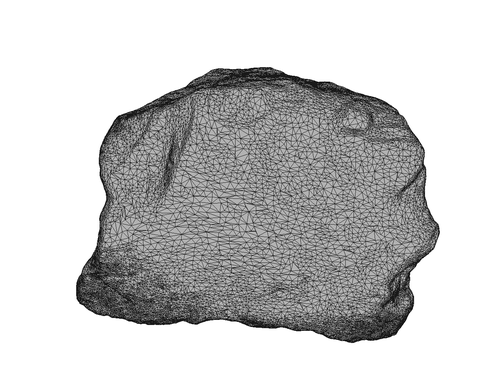}
		\caption{}
	\end{subfigure}
	\hfill
	\begin{subfigure}[b]{0.31\textwidth}
		\centering
		\includegraphics[width=\linewidth]{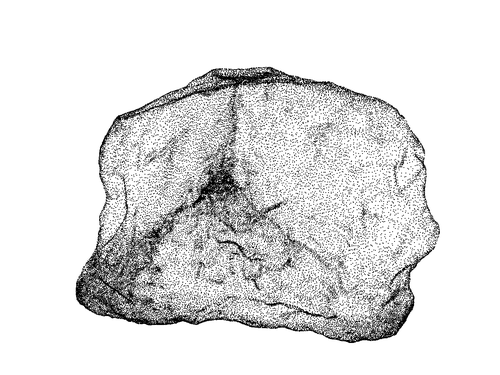}
		\caption{}
	\end{subfigure}
	\newline 
	\begin{subfigure}[b]{\textwidth}
		\centering
		\includegraphics[width=\linewidth]{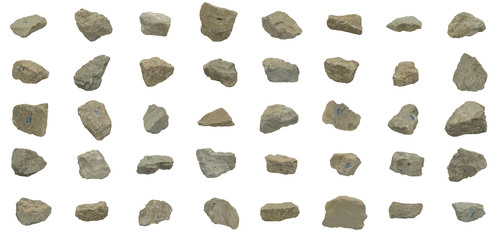}
		\caption{}
	\end{subfigure}
	\caption{(a) Textured model, (b) mesh model, (c) point cloud model of an example aggregate particle, and (d) collage of 40 reconstructed aggregate particles}
	\label{fig: 6-5}
\end{figure}

Example reconstruction results are visualized in \autoref{fig: 6-5}. The reconstructed models are available in different formats, such as the textured model that preserves the surface color information (\autoref{fig: 6-5}a), mesh model that shows the wireframe of vertex connectivity (\autoref{fig: 6-5}b), and point cloud model with discrete point coordinates (\autoref{fig: 6-5}c). An image collage of 40 RR3 aggregate samples reconstructed in this study is presented in \autoref{fig: 6-5}d. In terms of geological classification, these aggregate samples are dolomite rocks with white to yellowish colors, as shown in \autoref{fig: 6-5}d. 

The quality and fidelity of the reconstruction results were assessed from visual effects and quantitative methods. Qualitatively, it can be observed that the reconstructed aggregate models are of high quality and fidelity, as shown in \autoref{fig: 6-5}. The aggregate models reproduce the geometric features and texture features of the original aggregate samples. Quantitatively, the surface resolution (or point density) of the reconstructed results is considerably high in aggregate research. On average, each sample was exported at a resolution of around 100,000 vertices and 200,000 faces. The surface resolution and point density of ten example RR3 aggregate particles are listed in \autoref{tab:6-0}. The resolution is calculated based on the ratio between the number of points in the point cloud model and the surface area of the reconstructed mesh model. The average resolutions for all 46 RR3 aggregates and all 36 RR4 aggregates are 1.66 $points/mm^2$ and 0.93 $points/mm^2$ , respectively. The resolution statistics indicate that the aggregate models were reconstructed at a resolution of approximately 1 $point/mm^2$, i.e., the average distance between adjacent points is around 0.04 in. (1 mm). Although the reconstruction could be conducted at an even higher resolution, the necessity should be assessed in the context of aggregate research. First, the aggregate samples used in this study are relatively large-sized aggregates, with RR3 and RR4 samples having nominal sizes around 3.9 in. (10 cm) and 7.9 in. (20 cm), respectively. These large-sized aggregates are different from fine materials where particles may be at the micrometer level ($\mu m$). Therefore, the reconstruction resolution at the millimeter level ($mm$) is deemed as sufficient and considerably high for regular and large-sized aggregates. Moreover, the main purpose of the 3D aggregate particle library is to study the macro geometric features of aggregates and further investigate the aggregate assembly in stockpile forms. This also explains why the resolution of individual reconstructed models should be selected as considerably high instead of extremely high.
 
\begin{table}[!htb]
	\centering
	\caption{Surface Resolution and Point Density of Ten Example RR3 Particles}
	\label{tab:6-0}
	\begin{tabular}{L{2cm}L{2cm}L{2cm}L{2cm}L{2cm}}
		\hline
		\textbf{Rock ID} & \textbf{Surface Area ($cm^2$)} & \textbf{No. of Vertices} & \textbf{No. of Faces} & \textbf{Resolution ($points/mm^2$)} \\ \hline
		1  & 1308.69 & 99680  & 199356 & 0.76 \\
		2  & 2201.8  & 209440 & 418872 & 0.95 \\
		3  & 2586.81 & 261948 & 523884 & 1.01 \\
		4  & 2108.77 & 297392 & 594760 & 1.41 \\
		5  & 2257.61 & 151599 & 303190 & 0.67 \\
		6  & 1397.78 & 93369  & 186734 & 0.67 \\
		7  & 1664.52 & 86056  & 172108 & 0.52 \\
		8  & 1836.54 & 134359 & 268714 & 0.73 \\
		9  & 2154.91 & 223307 & 446594 & 1.04 \\
		10 & 1607.77 & 80549  & 161094 & 0.50 \\ \hline
		\multicolumn{5}{l}{Note: 1 $cm^2$ = 0.155 $in.^2$, 1 mm = 0.04 in.} 
	\end{tabular}
\end{table}

For each reconstructed aggregate particle, the basic 3D properties can be calculated from the 3D mesh model, including volume, surface area, and the shortest, intermediate, and longest dimensions in the three principal axes. The 3D properties of the ten selected RR3 aggregate particles are listed in \autoref{tab:6-1}. If the intermediate dimension is denoted as the nominal size of an aggregate, the sizes of these aggregate samples ranged from 3 in. (7.6 cm) to 6 in. (15.2 cm). For the ground-truth, the submerged volume of each aggregate sample was measured by a water displacement method following \textcite{astm_d6473_standard_2015}, listed as the measured volume in the second column in \autoref{tab:6-1}. 

\begin{table}[!htb]
	\centering
	\caption{Measured Volume, Reconstructed Volume, Area, and Principal Dimensions of Ten Example RR3 Particles}
	\label{tab:6-1}
	\begin{tabular}{L{2cm}L{2cm}L{2cm}L{2cm}L{2cm}L{2cm}L{2cm}}
		\hline
		\textbf{Rock   ID} &
		\textbf{Measured   Volume ($cm^3$)} &
		\textbf{Reconstructed   Volume ($cm^3$)} &
		\textbf{Surface   Area ($cm^2$)} &
		\textbf{Shortest   Dimension ($cm$)} &
		\textbf{Intermediate   Dimension ($cm$)} &
		\textbf{Longest   Dimension ($cm$)} \\ \hline
		1  & 1014.9 & 1042.3  & 685.32 & 7.682  & 13.142 & 22.695 \\
		2  & 763.5  & 786.33  & 537.87 & 9.308  & 12.519 & 17.412 \\
		3  & 601.8  & 605.04  & 418.69 & 9.477  & 10.075 & 14.572 \\
		4  & 791.4  & 795.69  & 558.41 & 9.118  & 10.133 & 19.925 \\
		5  & 727.6  & 744.83  & 503.13 & 9.803  & 10.649 & 18.842 \\
		6  & 688.1  & 691.96  & 478.72 & 7.497  & 9.987  & 15.925 \\
		7  & 644    & 662.47  & 465.96 & 11.614 & 13.867 & 14.041 \\
		8  & 1140.5 & 1165.03 & 704.29 & 10.617 & 12.213 & 21.923 \\
		9  & 592.7  & 601.1   & 435.01 & 8.068  & 11.517 & 17.851 \\
		10 & 890.8  & 920.92  & 590.14 & 10.374 & 14.513 & 17.37  \\ \hline
		\multicolumn{7}{l}{Note: 1 $cm$ = 0.4 in., 1 $cm^2$ = 0.16 $in.^2$,  1 $cm^3$ = 0.06 $in.^3$} 
	\end{tabular}
\end{table}

\begin{figure}[!htb]
	\centering
	\includegraphics[width=0.6\textwidth]{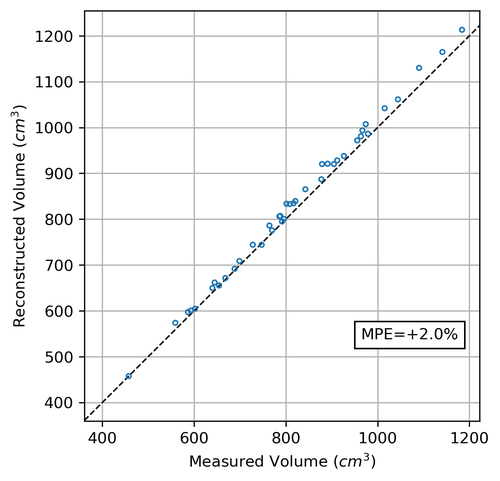}
	\caption{Comparison of reconstructed volume and measured volume of aggregate samples}
	\label{fig: 6-6}
\end{figure}

To validate the accuracy of the 3D reconstruction procedure, the reconstructed volume is compared against the measured ground-truth volume, as presented in \autoref{fig: 6-6}. A 45-degree line is plotted as a reference for the comparison. As the quantitative measure of accuracy, a statistical indicator, Mean-Percentage-Error (MPE), is calculated using \autoref{eqn: 6-3}. Note that different from Mean-Absolute-Percentage-Error (MAPE), MPE can have a positive or negative sign, indicating a systematic overestimate or underestimate behavior, respectively.

\begin{equation} \label{eqn: 6-3}
	MPE(\%)=\frac{\sum_{i=1}^N \frac{R_i-M_i}{M_i}}{N}
\end{equation}
\noindent where
$R_i$ is the reconstructed result of $i^{th}$ sample, $M_i$ is the ground-truth measurement of $i^{th}$ particle, and $N$ is the total number of particles.

\autoref{fig: 6-6} shows a very good agreement between the reconstructed volume from the marker-based reconstruction approach and the ground-truth measured volume, with a MPE of $+2.0\%$. The positive MPE also indicates a consistent, systematic overestimate of the reconstructed volumes. There are three potential reasons for this overestimation. First, the pixel locations of background markers are used to localize the marker in 3D coordinates. Therefore, pixel deviation when labeling the background markers may lead to a slight change of the scale reference. Second, a porous surface condition was observed on these dolomite aggregate particles, and the micro-texture areas that are filled with water during the measurement of the submerged volume may be reconstructed as flat faces. This could also lead to a systematic overestimate of the true submerged volume. Lastly, since SfM-based photogrammetry methods entail an optimization approach to jointly approximate the true object geometry, and cameras provide sparser representation (pixels) than laser scanning devices, it is reasonable to assume that certain systematic deviation may exist within acceptable accuracy. Also, the mesh reconstruction from point cloud is an approximation algorithm that may bring systematic deviation near the true surface of the aggregates.

When compared with the three-view reconstruction approach in \cref{chapter-4}, where MAPE of 5.1\% (before averaging) and 3.6\% (by averaging results from three repetitions) were obtained for the same aggregate particles, it is important to highlight the essential difference between the two approaches. First, the three-view reconstruction approach is a volumetric estimation approach rather than a true 3D reconstruction approach. The results from that approach are intersecting voxels (volume elements) that are simplified and approximated 3D representation of the sample, while the results from the approach developed in this study are true 3D mesh models of the sample. Second, the MAPE value with the three-view reconstruction approach is calculated after applying a complex volume correction step to the raw reconstructed volumes, while the volumes in this approach are raw volumes directly measured from the reconstructed models without any correction. Moreover, the three-view reconstruction approach targets quick size estimation in the field, yet the approach developed in this study focuses on the high-fidelity reconstruction of individual aggregates to obtain their true 3D models.

\section{Comparative Analyses of 2D and 3D Particle Morphologies}
Based on the multi-view images used during reconstruction and the resulting reconstructed 3D models, a comparative analysis is conducted to study the differences between 2D and 3D morphological properties of aggregates. The purpose of the comparative analysis is twofold: first, to check if major differences exist between 2D and 3D morphology indicators; and second, to investigate the extent to which the morphological properties from 2D analysis can represent or indicate the true 3D morphological properties.

\subsection{Morphological Indicators for Comparative Analysis}
Since the comparative analysis is between 2D and 3D morphology, the morphological indicators should ideally have both the 3D version and its counterpart in 2D. Therefore, for the aspect ratio of particle shape, 2D and 3D Flat and Elongated Ratio (FER) indices are selected as the indicator pair, and for the roundness of particles, 2D circularity and 3D sphericity are selected as the indicator pair. The descriptions of the morphological indicators is detailed herein.

\subsubsection{2D and 3D Flat and Elongated Ratios}
As the 2D indicator of particle aspect ratio, 2D FER is a widely used concept in both \textcite{astm_d4791_standard_2019} standard measurement and imaging-based approaches \parencite{gates_fhwa-hif-11-030_2011, masad_analysis_2007, moaveni_evaluation_2013, tutumluer_video_2000} . In image analysis, 2D FER is usually calculated from the particle silhouette after segmentation. Feret diameters \parencite{feret_grosseur_1930} are measured along two perpendicular directions from different orientations. The maximum or longest Feret diameter, $L_{max}$, is obtained by searching for the longest edge-to-edge distance within the silhouette in all possible orientations, while the minimum or shortest Feret diameter, $L_{min}$, is obtained by searching for the shortest edge-to-edge distance within the silhouette that is perpendicular to the $L_{max}$ direction. The 2D FER is then defined as follows (\autoref{eqn: 6-4}):
\begin{equation} \label{eqn: 6-4}
	FER_{2D}=\frac{L_{max}}{L_{min}}, (L_{max}\geq L_{min})
\end{equation}

As the 3D counterpart of the aspect ratio indicator, 3D FER can be calculated after finding the minimum volume bounding box of the particle. \textcite{orourke_finding_1985} developed algorithms to find the minimal enclosing box of a point set. First, for each possible direction originated from the particle centroid, a 3D local coordinate frame is formed in the orthogonal searching directions. Next, for each orthogonal pair, the three edge-to-edge distances (3D Feret diameters) within the point set are calculated. The volume of the bounding box can then be computed, and the Feret diameters of the minimum volume bounding box are denoted as the shortest dimension $a$, intermediate dimension $b$, and longest dimension $c$. Accordingly, the orthogonal pair associated with the minimum volume bounding box represents the three principal axes of the particle. The 3D FER can then be defined based on the principal dimensions (\autoref{eqn: 6-5}):
\begin{equation} \label{eqn: 6-5}
	FER_{3D}=\frac{c}{a}, (c\geq b\geq a)
\end{equation}

\subsubsection{2D Circularity and 3D Sphericity}
To compare the roundness of particles, a compactness measure of irregular shape is selected as the indicator, which takes the form of circularity in 2D and sphericity in 3D. Both indicators measure how closely a shape resembles a perfect circle in 2D or sphere in 3D, which serves as the unity with value 1.0.
Given the area A and the perimeter P of a 2D silhouette, 2D circularity can be calculated as shown in \autoref{eqn: 6-6}. As a reference, an equilateral triangle has a circularity of 0.605 and a square has a circularity of 0.785, with higher values indicating the 2D shape is closer to a perfect circle.
\begin{equation} \label{eqn: 6-6}
	Circularity_{2D}=\frac{4\pi A}{P^2}
\end{equation}

For 3D sphericity, \textcite{wadell_volume_1932} defined the sphericity as the ratio between the surface area of an equivalent sphere having the same volume as the particle, $S_e$, and the measured surface area of the particle, $S$. This is often called the true sphericity. Given the surface area A and the volume V of a 3D model, the 3D sphericity can be computed using \autoref{eqn: 6-7}. As a reference, a tetrahedron has a sphericity of 0.67 and a cube has a sphericity of 0.81, again with higher values indicating the 3D shape is closer to a perfect sphere.
\begin{equation} \label{eqn: 6-7}
	Sphericity_{3D}=\frac{S_e}{S}=\frac{\sqrt[3]{36\pi V^2}}{A}
\end{equation}

Note that although the circularity and sphericity are considered counterparts in 2D and 3D, the 2D and 3D versions of a shape may not necessarily have the same value. For example, if we consider cube is a 3D version of the 2D square, its 3D sphericity (0.81) differs slightly from the 2D circularity (0.785). When comparing 2D circularity values with 3D sphericity values, this intrinsic difference should be recognized. However, the overall trend from angular shape to round shape is consistent for sphericity and circularity.  With the morphological indicators introduced in this section, 2D and 3D morphology statistics can be compared quantitatively. 

\subsection{Comparison Results}
\begin{figure}[!htb]
	\centering
	\begin{subfigure}[b]{0.8\textwidth}
		\centering
		\includegraphics[width=\linewidth]{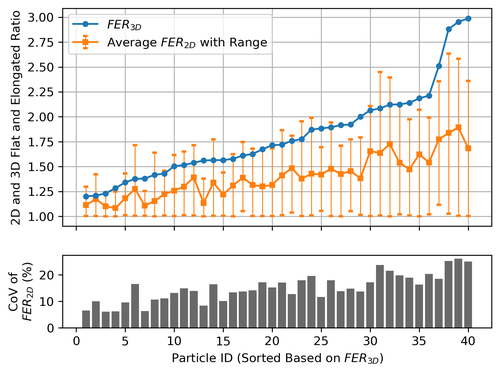}
		\caption{}
	\end{subfigure}
	\newline
	\begin{subfigure}[b]{0.8\textwidth}
		\centering
		\includegraphics[width=\linewidth]{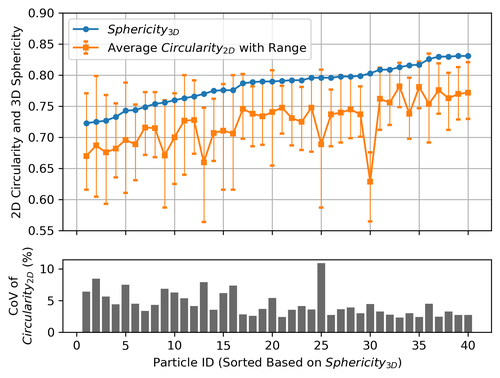}
		\caption{}
	\end{subfigure}
	\caption{(a) Comparison of 3D FER and 2D FER from multiple views, and (b) comparison of 3D sphericity and 2D circularity from multiple views}
	\label{fig: 6-7}
\end{figure}

For 2D morphology, 2D FER and 2D circularity are calculated from the multi-view images used in the reconstruction. The average value from multiple views is reported with range and standard deviation. Since for each aggregate particle, its 2D statistic is a distribution covering values from multiple views, directly comparing the standard deviation among different samples is not valid because each sample may have different averages. Therefore, the Coefficient of Variance (CoV), i.e., ratio between the standard deviation and the average, is used to characterize the variation of each sample. For 3D morphology, 3D FER and 3D sphericity are calculated from the reconstructed 3D model. \autoref{fig: 6-7} presents the comparison between 2D and 3D morphology. Note that the horizontal axis is sorted based on the 3D statistics to better illustrate the trend.

\autoref{fig: 6-7}a shows that 3D FER of an aggregate is consistently higher than the average 2D FER from multi-view images. The 3D FERs of the samples range from 1.0 to 3.0 with around $75\%$ of the samples having 3D FER less than 2.0. As for average 2D FERs, the values range from 1.0 to 2.0 with more than $75\%$ of the samples having an average 2D FER of less than 1.5. In addition to the average 2D FERs, the range bars illustrate the minimum and maximum 2D FER values across all multi-view images. It can be observed that the minimum 2D FERs usually reach 1.0 and the maximum 2D FERs have values close to the 3D FERs. This indicates that during multi-view 2D analysis, there are certain views that can better represent the true 3D FER than others. However, it should be stressed that 2D aggregate analysis is often times limited to a single-view analysis, e.g., in practical scenarios such as analyzing the aggregate shape on a conveyor belt, from top-views of aggregates spread on a table, or one angled face in aggregate stockpiles \parencite{gates_fhwa-hif-11-030_2011, masad_analysis_2007, moaveni_evaluation_2013, tutumluer_video_2000, cao_effects_2019}. Therefore, the average 2D FER could represent the value that is most likely to be obtained from a single-view analysis.

\autoref{fig: 6-7}b shows that the 3D sphericity of an aggregate is also consistently higher than the average 2D circularity from multi-view images. The 3D sphericities of the samples range between 0.70 and 0.85, and the average 2D circularities mostly lie between 0.65 and 0.80. Again, the range bars illustrate the minimum and maximum circularities across all multi-view images. Like the FER comparison, the maximum circularities may approach the 3D sphericities for several samples, but the average 2D circularities can be considered as the most common value that can be obtained from single-view analysis.

In \autoref{fig: 6-7}a and \autoref{fig: 6-7}b, the CoV chart is also presented below the main graph. The CoVs of 2D FER among all samples are mostly ranging between $10\%$ and $20\%$, while the CoVs of 2D circularity show less variation with most values being less than $10\%$. This may imply that the circularity indicator is usually less sensitive to varying views when compared to the FER indicator. Moreover, an opposite trend is observed in the FER CoV and circularity CoV: as the particle 3D FER increases, the CoV of 2D FERs tends to increase; as the particle 3D sphericities increase, the CoV of circularity tends to decrease. This is because higher 3D FER and lower 3D sphericity both indicate a 3D shape that is close to a rounded (less angular) sphere. Such uniform 3D shape results in less variance when projected into 2D silhouettes during multi-view analysis.

\begin{figure}[!htb]
	\centering
	\includegraphics[width=\textwidth]{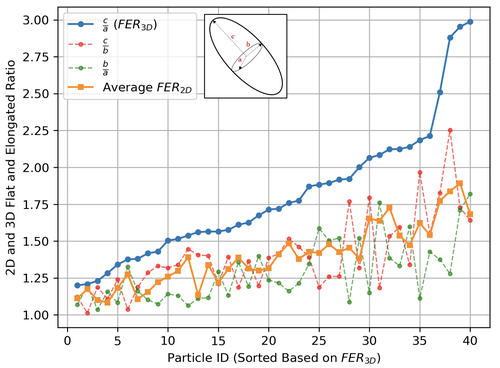}
	\caption{Comparison of average 2D FER and the ratios of 3D principal dimensions}
	\label{fig: 6-8}
\end{figure}

From \autoref{fig: 6-7} one can observe a consistent difference between the 2D and 3D morphologies, and the extent to which the morphological properties from 2D analysis can represent or indicate the 3D morphological properties needs to be further investigated. \autoref{fig: 6-8} illustrates the comparison between different ratios calculated from the 3D principal dimensions and the average 2D FER. In addition to the longest-to-shortest ratio (3D FER), the longest-to-intermediate ($c/b$) and intermediate-to-shortest ($b/a$) ratios are also plotted. Note that the $c/b$ and $b/a$ ratios must be both lower than the c/a ratio, but one ratio is not expected to be always lower or higher than the other, depending on the magnitude of the three principal dimensions. It is observed that in most cases, the average 2D FER falls within the envelope formed by the $c/b$ and $b/a$ ratios. This implies that the 2D FERs obtained from a single-view analysis are likely to capture the intermediate ratios among 3D principal dimensions rather than the true 3D FER. From an engineering practice, it is more likely the single-view 2D analysis misses the shortest dimension rather than the longest dimension. This is because a particle is mostly likely to settle along its shortest principal axis due to gravity. This observation also explains why the volume estimation step in pure 2D image analysis requires a 3D FER value to be given as an assumption. A correction factor could be applied to estimate the 3D FER from 2D FER upon further investigation using a comprehensive database of aggregate shapes.

\clearpage

\section{Summary}
This chapter reviewed existing imaging approaches for obtaining full 3D aggregate models and found the approaches usually require costly devices. To establish the 3D aggregate particle database, a marker-based 3D reconstruction approach was developed as a cost-effective and flexible procedure to allow full reconstruction of 3D aggregate shapes. The proposed approach is a photogrammetry-based method with auxiliary designs to achieve background suppression, robust point cloud stitching, and scale reference. The approach was demonstrated on relatively large-sized aggregates, and the reconstructed models showed good agreements with ground-truth measurements.
Comparative analysis was conducted between the 2D morphological properties from multi-view images and the 3D morphological properties from the reconstructed aggregate models. Significant differences were observed between the 2D and 3D statistics, which suggests that 2D morphological properties must be used carefully to infer the true 3D properties.

%% file: chapter07.tex
\chapter{Synthetic Data Generation of Aggregate Stockpiles for Deep Learning} \label{chapter-7}

In the development of 2D instance segmentation approaches, the training dataset consists of 2D images with individual aggregate boundaries manually labeled. Now evolving to the more advanced 3D instance segmentation and 3D shape completion approaches, the key challenge is that manual labeling on 3D data formats (e.g., 3D point clouds and/or 3D meshes) is known to be extremely time-consuming and likely not possible, especially for dense structures such as the aggregate stockpiles. Due to the difficulty of distinguishing highly-overlapped and mutually touching objects, the 3D manual labeling on dense structures could also be error-prone. In this regard, synthetic data generation techniques have demonstrated the power of generating realistic data with ground-truth labels for deep learning meanwhile conforming to the physics of real world.

This chapter first reviews the successful use of synthetic datasets among different tasks in the computer vision domain, as well as the graphics engines that empower the synthetic dataset preparation. After selecting the target graphics engine for realistic scene simulation, a synthetic data generation pipeline is designed to simulate densely-stacked aggregate stockpiles based on the assembly of instances from the 3D aggregate particle library. Finally, multi-view cameras and Light Detection and Ranging (LiDAR) sensors are simulated and raycasting techniques are developed to extract the 3D dense point clouds with ground-truth labels.

\section{The Success of Synthetic Datasets in Computer Vision Domain}

Since 2010s, deep learning based computer vision algorithms have gained great popularity and become prevailing in many computer vision tasks. This well explains the demand for various types of synthetic datasets nowadays since deep learning techniques are data-driven, and as the neural network models become more complex and heavy-weight, datasets purely based on manual labeling cannot easily scale up with the model development. However, the importance of synthetic datasets to be used as the benchmark has been recognized even before deep learning emerges. 

\begin{figure}[h]
	\centering
	\includegraphics[width=\textwidth]{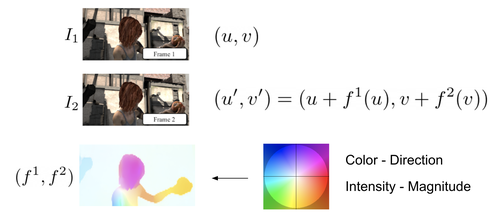}
	\caption{Optical flow estimation}
	\label{fig: sintel-optical}
\end{figure}

The pressing need for a synthetic dataset could actually date back to the 1980s in solving a low-level computer vision task named optical flow estimation \parencite{lucas_iterative_1981}. Given two consecutive image frames of a moving object or two stereo images with small disparity, ${I_1\ \text{and}\ I_2}$, the optical flow estimation task is to find the pixel-wise difference $(f^1, f^2)=(u'-u,v'-v)$ between each pixel $(u,v)$ in $I_1$  and its correspondence pixel $(u', v')$ in $I_2$. The optical flow concept is illustrated in \autoref{fig: sintel-optical}, where the color and intensity of the flow field represents the direction and magnitude of the flow vectors. The pixel differences form a optical flow field that captures the apparent velocities of pixel movement of an image. The seminal paper by \textcite{lucas_iterative_1981} developed the famous Lucas-Kanade algorithm as a computational photography approach and opened the optical flow research, but it was later on found that there do not exist a large-scale dataset that can quantify the accuracy of such algorithms. This is because it is nearly impossible for humans to identify and compute the flow field at the pixel level thus the ground-truth for this task is rarely available. As a result, optical flow datasets are commonly restricted in size, complexity, and diversity. Since then, the optical flow researchers have initiated the effort to generate image pairs with ground-truth optical flow in a synthetic way.

\begin{figure}[h]
	\centering
	\includegraphics[width=\textwidth]{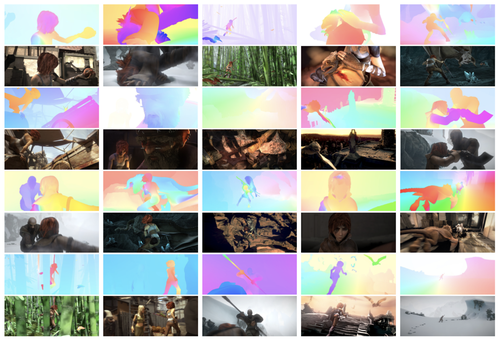}
	\caption{Ground-truth flow fields and corresponding images in MPI-Sintel optical flow dataset \parencite{butler_naturalistic_2012}}
	\label{fig: sintel}
\end{figure}

Recently, thanks to the rapid development in physics simulation, graphics rendering, and movie industry, one of the first and most famous synthetic datasets, Max Planck Institute Sintel dataset (MPI-Sintel, \cite{butler_naturalistic_2012}), was developed targeting the optical flow task, as shown in \autoref{fig: sintel}. MPI-Sintel dataset was derived from the open-source 3D animated short film Sintel, providing 35 film clips of 50 frames in length for each. Each clip involves small and large object motions in a naturalistic rendered scene. Internal motion blur pipeline in Blender software \parencite{blender_blender_2020} was modified to obtain the accurate motion vectors at the pixel level. This synthetic dataset also has control on the scene complexity thus revealed how several highly-ranked optical flow algorithms failed under different conditions, which vastly boosted the research development since then.

\begin{figure}
	\centering
	\hfill
	\begin{subfigure}[b]{0.4\textwidth}
		\centering
		\includegraphics[height=0.6\textwidth]{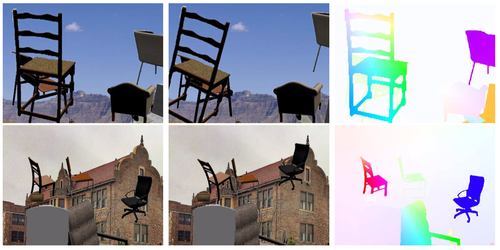}
		\caption{}
		\label{fig: flyingchairs}
	\end{subfigure}
	\hfill
	\begin{subfigure}[b]{0.4\textwidth}
		\centering
		\includegraphics[height=0.6\textwidth]{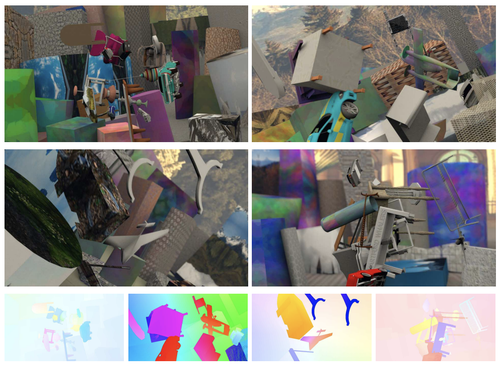}
		\caption{}
		\label{fig: flyingthings}
	\end{subfigure}
	\hfill
	\caption{(a) FlyingChairs dataset \parencite{dosovitskiy_flownet_2015} and (b) FlyingThings3D dataset \parencite{mayer_large_2016}}
	\label{fig: flying}
\end{figure}

The success of MPI-Sintel dataset has also spread the synthetic data concept to many high-level computer vision tasks and inspired several task-specific datasets \parencite{nikolenko_synthetic_2021}. \textcite{dosovitskiy_flownet_2015} introduced the FlyingChairs dataset that was specifically prepared for training CNNs architecture because the training usually requires a much larger amount of data than Sintel. The dataset contains 22,872 frame pairs and ground truth by randomly placing chairs in front of background scenes of cities, landscapes and mountains, with examples given in \autoref{fig: flyingchairs}. The follow-up research further extended the object categories as well as the task domain to 3D scene flow, resulting in the FlyingThings3D dataset with 35,927 object models \parencite{mayer_large_2016}, as shown in \autoref{fig: flyingthings}. An important finding the authors brought was that the network trained on such less-realistic datasets has achieved impressive generalization performance onto the realistic datasets (such as Sintel) and even real datasets (such as KITTI, \cite{geiger_vision_2013}). This indicates the potential of using synthetic dataset for challenging tasks and generalizing to tackle real-world problems based on the transfer learning mechanism \parencite{pratt_machine_1997}.

\begin{figure}
	\centering
	\hfill
	\begin{subfigure}[b]{0.3\textwidth}
		\centering
		\includegraphics[height=4cm]{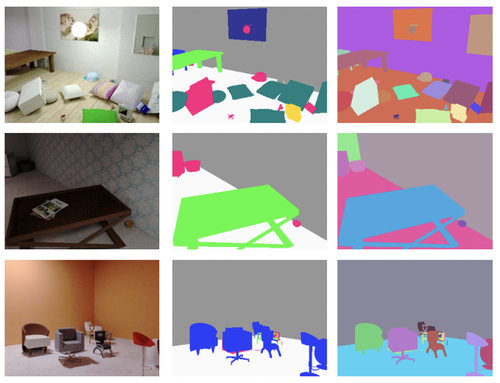}
		\caption{}
		\label{fig: scenenet}
	\end{subfigure}
	\hfill
	\begin{subfigure}[b]{0.6\textwidth}
		\centering
		\includegraphics[height=4cm]{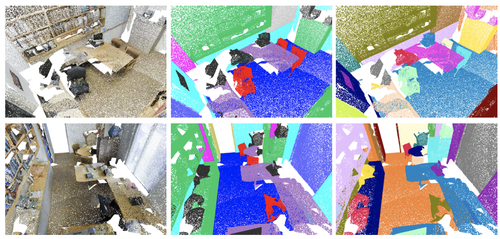}
		\caption{}
		\label{fig: s3dis}
	\end{subfigure}
	\hfill
	\caption{(a) SceneNet RGB-D synthetic dataset with semantic and instance labels \parencite{mccormac_scenenet_2017} and (b) S3DIS real dataset with semantic and instance labels \parencite{armeni_3d_2016}}
	\label{fig: indoor}
\end{figure}

Furthermore, advanced and challenging 3D computer vision tasks such as 3D object detection and semantic/instance segmentation also benefit from the use of synthetic datasets. For the task of indoor environment understanding, SceneNet RGB-D \parencite{handa_scenenet_2015, mccormac_scenenet_2017} dataset includes 57 richly annotated indoor scenes such as bedrooms, offices, kitchens, living rooms and bathrooms that render a total of 16,895 random configurations. To match with the real-world indoor settings, it introduced automatic furniture arrangement following a hierarchical approach to impose spatial constraints at both the object level and the object-group level. An illustration of SceneNet RGB-D dataset and the per-pixel semantic and instance labels synthetically generated are shown in \autoref{fig: scenenet}. As a comparison, the S3DIS dataset \parencite{armeni_3d_2016} consists of real scanned indoor scenes that are semantically parsed into disjoint spaces and building elements. The dataset is composed of five large-scale areas that cover a total of 6,020 square meters. As shown in \autoref{fig: indoor}, the synthetic SceneNet RGB-D dataset achieves almost the equivalent high-fidelity of the indoor scenes as the real S3DIS dataset, while offering more flexibility in increasing both the complexity and scale of the scene for training.

Synthetic data have also demonstrated success in many specific tasks such as understanding the autonomous driving environment, robotics, face recognition, and human pose estimation \parencite{dosovitskiy_carla_2017, hu_face_2016, varol_learning_2017}. The great success of synthetic data in the computer vision domain enlightens and inspires the author to explore the potential of solving the aggregate stockpile segmentation task. Therefore, a synthetic data generation pipeline that is specially designed for preparing aggregate stockpile dataset will be introduced next. 

\section{Data Generation Pipeline for Aggregate Stockpiles}

\subsection{Selection of Graphics Engine for Scene Simulation}

To generate the sufficient amount of training data for deep learning, the synthetic data generation process should be built as an automated pipeline. This requires the selection of an appropriate graphics and physics engine that provides application programming interface (API) and allows low-level control of the engine. The selection process started by surveying the most popular engines that have been used  for synthetic data generation. A partial list of simulation environments and their engine is presented in \autoref{tab:engine}. Virtual KITTI \parencite{gaidon_virtual_2016} and OpenAI Remote Rendering Backend (ORRB, \cite{chociej_orrb_2019}) both uses Unity$^{\textregistered}$ as their backend engine, while CARLA \parencite{dosovitskiy_carla_2017} and VRGym \parencite{xie_vrgym_2019} uses Unreal Engine (UE). These two popular graphics engines offer competing functionalities as well as high simulation quality. In terms of the programming language for the API, Unity$^{\textregistered}$ uses C$^\sharp$ and JavaScript and UE adopts C\texttt{++}. As a result, Unity$^{\textregistered}$ was selected as the final graphics engine for stockpile scene simulation.

\begin{table}[]
	\centering
	\caption{A Partial List of Simulation Platforms and Backend Engines for Synthetic Data Generation}
	\begin{tabular}{llll}
		\hline
		\textbf{Simulation Environment} & \textbf{Year} & \textbf{Domain}    & \textbf{Engine} \\ \hline
		Virtual KITTI \parencite{gaidon_virtual_2016}                  & 2016          & Autonomous Driving & Unity           \\
		CARLA \parencite{dosovitskiy_carla_2017}                           & 2017          & Autonomous Driving & Unreal Engine   \\
		VRGym \parencite{xie_vrgym_2019}                          & 2019          & Robotics           & Unreal Engine   \\
		ORRB \parencite{chociej_orrb_2019}                           & 2019          & Robotics           & Unity           \\ \hline
	\end{tabular}
	\label{tab:engine}
\end{table}

\subsection{Synthetic Data Generation Pipeline}

\begin{figure}[!htb]
	\centering
	\includegraphics[width=\textwidth]{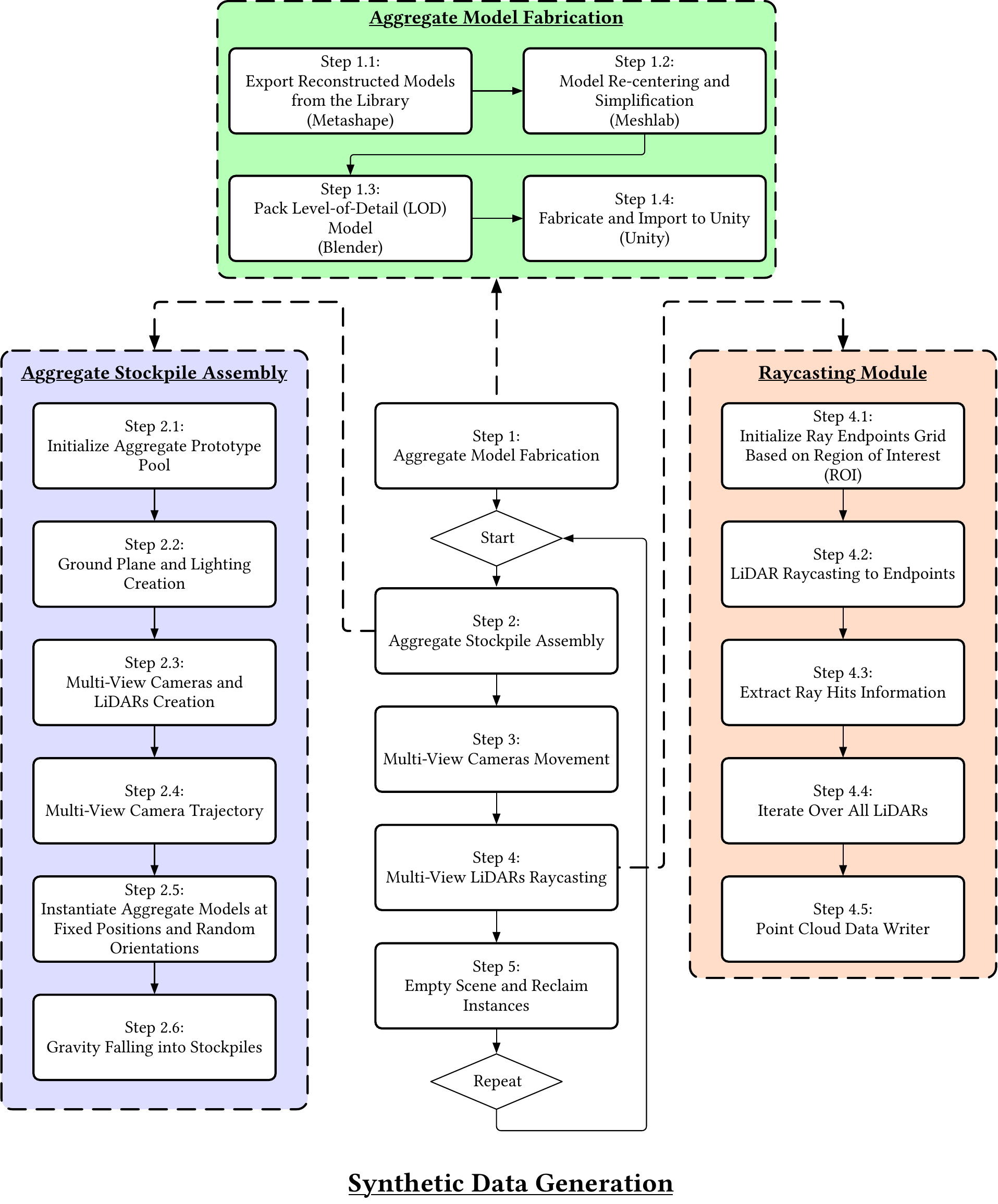}
	\caption{Synthetic data generation pipeline for aggregate stockpiles}
	\label{fig: synthetic-pipeline}
\end{figure}

The developed synthetic data generation pipeline comprises three main modules: aggregate model fabrication, aggregate stockpile assembly, and stockpile raycasting, as illustrated in \autoref{fig: synthetic-pipeline}. The aggregate model fabrication module involves necessary steps for pre-processing the aggregate model library into Unity$^{\textregistered}$'s kinematics object. The aggregate stockpile assembly module configures the scene environment by creating aggregate instances, sets up multi-view cameras and LiDARs, and enables gravity falling of the instances into a stockpile. Finally, the stockpile raycasting module simulates the mechanism of LiDAR sensors by casting rays to the stockpile and extract the 3D point cloud structure together with ground-truth labels. The model fabrication module and stockpile assembly modules are described in detail in \cref{sec-stockpile-assembly}, and the stockpile raycasting module and the results are presented in \cref{sec-raycasting}. As shown in \autoref{fig: synthetic-pipeline}, after fabricating aggregate models, the pipeline starts from the aggregate stockpile assembly module and simulates the multi-view camera movement and multi-view LiDAR raycasting once the gravity falling is completed. After finishing one stockpile, the scene is emptied and the aggregate instances are reclaimed to the pool, then new stockpile scenes are created by repeating the process until a target number of stockpiles are generated. Note that the entire pipeline was programmed and automated, allowing the data generation of arbitrary number of stockpile scenes in a very efficient way.

\section{Stockpile Assembly from the 3D Aggregate Particle Library} \label{sec-stockpile-assembly}

\subsection{Aggregate Model Fabrication}
As presented in \cref{chapter-6}, each individual aggregate model in the 3D aggregate particle library is dense mesh model that can have greater than 100,000 vertices and 200,000 faces. Note that for an aggregate stockpile to be simulated, the scene may have hundreds or even thousands of aggregate instances. Due to the computer memory restriction, modern graphics engines may not fluently handle such large quantity of dense models simultaneously. Therefore, the following pre-processing steps were conducted in the aggregate model fabrication module (see \autoref{fig: synthetic-pipeline}).

\begin{itemize}
	\item Step 1.1: Exporting reconstructed models from the 3D aggregate particle library. Aggregate models previously reconstructed with the Metashape software \parencite{agisoft_agisoft_2021} were exported in Wavefront OBJ format with vertex coordinates, face connections, vertex normals, and texture information. Wavefront OBJ is a standard mesh format that is commonly used in 3D modeling \parencite{marschner_fundamentals_2018}.
	
	\item Step 1.2: Re-centering and simplifying the raw models. Since all raw reconstructed models are not in a consistent coordinate system, the models were first re-centered at the origin following \cref{eqn: recentering}, where $N$ is the total number of vertices, $(\hat{x_i},\hat{y_i},\hat{z_i})$ are the re-centered coordinates, and $(x_i,y_i,z_i)$ are the original coordinates. 
	\begin{equation} \label{eqn: recentering}
		\forall\ i\in \{1,...,N\}, \ (\hat{x_i},\hat{y_i},\hat{z_i}) = (x_i,y_i,z_i) - \frac{1}{N} \sum_{j=1}^{N}(x_j, y_j, z_j)
	\end{equation}

	Next, the re-centered models were simplified (or downsampled) to a specific number of faces, resulting in the models with various Levels of Detail (LOD). The main reason for mesh simplification is that the necessary details of mesh vary by application, e.g., in our context of simulating an aggregate assembly, it is more desirable to use simplified versions of excessively detailed models. The simplification step utilizes the quadric based edge-collapse strategy in Meshlab \parencite{garland_surface_1997, cignoni_meshlab_2008}, which well preserves primary geometric features and topology of the raw mesh. The strategy is different from uniform vertex clustering methods and is also not identical to adaptive sampling where significant density variation exists. It is an iterative and shape-preserving sampling that contracts edges and maintains low surface error approximations. Three LOD levels were selected with LOD0 having 2,000 faces, LOD1 having 1,000 faces, and LOD2 having 500 faces, as lower LOD number indicating richer details. The demonstration of the LOD generation is presented in \autoref{fig: LODs}. Note that the simplified models have a smoother surface compared to the raw model, but still preserve the shape details of the aggregate models reasonably well for the simulation.
	\begin{figure}[!h]
		\begin{tikzpicture}
			\node[anchor=south west,inner sep=0] (image) at (0,0) {\includegraphics[trim=0 300 0 200, clip, width=\textwidth]{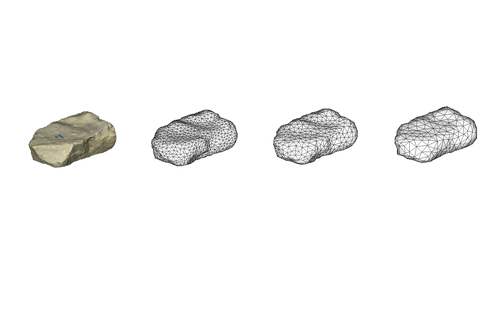}};
			\begin{scope}[
				x={(image.south east)},
				y={(image.north west)}] %
				\node[above, black, fill=white] at (0.125,0){Raw Model};
				\node[above, black, fill=white] at (0.375,0){LOD0 Model};
				\node[above, black, fill=white] at (0.625,0){LOD1 Model};
				\node[above, black, fill=white] at (0.875,0){LOD2 Model};
			\end{scope}
		\end{tikzpicture}
		\caption{Aggregate model simplification with different Levels of Detail (LOD)}
		\label{fig: LODs}
	\end{figure}
	
	\item Step 1.3: Packing LOD models. The main application of the LOD technique is to dynamically adjust the number of graphic operations to render models at a distance. For example, objects that are closer to the camera can be rendered with more details than those that are far away from the camera. This technique not only allows large-scale rendering with memory efficiency, but also reproduces the phenomenon of humans observing less details of distant objects \parencite{clark_hierarchical_1976}. To enable LOD in Unity$^{\textregistered}$, Blender software \parencite{blender_blender_2020} was used to pack the multiple LOD models into a hierarchical Autodesk FBX format. FBX is a proprietary format that provides good interoperability between modeling software such as Blender and Unity$^{\textregistered}$ \parencite{coumans_extensionspyscriptsmanualexportfbx_2009}. 
	
	\item Step 1.4: Fabricating and importing to Unity$^{\textregistered}$. With the packed LOD model, the geometry information of the aggregate models is ready for importing to Unity$^{\textregistered}$. However, since the process of individual aggregates forming a stockpile should follow the physics rule, it is necessary to fabricate the models with realistic physical properties. First, it is worth mentioning that the scale of the models was preserved throughout the previous steps, i.e. each model is to real-world scale. Second, the mass of each model was assigned as their weight measurement data. Finally, each model was imported as a rigid body with collision detection enabled. The collision detection mode was set to continuous dynamic which allows accurate results for fast moving actions such as the falling process. During the image rendering process in the graphics engine, the underlying LOD mesh used is automatically determined based on current camera viewing distance and the preset LOD transition band at $\{60\%, 30\%, 1\%\}$. During the simulation process in the physics engine, the most simplified mesh (i.e., LOD2) was used as the underlying collider to provide the most efficient collision detection for massive objects simulation.
\end{itemize}

\begin{figure}[!h]
	\centering
	\includegraphics[width=\textwidth]{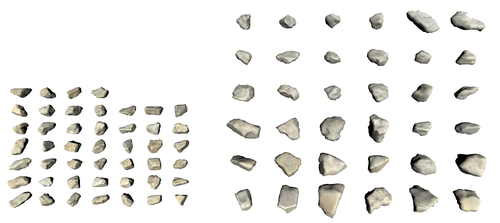}
	\begin{minipage}[t]{.5\linewidth}
		\centering
		\subcaption{RR3 Models}
	\end{minipage}%
	\begin{minipage}[t]{.5\linewidth}
		\centering
		\subcaption{RR4 Models}
	\end{minipage}
	\caption{Fabricated aggregate models in Unity$^{\textregistered}$}
	\label{fig: library-unity}
\end{figure}

As a result, the final fabricated models imported into Unity$^{\textregistered}$ with real-world physical properties are presented in \autoref{fig: library-unity}. It can be observed that the fabricated models are of the real-world scale showing that RR4 models are larger than the RR3 models.

\subsection{Aggregate Stockpile Assembly}
After the pre-processing steps, the simulation of the aggregate stockpile assembly entails aggregate model instantiation, multi-view cameras and LiDARs setup, and gravity falling. The details of the aggregate stockpile assembly module (see \autoref{fig: synthetic-pipeline}) are described as follows.

\begin{itemize}
	\item Step 2.1: Initializing aggregate prototype pool. As discussed previously, computer memory issue has always been an optimization bottleneck for fluent scene simulation. Considering the stockpile simulation process may induce continuous generation of hundreds of scenes with each scene containing hundreds of aggregates, the data management between the current scene and the next should be handled efficiently. The simplest approach that restarts a new scene by destroying all current instances was found to be problematic. This is because such a massive memory reclamation process takes long and fails to catch up with the simulation frame refresh rate. The final solution was to design an aggregate prototype pool, such that when an aggregate instance was created and no longer in use for the next scene, it is reclaimed to the pool and marked inactive rather than being destroyed in memory. Hence, when the same aggregate prototype is used repeatedly in new scenes, it is always maintained in the pool instead of subjected to repeated creation and destruction. This approach provides efficient data management and enables continuous simulation of stockpile scenes.
	
	\item Step 2.2: Ground plane and lighting creation. The ground plane of the scene was made infinite in scale to support aggregate stockpiles, and then it was textured with real ground images taken at quarries during the field site visits. To illuminate the scene, an upright directional light was added. The initial scene with only ground plane and the lighting source is illustrated in \autoref{fig: ground-and-light}. Note that the coordinate system in Unity$^{\textregistered}$ is a left-handed system. In the simulation, the $Y$ direction was denoted as the vertical direction and the ground plane is the $X-Z$ plane. The green square on the ground stands for the designated Region of Interest (ROI) for the synthetic data generation. Aggregate stockpiles are generated approximately within the ROI. It should be noted that the synthetic data generation herein mainly focuses on generating geometry data related to the aggregate stockpile rather than rendering photo-realistic images of the stockpiles, therefore the simplified lighting condition is considered sufficient for the task and does not aim to simulate the natural lighting condition (e.g., using ambient lighting).
	
	\begin{figure}[!htb]
		\centering
		\includegraphics[width=0.8\textwidth]{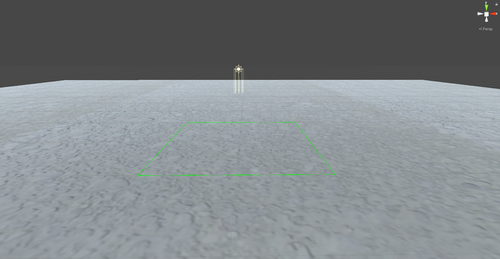}
		\caption{Ground plane and directional light of the stockpile scene}
		\label{fig: ground-and-light}
	\end{figure}

	\item Step 2.3: Multi-view cameras and LiDARs creation. To verify the correct assembly of the aggregate stockpiles and extract the point cloud data, multi-view cameras and LiDARs were designed, respectively. The multi-view cameras are positioned in a ring pattern at a specific height. Suppose the ROI center is $(c_x, c_z)$, the edge lengths of the ROI is $L_x$ and $L_z$, the camera height is $H$, and the number of multi-view cameras is $N$. The radius of the camera ring, $R$, is denoted as a factor $r$ times the half-diagonal length of the ROI, as given in \autoref{eqn: ring-radius}. 
	\begin{equation} \label{eqn: ring-radius}
		R = r\cdot \frac{\sqrt{L_x^2+L_z^2}}{2}
	\end{equation}
	
	The angle increment between two adjacent camera positions can be calculated based on the number of cameras evenly distributed on the ring:
	\begin{equation}
		\Delta \theta= \frac{2\pi}{N}
	\end{equation}
	
	Therefore, the $i^{th}$ camera position $(p_{i,x}, p_{i,y}, p_{i,z})$ can be computed by \autoref{eqn: cam-pos}. As for the camera orientation, all cameras were set by a look-at direction which points to the center of the ROI $(c_x, 0, c_z)$.
	\begin{equation} \label{eqn: cam-pos}
		\forall i \in {1,...,N}, 
		\begin{pmatrix}
			p_{i,x}\\
			p_{i,y}\\
			p_{i,z}
		\end{pmatrix} = 
		\begin{pmatrix}
			c_x + R\cdot cos((i-1)\cdot\Delta\theta) \\
			H \\
			c_z + R\cdot sin((i-1)\cdot\Delta\theta)
		\end{pmatrix}
	\end{equation}
	
	The multi-view LiDAR positions followed the similar ring pattern design, but experiments have indicated that LiDARs placed at various heights and ring radii yield better visibility than all LiDARs placed at the same height. Therefore, the LiDARs system consists of two rings, i.e., a narrow $N_1$-LiDAR ring with radius factor $r_1$ at a higher height $H_1$ and a wide $N_2$-LiDAR ring with radius factor $r_2$ at a lower height $H_2$, where $r_1 < r_2, H_1 > H_2$. An additional central LiDAR is also placed directly on top of the ROI center at height $H_1$. Hence, the total number of LiDARs is $N_1+N_2+1$.
	
	The multi-view cameras and LiDARs system used for a RR4 stockpile simulation is presented in \autoref{fig: cam-and-lidar}. During the simulation, the parameters used were $L_x=L_z=2, H=1, r=3, N=36, H_1=1.5, H_2=1, r_1=0.7, r_2=1.5, N_1=6, N_2=8$. Namely, a system of 36 multi-view cameras and 15 multi-view LiDARs was configured to inspect a $2m\times 2m$ ROI.
	
	\begin{figure}[!htb]
		\centering
		\hfill
		\begin{subfigure}[b]{0.6\textwidth}
			\centering
			\includegraphics[height=5.5cm]{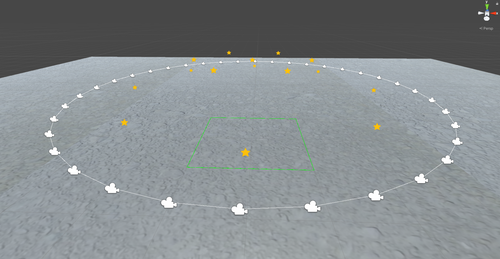}
			\caption{}
		\end{subfigure}
		\hfill
		\begin{subfigure}[b]{0.3\textwidth}
			\centering
			\includegraphics[height=5.5cm]{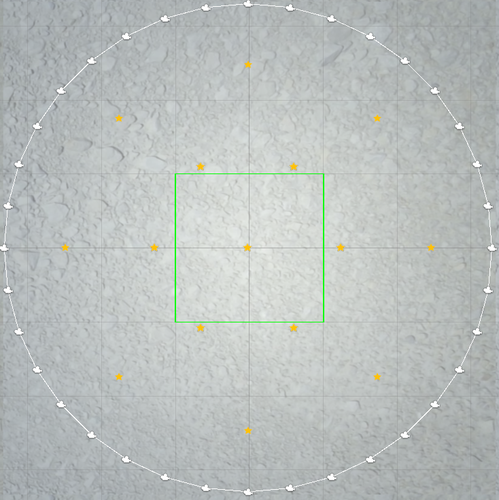}
			\caption{}
		\end{subfigure}
		\caption{(a) Side view and (b) top view of the multi-view cameras and LiDARs system}
		\label{fig: cam-and-lidar}
	\end{figure}

	\item Step 2.4: Multi-view camera motion along trajectory. To inspect the correctness and quality of the synthetic scene, camera motion was also programmed to control the camera moving along a trajectory, generating both video and multi-view images of the scene. The white curve in \autoref{fig: cam-and-lidar} is the trajectory the camera would follow once the simulation started. During the camera motion, the look-at direction was always fixed to be the ROI center, which keeps the stockpile scene in focus.
	
	\item Step 2.5: Instantiating aggregate models. Once the camera and LiDAR system has been configured, the aggregate instances are created at fixed initial positions and random orientations. First, aggregate instances were arranged in a grid with $80\%$ size of the ROI. The number of aggregates along each grid dimension, $N_g$, depends on the relative size of the aggregate and ensures they are not contacting with each other initially. Next, more instances were generated by adding up layers of such grid. The number of grid layers was set to be a random number within given range $(L_{min}, L_{max})$ to guarantee the uniqueness and randomness of each scene. \autoref{fig: stockpile-before} demonstrates the initial arrangement of aggregate instances before the gravity falling process. This arrangement consists of 10 layers of RR4 instances on a $6\times 6$ grid, resulting in 360 instances to form a stockpile. Note that the aggregate morphology (size and shape) was strictly preserved by only varying the position and orientation of the models. This guarantees the consistency in engineering properties between the 3D particle library and the simulated instances, therefore provides reliable benchmark for any further per-instance level validation. Also, this ensures every aggregate instance corresponds to a real shape of natural rocks rather than being virtual and irregularly-deformed shapes.
	
	\begin{figure}[!htb]
		\centering
		\hfill
		\begin{subfigure}[b]{\textwidth}
			\centering
			\includegraphics[width=\linewidth]{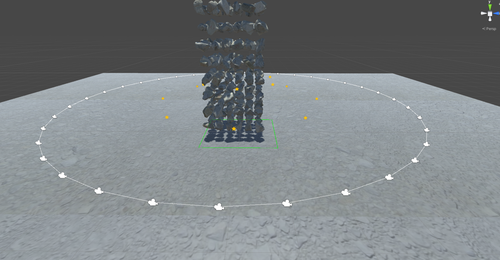}
			\caption{}
			\label{fig: stockpile-before}
		\end{subfigure}
		\hfill
		\begin{subfigure}[b]{\textwidth}
			\centering
			\includegraphics[width=\linewidth]{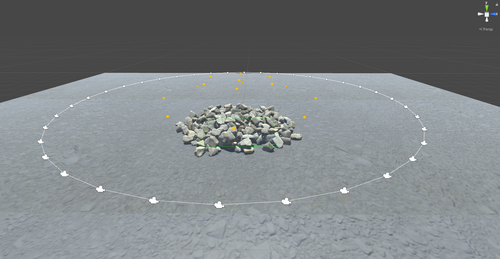}
			\caption{}
			\label{fig: stockpile-after}
		\end{subfigure}
		\caption{(a) Instantiated aggregates before gravity falling and (b) aggregate stockpile after gravity falling}
		\label{fig: unity-stockpile}
	\end{figure}
	
	\item Step 2.6: Gravity falling into stockpiles. Since each aggregate instance was fabricated with the real physical properties, the gravity falling of aggregates is enabled by setting the dynamics mode and starting the scene simulation. Collision detection are performed among instances based on their mesh structures. Typically, the falling process takes one or two seconds until the entire scene stabilizes. As shown in \autoref{fig: stockpile-after}, the formed stockpile presents a cone-shaped structure that is similar to common stockpiles at the quarry. Note that although the simulated gravity falling process does not perfectly reproduce the realistic formulation of a stockpile at the quarry (e.g., dumping from a haul truck), the particle arrangement of the stockpile is expected to have no significant difference from a real stockpile. This is because the dumping operation at the quarry also only involves falling due to the rock's gravity, and there is usually little or no packing involved during the real formulation of a stockpile.
\end{itemize}

\section{Synthetic Data Generation with Ground Truth Labels} \label{sec-raycasting}

\begin{figure}[!htb]
	\centering
	\includegraphics[width=\textwidth]{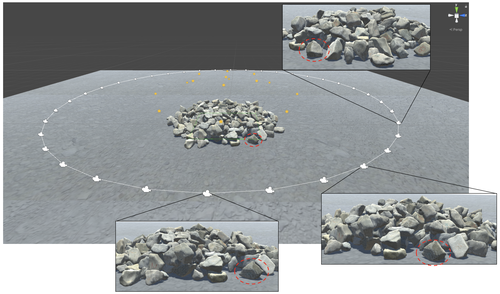}
	\caption{Multi-view camera images of the aggregate stockpile}
	\label{fig: unity-stockpile-multiview}
\end{figure}

After the stockpile assembly is completed, synthetic data can be generated from the stockpile scene. First, multi-view camera images are saved as the camera moves along the designated trajectory. These images help to inspect the correctness and quality of the stockpile assembly process. For example, as shown in \autoref{fig: unity-stockpile-multiview}, the multi-view images agree well with the stockpile scene, with a certain aggregate instance (marked by a red circle) being consistently visible across multiple cameras. The cone-shaped structure of the stockpile can also be better observed in the camera views.

Next, 3D point cloud data should be extracted from the stockpile scene together with the ground-truth labels that are necessary for deep learning. This step was developed as a standalone module using the raycasting techniques, as presented in \autoref{fig: synthetic-pipeline}. Raycasting (or ray-tracing) is a core technique in computer graphics for rendering 3D scenes onto 2D plane, where virtual light rays are cast or traced from a focal point to decide the visibility of objects along the ray paths \parencite{marschner_fundamentals_2018}. Details of the raycasting implementation in the synthetic data generation are described as follows.

\begin{itemize}
	\item Step 4.1: Initializing ray endpoints based on ROI. From each of the multi-view LiDAR's position, the endpoints of the rays were first calculated based on the ROI. To capture a larger Field of View (FOV) of the stockpile after falling, the raycasting region was set as a $120\%$ enlarged region of the ROI. Then, dense 2D grid points were generated on the enlarged ROI based on a ray density parameters $d$. The ray density parameter controls the spacing between the grid points. During the raycasting, the ray density was set as $d=0.02m=2cm$, providing 14,641 grid points per-LiDAR on a $2.4m\times 2.4m$ enlarged ROI.
	
	\item Step 4.2: LiDAR raycasting to endpoints. The ray vector $Ray=\{\overrightarrow{pos}, \overrightarrow{dir}\}$ is defined by the start position $\overrightarrow{pos}$ and the ray direction $\overrightarrow{dir}$. Suppose the $k^{th}$ LiDAR position is at $\overrightarrow{pos_k}=(l_{k,x}, l_{k,y}, l_{k,z})$ and the ${ij}^{th}$ endpoint is at $\overrightarrow{endpoint_{ij}}=(e_{ij,x}, e_{ij,y}, e_{ij,z})$, the ray vectors of the $k^{th}$ LiDAR can be formulated following \autoref{eqn: ray-vector}, where $N_x$ and $N_z$ are the number of rays in the $X$ and $Z$ directions based on the ray density. Note that the ray direction is a unit vector normalized with its magnitude. As a result, these ray vectors form a cluster of rays per-LiDAR, as illustrated in \autoref{fig: unity-raycasting}.
	\begin{equation} \label{eqn: ray-vector}
			\begin{aligned}
				\forall i\in \{1,...,N_x\}, & j\in \{1,...,N_z\},\\
				Ray_k&=
				\begin{pmatrix}
					\overrightarrow{pos_k}\\
					\overrightarrow{dir_k}
				\end{pmatrix}= 
				\begin{pmatrix}
					(l_{k,x}, l_{k,y}, l_{k,z})\\
					\frac{\vec{n}}{\lVert \vec{n} \rVert}, \vec{n}=(e_{ij,x}, e_{ij,y}, e_{ij,z})-(l_{k,x}, l_{k,y}, l_{k,z})
				\end{pmatrix}
			\end{aligned}
	\end{equation}

	\begin{figure}[!htb]
		\centering
		\includegraphics[width=\textwidth]{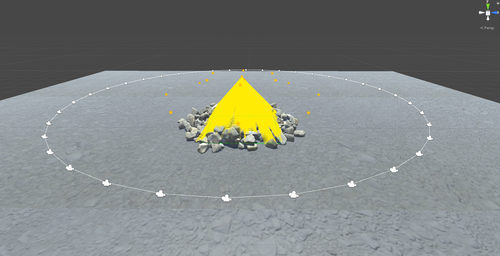}
		\caption{Cluster of rays cast from one of the LiDARs}
		\label{fig: unity-raycasting}
	\end{figure}

	\item Step 4.3: Extracting ray hits information. By casting the cluster of rays onto the stockpile scene, ray-instance intersection checks were conducted. If a ray hits the surface of an aggregate instance, the ray is marked as active. Otherwise, the ray is removed from the active ray list. Note that the ground plane is excluded for the ray intersection check since only the stockpile surface is relevant. As illustrated in \autoref{fig: raycasting-concept}, for an active ray hit on an instance surface, the following information can be extracted:
	\begin{itemize}
		\item $(x,y,z)$, 3D coordinates of the ray hit point,
		\item $(R,G,B)$, Red/Green/Blue color value on the instance surface,
		\item $LID$, ID of the LiDAR that casts this ray,
		\item $IID$, ID of the aggregate instance this ray hits.
	\end{itemize} 
	Note that although other per-point features (such as surface normals, point colors, etc.) may also be helpful information for the instance segmentation task, the data used in this research only involve the point coordinates. The reason is that the point coordinates are the most fundamental features of aggregate stockpiles. Other features may be difficult to estimate or exhibit high uncertainty for field stockpile data. For example, the surface normals are easy to extract from the synthetic stockpiles but may be challenging to estimate on real stockpile data. Since aggregate stockpiles are not highly-structured objects, the direction of surface normals in certain regions may be confusing or uncertain, especially at the boundaries of adjacent aggregates. In such case, incorrect surface normals may bring strong bias in the instance segmentation task.
	
	\begin{figure}[!htb]
		\centering
		\begin{tikzpicture}
			\node[anchor=south west,inner sep=0] (image) at (0,0) {\includegraphics[trim=200 150 400 200, clip, width=\textwidth]{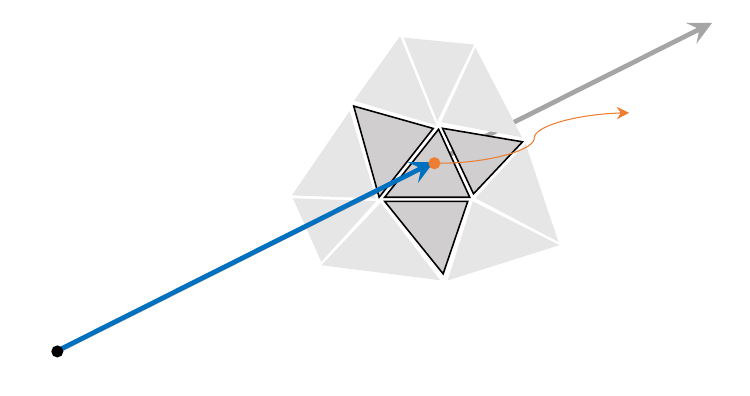}};
			\begin{scope}[
				x={(image.south east)},
				y={(image.north west)}] %
				\node[right, black] at (0.85,0.7){$\begin{aligned}
						&(x,y,z)\\
						&(R,G,B)\\
						&\text{LiDAR ID}\\
						&\text{Instance ID}
					\end{aligned}$};
			\end{scope}
		\end{tikzpicture}
		\caption{Raycasting mechanism}
		\label{fig: raycasting-concept}
	\end{figure}

	\item Step 4.4: Iterating over all multi-view LiDARs. The raycasting process was conducted for all multi-view LiDARs, with each LiDAR covering a partial view of the stockpile surface. By accumulating the extracted ray hits information for multi-view LiDARs, an all-around representation of the stockpile can be obtained.
	
	\item Step 4.5: Writing 3D point cloud data with ground-truth labels. The final synthetic data per scene is a 3D point cloud with each point having a data entry of $(x,y,z,R,G,B,LID,IDD)$. Note that how the developed synthetic data generation pipeline can generate per-point ground-truth labels. This is almost impossible by manually labeling such dense stockpile assemblies.
\end{itemize}

\begin{figure}[!htb]
	\centering
	\includegraphics[trim=200 0 200 0, clip, width=0.9\textwidth]{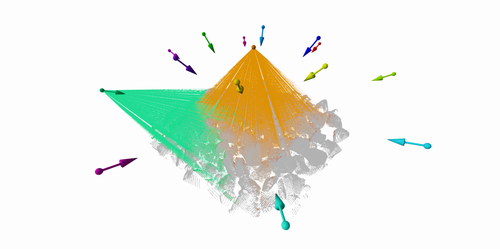}
	\caption{Point cloud coordinates by raycasting of multi-view LiDARs}
	\label{fig: raycasting-lidar}
\end{figure}

\begin{figure}[!htb]
	\centering
	\hfill
	\begin{subfigure}[b]{0.48\textwidth}
		\centering
		\includegraphics[trim=0 100 0 200, clip, width=\linewidth]{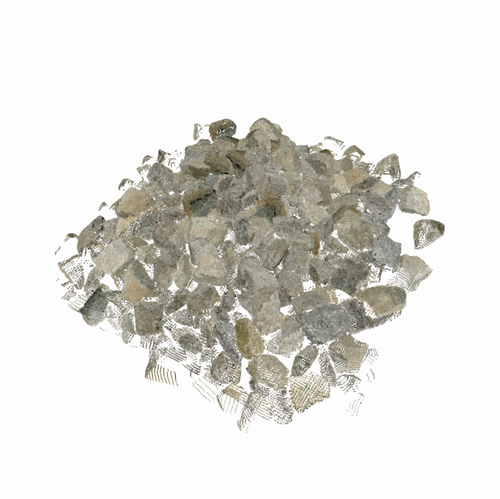}
		\caption{}
		\label{fig: stockpile-pointcloud}
	\end{subfigure}
	\hfill
	\begin{subfigure}[b]{0.48\textwidth}
		\centering
		\includegraphics[trim=0 100 0 200, clip, width=\linewidth]{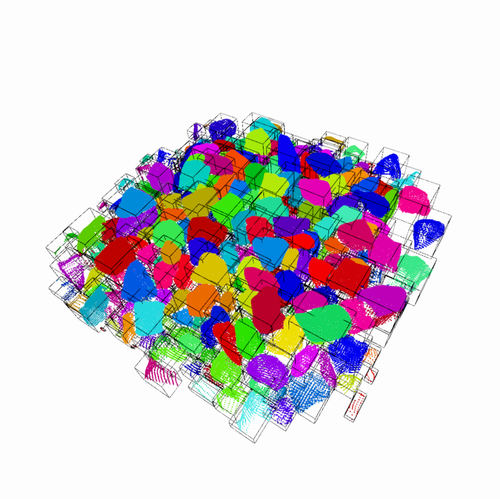}
		\caption{}
		\label{fig: stockpile-inslabel}
	\end{subfigure}
	\caption{(a) Point cloud and (b) ground-truth instance labels obtained from the raycasting step}
	\label{fig: stockplile-gt-label}
\end{figure}

As an example, the synthetically generated data of the RR4 stockpile is visualized with the ground-truth labels. First, the rays and ray hits of the $1^{st}$ and $13^{th}$ LiDAR are illustrated in \autoref{fig: raycasting-lidar}. The points hit by the $1^{st}$ LiDAR are colored in orange and those by the $13^{th}$ LiDAR are colored in cyan, while the points not visible to these two LiDARs are assigned a gray color. It can be seen that multi-view LiDARs are collaboratively capturing the different perspectives of individual aggregate surfaces. With all 15 LiDARs placed at different heights and radii, a comprehensive coverage of the whole aggregate stockpile was successfully achieved. The full point cloud of the aggregate stockpile (in \autoref{fig: stockpile-pointcloud}) well represents the simulated stockpile scene as previously presented in \autoref{fig: unity-stockpile-multiview}.

Moreover, the most important synthetic data, i.e., the per-point instance labels, are demonstrated in \autoref{fig: stockplile-gt-label}. For the example RR4 stockpile, a total of 323 aggregate instances were labeled by the synthetic data generation, with each instance displaying a different color and a black bounding box as shown in \autoref{fig: stockpile-inslabel}. Depending on the LiDAR positions, the number of ray hits of each LiDAR ranges from 13,832 to 14,496, which generates a 3D point cloud of 213,415 points. This leads to an average of 660 points per instance as a detailed shape representation.

Based on the developed synthetic data generation pipeline, scenes with various configurations were generated. The three types of stockpile configurations are: (a) RR3 stockpiles, (b) RR4 stockpiles, and (c) RR3-RR4 mix stockpiles that contain aggregate instances from both RR3 and RR4 categories. Because the sizes of the RR3 and RR4 aggregates are different, the scene control parameters need to be adjusted accordingly to ensure the multi-view cameras and LiDARs have the best coverage. The hyper-parameters used in the scene control are listed in \autoref{tab:synthetic-parameters}. 

\begin{table}[!htb]
	\centering
	\caption{Hyper-Parameters of the Stockpile Scene Control During the Synthetic Data Generation}
	\label{tab:synthetic-parameters}
	\begin{tabular}{c|c|ccc}
		\hline
		\multicolumn{2}{c|}{\textbf{Hyper-Parameters}} & \textbf{RR3 Stockpile} & \textbf{RR4 Stockpile} & \textbf{RR3-RR4 Mix Stockpile} \\ \hline
		\multirow{2}{*}{ROI}                      & $L_x\ (m)$ & \multicolumn{3}{c}{2.0}    \\ \cline{2-5} 
		& $L_z\ (m)$ & \multicolumn{3}{c}{2.0}    \\ \hline
		\multirow{3}{*}{Aggregates Instantiation} & $N_g$      & 9       & 7      & 7     \\ \cline{2-5} 
		& $L_{min}$  & \multicolumn{3}{c}{6}    \\ \cline{2-5} 
		& $L_{max}$  & \multicolumn{3}{c}{8}    \\ \hline
		\multirow{3}{*}{Multi-View Cameras}       & $N$          & \multicolumn{3}{c}{36}   \\ \cline{2-5} 
		& $H\ (m)$   & 0.5     & 1.0    & 0.8   \\ \cline{2-5} 
		& $r$          & 2.5     & 3.0    & 3.0   \\ \hline
		\multirow{7}{*}{Multi-View LiDARs}        & $N_1$      & \multicolumn{3}{c}{6}    \\ \cline{2-5} 
		& $N_2$      & \multicolumn{3}{c}{8}    \\ \cline{2-5} 
		& $H_1\ (m)$ & 0.8     & 1.5    & 1.2   \\ \cline{2-5} 
		& $H_2\ (m)$ & 0.6     & 1.0    & 0.7   \\ \cline{2-5} 
		& $r_1$      & 0.5     & 0.7    & 0.7   \\ \cline{2-5} 
		& $r_2$      & 1.3     & 1.5    & 1.5   \\ \cline{2-5} 
		& $d\ (m)$   & \multicolumn{3}{c}{0.02} \\ \hline
	\end{tabular}
\end{table}

As a result, a total of 300 stockpile scenes were generated with 100 scenes for each configuration. The average number of points per scene and the total number of instances are listed in \autoref{tab:synthetic-scenes}. These 300 stockpile scenes with in total 105,054 aggregate instances constitute the synthetic dataset that is essential for the development of the 3D instance segmentation pipeline in \cref{chapter-8}. Lastly, it is important to note that the term ``synthetic" in the context of this research means the stockpile scenes and aggregate particle arrangement are from a virtual yet realistic (i.e., emulated reality) gravity falling simulation, but every single aggregate instance used in the simulation are from high-fidelity reconstruction of the real, natural rocks. Therefore, the synthetic stockpile data can be considered as a reasonably good reproduction of real-world stockpiles.

\begin{table}[!htb]
	\centering
	\caption{Statistics of Synthetically Generated Scenes and Point Clouds}
	\label{tab:synthetic-scenes}
	\begin{tabular}{lp{0.2\linewidth}lp{0.15\linewidth}lp{0.25\linewidth}lp{0.3\linewidth}}
		\hline
		\textbf{Scene Type} & \textbf{Number of Scenes} & \textbf{Average Number of  Points} & \textbf{Total Number of Instances} \\ \hline
		RR3 Stockpile         & 100 & 202,954  & 56,486  \\
		RR4 Stockpile         & 100 & 200,808 & 23,766  \\
		RR3-RR4 Mix Stockpile & 100 & 177,600 & 24,802  \\ \hline
		Total                 & 300 & -       & 105,054 \\ \hline
	\end{tabular}
\end{table}

\section{Summary}

This chapter reviewed the successful use of synthetic datasets among low-level and high-level computer vision tasks, especially for the data-driven deep learning development. To develop the 3D analyses of aggregate stockpiles, using a synthetic dataset was deemed necessary considering the prohibitively time-consuming and error-prone 3D manual labeling process. A synthetic data generation pipeline was designed and developed, which comprises three main modules: aggregate model fabrication, aggregate stockpile assembly, and stockpile raycasting. After instantiating aggregate instances and enabling gravity falling to form stockpiles, multi-view cameras and LiDARs were programmed to automatically extract the scene information with ground-truth using the raycasting techniques. Following the pipeline, a total of 300 densely-stacked aggregate stockpiles containing 105,054 aggregates were successfully simulated based on the assembly of instances from the 3D aggregate particle library. This synthetic dataset serves as the cornerstone for the 3D instance segmentation development which will be discussed in the next chapter.

%% file: chapter08.tex
\chapter{Automated 3D Instance Segmentation of Aggregate Stockpiles} \label{chapter-8}

A 3D representation of aggregate stockpiles can be obtained from the 3D reconstruction techniques previously discussed, however, a more in-depth morphological analysis of the stockpile requires detailed information of individual aggregate particles on the stockpile surface. The automated 2D instance segmentation approach presented in \cref{chapter-5} provides a good solution for 2D stockpile images. Nevertheless, the development of a 3D instance segmentation approach for dense stockpiles remains a very challenging task. This is usually due to the lack of high-quality instance labeling dataset as well as the irregularity of 3D data representation.

This chapter first reviews the state-of-the-art advancements in computer vision regarding the 3D instance segmentation task, and analyzes the most suitable strategy for the application of dense stockpile segmentation. Then, a selected deep learning framework is implemented with necessary modifications for the automated stockpile segmentation. Based on the synthetic dataset established in \cref{chapter-7}, the framework is trained to learn the segmentation of individual aggregate instances from the stockpile. 

\section{Review of 3D Instance Segmentation Task in Computer Vision}
Similar to 2D instance segmentation, 3D instance segmentation task focuses on detecting and separating objects at the instance level, which is a much harder task than 3D object detection and semantic segmentation. This makes 3D instance segmentation a fundamental yet challenging topic in computer vision that facilitates various types of applications in autonomous driving, robotics, medical imaging, etc. \parencite{guo_deep_2020, he_deep_2021}. On the one hand, 3D data provides more comprehensive geometric and scale information than 2D images, especially in the understanding of spatial features and relations. But on the other hand, unlike 2D images represented in a pixel grid that can naturally be handled by the convolutional CNN design, the typical 3D data representation (i.e., point clouds, meshes, voxels) usually presents high unorderedness and irregularity than 2D images. To handle the challenges in 3D instance segmentation, two major categories of methods, i.e., detection-based and detection-free, were developed in the computer vision community.

\subsection{Detection-Based Methods}

Detection-based methods are essentially a two-stage approach, which first detects object proposals and then refines the proposals by generating the instance masks. These methods usually propose 3D bounding box of object instances in an explicit way. Since this type of approach imitates the mechanism of human attention by refining from a high-level perception, it is usually depicted as top-down methods.

The 3D Semantic Instance Segmentation network (3D-SIS) developed by \textcite{hou_3d-sis_2019} jointly learns the geometric and color signals from multi-view RGB-D scan data. The 2D image features are extracted using 2D CNNs and back-projected to the associated 3D voxel grid. The geometric and color features are processed by 3D convolutions and form a global semantic feature map. Then, a 3D Region Proposal Network (3D-RPN) and a 3D Region of Interest (3D-RoI) pooling layer are used to generate bounding boxes, semantic labels, and instance masks. This approach generates accurate instance predictions, but it is in fact a 2.5D method rather than true 3D method because it does not directly process the 3D data format. 

As a true 3D method that directly processes point cloud data, Generative Shape Proposal Network (GSPN) and the framework Region-based PointNet (R-PointNet) proposed by \textcite{yi_gspn_2019} generate object proposals by first predicting the shapes of a potential object. It has strong emphasis on the geometric understandings and the ``objectness'' of the proposals. The first component in this architecture is a center prediction network, which predicts the potential object centroids as the starting point. Such center prediction design is essentially a detection-based method and is later on adopted in other networks such as the Gaussian Instance Segmentation Network (GICN, \cite{liu_learning_2020}). However, designs relying on a center prediction step may have several limitations. First, the predicted centroid plays a critical role in the following instance proposal steps. In GICN, for example, the centroids it predicts are forced to be certain points in the point cloud, which does not apply in the aggregate stockpile context since most of the aggregate center will lie out of the stockpile surface. Furthermore, if the center prediction step generates less accurate predictions, the error may propagate throughout the instance segmentation process.

Another representative detection-based method is the 3D-BoNet proposed by \textcite{yang_learning_2019} which merges the two-stage method into a single-stage trainable method with two branches. 3D-BoNet learns a fixed number of 3D bounding boxes with confidence scores, and estimates per-box instance masks for object proposal. Again, the major concern of applying such bounding box oriented method in the stockpile segmentation context is it excessively relies on the correctness of bounding box prediction. Suppose the bounding box intersects with an aggregate surface and fails to encompass the instance, the resulting shape of the aggregate instance will not only be incomplete but also inaccurate.

\subsection{Detection-Free Methods}

Different from the detection-based methods, detection-free methods often learn the point-wise features and then apply clustering (or grouping) to obtain instance information.
This type of approach works in the reverse direction of human perception which first focuses on fine-grained details, therefore, it is also called bottom-up methods.

PanopticFusion network developed by \textcite{narita_panopticfusion_2019} first predicts pixel-wise panoptic labels on image frames by 2D instance segmentation network Mask R-CNN \parencite{he_mask_2017} and then integrates the labels into 3D volumetric map together with depth measurement. Similar to 3D-SIS in the detection-based category, this method is essentially a 2.5D approach that does not directly work on 3D data representation.

The Similarity Group Proposal Network (SGPN) proposed by \textcite{wang_sgpn_2018} assumes that the points belonging to the same object instance should share similar features. Based on this assumption, it learns a similarity matrix that indicates the similarity between each point pairs in the feature space. Despite this assumption may be reasonable, it was found that the similarity measure may be over-simplified such that adjacent objects of the same class are not easily separable by SGPN. Different from the datasets that contain many object categories, in the context of aggregate stockpile analysis, all instances will be of the same class, which makes SGPN less competent for our task.

Multi-scale Affinity with Sparse Convolution (MASC) proposed by \textcite{liu_masc_2019} builds upon the Submanifold Sparse Convolution Network (SSCN, \cite{graham_3d_2018}) operations to predict the semantic scores and the affinity scores between neighboring points at different scales. A clustering algorithm was used to segment points into instances based on the semantic and affinity information. This method opens a thread of follow-up research that brings great improvement in instance segmentation. For example, PointGroup \parencite{jiang_pointgroup_2020} learns a shifted coordinate space by moving points closer to its potential center and also applies a cluster algorithm on the original coordinates and shifted coordinates. A ScoreNet module is designed to judge and guide the proposal generation after the clustering step. Likewise, Dyco3D \parencite{he_dyco3d_2021} further extends PointGroup by incorporating dynamic convolution operations and transformer layers to better capture the shape context around each point.  OccuSeg \parencite{han_occuseg_2020} tackles the instance segmentation as a multi-task learning task to produce both occupancy signals and spatial features, based on which an object occupancy-aware segmentation approach is applied. The occupancy signal represents the number of voxels occupied by each instance thus improves the robustness of the clustering step. 

\section{Deep Learning Framework for Automated 3D Stockpile Segmentation}

\subsection{Synthetic Dataset for Stockpile Segmentation} 

As previously described in \cref{chapter-7}, the synthetic dataset was designed for the learning of stockpile segmentation task. Each synthetic scene is the 3D point cloud of a stockpile formed from different sizes of aggregates, prepared with ground-truth instance labels at each point. The entire dataset was divided into train and test splits with test set being an independent set never used during training. The layout of the synthetic dataset used in the stockpile segmentation task is listed in \autoref{tab:synthetic-split}.

\begin{table}[!htb]
	\centering
	\caption{Number of Scenes in the Synthetic Dataset Used in Train and Test}
	\label{tab:synthetic-split}
	\begin{tabular}{lll}
		\hline
		Scene Type            & Number of Scenes in Train Split & Number of Scenes in Test Split \\ \hline
		RR3 Stockpile         & 90                              & 10                             \\
		RR4 Stockpile         & 90                              & 10                             \\
		RR3-RR4 Mix Stockpile & 90                              & 10                             \\ \hline
		Total                 & 270                             & 30                             \\ \hline
	\end{tabular}
\end{table}

\subsection{Point Grouping Framework with Shifted Coordinates}

Based on the review of 3D instance segmentation research in computer vision, it was concluded that detection-free methods are more suitable for the task of aggregate stockpile segmentation. The most important reason is the salience of the aggregate stockpile structure: aggregate stockpiles are point clouds with very densely-stacked instances. The 3D instance segmentation datasets in computer vision are mostly available from autonomous driving \parencite{geiger_vision_2013} and indoor environments \parencite{armeni_3d_2016, mccormac_scenenet_2017}, where the separation among object instances is considerably higher than that in a stockpile. Therefore, detection-based methods that strongly rely on the precision of predicted bounding boxes are likely to fail or produce inaccurate results on the densely-stacked structure. This observation also agrees with the nature of human's top-down perception, i.e., can easily distinguish objects sparsely separated but fail to disentangle small pieces from a pile. Detection-free methods, on the other hand, follow a bottom-up strategy that builds up high-level segmentation from fine-grained details and may better handle the stockpile structure. As a result, a state-of-the-art network, PointGroup \parencite{jiang_pointgroup_2020}, was selected, implemented, and customized for the 3D stockpile segmentation task.

\begin{figure}[!htb]
	\centering
	\includegraphics[trim=0 100 0 100, clip, width=\linewidth]{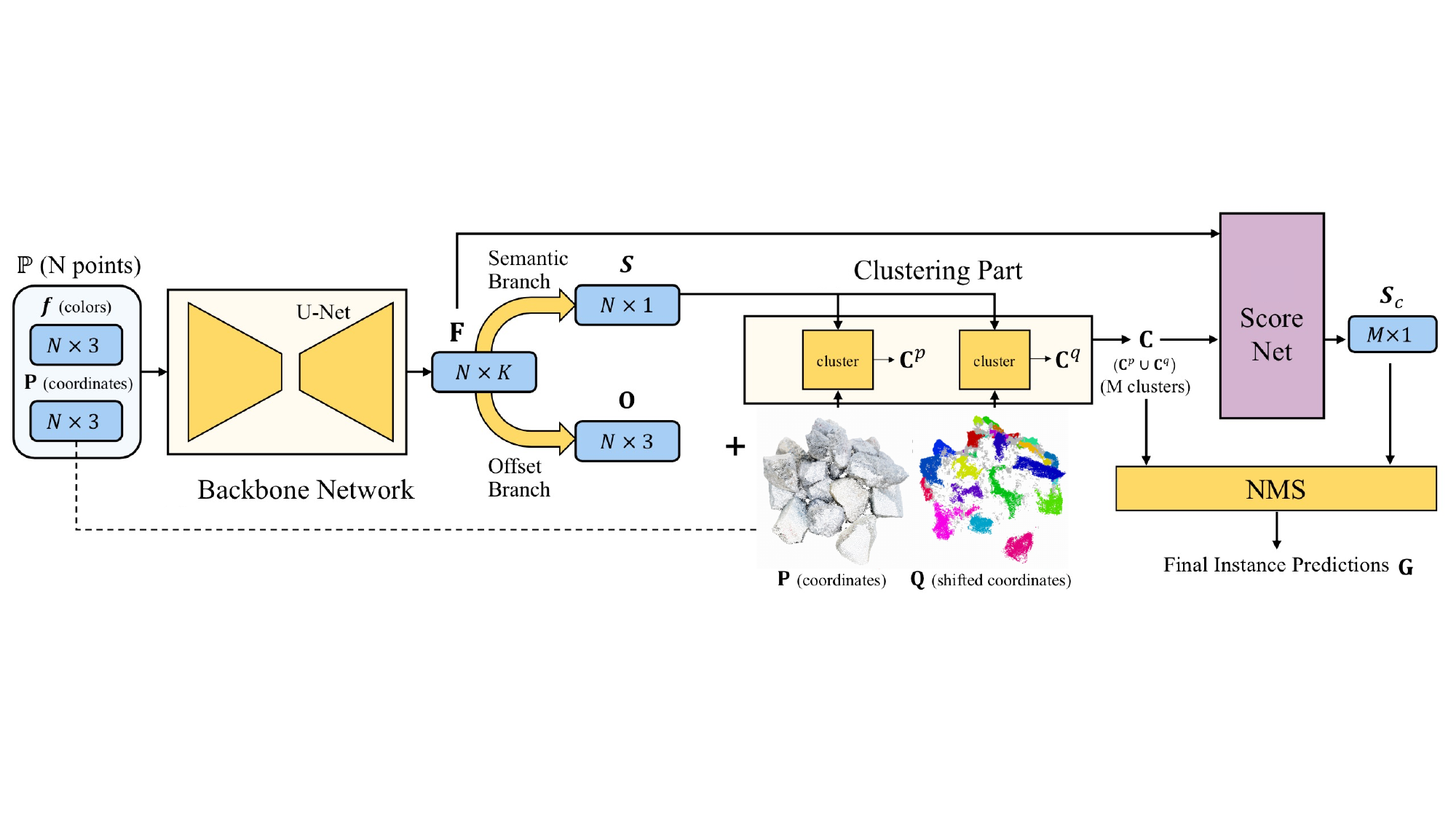}
	\caption{PointGroup architecture for instance segmentation}
	\label{fig: pointgroup}
\end{figure}

The overall architecture of the PointGroup network is illustrated in \autoref{fig: pointgroup}. The network consists of three main components: feature extraction by a backbone network, point clustering on dual coordinate sets, and cluster scoring. The key design in the network is to learn per-point offset vectors to shift the original coordinates into a more compact coordinate space, such that the clustering process will be more robust.

The backbone feature extraction network follows a U-Net structure \parencite{ronneberger_u-net_2015} with Submanifold Sparse Convolution (SSC, \cite{graham_3d_2018}) layers. The input of the network is a point cloud with fixed data dimension, denoted as $\mathbb{P}=\{p_i=(x_i, y_i, z_i)\in \mathbb{R}^3|\ i\in\ \{1,...,N\}\}$ where $N$ is the number of input points. The original PointGroup architecture uses point colors as additional input features, but in the context of aggregate stockpile segmentation, it was decided the learning should be based on point coordinates only. This is because unlike the datasets in autonomous driving and indoor environment, aggregates can have high in-class variation in terms of the particle color. Aggregates from different geological origins and experiencing various weathering conditions may have very distinct colors. Considering the instance segmentation of aggregate stockpile is theoretically plausible by exploring the void space between instances, the geometry information (i.e., point coordinates) is expected to be the most important input. Therefore, the network was customized to only take point coordinates as the input. After feature extraction, the geometry information of the stockpile is encoded as a per-point feature matrix $\textbf{F}={f_i}\in \mathbb{R}^{N\times K}$, where $K$ is the number of feature channels.

The per-point offset vectors are then predicted from the feature matrix to shift points towards the centroid of its potential instance, as illustrated in the offset branch. Note that the original PointGroup architecture design uses two branches, one for semantic segmentation and the other for predicting the per-point offset vector. In the context of aggregate stockpile segmentation, the semantic branch was removed since the point cloud is expected to contain only single-class aggregate instances. The offset branch predicts per-point offset vector $o_i = (\Delta x_i , \Delta y_i , \Delta z_i)$, and the shifted coordinates space can be obtained by applying the per-point offset to the original coordinates, denoted as $\mathbb{Q}=\{q_i=(x_i+\Delta x_i, y_i+\Delta y_i, z_i+ \Delta z_i)\in \mathbb{R}^3|\ i\in\ \{1,...,N\}\}$. The shifted coordinates was found to be more efficient for clustering and grouping since the points have now been re-arranged in an instance-aware pattern. Based on the original coordinates space $\mathbb{P}$ and the shifted coordinates space $\mathbb{Q}$, a clustering step is performed to generate instance proposals. Since the semantic branch is removed from the design, the clustering step is also customized to be a coordinate-based clustering. The clustering algorithm follows a breath-first search mechanism by grouping adjacent points within a given radius. The clustering radius $r$ is a hyper-parameter that influences the clustering performance. During experiments, it was found that the shifted coordinates are more effective for generating instance proposals. This may be explained by the nature of a dense structure such as an aggregate stockpile, where the segmentation on a more compact shifted space is easier than segmentation on the uniformly-spaced original representation. This observation agrees with the findings in PointGroup development that shifted coordinates is more suitable for separating nearby objects. As a result, the clustered instance proposals from both the original coordinates $\mathbb{P}$ and the shifted coordinates space $\mathbb{Q}$ are denoted as $\textbf{C}^{\mathbb{P}}$ and $\textbf{C}^{\mathbb{Q}}$, respectively. 

The raw instance proposals may contain many overlapped prediction duplicates as well as low-confidence predictions, therefore the ScoreNet module is used to rank the clusters $\textbf{C}^{\mathbb{P}}\cup \textbf{C}^{\mathbb{Q}}$. The ScoreNet is a sub-network that applies another U-Net structure on the per-point coordinates and feature vectors. As a final step, a 3D version of Non-Maximum Suppression (NMS, \cite{hosang_learning_2017}) is applied to condense highly-overlapped instance proposals by selecting the proposal with highest confidence score among overlapping proposals.

\section{Evaluation of Stockpile Segmentation Performance}

\begin{figure}[!htb]
	\centering
	\begin{tabular}{C{2cm}C{6cm}C{6cm}}
		Stockpile Type & Original Coordinates $\mathbb{P}$ & Shifted Coordinates $\mathbb{Q}$  \\
		RR3 & 
		\includegraphics[trim=0 50 0 200, clip, height=5cm]{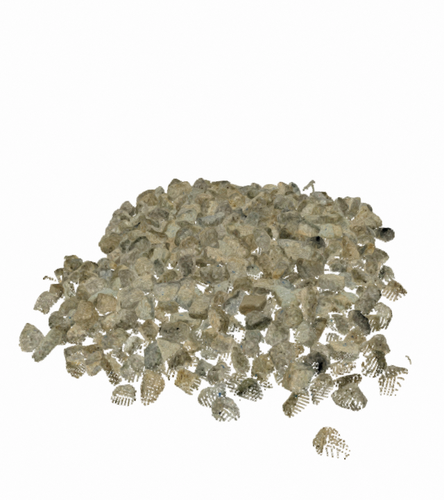} &
		\includegraphics[trim=0 50 0 200, clip, height=5cm]{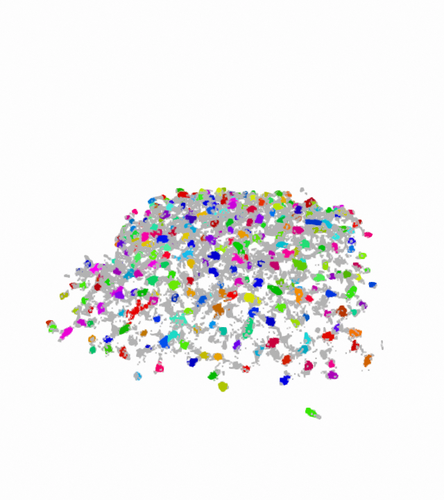} \\
		RR4 & 
		\includegraphics[trim=0 50 0 200, clip, height=5cm]{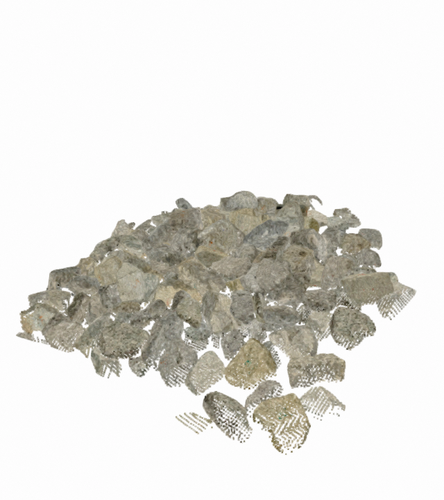} &
		\includegraphics[trim=0 50 0 200, clip, height=5cm]{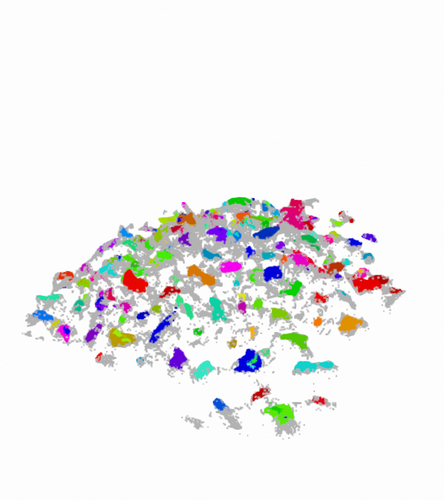} \\
		RR3-RR4 Mix & 
		\includegraphics[trim=0 50 0 200, clip, height=5cm]{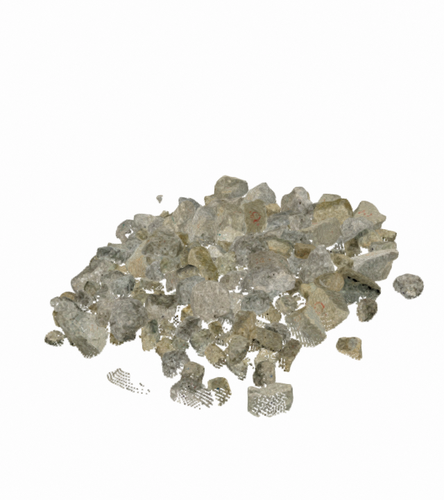} &
		\includegraphics[trim=0 50 0 200, clip, height=5cm]{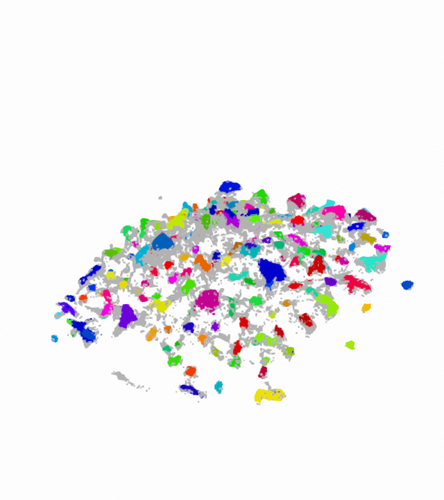} \\
	\end{tabular}
	\caption{Original coordinates $\mathbb{P}$ and shifted coordinates $\mathbb{Q}$ by applying the per-point offset}
	\label{fig:shift-results}
\end{figure}

\begin{figure}[!htb]
	\centering
	\begin{tabular}{C{1.5cm}C{4.5cm}C{4.5cm}C{4.5cm}}
		Stockpile Type & Input Point Cloud & Ground-Truth Instances  & Segmented Instances \\
		RR3 & 
		\includegraphics[trim=0 50 0 200, clip, height=4cm]{c8/RR3_090/raw} &
		\includegraphics[trim=0 50 0 200, clip, height=4cm]{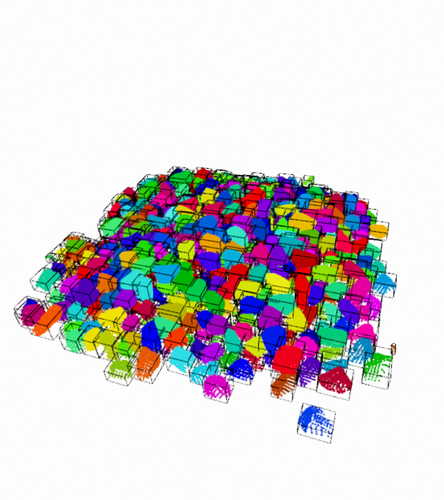} &
		\includegraphics[trim=0 50 0 200, clip, height=4cm]{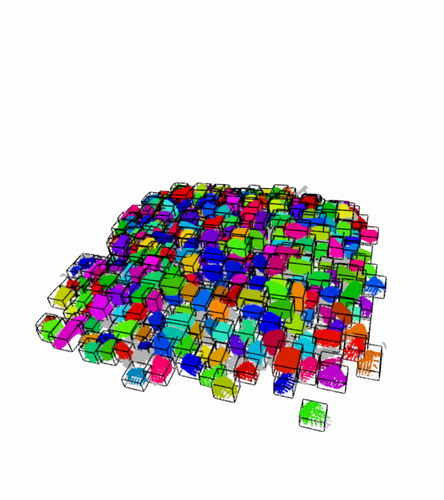} \\
		RR4 & 
		\includegraphics[trim=0 50 0 200, clip, height=4cm]{c8/RR4_090/raw} &
		\includegraphics[trim=0 50 0 200, clip, height=4cm]{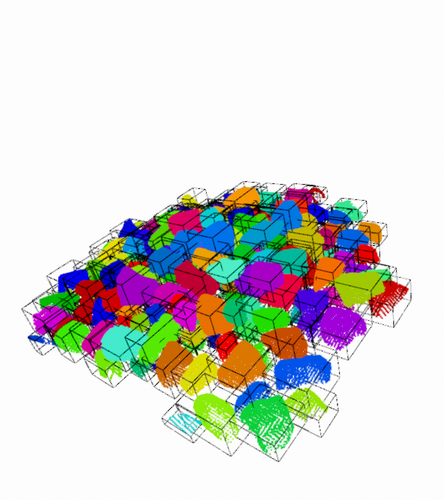} &
		\includegraphics[trim=0 50 0 200, clip, height=4cm]{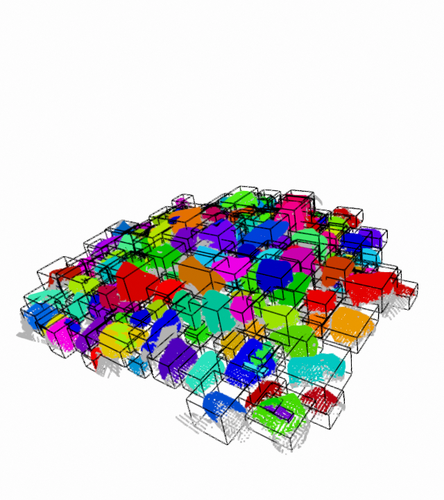} \\
		RR3-RR4 Mix & 
		\includegraphics[trim=0 50 0 200, clip, height=4cm]{c8/RR3_RR4_Mix_090/raw} &
		\includegraphics[trim=0 50 0 200, clip, height=4cm]{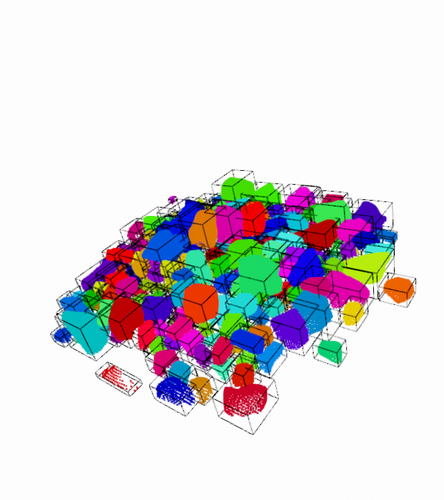} &
		\includegraphics[trim=0 50 0 200, clip, height=4cm]{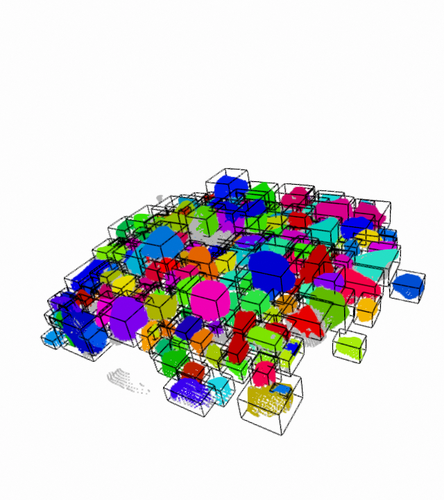} \\
	\end{tabular}
	\caption{Comparisons of segmentation results and ground-truth instances}
	\label{fig:gt-results}
\end{figure}

The network was trained on the synthetic dataset, and the performance of the instance segmentation was evaluated on the test set of the dataset. Qualitative results are presented in \autoref{fig:shift-results} and \autoref{fig:gt-results}. First, the original and shifted coordinates space are visualized to indicate the effectiveness of learning the per-point offset. One example is given for each of the stockpile scene type in the dataset. As shown in \autoref{fig:shift-results}, the network successfully learned the per-point offset prediction by showing reasonable clustering of the points in the shifted coordinates. Note that each different color in the shifted coordinates $\mathbb{Q}$ represents the clustered points belonging to individual instances. The hyper-parameter, clustering radius $r$ was found to be 0.008 for providing the best performance on the dataset. With the more compact clustered coordinates, the generation of instance proposals is expected to be more robust and reasonable. It is also observed that across different stockpile scene types in the test set, the network demonstrates consistent effectiveness and performance in predicting the per-point offset.
 
Next, the segmentation results were compared with the ground-truth labels in test set to qualitatively evaluate the segmentation effect, as shown in \autoref{fig:gt-results}. The final instance proposals are visualized with enclosing bounding boxes to better show the location of the segmented instances. It can be seen that the segmentation results are reasonably good compared to the ground-truth instances, with most of the aggregate particles identified and successfully segmented. Although some over-segmentation and under-segmentation effects can be observed from the segmentation results, it is considered an efficient and high-quality segmentation surpassing human vision's capability of handling such dense structures.
 
Quantitative measurement was also conducted on the quality of segmentation. Before introducing the metrics used for stockpile segmentation, a brief overview is presented for the popular evaluation standard used in machine learning and computer vision research \parencite{powers_evaluation_2011}. By comparing the predictions of a machine learning model and the ground-truth, the results can be categorized into four groups: True Positives (TP), False Positives (FP), True Negatives (TN), and False Negatives (FN). The positive/negative part represents the prediction results, while the true/false part indicates the correctness of the predictions when compared with ground-truth. For example, TP means a sample is predicted as positive and the prediction is true, i.e., the prediction is consistent with the ground truth. For 2D and 3D instance segmentation tasks that do not have clear true/false correspondences, the definition of the ``match'' between a prediction and a ground-truth commonly follows the Intersection over Union (IoU) concept. 2D IoU for instance segmentation is the number of pixels in common between the segmented and ground-truth masks divided by the total number of pixels present across both masks, as previously given in \autoref{eqn: 5-3}. Similar to 2D IoU, 3D IoU for point cloud data is commonly defined by the intersection and union volumes between two axis-aligned bounding boxes of the instance \parencite{zhou_iou_2019}:

\begin{equation} \label{eqn: 8-1}
	\mathbf{IoU_{3D}}(\%)=\frac{V_{\text{Segmented}}\cap V_{\text{Ground-Truth}}}{V_{\text{Segmented}}\cup V_{\text{Ground-Truth}}}
\end{equation}

Therefore, by setting a IoU threshold, the correspondence between prediction and ground-truth can be determined thus the TP, FP, TN, FN can be defined. Typically, precision and recall metrics are used to measure the performance of model, as shown in \autoref{eqn: 8-2}. Next, to capture the precision-recall behavior at different threshold IoU values, a precision-recall curve is usually generated by varying IoU thresholds, and an Average Precision (AP) is defined as the area integral under the precision-recall curve.
\begin{equation} \label{eqn: 8-2}
	Precision=\frac{TP}{TP+FP}, \ Recall=\frac{TP}{TP+FN}
\end{equation}

\begin{figure}[!htb]
	\centering
	\includegraphics[trim=0 150 80 0, clip, width=\linewidth]{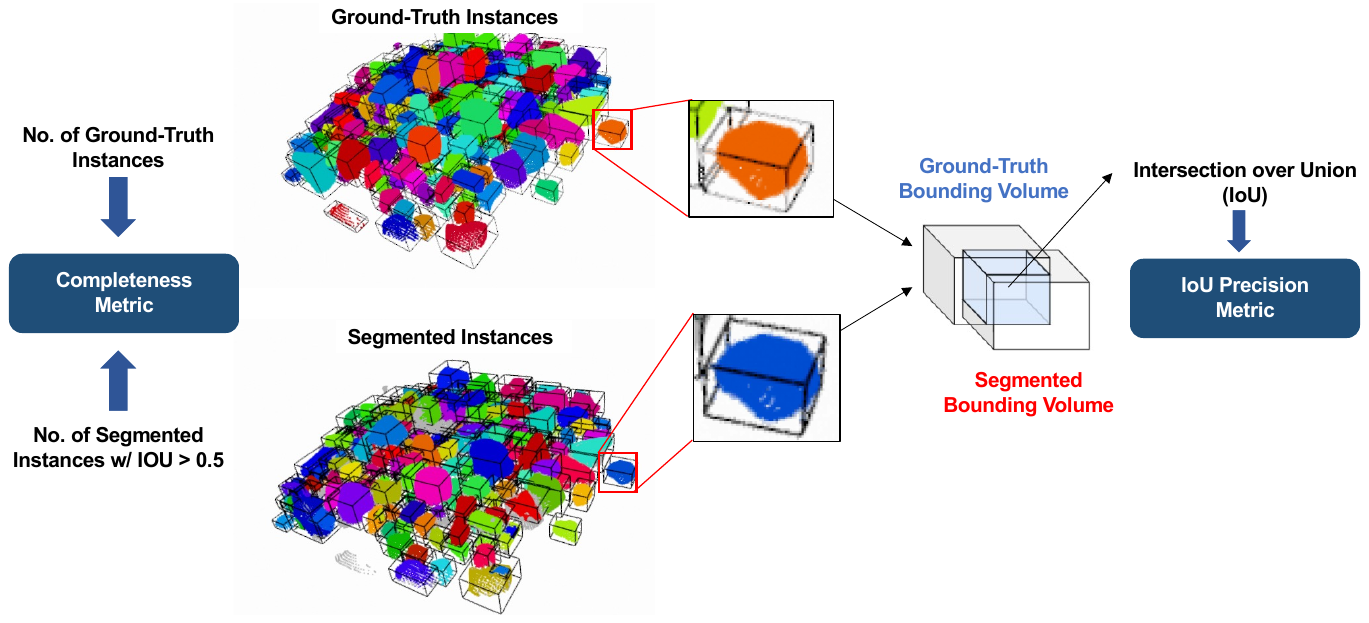}
	\caption{Completeness and IoU precision metrics used to compare the segmentation instances with the ground-truth labels}
	\label{fig: seg-metrics}
\end{figure}

In the context of aggregate stockpile segmentation task, however, the metrics are customized based on the standard metrics to better indicate the most relevant performance in the context of stockpile segmentation. First, the IoU threshold is fixed at 0.5 to determine the prediction and ground-truth correspondence. At this threshold, the ``completeness'' is defined as the ratio between the number of segmented instances (TP) and the number of ground-truth instances (TP+FN). This ratio describes the percentage of aggregate instances correctly detected as compared to the ground-truth labeling. In fact, the completeness metric herein is identical to the standard recall metric but is renamed to distinguish in the context of stockpile segmentation. Since a fixed IoU threshold is used, the AP concept no longer applies. However, a metric is needed to further indicate how closely the segmented instances align with the ground-truth, even if they all have IoUs beyond the threshold. Therefore, a IoU precision metric is defined as the per-instance 3D IoU score that calculates the percent overlap between the segmentation and the corresponding ground-truth. Then, for the entire stockpile, an IoU average precision (IoU AP) metric can be calculated that measures the overall volumetric similarity between the segmented and ground-truth instances. The definition of the two metrics are given in \autoref{eqn: 8-3} and the demonstrations are shown in \autoref{fig: seg-metrics}. Note that the performance of the instance segmentation network is evaluated using these newly-defined metrics to provide more practical interpretation of the aggregate stockpile segmentation task, and will be further evaluated against ground-truth morphological properties in \cref{chapter-10}. These metrics were selected over the standard metrics mainly because they are better linked to the next shape completion and field validation tasks.

\begin{equation} \label{eqn: 8-3}
	Completeness=\frac{TP}{TP+FN},\ IoU AP = \frac{\sum_{i=1,...,N}IoU_{3D,i}}{N}
\end{equation}

Following the completeness and IoU AP metrics, the network performance was evaluated on 30 stockpiles from the test set. As listed in \autoref{tab:segmentation-results}, the average completeness and IoU AP values are $78.4\%$ and $82.2\%$, respectively, which are considered high for the dense stockpile segmentation task. The average completeness value shows that above $75\%$ of aggregates can be successfully identified as compared to the ground truth, with individual instances segmented at a relatively good IoU AP of $82\%$ on average.  Note that the completion metric also indicates there were about $20\%$ of aggregates not segmented. In addition to the fact that the network tends to only segment instances with high confidence, it is critical to understand that the test set is also from synthetic data generation. Under the synthetic setting, the ground-truth labels were generated in an omniscient and omnipotent way. Namely, even for instances that are deeply occluded from the surrounding aggregates and mostly buried in the stockpile, the synthetic raycasting is able to obtain its ground-truth label. Such labeling is not expected to be plausible for a labeling process based on human perception. Hence, some of non-segmented points are likely to not be recognized as a true instance by human vision either. Overall, the network demonstrates good performance on the stockpile segmentation task. Moreover, the standard deviation values for completeness and precision are $6.3\%$ and $4.8\%$ respectively, which implies good generality and robustness of the network among different stockpile scene types. 

\begin{table}[!htb]
	\centering
	\caption{Completeness and IoU AP of the Instance Segmentation Results on Test Set}
	\label{tab:segmentation-results}
	\begin{tabular}{C{2cm}C{2cm}C{2cm}C{3cm}C{3cm}C{2cm}}
		\hline
		\textbf{Scene Type} &
		\textbf{Stockpile ID} &
		\textbf{Number of Segmented Instances (with IoU > 0.5)} &
		\textbf{Number of Ground-Truth Instances} &
		\textbf{Completeness (\%)} &
		\textbf{IoU AP (\%)} \\ \hline
		\multirow{10}{2cm}{\textbf{RR3}}         & 1  & 407 & 564       & 72.2 & 80.5 \\
		& 2  & 410 & 563       & 72.8 & 79.4 \\
		& 3  & 420 & 559       & 75.1 & 82.1 \\
		& 4  & 386 & 485       & 79.6 & 83.6 \\
		& 5  & 404 & 564       & 71.6 & 72.5 \\
		& 6  & 399 & 562       & 71.0 & 74.0 \\
		& 7  & 392 & 486       & 80.7 & 78.9 \\
		& 8  & 402 & 562       & 71.5 & 84.5 \\
		& 9  & 391 & 486       & 80.5 & 82.3 \\
		& 10 & 405 & 561       & 72.2 & 76.7 \\ \hline
		\multirow{10}{2cm}{\textbf{RR4}}         & 1  & 204 & 240       & 85.0 & 85.6 \\
		& 2  & 192 & 209       & 91.9 & 87.9 \\
		& 3  & 184 & 208       & 88.5 & 88.0 \\
		& 4  & 196 & 245       & 80.0 & 82.4 \\
		& 5  & 182 & 210       & 86.7 & 89.8 \\
		& 6  & 174 & 212       & 82.1 & 88.8 \\
		& 7  & 191 & 211       & 90.5 & 90.5 \\
		& 8  & 180 & 239       & 75.3 & 84.5 \\
		& 9  & 193 & 237       & 81.4 & 87.4 \\
		& 10 & 213 & 240       & 88.8 & 89.6 \\ \hline
		\multirow{10}{2cm}{\textbf{RR3-RR4 Mix}} & 1  & 184 & 251       & 73.3 & 80.1 \\
		& 2  & 196 & 249       & 78.7 & 79.3 \\
		& 3  & 163 & 214       & 76.2 & 78.7 \\
		& 4  & 190 & 251       & 75.7 & 76.5 \\
		& 5  & 172 & 214       & 80.4 & 82.4 \\
		& 6  & 187 & 251       & 74.5 & 80.5 \\
		& 7  & 198 & 251       & 78.9 & 79.5 \\
		& 8  & 189 & 250       & 75.6 & 83.1 \\
		& 9  & 148 & 214       & 69.2 & 75.6 \\
		& 10 & 184 & 250       & 73.6 & 80.3 \\ \hline
		&    &     & Average   & 78.4 & 82.2 \\
		&    &     & Deviation & 6.3  & 4.8 
	\end{tabular}
\end{table}
\clearpage

\section{Summary}

This chapter reviewed the state-of-the-art advancements in computer vision regarding the 3D instance segmentation task and selected the most suitable strategy for the application of dense stockpile segmentation. A state-of-the-art deep learning framework was implemented with necessary modifications for automated stockpile segmentation and trained on the synthetic dataset. Based on the qualitative and quantitative evaluation results, the network demonstrated good performance on segmenting individual aggregate instances from dense stockpiles with considerably high completeness and precision. A more realistic evaluation of the 3D instance segmentation network would be conducted on field stockpiles, which will be presented in the next two chapters upon integration with the 3D shape completion component.

%% file: chapter09.tex
\chapter{3D Aggregate Shape Completion by Learning Partial-Complete Shape Pairs} \label{chapter-9}

Unlike in 2D segmentation where each segmented instance is a valid view of the aggregate, the results from 3D instance segmentation are partial shapes that contain missing parts not visible from any of the viewing angle. Although morphological analysis could still be performed on the incomplete shapes, it is believed that a 3D aggregate shape completion step would be beneficial towards understanding the potential shape of the underlying part based on partial observations.

This chapter first reviews the current research developments of 3D shape completion in the computer vision domain and selects the state-of-the-art strategy that is applicable to learning irregular aggregate shapes. Partial and complete shape pairs are then generated from the 3D aggregate particle library based on varying-visibility and varying-view raycasting techniques. The selected deep learning framework is implemented and trained on the partial-complete shape pairs to learn the shape completion of aggregates. Finally, the shape completion framework is evaluated on several unseen aggregate shapes for its robustness and reliability.

\section{Review of 3D Shape Completion Task in Computer Vision}

The 3D shape completion task in computer vision mainly involves three lines of research approaches: geometry primitives-based, template matching-based, and deep-learning based \parencite{berger_state_2014, han_high-resolution_2017}. The geometry primitive-based methods usually employ hand-crafted features for specific shape categories. \textcite{schnabel_completion_2009} developed a reconstruction approach by detecting primitive shapes (i.e., planes, cylinders, etc.) on the incomplete point cloud and using them as a guidance to fill large gaps in the incomplete shape. The completion results were suitable for use in a Computer-Aided Design (CAD) system, which indicates the complete shapes tend to follow a manufactured shape instead of a natural shape such as rocks. Similarly, \textcite{lafarge_surface_2013} proposed an approach to detect and resample the structural components such as planes to guide the completion process. These types of methods are more suitable for objects with regularized structures rather than natural random shapes.

Template matching-based approaches perform nearest neighbor search in a shape database and attempt to deform and fit the template with the incomplete input shape. \textcite{pauly_example-based_2005} presented an approach to retrieve a suitable shape template from a database, warp the template to conform with the input, and consistently blend the warped models to obtain the final shape. Such template-based methods are often limited by the shape prior provided by the database. For inputs with complicated structures, the best match template in the database may still deviate greatly from the ideal complete shape. In the context of aggregate shape completion, it is even more complicated since aggregates are natural shapes formed from stochastic process.

Deep learning-based methods, in contrast, focus on learning the abstract shape features and capturing the global and local shape context rather than fitting the shapes with certain prior such as primitives or templates. Point Completion Network (PCN) proposed by \textcite{yuan_pcn_2018} is a pioneering work that directly learns on point clouds without any structural assumption or annotation about the unseen shape. It handles the shape completion following an encoder-decoder approach, where the partial shape is condensed into a high-dimensional feature vector by an encoder, and the decoder generates fine-grained completion by enriching the feature space. Such an encoder-decoder design has proven to be efficient and inspires many follow-up works. Point Fractal Network (PF-Net) developed by \textcite{huang_pf-net_2020} designed a multi-resolution encoder and decoder to learn the shape features at different scales, which recovers the missing regions while preserving the partial input. \textcite{wen_pmp-net_2021} proposed Point Moving Path Network (PMP-Net) that provides a new perspective of treating shape completion as a dynamic deformation process. Each point is moved to complete the point cloud while the total distance of point movement is optimized. Different from modeling the point moving process, SnowflakeNet developed by \textcite{xiang_snowflakenet_2021} models the shape completion as a snowflake-like growth of points in space. The SnowflakeNet is able to capture the local and global structure characteristics as well as predict geometries with fine details.

\section{Partial-Complete Aggregate Shape Pairs from Varying-Visibility and Varying-View Raycasting}

To serve as the dataset for learning a 3D shape completion task, aggregate shapes from partial observations associated with their corresponding ground-truth complete views should be generated and learned in pairs. Establishing such dataset is usually challenging due to the fact that it is difficult to obtain partial views and complete views for an aggregate at the same time. 

A simplified approach could be randomly removing parts from the complete aggregate models thus generating incomplete views of the shape. However, this approach is likely to suffer from the following issues. First, with point cloud being an unordered and irregular data format, randomly removing points by index may result in inconsistent effect of removal, i.e. the removed points could be clustering around a certain region or randomly distributed on the original surface. The former emulates the missing parts of partial observations, but the latter merely leads to a nearly uniform downsampling of the complete shape without missing parts. This limitation could be addressed by intersecting certain shape primitives (e.g., sphere, cylinder, etc.) with the complete aggregate models. But even with this approach, the missing regions of the partial shapes are expected to have many artifacts, such as very unnatural cut along the shape boundaries.

A more realistic approach was developed by further investigating the cause of partial observations of aggregate shapes. During the reconstruction of an aggregate stockpile, multi-view sensors (i.e., cameras, LiDARs) are commonly used to observe the stockpile surface. Individual aggregates on the stockpile surface may be visible to several sensors simultaneously from different viewing angles. However, the sensors can only occupy the open space around the stockpile with viewing angles from the other side of the stockpile being missing observations. Based on this fact, the proposed approach was to simulate the sensing process by varying visibility and varying view and generate realistic partial views that are possible in a real observation.

\subsection{Configuration of Multi-View Sensors}

\begin{figure}[!htb]
	\centering
	\hfill
	\begin{subfigure}[b]{0.65\textwidth}
		\centering
		\includegraphics[trim=0 0 500 300, clip, height=5cm]{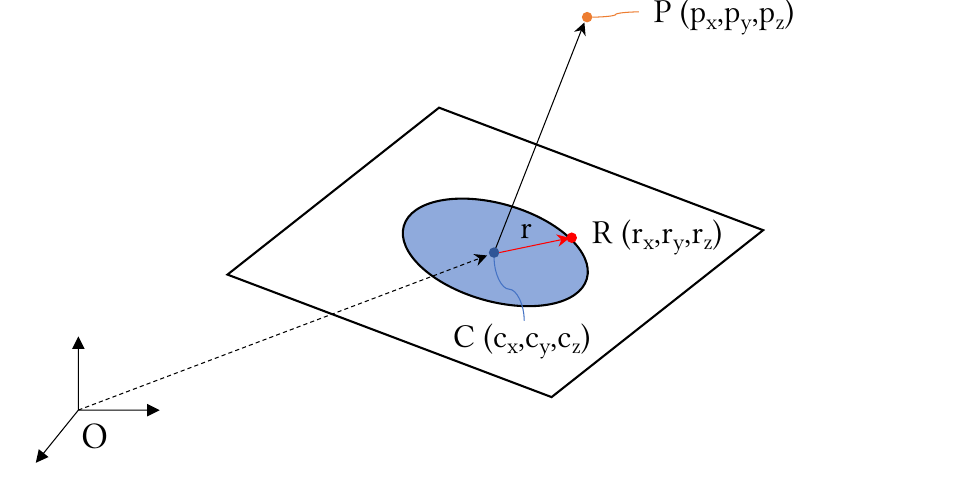}
		\caption{}
		\label{fig: raydisk-a}
	\end{subfigure}
	\hfill
	\begin{subfigure}[b]{0.3\textwidth}
		\centering
		\includegraphics[trim=800 0 0 380, clip, height=5cm]{c9/raydisk}
		\caption{}
		\label{fig: raydisk-b}
	\end{subfigure}
	\hfill
	\caption{(a) Coordinate system of the raycasting space and (b) ray endpoints on the disk circumference}
	\label{fig: raydisk}
\end{figure}

The aggregate models in the 3D aggregate particle library were placed individually in 3D space. To extract the point cloud representation of the aggregate shape, the multi-view sensors were configured as virtual LiDAR sensors with raycasting capability. Each sensor was specifically programmed as a LiDAR that can project a disk of rays to the plane perpendicular to its viewing angle. Suppose the sensor is positioned at $P=(p_x, p_y, p_z)$, the centroid of the aggregate model is at $C=(c_x, c_y, c_z)$, and an arbitrary ray endpoint on the disk circumference is at $R=(r_x, r_y, r_z)$. The sensor should cast rays in a ring pattern that can form a disk of radius $r$, as illustrated in \autoref{fig: raydisk-a}. 

The geometry problem is then to find $\overrightarrow{OR}$ by given $\overrightarrow{OP}$, $\overrightarrow{OC}$ and $\lVert\overrightarrow{CR}\rVert$. First, the vector $\overrightarrow{CR}$ was decomposed in orthogonal directions $\vec{u}$ and $\vec{v}$, as demonstrated in the plane view in \autoref{fig: raydisk-b}. Then, a parametric representation of any arbitrary point on the disk circumference can be expressed in \autoref{eqn: raydisk-1}. 
\begin{equation} \label{eqn: raydisk-1}
	\overrightarrow{CR}(\theta) =rcos(\theta)\cdot \vec{u} + rsin(\theta)\cdot \vec{v} 
\end{equation}

Accordingly, the coordinate of point $R$ can be solved by \autoref{eqn: raydisk-2}. 
\begin{equation} \label{eqn: raydisk-2}
	\begin{aligned}
		\overrightarrow{OR}(\theta) &=\overrightarrow{OC} + \overrightarrow{CR}(\theta)\\
		&=(c_x, c_y, c_z)+rcos(\theta)\cdot \vec{u} + rsin(\theta)\cdot \vec{v} 
	\end{aligned}
\end{equation}

The problem is now simplified as finding a valid orthonormal basis $\vec{u}$ and $\vec{v}$ on the disk plane. A general approach is to find two arbitrary linear independent (i.e., not co-linear) vectors on the plane, and apply Gram-Schmidt orthonormalization \parencite{beilina_numerical_2017} to construct an orthonormal basis. Under the above setting, however, the process can be further simplified since we know the normal of the plane and thus the plane equation.

The normal of the disk plane is the vector $\vec{n}=\overrightarrow{CP}=(p_x-c_x, p_y-c_y, p_z-c_z)=(n_x, n_y, n_z)$. Therefore, the disk plane passing point $C=(c_x, c_y, c_z)$ with a normal $\vec{n}=(n_x, n_y, n_z)$ is expressed by \autoref{eqn: raydisk-3}.
\begin{equation} \label{eqn: raydisk-3}
	n_x(x-c_x)+n_y(y-c_y)+n_z(z-c_x)=0
\end{equation}

By first finding an arbitrary vector $\vec{u}$ on the plane (e.g., by forcing two fields $u_x=u_y=0$ and solving for $u_z$), the other component $\vec{v}$ in the orthonormal basis can be found by the cross product of $\vec{u}$ and $\vec{n}$. After normalization, the vectors $\vec{u}$ and $\vec{v}$ form the orthonormal basis and can be substituted into \autoref{eqn: raydisk-2} to obtain the ray endpoints on the disk.

Next, ray endpoints are uniformly generated in multiple rings with different radii $r$. One valid approach is to generate an equal number of endpoints on each ring. However, this may lead to non-uniform spacing between the ray endpoints of inner rings and outer rings, since the ray spacing is proportional to the ring radius. To address this issue, an improved approach was used that maintain a constant arc spacing $ar$ between ray endpoints among different rings. For a ring with radius $r$, the central angle between two adjacent ray endpoints can be calculated from the arc length equation:
\begin{equation} 
	ar=r\cdot \Delta\theta \rightarrow \Delta\theta=\frac{ar}{r}
\end{equation}

Then, the central angle increment can be determined by calculating the number of ray endpoints on the ring:
\begin{equation} 
	\Delta\hat{\theta}=\frac{2\pi}{\lfloor 2\pi/\Delta\theta\rfloor}
\end{equation}

\begin{figure}[!htb]
	\centering
	\hfill
	\begin{subfigure}[b]{0.45\textwidth}
		\centering
		\includegraphics[trim=200 200 200 0, clip, width=\textwidth]{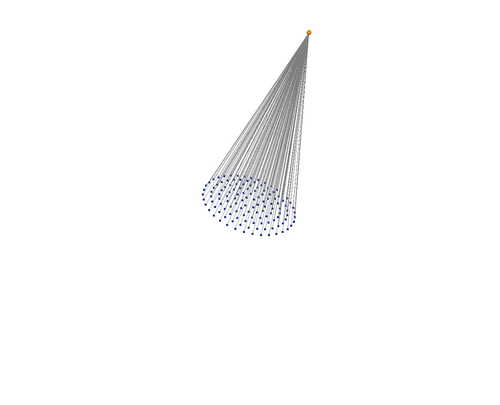}
		\caption{}
		\label{fig: lidar-raydisk-a}
	\end{subfigure}
	\hfill
	\begin{subfigure}[b]{0.45\textwidth}
		\centering
		\includegraphics[trim=200 200 200 0, clip, width=\textwidth]{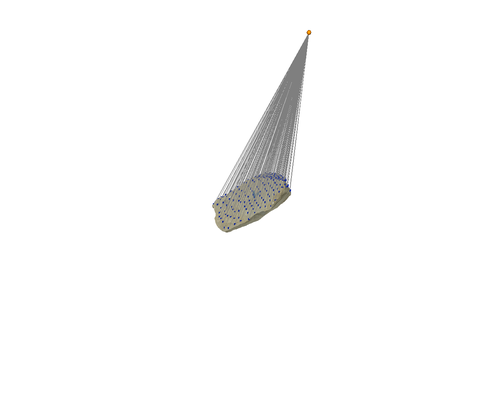}
		\caption{}
		\label{fig: lidar-raydisk-b}
	\end{subfigure}
	\hfill
	\caption{(a) Sensor configured with a disk raycasting pattern (b) sensor raycasting on an aggregate model}
	\label{fig: lidar-raydisk}
\end{figure}

Based on the formulation above, each multi-view sensor was configured with a disk raycasting pattern, as illustrated in \autoref{fig: lidar-raydisk-a}. By implementing a similar raycasting technique described in \cref{sec-raycasting}, the sensor raycasting results on an example aggregate model is demonstrated in \autoref{fig: lidar-raydisk-b}. As shown in blue points, the shape of the aggregate model was accurately captured by such raycasting technique, and the orthogonal disk plane ensured the maximum visibility from the sensor position. Note that \autoref{fig: lidar-raydisk} uses a sparser ray density (i.e. larger arc spacing and ring spacing) for illustration purpose. The real density parameters used in the following steps were arc spacing of 0.2 cm and ring spacing of 0.2 cm, and the total number of rings were calculated to cover a disk plane with a radius $150\%$ of the aggregate model size.

\subsection{Varying-Visibility Raycasting for Shape Observation}
With the raycasting capability programmed into the sensor, a varying-visibility raycasting scheme was designed for partial shape observation. First, a total of $N$ sensors were initialized at positions uniformly distributed on a $r$-radius sphere. For each aggregate model, $N$ was fixed at 16 and $r$ was set as $5$ times the model's equivalent radius. When all $N$ sensors are active, the accumulative raycasting results represent the complete shape (or ground-truth shape) of the aggregate model.

\begin{figure}[!htb]
	\centering
	\hfill
	\begin{subfigure}[b]{0.24\textwidth}
		\centering
		\includegraphics[width=\textwidth]{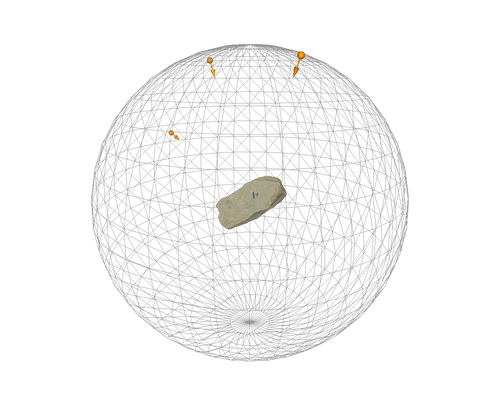}
		\caption{N=3}
	\end{subfigure}
	\hfill
	\begin{subfigure}[b]{0.24\textwidth}
		\centering
		\includegraphics[width=\textwidth]{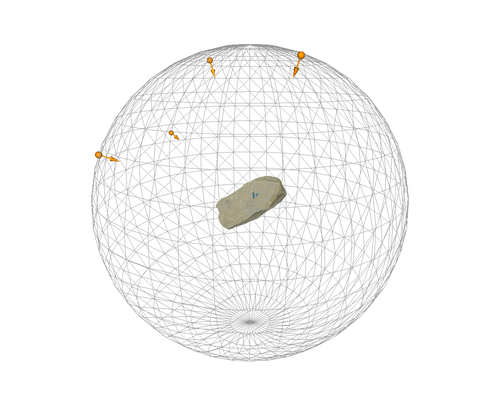}
		\caption{N=4}
	\end{subfigure}
	\hfill
	\begin{subfigure}[b]{0.24\textwidth}
		\centering
		\includegraphics[width=\textwidth]{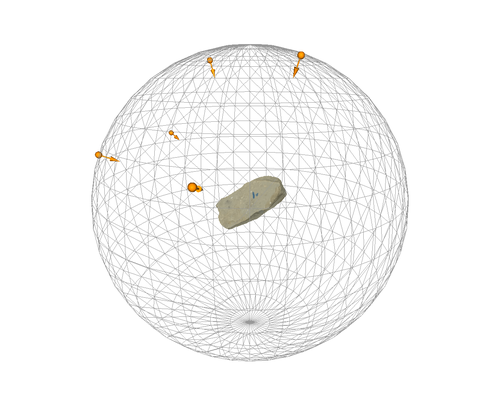}
		\caption{N=5}
	\end{subfigure}
	\hfill
	\begin{subfigure}[b]{0.24\textwidth}
		\centering
		\includegraphics[width=\textwidth]{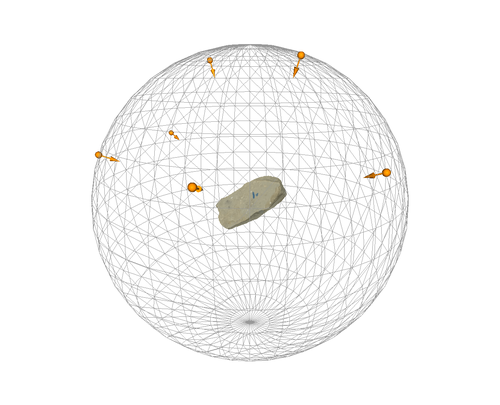}
		\caption{N=6}
	\end{subfigure}
	\newline 
	\hfill
	\begin{subfigure}[b]{0.24\textwidth}
		\centering
		\includegraphics[width=\textwidth]{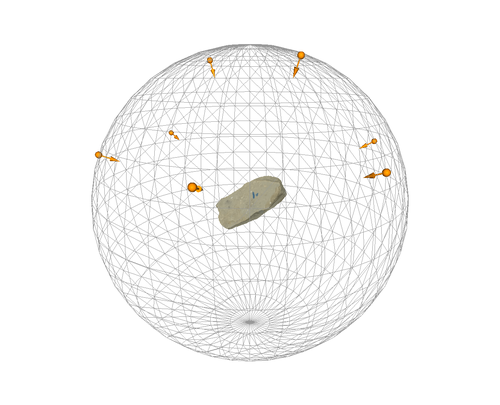}
		\caption{N=7}
	\end{subfigure}
	\hfill
	\begin{subfigure}[b]{0.24\textwidth}
		\centering
		\includegraphics[width=\textwidth]{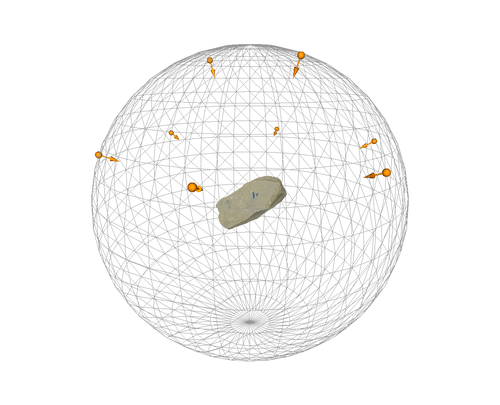}
		\caption{N=8}
	\end{subfigure}
	\hfill
	\begin{subfigure}[b]{0.24\textwidth}
		\centering
		\includegraphics[width=\textwidth]{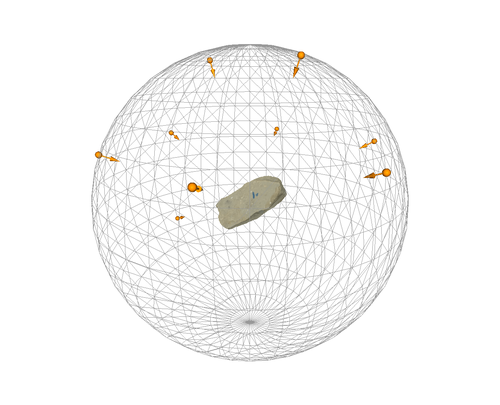}
		\caption{N=9}
	\end{subfigure}
	\hfill
	\begin{subfigure}[b]{0.24\textwidth}
		\centering
		\includegraphics[width=\textwidth]{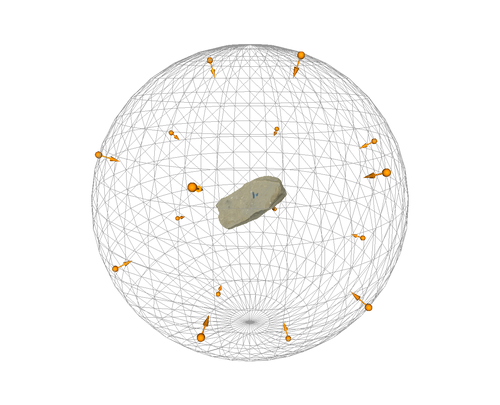}
		\caption{N=16 (complete)}
	\end{subfigure}
	\caption{Sensor sets with increasing number of active sensor views (N)}
	\label{fig: sensor-set}
\end{figure}

To simulate the partial shape observation process, multiple sensor sets consisting of different number of active sensors were created, as shown in \autoref{fig: sensor-set}. These sensor sets are expected to represent different levels of visibility of the aggregate model in a multi-view setting. The specific number of active sensors in the sensor ranges from 3 to 9 for partial views and 16 for the complete view, resulting in seven visibility levels of the partial observations.

\begin{figure}[!htb]
	\centering
	\hfill
	\begin{subfigure}[b]{0.45\textwidth}
		\centering
		\includegraphics[trim=100 300 200 0, clip, width=\textwidth]{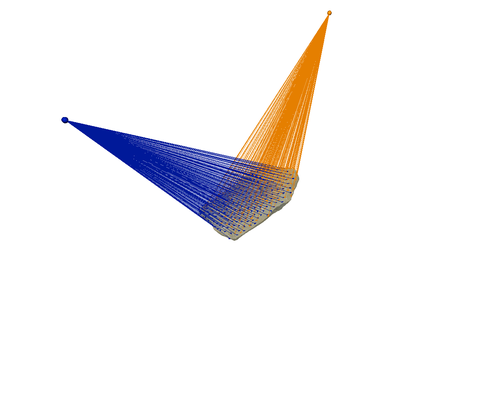}
		\caption{}
	\end{subfigure}
	\hfill
	\begin{subfigure}[b]{0.45\textwidth}
		\centering
		\includegraphics[trim=100 300 200 0, clip, width=\textwidth]{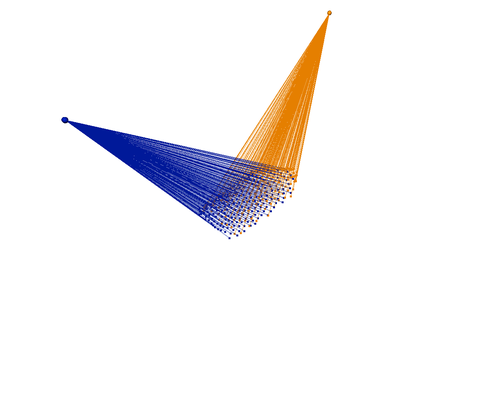}
		\caption{}
	\end{subfigure}
	\newline
	\begin{subfigure}[b]{0.8\textwidth}
		\centering
		\includegraphics[trim=100 300 200 0, clip, width=\textwidth]{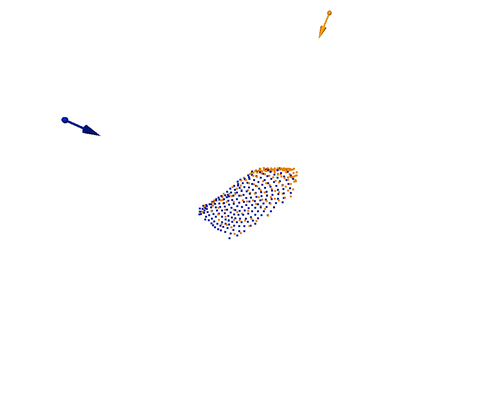}
		\caption{}
	\end{subfigure}
	\hfill
	\caption{Collaborative raycasting of two sensors: (a) ray hits on aggregate surface, (b) extracted points of each sensor, and (c) accumulated point clouds from both sensors}
	\label{fig: sensor-colab}
\end{figure}

\begin{figure}[!htb]
	\centering
	\hfill
	\begin{subfigure}[b]{0.24\textwidth}
		\centering
		\includegraphics[width=\textwidth]{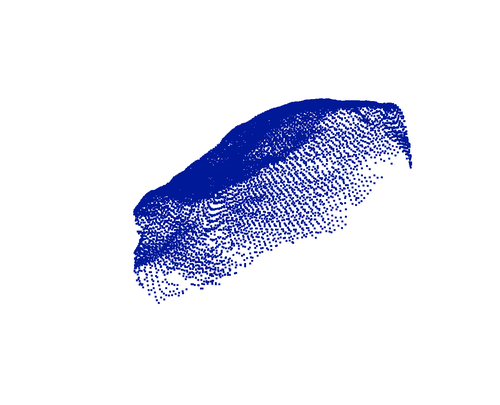}
		\caption{N=3}
	\end{subfigure}
	\hfill
	\begin{subfigure}[b]{0.24\textwidth}
		\centering
		\includegraphics[width=\textwidth]{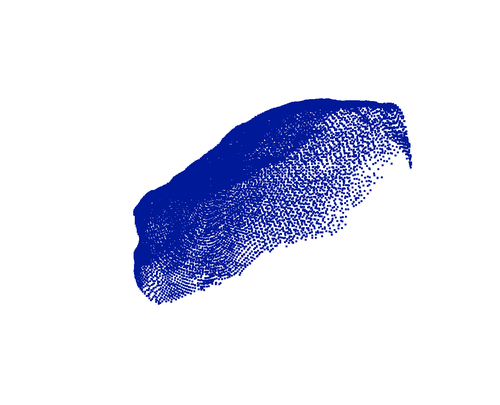}
		\caption{N=4}
	\end{subfigure}
	\hfill
	\begin{subfigure}[b]{0.24\textwidth}
		\centering
		\includegraphics[width=\textwidth]{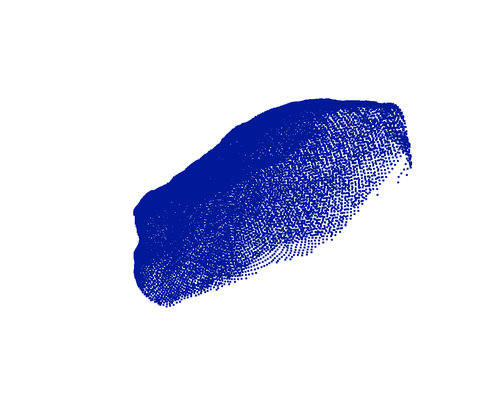}
		\caption{N=5}
	\end{subfigure}
	\hfill
	\begin{subfigure}[b]{0.24\textwidth}
		\centering
		\includegraphics[width=\textwidth]{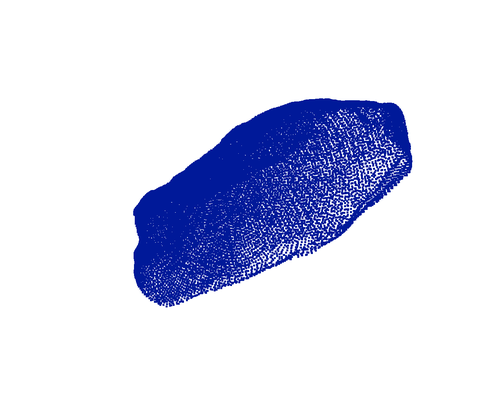}
		\caption{N=6}
	\end{subfigure}
	\newline 
	\hfill
	\begin{subfigure}[b]{0.24\textwidth}
		\centering
		\includegraphics[width=\textwidth]{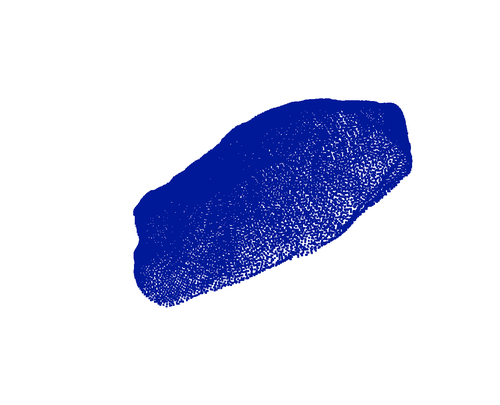}
		\caption{N=7}
	\end{subfigure}
	\hfill
	\begin{subfigure}[b]{0.24\textwidth}
		\centering
		\includegraphics[width=\textwidth]{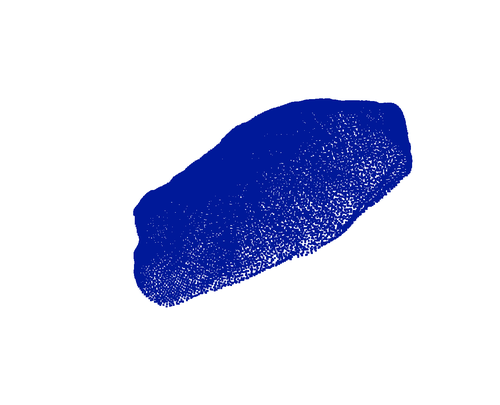}
		\caption{N=8}
	\end{subfigure}
	\hfill
	\begin{subfigure}[b]{0.24\textwidth}
		\centering
		\includegraphics[width=\textwidth]{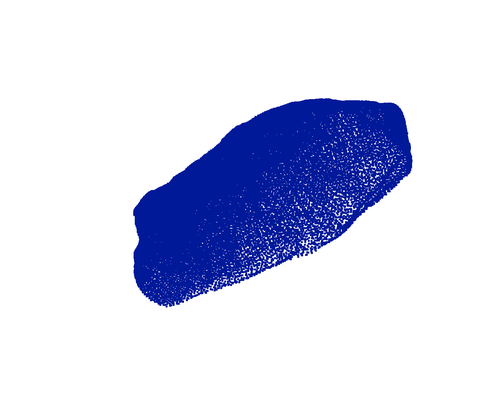}
		\caption{N=9}
	\end{subfigure}
	\hfill
	\begin{subfigure}[b]{0.24\textwidth}
		\centering
		\includegraphics[width=\textwidth]{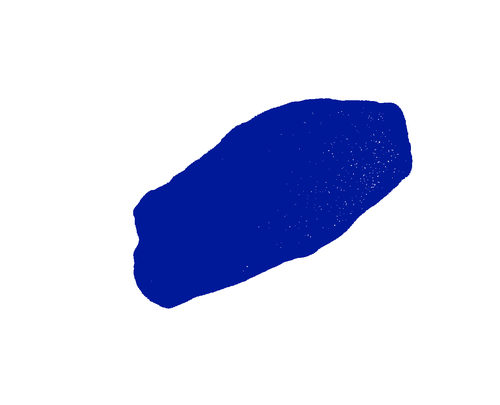}
		\caption{N=16 (complete)}
	\end{subfigure}
	\caption{Varying-visibility shapes with increasing number of active sensor views (N)}
	\label{fig: sensor-views}
\end{figure}

Active sensors in the sensor set collaboratively extract point cloud of the aggregate surface and accumulate to get the partial representation of the shape. The concept of the collaborative raycasting is demonstrated in \autoref{fig: sensor-colab}. Accordingly, based on the sensor sets in \autoref{fig: sensor-set}, the extracted point clouds of the varying-visibility partial views and the complete view are presented in \autoref{fig: sensor-views}. It is clearly shown that the varying-visibility sensor raycasting scheme is able to effectively capture the partial shapes at different visibility levels.

\subsection{Varying-View Raycasting from Different Orientations}
\begin{figure}[!htb]
	\centering
	\hfill
	\begin{subfigure}[b]{0.24\textwidth}
		\centering
		\includegraphics[width=\textwidth]{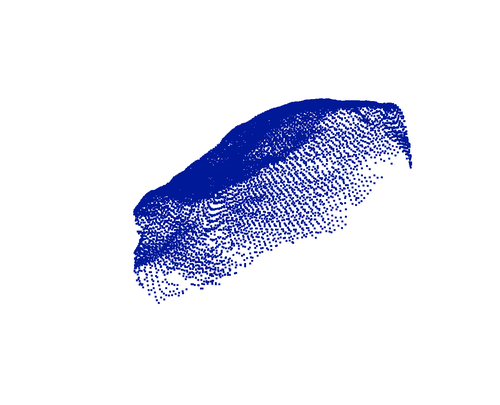}
		\caption{m=1}
	\end{subfigure}
	\hfill
	\begin{subfigure}[b]{0.24\textwidth}
		\centering
		\includegraphics[width=\textwidth]{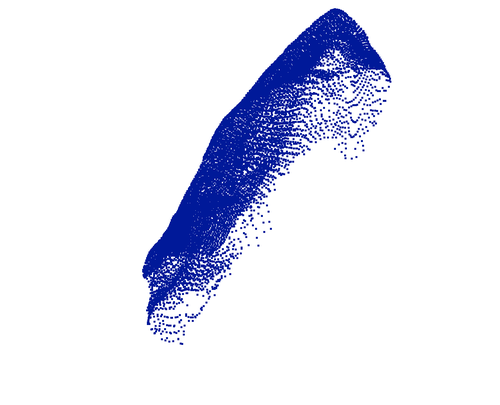}
		\caption{m=2}
	\end{subfigure}
	\hfill
	\begin{subfigure}[b]{0.24\textwidth}
		\centering
		\includegraphics[width=\textwidth]{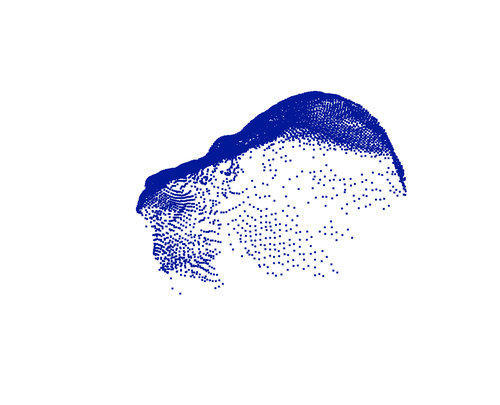}
		\caption{m=3}
	\end{subfigure}
	\hfill
	\begin{subfigure}[b]{0.24\textwidth}
		\centering
		\includegraphics[width=\textwidth]{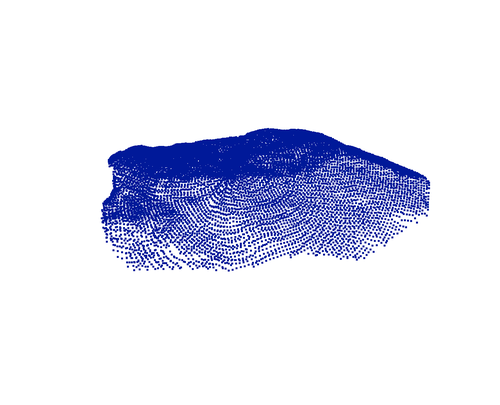}
		\caption{m=4}
	\end{subfigure}
	\newline 
	\hfill
	\begin{subfigure}[b]{0.24\textwidth}
		\centering
		\includegraphics[width=\textwidth]{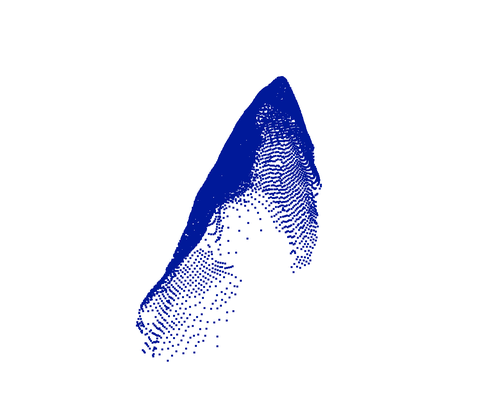}
		\caption{m=5}
	\end{subfigure}
	\hfill
	\begin{subfigure}[b]{0.24\textwidth}
		\centering
		\includegraphics[width=\textwidth]{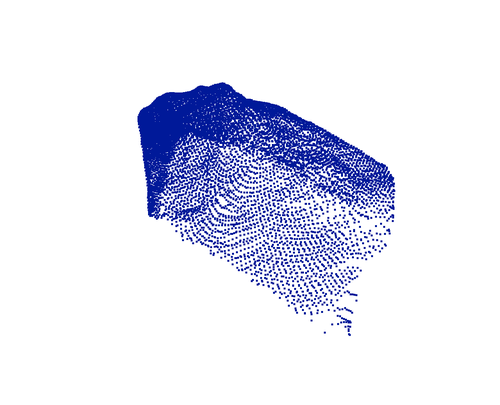}
		\caption{m=6}
	\end{subfigure}
	\hfill
	\begin{subfigure}[b]{0.24\textwidth}
		\centering
		\includegraphics[width=\textwidth]{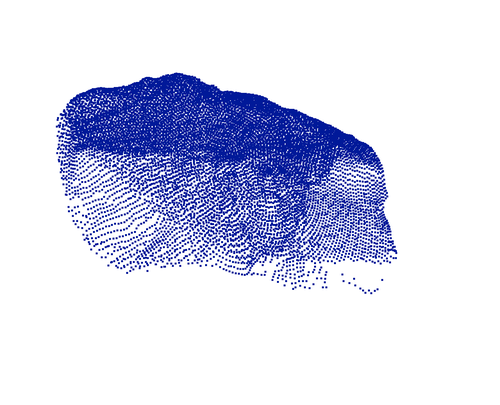}
		\caption{m=7}
	\end{subfigure}
	\hfill
	\begin{subfigure}[b]{0.24\textwidth}
		\centering
		\includegraphics[width=\textwidth]{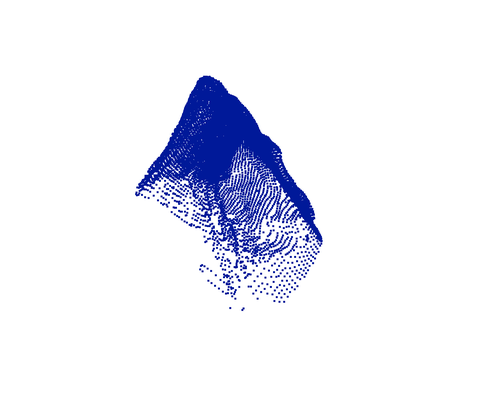}
		\caption{m=8}
	\end{subfigure}
	\newline 
	\hfill
	\begin{subfigure}[b]{0.24\textwidth}
		\centering
		\includegraphics[width=\textwidth]{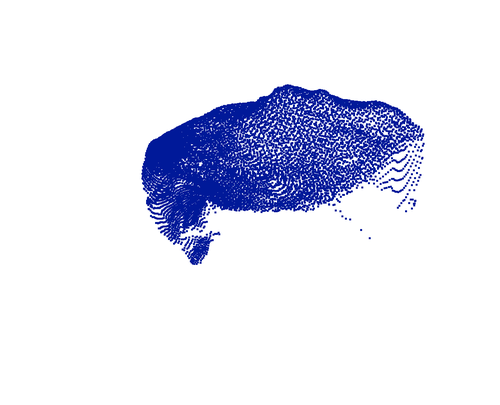}
		\caption{m=9}
	\end{subfigure}
	\hfill
	\begin{subfigure}[b]{0.24\textwidth}
		\centering
		\includegraphics[width=\textwidth]{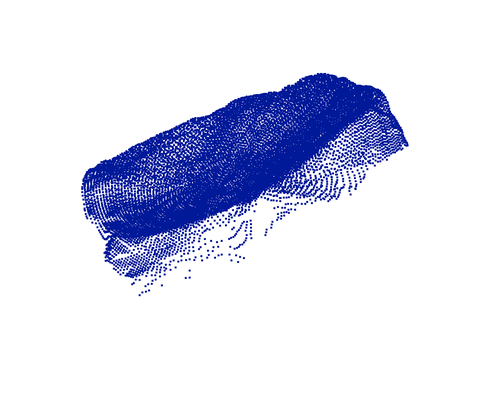}
		\caption{m=10}
	\end{subfigure}
	\hfill
	\begin{subfigure}[b]{0.24\textwidth}
		\centering
		\includegraphics[width=\textwidth]{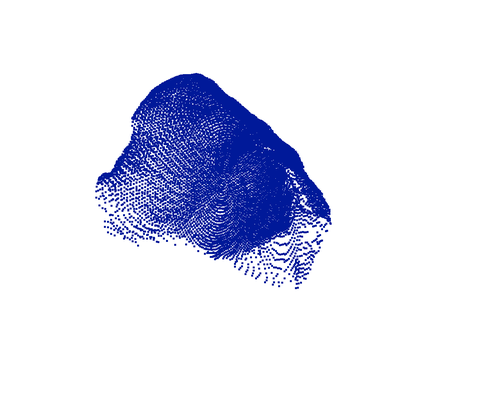}
		\caption{m=11}
	\end{subfigure}
	\hfill
	\begin{subfigure}[b]{0.24\textwidth}
		\centering
		\includegraphics[width=\textwidth]{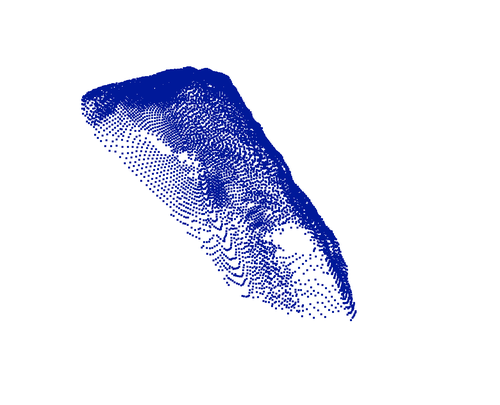}
		\caption{m=12}
	\end{subfigure}
	\caption{Varying-view shapes at $m^{th}$ aggregate model orientation}
	\label{fig: sensor-orientations}
\end{figure}

The varying-visibility raycasting scheme mainly focuses on generating partial views at different visibility levels, yet with all the partial views under the same orientation of the aggregate model. Therefore, a separate scheme was developed to vary the model orientation, which is named the varying-view raycasting scheme. Note that varying the orientation of the model has the same effect of varying the orientation of the entire sensor set based on the principle of relative motion. The orientation of each aggregate model was permuted $M$ times, where each orientation was computed by finding the $M$ uniformly distributed positions on a unit sphere and using them as the directional vector for rotation. At each orientation, the entire sensor sets first perform the varying-visibility raycasting scheme, permute the model orientation using this scheme, and repeat the previous steps. The effect of the varying-view raycasting scheme is demonstrated in \autoref{fig: sensor-orientations}, showing only the first sensor set (i.e., $N=3$) views per each orientation. $M$ is set as 12 for demonstration purpose but was set to 16 during the dataset generation process. Note that these are all partial views (with the lowest visibility) of the same aggregate model.

\subsection{Dataset of Partial-Complete Aggregate Shape Pairs}

The varying-visibility and varying-view raycasting schemes together simulate the partial observation process in a comprehensive way. By treating the all-around sensor view ($N=16$) as the ground-truth complete shape, a dataset of partial-complete aggregate shape pairs can be efficiently established. For each of the 82 rock models (46 RR3 rocks and 36 RR4 rocks) in the 3D aggregate particle library, a total of 9,184 partial-complete shape pairs were generated, since one model has seven visibility levels ($N=\{3,4,5,6,7,8,9\}$) and $M=16$ model orientations. The dataset was further divided into 9,000 training pairs and 184 validation pairs. The validation pairs were randomly selected and separated from the dataset. In addition, to further check the network performance on unseen aggregate shapes, six extra aggregate models were used to generate 672 partial-complete shape pairs. These are RR3 models that were not included in the 3D aggregate library and are therefore considered as an independent test set of the shape completion network. The dataset organization is listed in \autoref{tab:completion-split}. Note that the dataset was regularized by uniform downsampling to 2,048 points per partial shape and 16,384 points per complete shape, which is common fixed data sizes in other popular datasets such as ShapeNet \parencite{chang_shapenet_2015} and Completion3D \parencite{tchapmi_topnet_2019}.

\begin{table}[!htb]
	\centering
	\caption{Dataset Organization for Learning the Shape Completion}
	\label{tab:completion-split}
	\begin{tabular}{l L{0.35\linewidth} L{0.35\linewidth}}
		\hline
		\textbf{Dataset Split} & \textbf{Number of Prototype Aggregate Models} & \textbf{Number of Partial-Complete Shape Pairs} \\ \hline
		Train      & 82 & 9,000 \\
		Validation & 82 & 184   \\
		Test       & 6  & 672   \\ \hline
	\end{tabular}
\end{table}

\section{Deep Learning Framework for Learning 3D Shape Completion}

\begin{figure}[!htb]
	\centering
	\includegraphics[trim=50 150 50 50, clip, width=\textwidth]{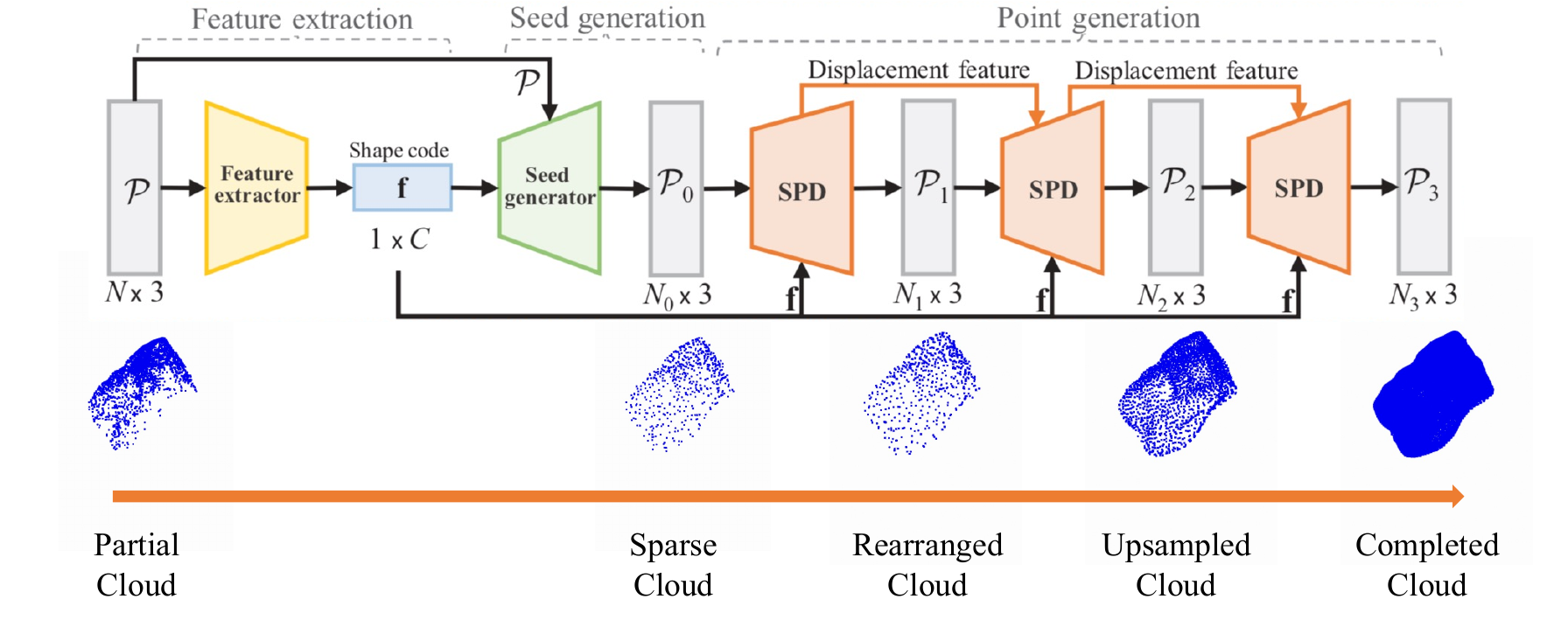}
	\caption{SnowflakeNet architecture for 3D shape completion}
	\label{fig: snowflake}
\end{figure}

Based on the review of 3D shape completion approaches in computer vision, a state-of-the-art network, SnowflakeNet \parencite{xiang_snowflakenet_2021}, was selected and implemented for learning the 3D shape completion of aggregates. The overall architecture of SnowflakeNet is presented in \autoref{fig: snowflake}. The network models the 3D shape completion process as a multi-stage snowflake-like growth of points in space, which consists of three major modules: feature extraction, seed generation, and point generation.

\subsection{Feature Extraction Module}
The input of the network is a point cloud with fixed data dimension, denoted as $\mathbb{P}=\{p_i=(x_i, y_i, z_i)\in \mathbb{R}^3|\ i\in\ \{1,...,N\}\}$ where $N$ is the number of input points. The overall shape completion process follows an encoder-decoder approach, where the partial input cloud is condensed into a high-dimensional feature vector by an encoder, and then the decoder generates fine-grained completion by enriching the feature space. 

For the encoder part, the network uses set abstraction layers developed in PointNet++ \parencite{qi_pointnet_2017-1} and Point Transformer layers \parencite{zhao_point_2021} together to encode the global and local shape context into a linear feature vector or shape latent code of size $1\times C$. This step is denoted as the feature extraction process to obtain high-level shape characteristics with a condensed representation. Although all the training data have 2,048 points, the network can actually take any arbitrary data size since this feature extraction step will first perform regularization to sample the data size down to 512 points following the Farthest Point Sampling (FPS) algorithm proposed in PointNet++ \parencite{qi_pointnet_2017-1}. FPS is a shape feature-preserving technique to efficiently reduce the 3D data size while maintain the prominent features. 

\subsection{Seed Generation Module}
After the feature extraction, a two-stage decoder in the network conducts the shape completion task. First, a coarse-grained decoder denoted as the seed generator predicts a sparse version of the complete cloud with $N_c=256$ points/seeds. This decoder consists of 1D deconvolution (i.e., transposed convolution) layers and Multi-Layer Perceptron (MLP) layers to learn the seed generation, which is referred to as the point splitting operation in the network. The point splitting operation is essentially the 1D deconvolution operation with large receptive field such that it can capture both existing and missing shape characteristics. The generated seeds are then merged with the input partial cloud to fill the missing portions. However, the merged cloud has non-uniform point density with fewer points in the missing regions. Therefore, FPS is used to re-sample the cloud into a uniform sparse cloud of $N_0=512$ points with the complete shape, denoted as $\mathbb{P}_0$. The overall design concept is similar to a seeded region growing approach, which first focuses on capturing the high-level shape characteristics with sparse representation and then enhancing the shape details as the next step.

\subsection{Point Generation Module}
Based on the coarse cloud with complete shape, a fine-grained decoder is designed to predict high-quality complete cloud while preserving the shape features. This decoder uses the Snowflake Point Deconvolution (SPD) layers to upsample the points by splitting each parent point into multiple child points, which is done by first duplicating the parent points and then adding variations to the duplicates. Different from previous methods that ignore the local shape characteristics around the parent point, SPD utilizes point-wise splitting operation to fully leverage the local geometric information around the parent point. The key design in the SPD is the Skip Transformer (ST). With an upsampling factor of the SPD $r$, all parent points are first duplicated with $r$ copies. Each point is passed through the ST layer to get per-point displacement feature vectors $K$. Then, an MLP layer computes a per-point coordinate shift $\Delta p_i$, which is added to the original coordinates to get the upsampled points.

ST uses the PointNet \parencite{qi_pointnet_2017} features as query $Q$, generates the shape context feature $H$, and further conducts deconvolution to get the internal displacement features as key $K$. Following the general design of transformer, per-point query and key vectors are concatenated to form the value vector, and the attention vector is estimated based on the key and value vectors. Note that the attention vector denotes how much attention the old shape characteristics receive during the upsampling process. The displacement features $K$ are carried between SPD operations, which allows the shape context to propagate along the sequential upsampling process. 

By applying SPD with different upsampling factors, a sequence of gradually-refined point clouds can be generated. The upsampling factors used in the network are $r_1=1, r_2=4, r_3=8$. The first SPD with $r_1=1$ generates a rearranged point cloud $\mathbb{P}_1$ as the same size of the sparse cloud ($N_1=512$) but with points slightly rearranged to form a more reasonable shape. The following two SPDs with $r_2=4$ and $r_3=8$ predicts the upsampled cloud $\mathbb{P}_2$ ($N_2=2048$) and the final completed cloud $\mathbb{P}_3$ ($N_3=16,384$), respectively.

\section{Evaluation of 3D Shape Completion Results}
To evaluate the performance of the 3D shape completion network, both qualitative and quantitative evaluation were conducted. First, the effect of the point splitting operation during the upsampling step was visualized and inspected. Next, quantitative metrics at both micro-scale (i.e., per-point level) and macro-scale (i.e., per-instance level) were used to validate the effectiveness of the network. Lastly, additional tests on unseen aggregate shapes were performed to further check the potential performance of the network in field application.

\subsection{Effect of Point Splitting Operation}

\begin{figure}
	\centering
	\includegraphics[width=\textwidth]{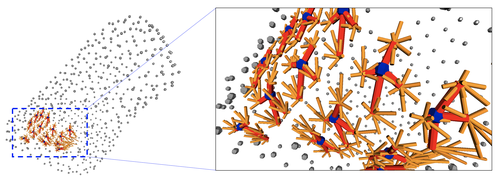}
	\caption{Effect of the point splitting operation during upsampling}
	\label{fig: point-splitting}
\end{figure}

As described above, the SPD upsampling process is a key step to generate a high-quality dense cloud from the sparse seed cloud. Therefore, the point splitting effect is illustrated in \autoref{fig: point-splitting} for qualitative inspection. The point cloud on the left is the rearranged cloud $\mathbb{P}_1$ with $N_1=512$ points colored in gray. The effect of point splitting was visualized on a selected region on the cloud with an enlarged view on the right. The parent points on rearranged cloud $\mathbb{P}_1$ were colored in blue, with the connection paths to the first-level SPD splitting (in red) and the second-level SPD splitting (in orange). The endpoints of the red paths represent the splitted points that form the upsampled cloud $\mathbb{P}_2$ and the orange paths represent the points in the final completed cloud $\mathbb{P}_3$. With the upsampling factor $r_2=4$ and $r_3=8$, the splitting paths construct a quad-tree (i.e., tree structure with four children per node) and octree (i.e., tree structure with eight children per node) at each splitting, respectively. From the visualization in \autoref{fig: point-splitting}, it can be seen that the point splitting operation generates reasonable upsampling of the cloud by preserving the local shape context. Note that although the splitting results may look similar to a linear or bilinear interpolation between the points in $\mathbb{P}_1$, the mechanism behind the splitting is completely different from an interpolation operation. Interpolation is a deterministic approach that does not guarantee shape-preserving results and may often smooth the surface, while point splitting is a learning-based approach that adds fine-grained details yet preserves local shape characteristics.

\subsection{Point-wise Discrepancy with the Ground-Truth Shapes}
As a measure of the point-wise discrepancy between the completed shape $\mathbb{P}_3$ and the ground-truth shape, $L_1$-Chamfer Distance (CD, \textcite{chang_shapenet_2015}) was used as the metric. Given two point sets, $S_1$ and $S_2$, Chamfer distance measures the average distance from every point in $S_1$ to its nearest point in $S_2$, and vice versa. $L_1$ stands for the least absolute distance. The calculation of CD follows \autoref{eqn: chamfer}, where $N_1$ and $N_2$ are the number of points in $S_1$ and $S_2$, respectively.

\begin{equation} \label{eqn: chamfer}
	d_{CD, L_1}(S_1,S_2)=\frac{1}{N_1}\sum_{x\in S_1}\min_{y\in S_2} \lVert x-y \rVert + \frac{1}{N_2}\sum_{y\in S_2}\min_{x\in S_1} \lVert y-x) \rVert
\end{equation}

CD metric depicts the quality of shape completion at the micro-scale (per-point level). The CDs calculated on different dataset splits are listed in \autoref{tab:chamfer-split}. The per-point average CD is very small, which indicates the overall completed shape agrees well with the ground-truth shape, in training set, validation set, and test set. However, it should be noticed that the distance calculated is an average value, therefore, a small portion of points with large deviation may not reflect significantly on the CD metric. 

The increase in CD from training to test set indicates that the shape prediction demonstrates less accuracy on uncertain shapes. Recall that the training set contains the known shapes the network is supposed to learn from, and the validation set includes shapes that are generated based on the same set of known aggregate models but with unique visibility/view. Namely, the shapes in the validation set are not used during training but are not considered as completely novel shapes either. The shapes in the test set, in contrast, are all unseen shapes.
\begin{table}[!htb]
	\centering
	\caption{Chamfer Distance on Different Dataset Splits}
	\label{tab:chamfer-split}
	\begin{tabular}{ll}
		\hline
		\textbf{Dataset Split} & \textbf{Chamfer Distance (mm)} \\ \hline
		Train                  & 0.00391                        \\
		Validation             & 0.00483                        \\
		Test                   & 0.00559                        \\ \hline
	\end{tabular}
\end{table}

\subsection{Shape Completion Results on Novel Views of Known Shapes}

\begin{figure}[!htb]
	\centering
	\begin{tabular}{C{2.2cm}C{2.2cm}C{2.2cm}C{2.2cm}C{2.2cm}C{2.2cm}}
		Partial Cloud $\mathbb{P}$ & Sparse Cloud $\mathbb{P}_0$ & Rearranged Cloud $\mathbb{P}_1$ & Upsampled Cloud $\mathbb{P}_2$ & Completed Cloud $\mathbb{P}_3$ & Ground-Truth \\
		\includegraphics[trim=0 200 0 200, clip, height=1.5cm]{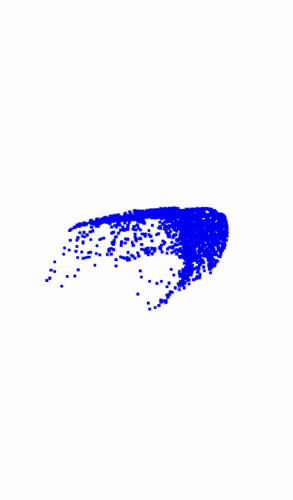} &
		\includegraphics[trim=0 200 0 200, clip, height=1.5cm]{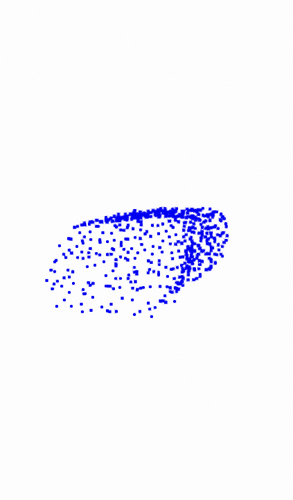}  &  
		\includegraphics[trim=0 200 0 200, clip, height=1.5cm]{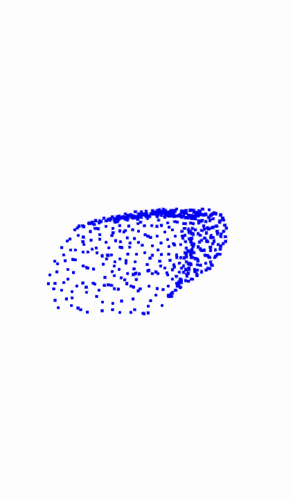}  &  
		\includegraphics[trim=0 200 0 200, clip, height=1.5cm]{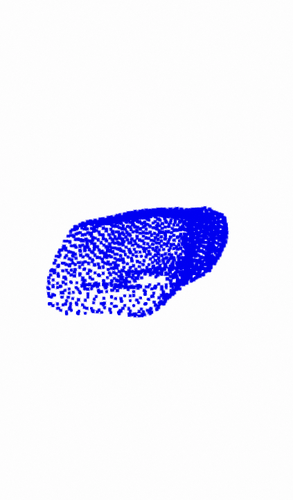}  &  
		\includegraphics[trim=0 200 0 200, clip, height=1.5cm]{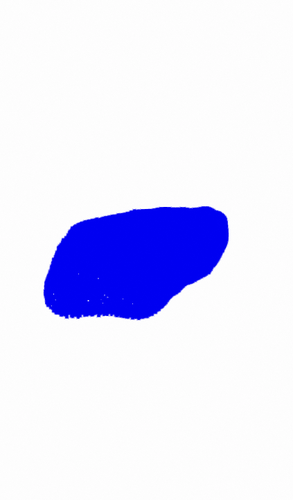}  & 
		\includegraphics[trim=0 200 0 200, clip, height=1.5cm]{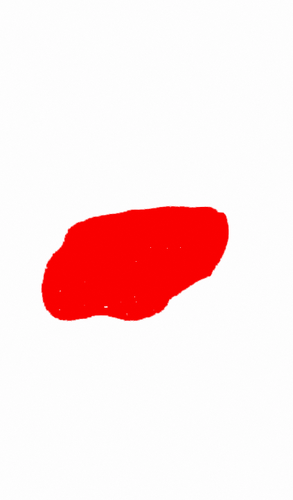}  \\ 
		\includegraphics[trim=0 200 0 150, clip, height=1.5cm]{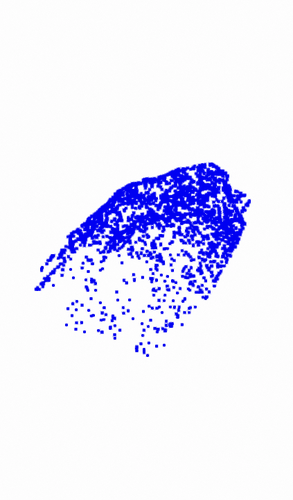} &
		\includegraphics[trim=0 200 0 150, clip, height=1.5cm]{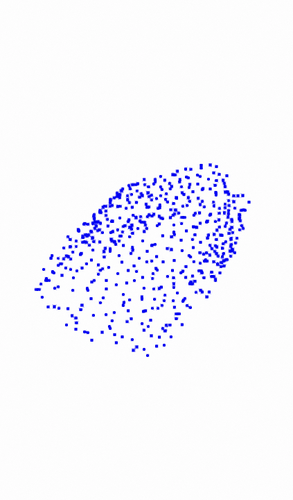}  &  
		\includegraphics[trim=0 200 0 150, clip, height=1.5cm]{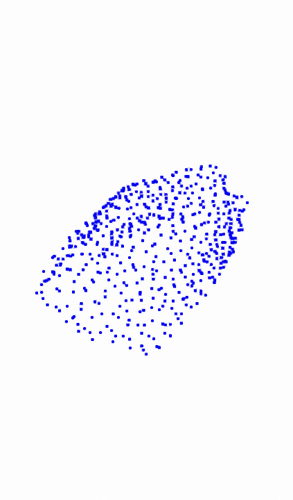}  &  
		\includegraphics[trim=0 200 0 150, clip, height=1.5cm]{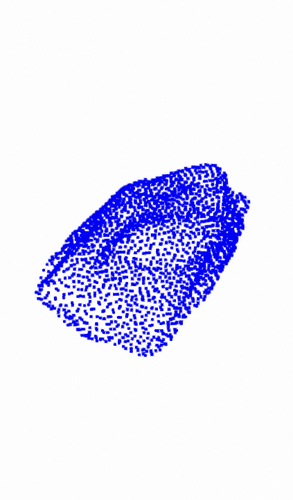}  &  
		\includegraphics[trim=0 200 0 150, clip, height=1.5cm]{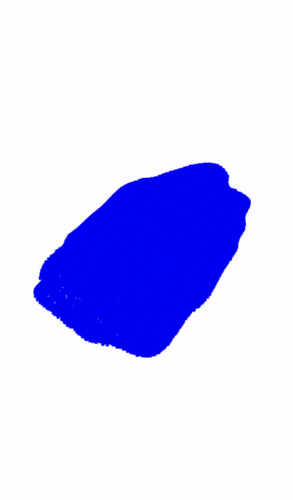}  & 
		\includegraphics[trim=0 200 0 150, clip, height=1.5cm]{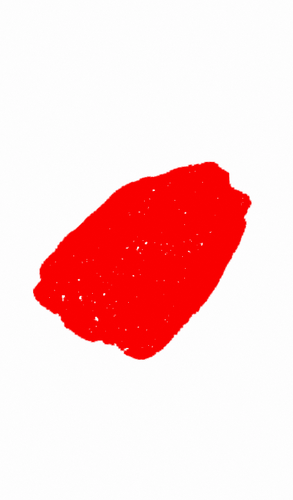}  \\ 
		\includegraphics[trim=0 150 0 200, clip, height=1.5cm]{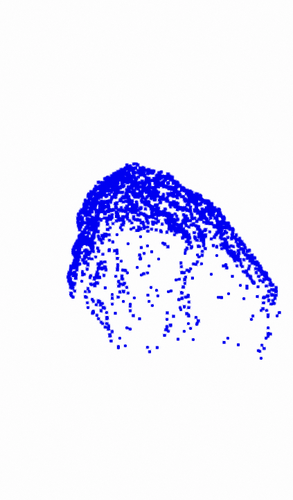} &
		\includegraphics[trim=0 150 0 200, clip, height=1.5cm]{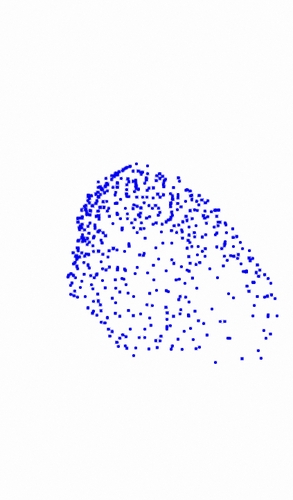}  &  
		\includegraphics[trim=0 150 0 200, clip, height=1.5cm]{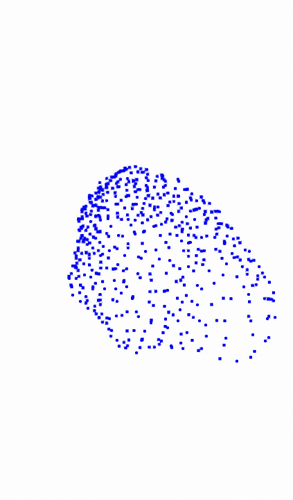}  &  
		\includegraphics[trim=0 150 0 200, clip, height=1.5cm]{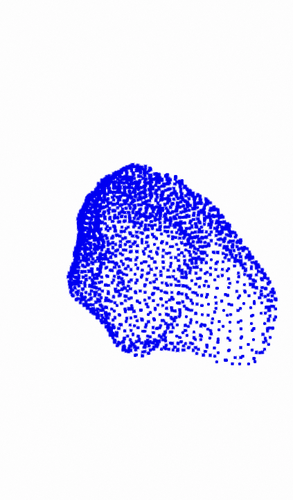}  &  
		\includegraphics[trim=0 150 0 200, clip, height=1.5cm]{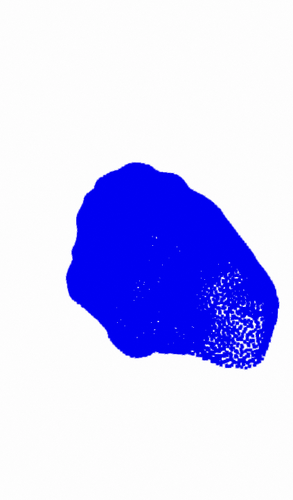}  & 
		\includegraphics[trim=0 150 0 200, clip, height=1.5cm]{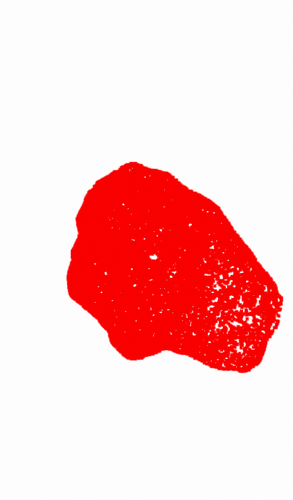}
	\end{tabular}
	\caption{Intermediate and final shape completion results for shapes in the validation set}
	\label{fig:val-results}
\end{figure}

For the partial shapes in the validation set, the shape completion results of three randomly selected inputs are presented in \autoref{fig:val-results}. It can be observed that the network can generate high-quality results that agree well with the ground-truth shape for those novel views of the shapes known in the training dataset. This may indicate the network effectively learn the high-level shape representation rather than behaving similarly to a template matching-based approach.

In addition to the micro-scale CD metric, macro-scale metrics that describe the particle shape at the instance level were also used for evaluation. The metrics include ESD, shortest/intermediate/longest dimensions, 3D FER, surface area, and volume. Comparisons were made between the completed shape and the ground-truth shape for each of the metrics, as shown in \autoref{fig: val-results-macro}. The comparisons of macro-scale metrics demonstrate that the completed shapes in the validation set achieve a good match in terms of aggregate morphological properties, where the MAPE error between the prediction and the ground-truth is less than $2.5\%$ for all metrics.

\begin{figure}[!htb] %
	\centering
	\begin{subfigure}[b]{0.45\textwidth}
		\centering
		\includegraphics[width=\textwidth]{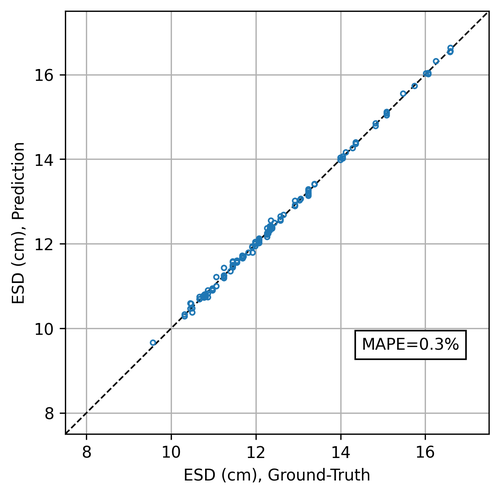}
		\caption{ESD}
	\end{subfigure}
	\hfill
	\begin{subfigure}[b]{0.45\textwidth}
		\centering
		\includegraphics[width=\textwidth]{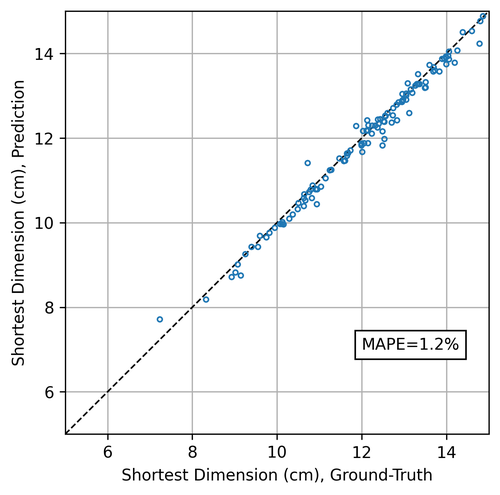}
		\caption{Shortest Dimension}
	\end{subfigure}
	\newline 
	\hfill
	\begin{subfigure}[b]{0.45\textwidth}
		\centering
		\includegraphics[width=\textwidth]{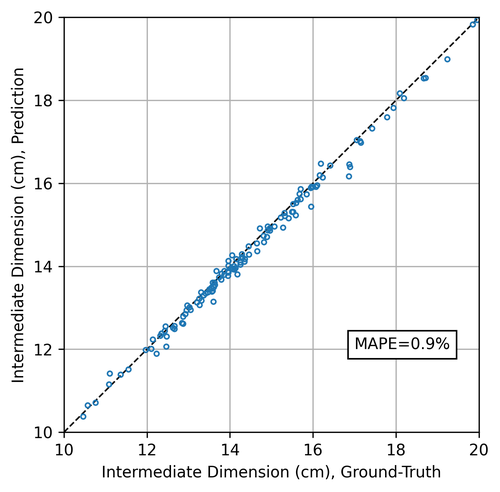}
		\caption{Intermediate Dimension}
	\end{subfigure}
	\hfill
	\begin{subfigure}[b]{0.45\textwidth}
		\centering
		\includegraphics[width=\textwidth]{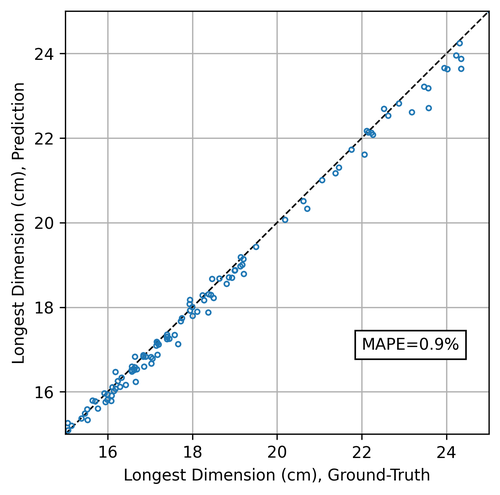}
		\caption{Longest Dimension}
	\end{subfigure}
	\caption{Comparisons of macro-scale metrics between the completed shapes and ground-truth shapes in the validation set}
	\label{fig: val-results-macro}
\end{figure}

\begin{figure}[!htb] \ContinuedFloat
	\captionsetup{list=off,format=continued} %
	\caption{}
	\begin{subfigure}[b]{0.45\textwidth}
		\centering
		\includegraphics[width=\textwidth]{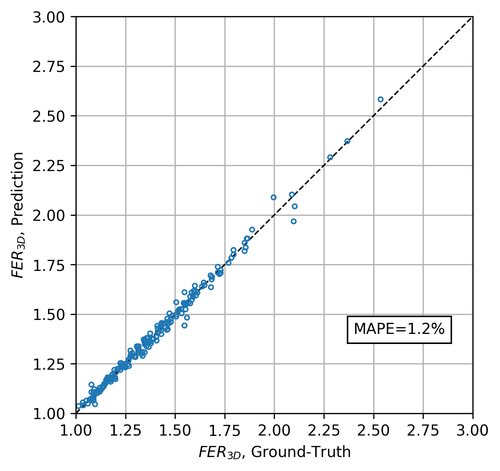}
		\caption{3D FER}
	\end{subfigure}
	\hfill
	\begin{subfigure}[b]{0.45\textwidth}
		\centering
		\includegraphics[width=\textwidth]{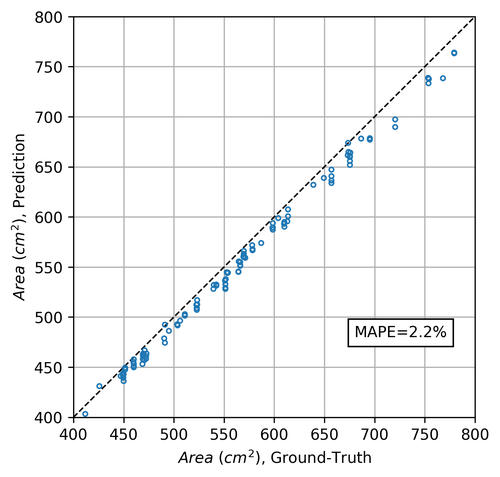}
		\caption{Surface Area}
	\end{subfigure}
	\newline 
	\begin{subfigure}[b]{0.45\textwidth}
		\centering
		\includegraphics[width=\textwidth]{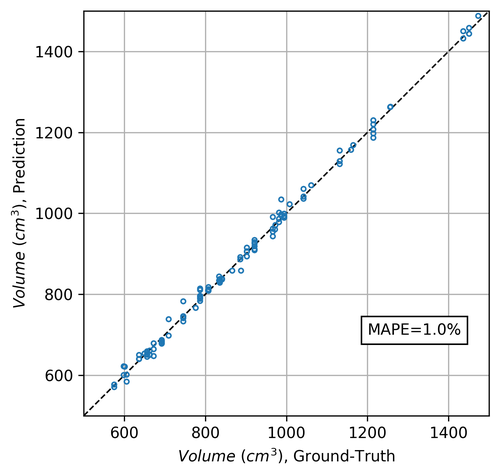}
		\caption{Volume}
	\end{subfigure}
\end{figure}
\clearpage

\subsection{Shape Completion Results on Unseen Aggregate Shapes}

\begin{figure}[!htb]
	\centering
	\begin{tabular}{C{2.2cm}C{2.2cm}C{2.2cm}C{2.2cm}C{2.2cm}C{2.2cm}}
		Partial Cloud $\mathbb{P}$ & Sparse Cloud $\mathbb{P}_0$ & Rearranged Cloud $\mathbb{P}_1$ & Upsampled Cloud $\mathbb{P}_2$ & Completed Cloud $\mathbb{P}_3$ & Ground-Truth \\
		\includegraphics[trim=0 200 0 200, clip, height=1.5cm]{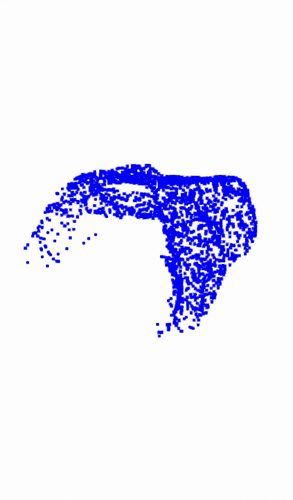} &
		\includegraphics[trim=0 200 0 200, clip, height=1.5cm]{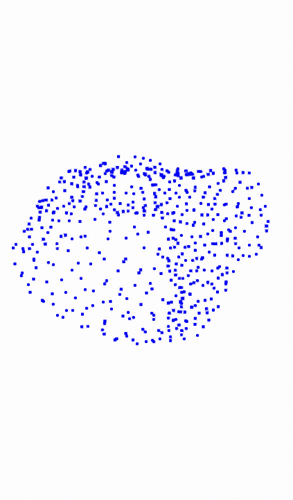}  &  
		\includegraphics[trim=0 200 0 200, clip, height=1.5cm]{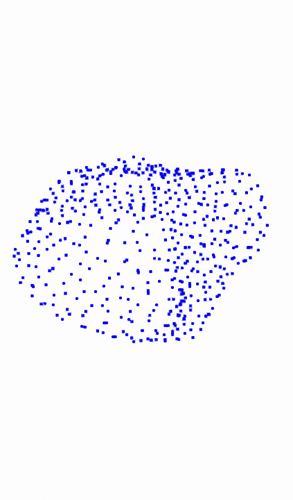}  &  
		\includegraphics[trim=0 200 0 200, clip, height=1.5cm]{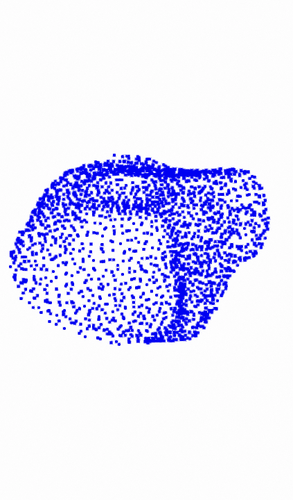}  &  
		\includegraphics[trim=0 200 0 200, clip, height=1.5cm]{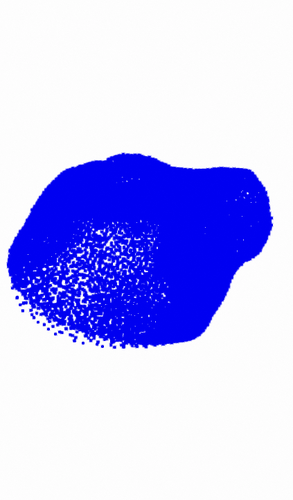}  & 
		\includegraphics[trim=0 200 0 200, clip, height=1.5cm]{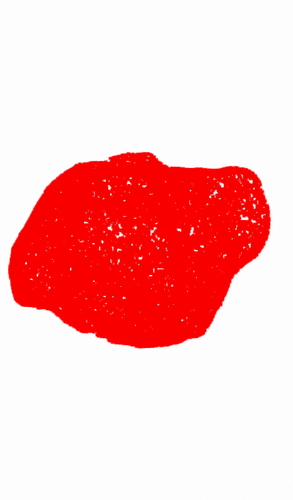}  \\ 
		\includegraphics[trim=0 200 0 150, clip, height=1.5cm]{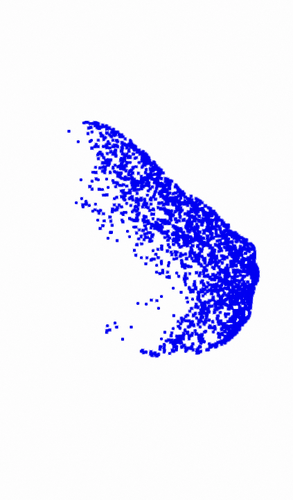} &
		\includegraphics[trim=0 200 0 150, clip, height=1.5cm]{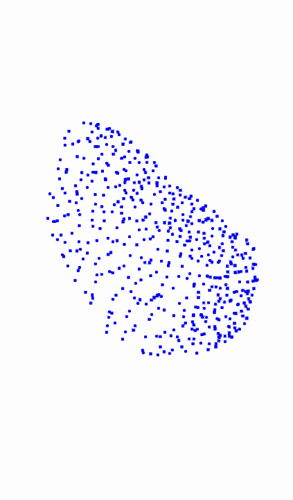}  &  
		\includegraphics[trim=0 200 0 150, clip, height=1.5cm]{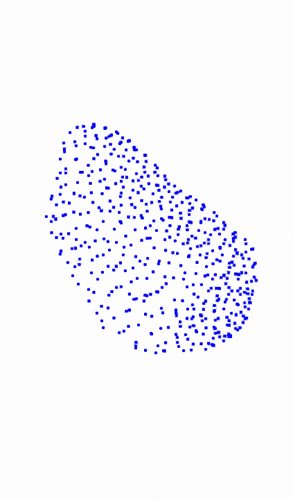}  &  
		\includegraphics[trim=0 200 0 150, clip, height=1.5cm]{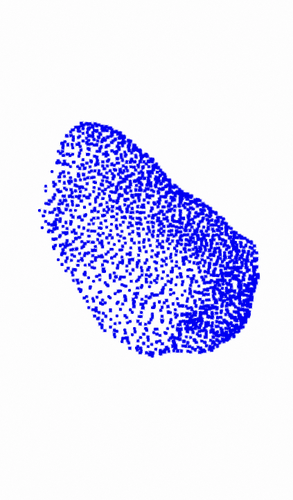}  &  
		\includegraphics[trim=0 200 0 150, clip, height=1.5cm]{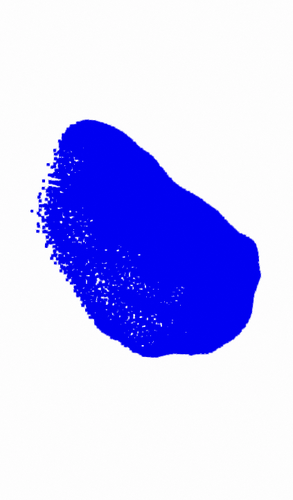}  & 
		\includegraphics[trim=0 200 0 150, clip, height=1.5cm]{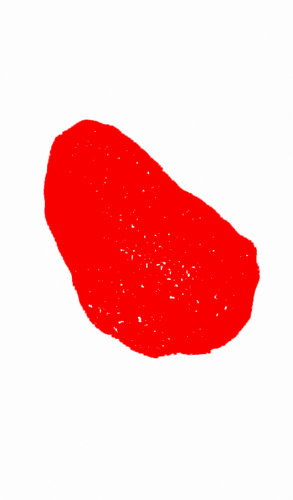}  \\ 
		\includegraphics[trim=0 200 0 200, clip, height=1.5cm]{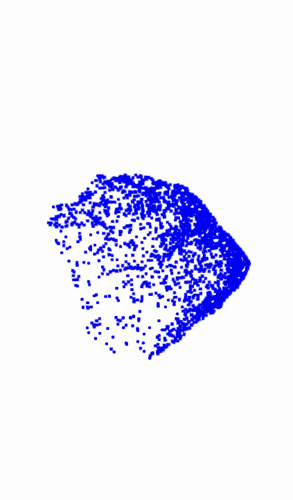} &
		\includegraphics[trim=0 200 0 200, clip, height=1.5cm]{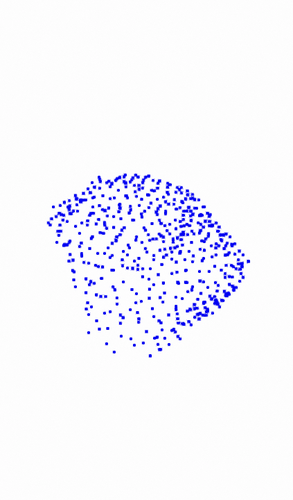}  &  
		\includegraphics[trim=0 200 0 200, clip, height=1.5cm]{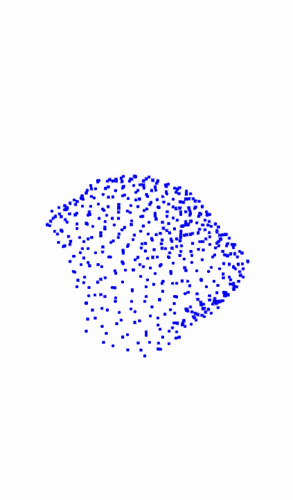}  &  
		\includegraphics[trim=0 200 0 200, clip, height=1.5cm]{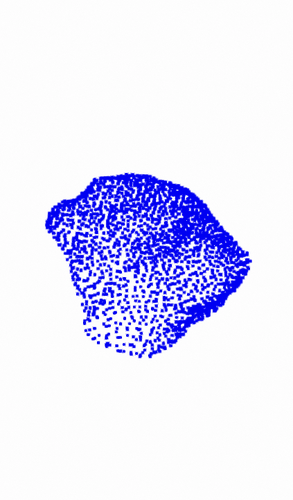}  &  
		\includegraphics[trim=0 200 0 200, clip, height=1.5cm]{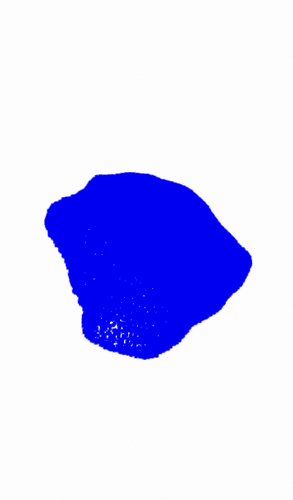}  & 
		\includegraphics[trim=0 200 0 200, clip, height=1.5cm]{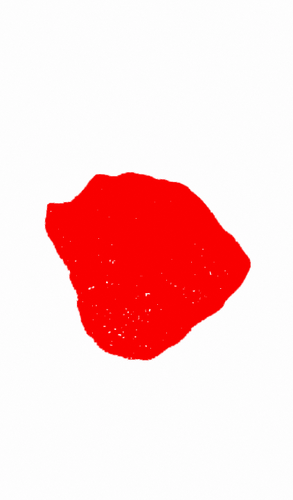}
	\end{tabular}
	\caption{Intermediate and final shape completion results for shapes in the test set}
	\label{fig:test-results}
\end{figure}

The comparisons above demonstrate the good performance of the network in handling novel views of known shapes, but the network's ability for predicting reasonable shapes for a completely unseen particle has not been verified. In this regard, the same type of comparison was made for the unseen shapes in the test set. First, the shape completion results of three randomly selected inputs from the test set are presented in \autoref{fig:test-results}. By comparing with the completion results from the validation set, two major observations were made between the validation and the test sets. First, the results from the test set shows more uncertain prediction towards the missing region of the shape. The test set results demonstrate a more scattered pattern among the predicted points that are near the missing region, meanwhile the validation set results generate sharper and more confident completion in the missing space. This is actually a good sign showing that given a completely unseen shape, the network is trying to predict the missing part in a probabilistic manner instead of forcing to fit certain shape primitives.

In terms of the macro-scale metrics, it is observed that the network is still able to predict shapes with reasonably good matches regarding the morphological properties, as shown in \autoref{fig: test-results-macro}. The MAPE errors of the results are consistently higher than the validation set results, which aligns with the fact that these are completely unseen shapes. The MAPE errors still lie within $5\%$, but it should be noted that the MAPE error describes an error average rather than extremes. It can be seen that the maximum percentage error of the morphological properties could reach $10\%$ or $20\%$ for certain predictions. Considering the fact that predicting the unseen shape should always been a probabilistic approach, the author concludes that the network presents good performance in predicting reasonable shapes for unseen aggregates.

\begin{figure}[!htb] %
	\centering
	\begin{subfigure}[b]{0.45\textwidth}
		\centering
		\includegraphics[width=\textwidth]{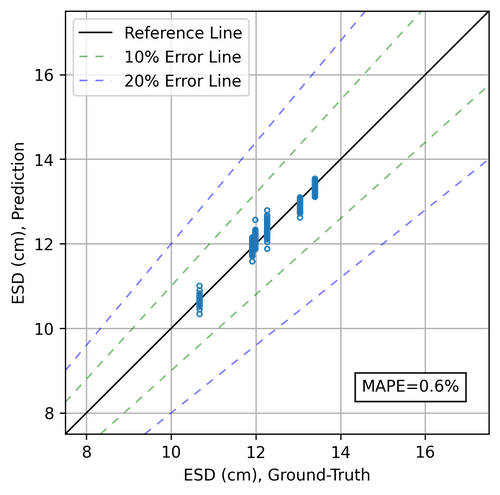}
		\caption{ESD}
	\end{subfigure}
	\hfill
	\begin{subfigure}[b]{0.45\textwidth}
		\centering
		\includegraphics[width=\textwidth]{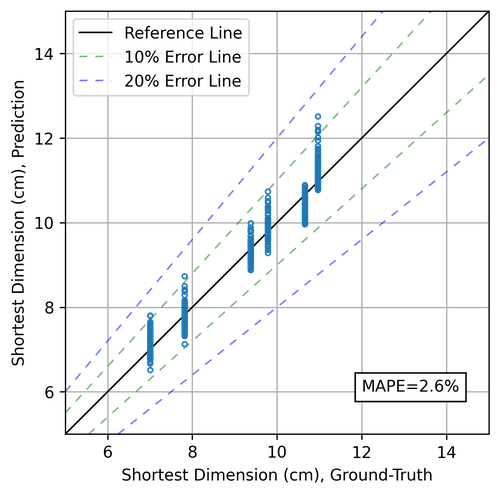}
		\caption{Shortest Dimension}
	\end{subfigure}
	\newline 
	\hfill
	\begin{subfigure}[b]{0.45\textwidth}
		\centering
		\includegraphics[width=\textwidth]{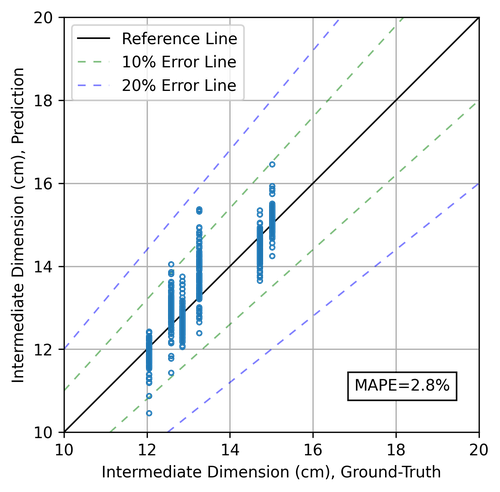}
		\caption{Intermediate Dimension}
	\end{subfigure}
	\hfill
	\begin{subfigure}[b]{0.45\textwidth}
		\centering
		\includegraphics[width=\textwidth]{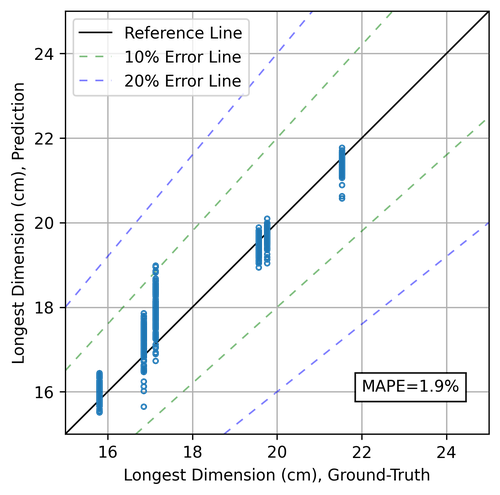}
		\caption{Longest Dimension}
	\end{subfigure}
	\caption{Comparisons of macro-scale metrics between the completed shapes and ground-truth shapes in the test set}
	\label{fig: test-results-macro}
\end{figure}

\begin{figure}[!htb] \ContinuedFloat
	\captionsetup{list=off,format=continued} %
	\caption{}
	\begin{subfigure}[b]{0.45\textwidth}
		\centering
		\includegraphics[width=\textwidth]{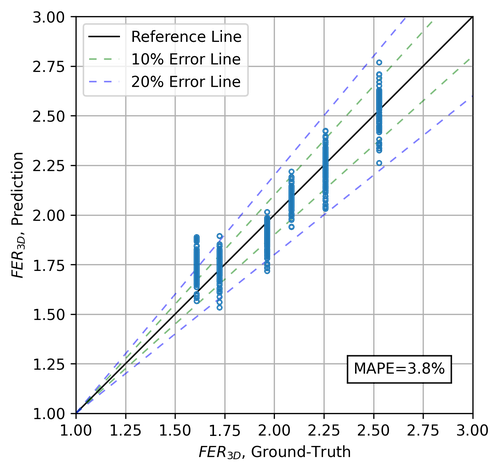}
		\caption{3D FER}
	\end{subfigure}
	\hfill
	\begin{subfigure}[b]{0.45\textwidth}
		\centering
		\includegraphics[width=\textwidth]{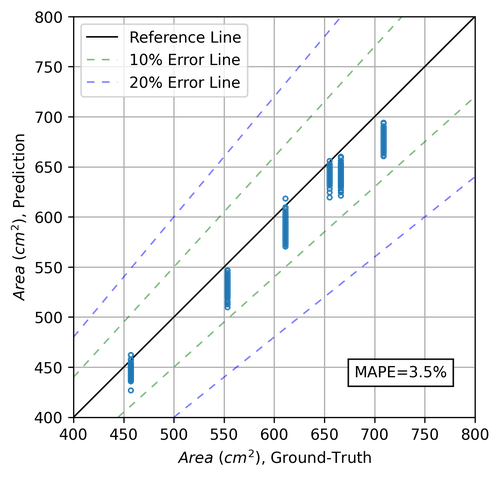}
		\caption{Surface Area}
	\end{subfigure}
	\newline 
	\begin{subfigure}[b]{0.45\textwidth}
		\centering
		\includegraphics[width=\textwidth]{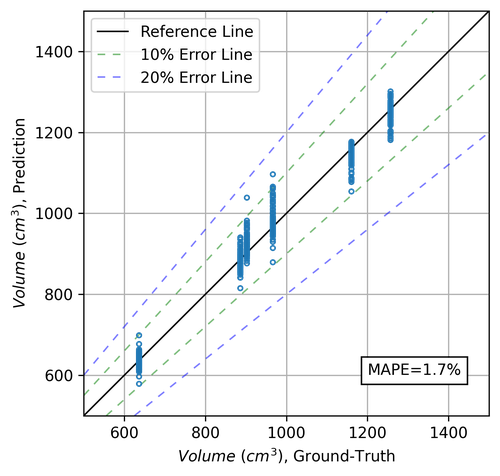}
		\caption{Volume}
	\end{subfigure}
	
\end{figure}
\clearpage 

\section{Summary}

This chapter reviewed both traditional and most recent computer vision approaches for the 3D shape completion task and selected a state-of-the-art strategy that is applicable to learning irregular aggregate shapes. To generate partial-complete shape pairs for deep learning, the varying-visibility and varying-view raycasting schemes were developed and an aggregate shape completion dataset containing 9,000 training pairs, 184 validation pairs, and 672 test pairs was prepared. The selected deep learning framework was implemented and trained on the partial-complete shape pairs to learn the shape completion of aggregates. Multi-dimensional evaluation metrics were used to validate the performance of the shape completion network, from micro-scale metrics (per-point deviation) to macro-scale metrics (morphological properties). As a result, the shape completion network demonstrated good performance on both novel views of the known shapes as well as completely unseen shapes. In the next chapter, the end-to-end reconstruction-segmentation-completion pipeline will be established and tested in the field.

%% file: chapter10.tex
\chapter{Field Application and Validation of the 3D Reconstruction-Segmentation-Completion Framework} \label{chapter-10}

Analyzing the morphological properties (i.e., size and shape) of aggregates in a stockpile has always been a very challenging task. State-of-the-practice methods commonly involve prohibitively time-consuming and labor-intensive measurement of individual aggregate particles or rocks and/or rough estimates based on visual inspection. The volumetric reconstruction approach (\cref{chapter-4}) and 2D stockpile segmentation approach (\cref{chapter-5}) developed in previous chapters help to improve the characterization methods for individual aggregates and aggregate stockpile images, respectively. Nonetheless, upon further comparative analyses on 2D and 3D particle morphologies (\cref{chapter-6}), it was noted that 3D representation of aggregates and stockpiles is considered as a more advanced and realistic characterization than 2D images. In this aspect, a series of closely-related developments was conducted regarding the 3D reconstruction of aggregates and stockpiles (\cref{chapter-6}), 3D segmentation of stockpiles (\cref{chapter-7} and \cref{chapter-8}), and 3D shape completion of partial aggregate shapes (\cref{chapter-9}).

This chapter presents the integration of the developed key components as an end-to-end 3D reconstruction-segmentation-completion framework (RSC-3D) that applies 3D reconstruction, 3D stockpile segmentation, and 3D shape completion for the morphological characterization of aggregates in dense stockpiles. Field applications of the framework are demonstrated and tested on re-engineered stockpiles from collected aggregate samples as well as field stockpiles at the quarry. The performance of the 3D segmentation and shape completion is further validated with field data during 3D stockpile analysis. The robustness and reliability of potential applications using this framework are evaluated based on the correctness of predicted engineering properties of the analyzed aggregate stockpiles.

Preliminary outcomes of this chapter are published in \textcite{huang20223d, huang2023riprap, tutumluer2022three}. Follow-up studies built on top of the developed framework and methodologies are published in \textcite{luo2023toward, luo2024toward, luo2024towards, ding2024augmented, ding2024augmented-1, ding2024ballast}, specifically in the application domain of railway ballast evaluation.
 
\section{Description of Re-engineered Stockpiles and Field Stockpiles}

The field validation of the integrated framework requires stockpiles with valid ground-truth information. This ground-truth information would preferably be the comprehensive morphological properties (i.e., ESD, 3D FER, volume, surface area, etc.) for a multi-dimensional validation, or only the weight measurement under restricted field conditions. 

For the first comprehensive type of ground-truth, individual aggregates in the stockpiles should be the ones that have been fully reconstructed and analyzed, with 3D models available. To this end, all aggregate samples in the 3D aggregate particle library were used to build different stockpiles. These stockpiles are categorized as ``re-engineered'' stockpiles. On the other hand, several field stockpiles prepared by the collaborated aggregate producers were also used for validating the integrated framework. For these stockpiles, due to the restricted activities at the quarry site for individually reconstructing aggregate samples or collecting heavy-weight samples for reconstruction in laboratory, only weight measurement has been available as ground-truth validation.

One major difference between the re-engineered stockpiles and the field stockpiles is whether the aggregate sources have been used during the training of the 3D segmentation and shape completion networks. Aggregate source of the re-engineered stockpiles is the 3D aggregate particle library, which was used extensively in generating the datasets for the instance segmentation and the shape completion tasks. Although no over-fitting effect was observed during the development of the two networks, it is a completely valid concern that the networks may perform better on the re-engineered stockpiles and generalize poorly towards unseen data, i.e., field stockpiles. Therefore, the ground-truth validation on the field stockpiles plays a crucial rule in evaluating the generalization ability of the developed framework.

\subsection{Re-engineered Stockpiles}

\begin{figure}[!htb]
	\centering
	\includegraphics[width=\textwidth]{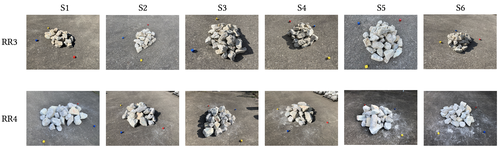}
	\caption{Photos of re-engineered RR3 and RR4 stockpiles}
	\label{fig: photo-reengineer}
\end{figure}

The 3D aggregate particle library contains 46 RR3 rocks and 36 RR4 rocks. Based on these rocks, stockpiles of different size categories were re-engineered at the Advanced Transportation Research and Engineering Laboratory (ATREL) in Rantoul, IL.  
At each time all 46 RR3 aggregate samples were used to build a re-engineered RR3 stockpile, and multi-view images were taken for 3D reconstruction. After the image acquisition step was completed, the aggregate samples were randomly permuted (e.g., rocks buried inside the current stockpile were placed preferably on the surface for the next stockpile) to vary the stockpile configuration. As a result, six re-engineered RR3 stockpiles were built and the multi-view image data was acquired, denoted as stockpiles $S1$ to $S6$. The same process was repeated to build six RR4 re-engineered stockpiles based on the 36 RR4 rocks. To establish correspondences between the segmentation results and the ground truth, IDs of the aggregate samples were marked on multiple faces of the rock, such that the samples could be identified later on from the multi-view image and/or the reconstruction point cloud. The numbering of the samples allows finding the associated ground-truth statistics for detailed comparisons. Recall that the ground-truth information for these re-engineered stockpiles includes complete morphological properties from the particle library. The information of the re-engineered stockpiles is listed as the first two rows in \autoref{tab:num-stockpiles}, and the photos of re-engineered stockpiles are presented in \autoref{fig: photo-reengineer}.

\begin{table}[!htb]
	\centering
	\caption{Information of Re-engineered and Field Stockpiles}
	\label{tab:num-stockpiles}
	\begin{tabular}{cC{5cm}C{3cm}C{3cm}}
		\hline
		\textbf{Size Category} & \textbf{Number of Aggregate Samples in Stockpile} & \textbf{Number of Stockpiles} & \textbf{Ground-Truth} \\ \hline
		RR3 (Re-engineered) & 46 & 6 & Morphological Properties         \\
		RR4 (Re-engineered) & 36 & 6  & Morphological Properties        \\ \hline
		RR3R (Field) & 24 & 3 & Weight Measurement \\
		RR4K (Field) & 16 & 3  & Weight Measurement        \\
		RR5K (Field) & 20 & 3 & Weight Measurement \\ \hline
	\end{tabular}
\end{table}

\subsection{Field Stockpiles}

\begin{figure}[!htb]
	\centering
	\includegraphics[width=\textwidth]{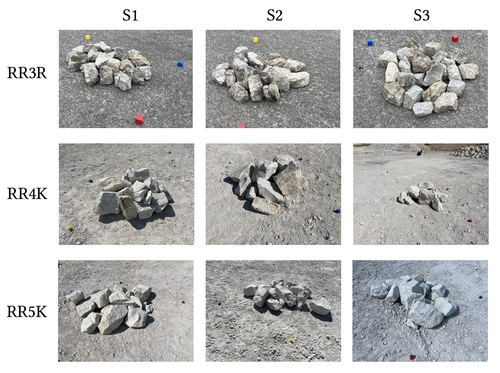}
	\caption{Photos of field RR3R, RR4K and RR5K stockpiles}
	\label{fig: photo-field}
\end{figure}

Field stockpile images were collected during site visits to aggregate quarries at Rantoul, IL and Kankakee, IL. RR3, RR4 and RR5 stockpiles were prepared at these sites manually (for RR3) and by front loader trucks (for RR4 and RR5). At the beginning, 24 RR3 rocks, 16 RR4 rocks and 20 RR5 rocks were selected in the field. Similar to the re-engineered stockpiles, numbers were marked on many faces of each rock, and weight measurement was performed to obtain the ground-truth data. Then, the front loader truck moved the aggregate rocks to form a stockpile, and permute the stockpile after the multi-view images were collected. This process was repeated for three times in each category. To distinguish from the re-engineered stockpiles, these stockpiles are denoted as RR3R, RR4K and RR5K with letters ``R'' and ``K'' indicating the source locations of the field stockpiles. The information of the field stockpiles is listed as the last three rows in \autoref{tab:num-stockpiles}, and photos of field stockpiles are given in \autoref{fig: photo-field}.

\section{3D Reconstruction of Aggregate Stockpiles with Scale Reference}

Using a similar 3D reconstruction approach previously discussed in \cref{chapter-6}, the aggregate stockpiles can be reconstructed based on multi-view stereo photography. Different from the previous approach that is specially designed to obtain a complete model (i.e. two-side reconstruction) of aggregates, the reconstruction of an aggregate stockpile only requires one pass of 3D reconstruction from the multi-view images collected by walking around the stockpile. The object markers are not needed for this one-side reconstruction either. The background markers, though, are still necessary for providing the scale reference as GCPs. 

\begin{figure}
	\centering
	\begin{subfigure}[b]{0.3\textwidth}
		\centering
		\includegraphics[height=6cm]{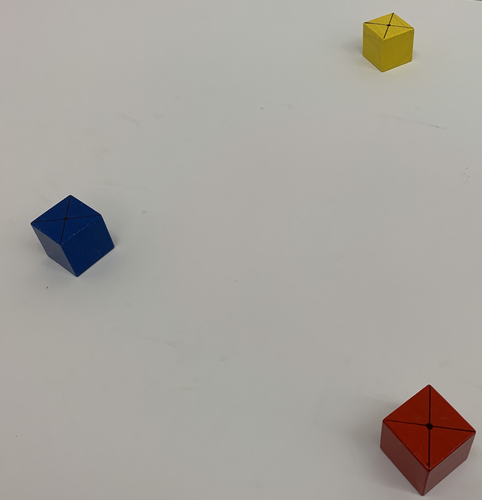}
		\caption{}
		\label{fig: marker-a}
	\end{subfigure}
	\hfill
	\begin{subfigure}[b]{0.65\textwidth}
		\centering
		\includegraphics[height=6cm]{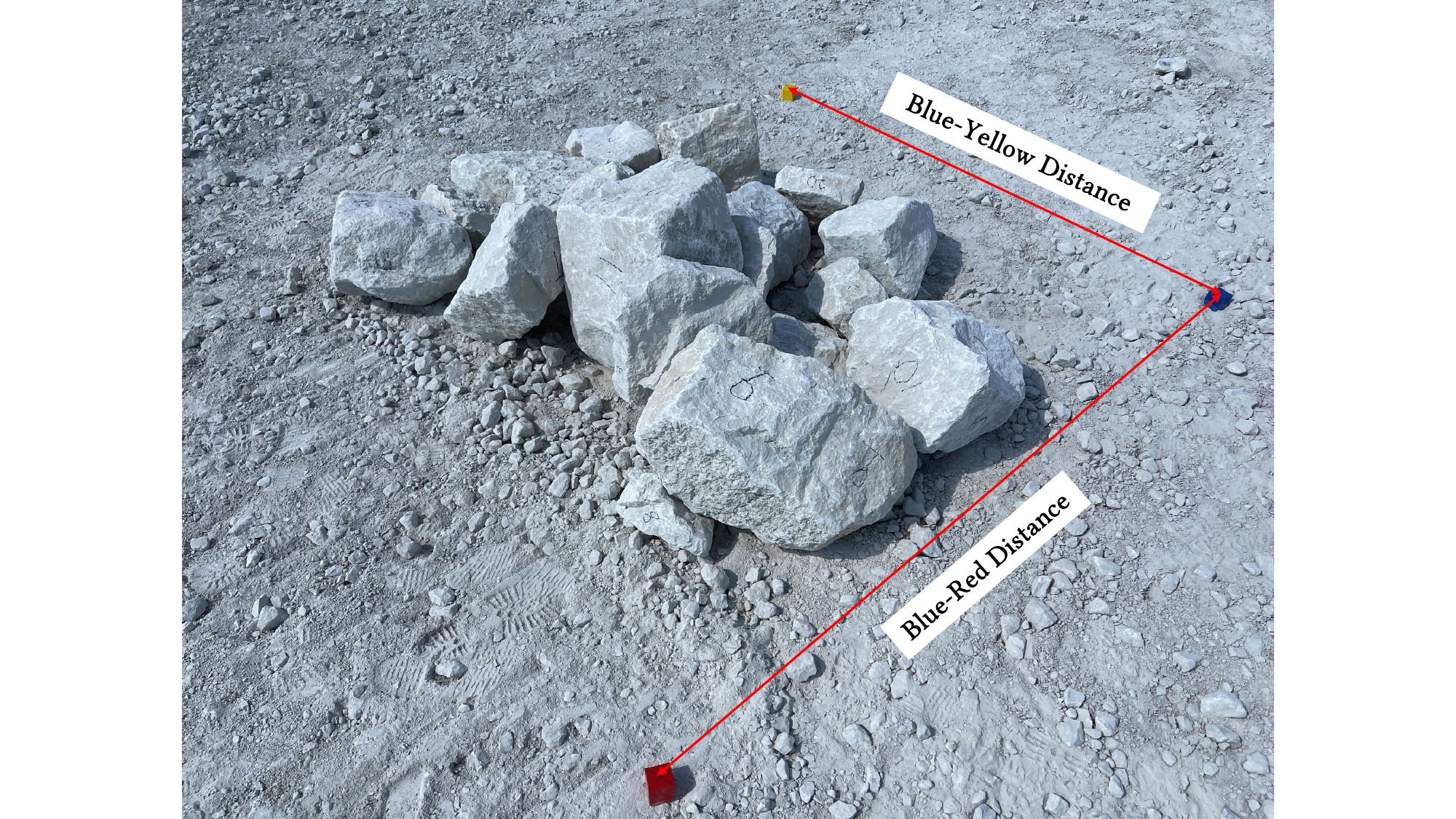}
		\caption{}
		\label{fig: marker-b}
	\end{subfigure}
	\caption{Field marker system for scale reference}
	\label{fig: marker}
\end{figure}

Under the field condition where the aggregate stockpile could vary in size, the previous design of a fixed-distance marker system no longer applies. To address this issue, a new marker system was designed to provide a flexible scale reference in the field. The marker system consists of three colored blocks with red, blue, and yellow colors, as shown in \autoref{fig: marker-a}. The top surface of the colored block was marked with a cross sign intersecting at a center point, which can be conveniently identified in an image. \autoref{fig: marker-a} demonstrates the use of the marker system during the 3D reconstruction approach. Before taking the multi-view images, the marker system is placed near the stockpile to form an angle. Note that the principle of using the marker system as GCPs is to form a plane that can be localized in the reconstruction coordinate space, therefore the three markers should not be co-linear (i.e., approximately lie on the same line). Next, the distances between the markers are measured. In the field experiments, the blue block was used as a pivot marker, and the blue-red and the blue-yellow distances were measured. During the 3D reconstruction approach, by identifying the marker pixel location on a subset of multi-view images and taking the two measured distances as inputs, the reconstructed point cloud of the stockpile can be accurately resized to match the real-world scale.

\begin{figure}[!htb]
	\centering
	\includegraphics[width=0.9\textwidth]{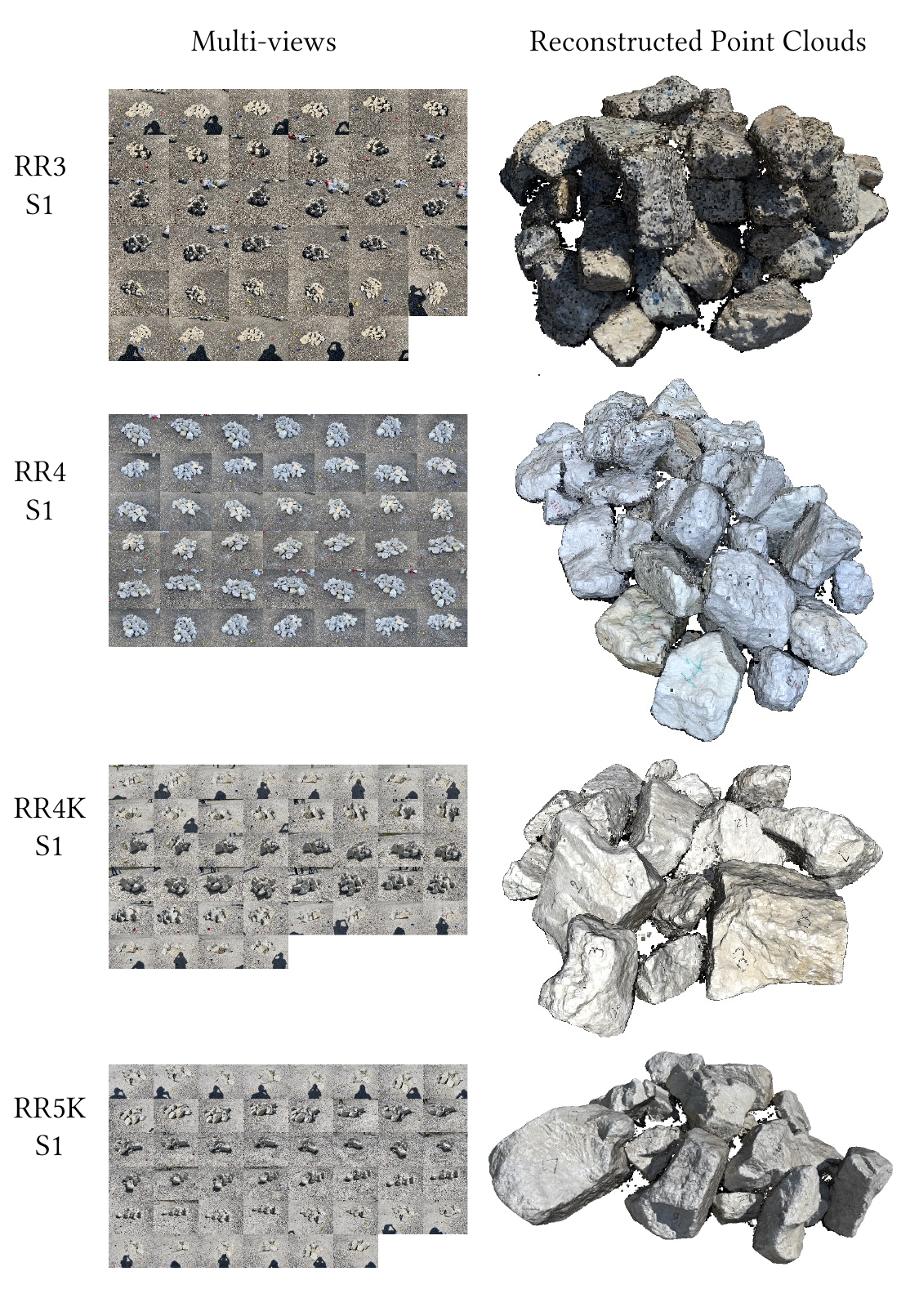}
	\caption{3D reconstruction results of stockpiles in different categories}
	\label{fig: reconstruction}
\end{figure}

Examples of the 3D reconstruction results are presented in \autoref{fig: reconstruction} for S1 stockpiles in each category. The ground surface was also reconstructed along with the stockpile but was manually removed as a pre-processing operation for the 3D segmentation step. Depending on the operator, the number of multi-view images collected per stockpile ranges from 26 images to 50 images. Note that the reconstructed clouds are of consistently high quality for the presented results and for other stockpiles. Based on this practice, around 36 to 50 multi-view images are considered as the recommended number of images to be collected for an all-around inspection of aggregate stockpiles.

\section{3D Stockpile Segmentation and Aggregate Shape Completion Based on Deep Learning}

The 3D reconstructed point clouds of the aggregate stockpiles were then used as the input to the 3D instance segmentation network with the following pre-processing steps. First, the color information was suppressed for the point cloud, since the customized 3D instance segmentation network was designed to be less affected by varying aggregate colors and focused more on the geometry features of a stockpile. Second, the point cloud density was sampled to be consistent with the training configuration. The dense point clouds from the 3D reconstruction was uniformly downsampled to an approximate density of 25,000 points per square meter.

\begin{figure}[!htb]
	\centering
	\begin{tabular}{C{2cm}C{4.5cm}C{4.5cm}C{4.5cm}}
		Stockpile ID & Input Point Cloud $\mathbb{P}$ & Shifted Coordinates $\mathbb{Q}$  & Segmented Instances \\
		RR3-S1 & 
		\includegraphics[trim=0 50 0 200, clip, height=4cm]{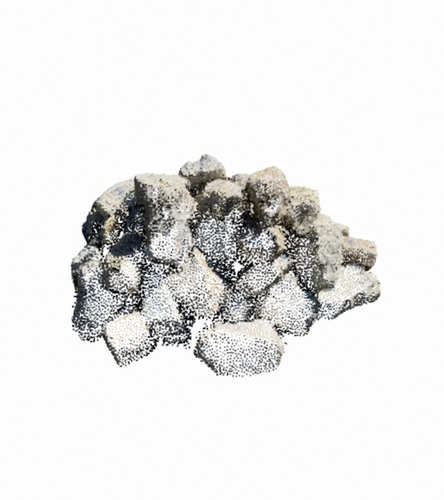} &
		\includegraphics[trim=0 50 0 200, clip, height=4cm]{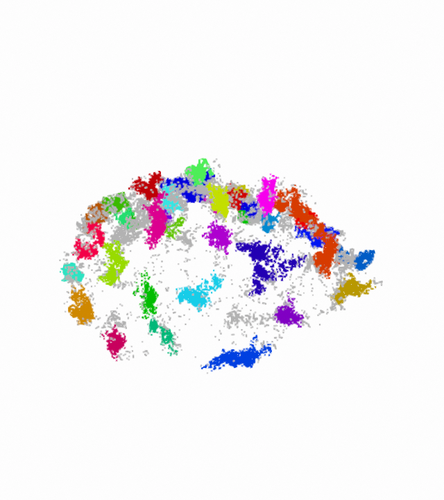} &
		\includegraphics[trim=0 50 0 200, clip, height=4cm]{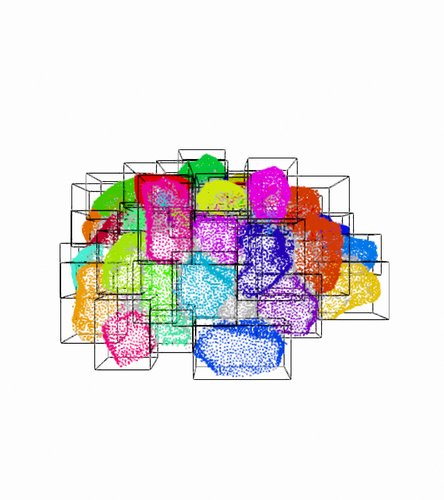} \\
		RR4-S1 & 
		\includegraphics[trim=0 0 0 200, clip, height=4cm]{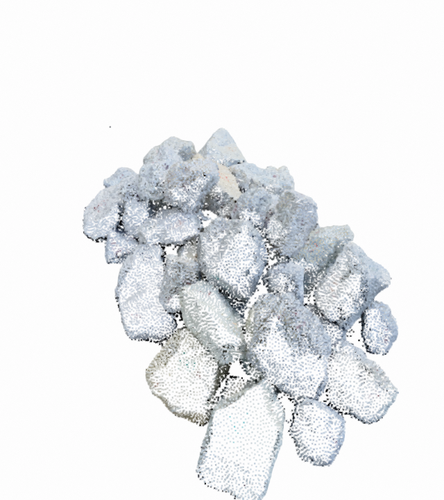} &
		\includegraphics[trim=0 0 0 200, clip, height=4cm]{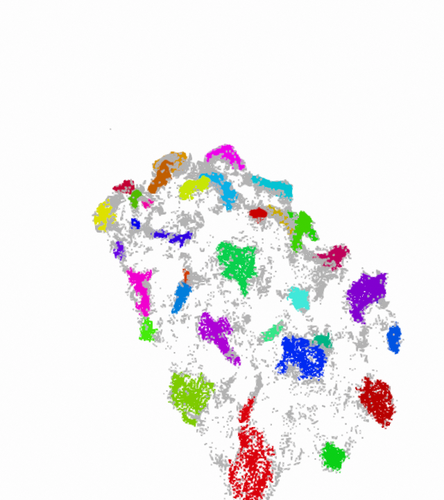} &
		\includegraphics[trim=0 0 0 200, clip, height=4cm]{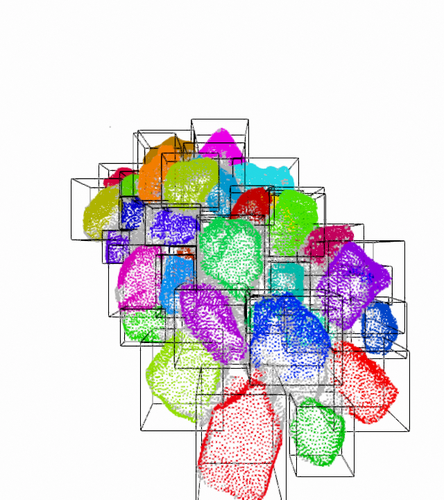} \\
		RR4K-S1 & 
		\includegraphics[trim=0 50 0 80, clip, height=4cm]{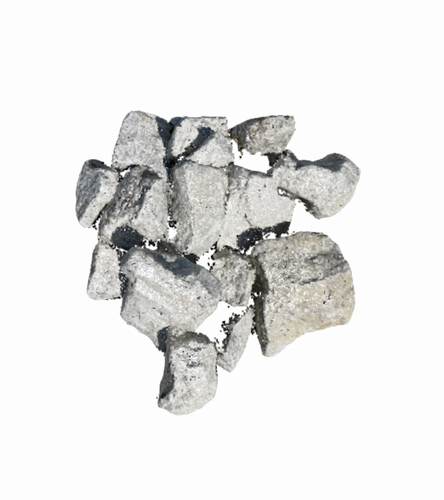} &
		\includegraphics[trim=0 50 0 80, clip, height=4cm]{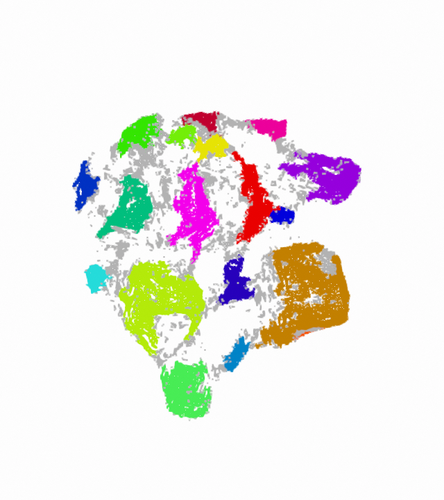} &
		\includegraphics[trim=0 50 0 80, clip, height=4cm]{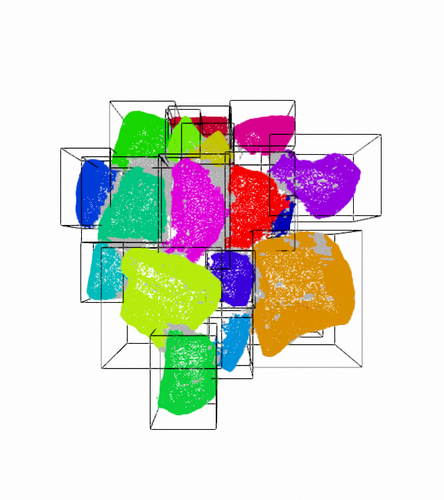} \\
		RR5K-S1 & 
		\includegraphics[trim=0 50 0 50, clip, height=4cm]{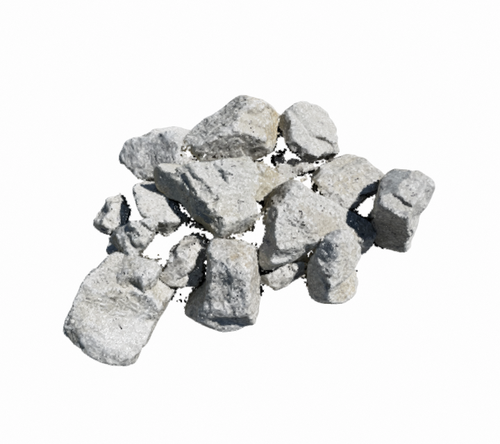} &
		\includegraphics[trim=0 50 0 50, clip, height=4cm]{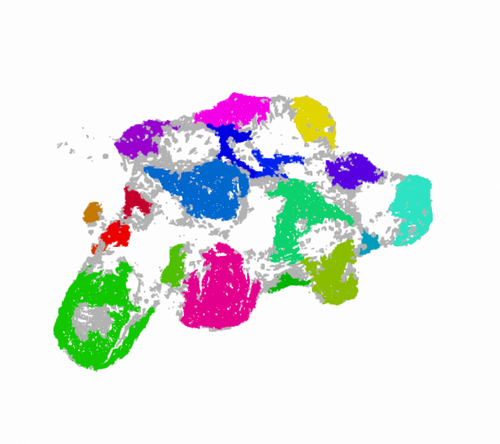} &
		\includegraphics[trim=0 50 0 50, clip, height=4cm]{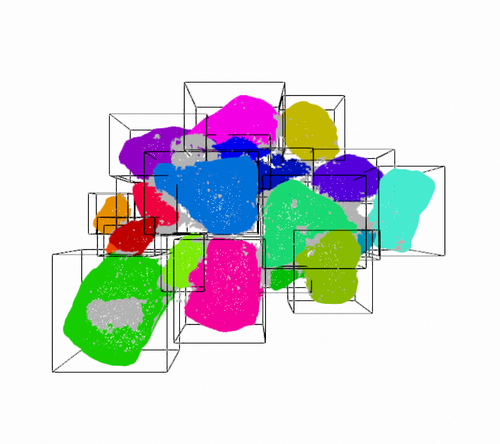} \\
	\end{tabular}
	\caption{3D instance segmentation results of re-engineered and field stockpiles}
	\label{fig:segmentation}
\end{figure}

The segmentation results for S1 stockpiles in each category are illustrated in \autoref{fig:segmentation}. Comparing to the segmentation results on the synthetic dataset (\autoref{fig:gt-results}), the results herein demonstrate a very good generalized performance of the network on real field data. The network conducts effective offset to obtain the shifted coordinates and segments out aggregate instances with reasonable boundaries and bounding boxes. It can also be observed that the results on the field data with completely unseen rocks are still of high quality, although certain cases of under-segmentation and over-segmentation were observed in the field results. 
 
\begin{figure}[!htb]
	\centering
	\begin{tabular}{C{2.2cm}C{2.2cm}C{2.2cm}C{2.2cm}C{2.2cm}}
		Partial Cloud $\mathbb{P}$ & Sparse Cloud $\mathbb{P}_0$ & Rearranged Cloud $\mathbb{P}_1$ & Upsampled Cloud $\mathbb{P}_2$ & Completed Cloud $\mathbb{P}_3$ \\
		\includegraphics[height=1.5cm]{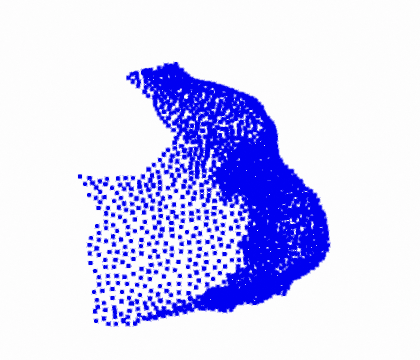} &
		\includegraphics[height=1.5cm]{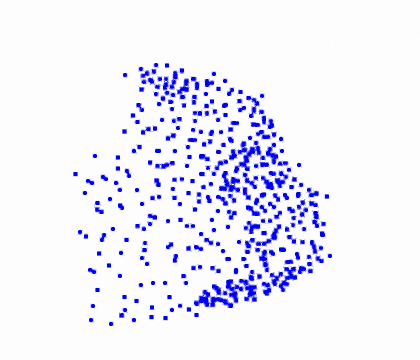}  &  
		\includegraphics[height=1.5cm]{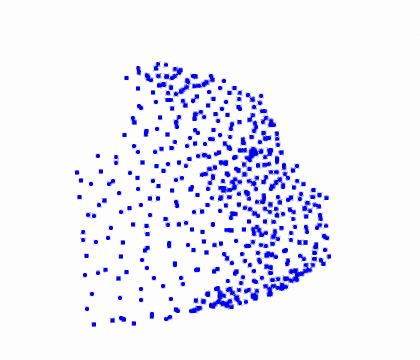}  &  
		\includegraphics[height=1.5cm]{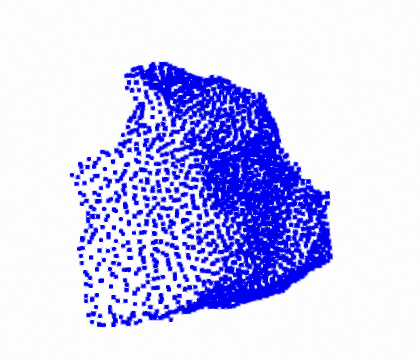}  &  
		\includegraphics[height=1.5cm]{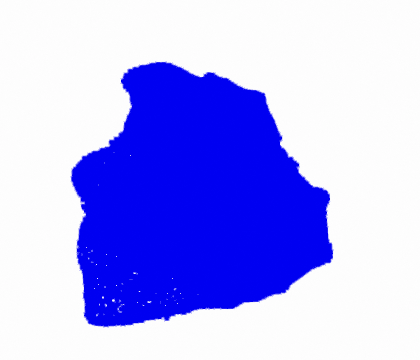}  \\ 
		\includegraphics[height=1.5cm]{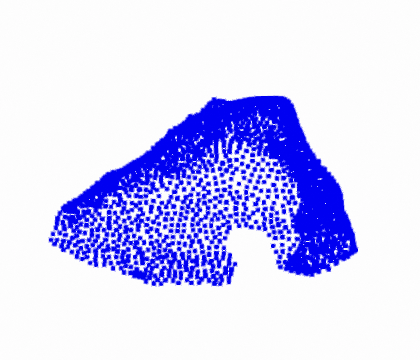} &
		\includegraphics[height=1.5cm]{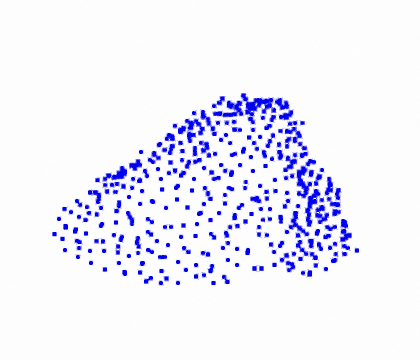}  &  
		\includegraphics[height=1.5cm]{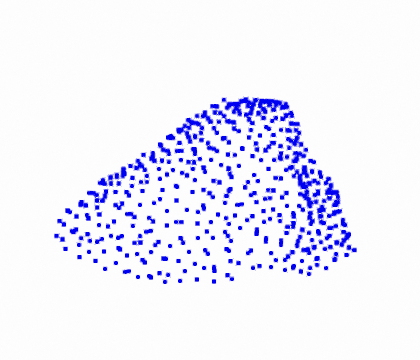}  &  
		\includegraphics[height=1.5cm]{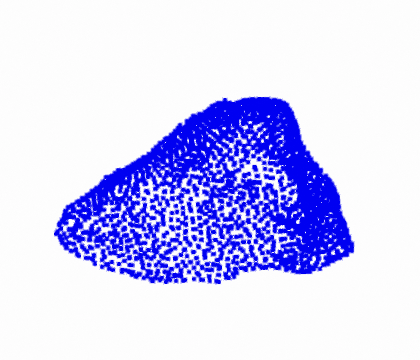}  &  
		\includegraphics[height=1.5cm]{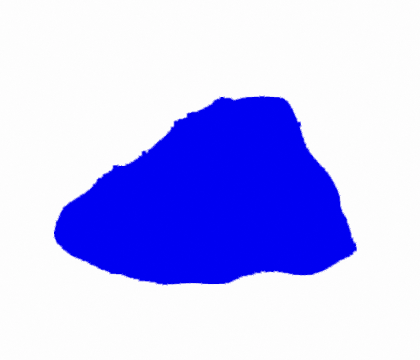}  \\ 
		\includegraphics[height=1.5cm]{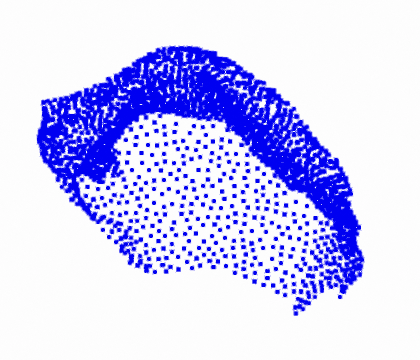} &
		\includegraphics[height=1.5cm]{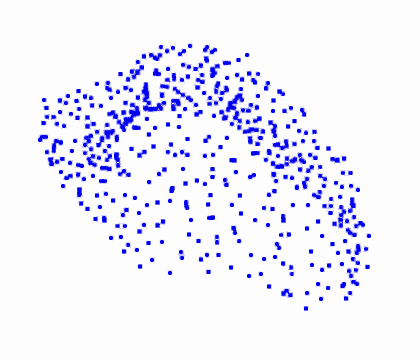}  &  
		\includegraphics[height=1.5cm]{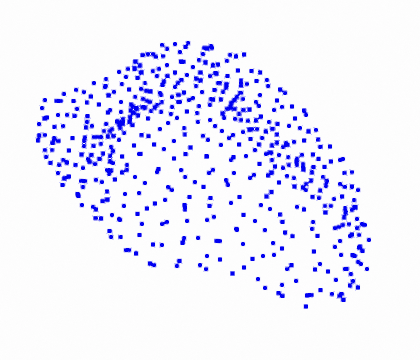}  &  
		\includegraphics[height=1.5cm]{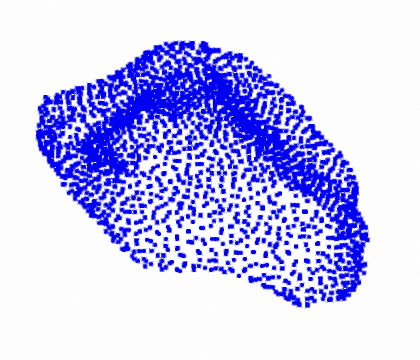}  &  
		\includegraphics[height=1.5cm]{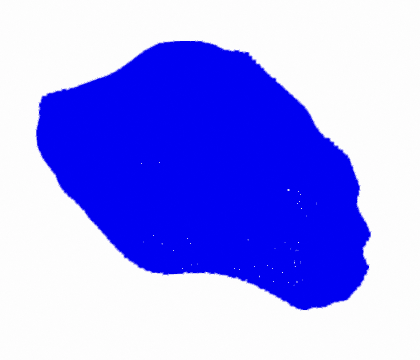}  	
	\end{tabular}
	\caption{Intermediate and final shape completion results for segmented aggregates from the instance segmentation step}
	\label{fig:completion}
\end{figure}

The shape completion results for several representative shapes from the segmentation results of RR3-S1 stockpile are presented in \autoref{fig:completion}. The partial shapes in almost random forms were successfully completed by the shape completion network by first predicting a very reasonable sparse cloud and then progressively completing the cloud with fine details. 

Overall, the integrated reconstruction-segmentation-completion framework works robustly to generate meaningful segmented instances and completed shapes, without exhibiting differences in performance between the prepared synthetic dataset and real field data. This consistency in performance implies the successful use of synthetic stockpile dataset in 3D aggregate stockpile analysis.

\section{3D Morphological Analysis with Ground-Truth Validation}

With the completed shapes of each segmented instance, 3D morphological analyses were conducted and the results were compared against ground-truth for validation. For RR3 and RR4 stockpiles, the ground-truth values are the aggregate morphological properties. For RR3R, RR4K and RR5K stockpiles, the ground-truth is the weight measurement. To obtain the correct mapping between a segmented instance and its ground-truth model, the raw point clouds with color were inspected carefully to first identify the numbers on the aggregate surfaces and then query the 3D particle library for the ground-truth properties. It should be noted that, although the numbers were marked on the aggregate surfaces as clear as possible, it is still very likely that the ground-truth particle ID cannot be recognized from the color point cloud. This is because certain aggregates in the stockpile are highly occluded by the surrounding ones and only a less meaningful portion is visible. In such a non-matching case, the segmented instance is not associated with a ground-truth. The validation was conducted on those aggregates with clearly identifiable numbering.

\subsection{Re-engineered Stockpile Validation with Ground-Truth Morphological Properties}

Comparisons were made between the morphological analysis results of the completed shapes and the known morphological properties of their ground-truth correspondences.The morphological properties involved in the comparison were ESD, shortest/intermediate/longest dimensions, 3D FER, surface area, and volume. The statistics on six RR3 stockpiles were graphed together to observe general trends, as presented in \autoref{fig: rr3-all}. The same validation was given for six RR4 stockpiles in \autoref{fig: rr4-all}.

For the dimensional metrics (i.e., ESD, shortest/intermediate/longest dimensions), \autoref{fig: rr3-all}a to \autoref{fig: rr3-all}d demonstrate reasonably good agreements between the completion results and the ground-truth. The intermediate and longest dimensions have lower MAPE errors (less than $10\%$) which indicates the dimensions of aggregates could be reliably captured from the stockpile analysis using the integrated framework. The shortest dimension is harder to precisely capture since it is sensitive to the shape completion process. \autoref{fig: rr4-all}a to \autoref{fig: rr4-all}d demonstrate similar trends for RR4 stockpiles.

For the shape metrics (i.e., 3D FER), it was observed that relatively high deviations exist which could either over-estimate or under-estimate the FER ratio by predicting based on a partial shape. Although no consistent trend was noticed regarding the FER, the author believes the explanation is two-fold:
\begin{itemize}
	\item First, due to the nature of the shape prediction/completion task, the results are predicted in a probabilistic manner. Despite the fact that 3D approach is able to acquire much more comprehensive geometric information from the stockpile than a 2D approach, the essence of shape prediction is still to find the most likely shape based on prior knowledge of particle shapes. Behind the visible parts, no unique underlying shape exists that is more probable than all other possibilities. The variation in natural aggregate shapes is much higher than common objects (such as cars, chairs, etc.), which makes the shape completion task quite challenging in the first place.
	\item On the other hand, the co-existence of over-estimation and under-estimation is in fact a meaningful feature in terms of shape completion. As previously shown in \autoref{fig: test-results-macro}, the shape completion study on unseen aggregate shapes has predicted a FER range that involves deviation on both sides. In this regard, the author believes this behavior is meaningful and reasonable. In contrast, a consistent over- or under-estimation in terms of the particle shape would be considered problematic for its strong bias on shape prediction.
\end{itemize}

As for the volumetric properties (i.e., volume and surface area), a consistent under-estimation was observed from both RR3 and RR4 stockpiles. The results will be discussed shortly after similar trends are presented for RR3R, RR4K and RR5K stockpiles in the next section. Overall, the integrated framework shows good performance in 3D stockpile analysis and is able to generate meaning morphological results, especially for the aggregate dimensions.

\begin{figure}[!htb] %
	\centering
	\begin{subfigure}[b]{0.45\textwidth}
		\centering
		\includegraphics[width=\textwidth]{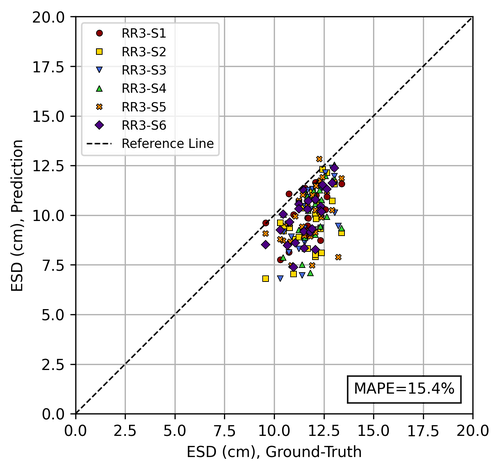}
		\caption{ESD}
	\end{subfigure}
	\hfill
	\begin{subfigure}[b]{0.45\textwidth}
		\centering
		\includegraphics[width=\textwidth]{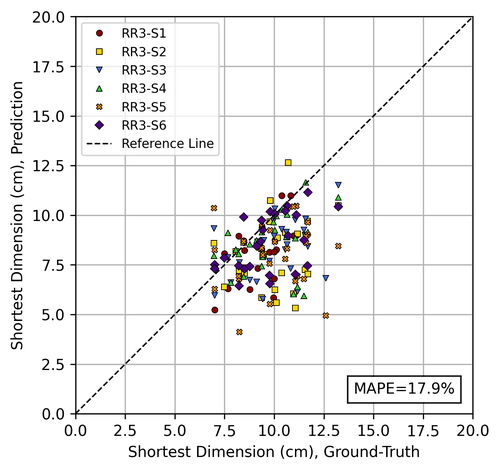}
		\caption{Shortest Dimension}
	\end{subfigure}
	\newline 
	\hfill
	\begin{subfigure}[b]{0.45\textwidth}
		\centering
		\includegraphics[width=\textwidth]{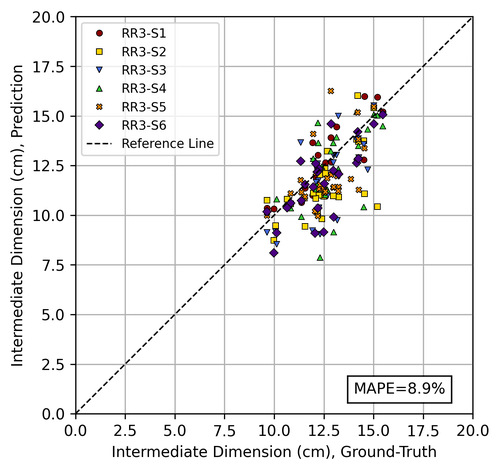}
		\caption{Intermediate Dimension}
	\end{subfigure}
	\hfill
	\begin{subfigure}[b]{0.45\textwidth}
		\centering
		\includegraphics[width=\textwidth]{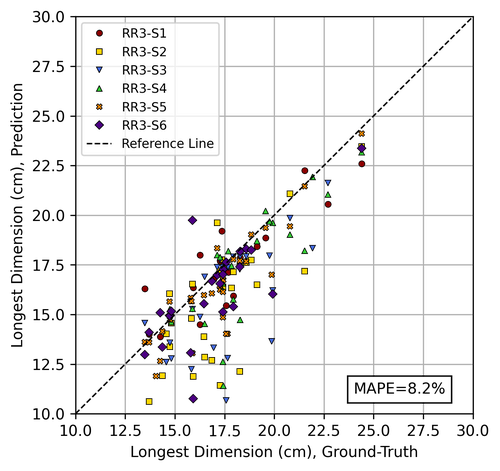}
		\caption{Longest Dimension}
	\end{subfigure}
	\caption{Comparisons of morphological properties between the completed aggregates and ground-truth aggregates for all RR3 stockpiles}
	\label{fig: rr3-all}
\end{figure}

\begin{figure}[t] \ContinuedFloat
	\captionsetup{list=off,format=continued}
	\caption{}
	\begin{subfigure}[b]{0.45\textwidth}
		\centering
		\includegraphics[width=\textwidth]{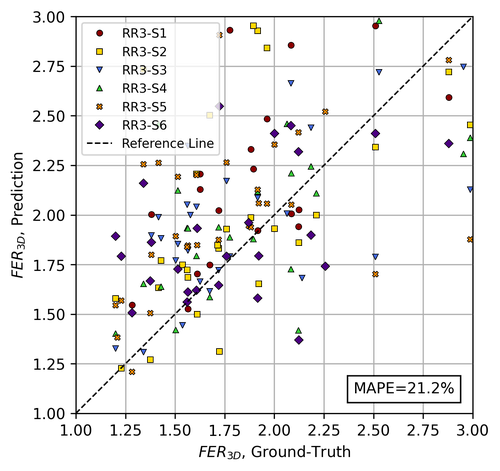}
		\caption{3D FER}
	\end{subfigure}
	\hfill
	\begin{subfigure}[b]{0.45\textwidth}
		\centering
		\includegraphics[width=\textwidth]{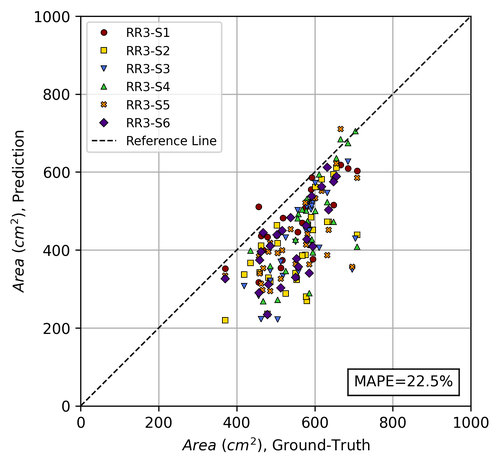}
		\caption{Surface Area}
	\end{subfigure}
	\newline 
	\begin{subfigure}[b]{0.45\textwidth}
		\centering
		\includegraphics[width=\textwidth]{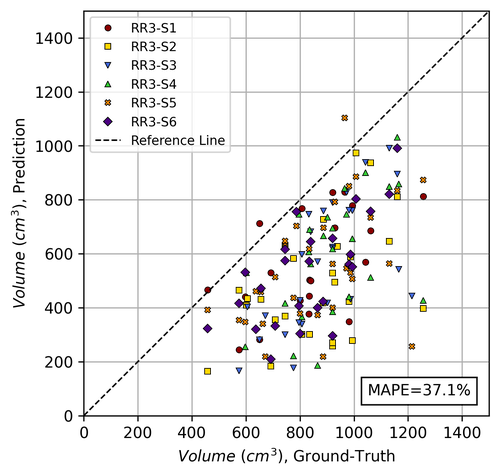}
		\caption{Volume}
	\end{subfigure}
\end{figure}
\clearpage

\begin{figure}[!htb] %
	\centering
	\begin{subfigure}[b]{0.45\textwidth}
		\centering
		\includegraphics[width=\textwidth]{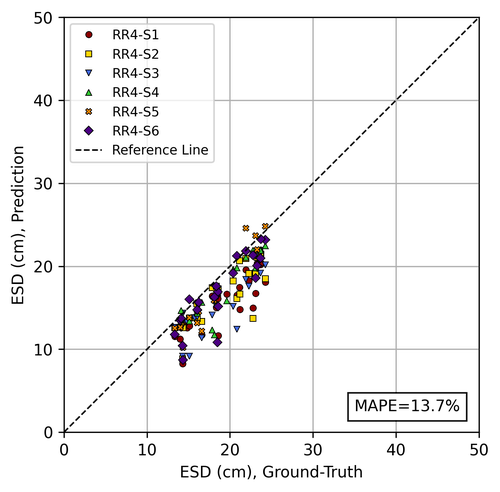}
		\caption{ESD}
	\end{subfigure}
	\hfill
	\begin{subfigure}[b]{0.45\textwidth}
		\centering
		\includegraphics[width=\textwidth]{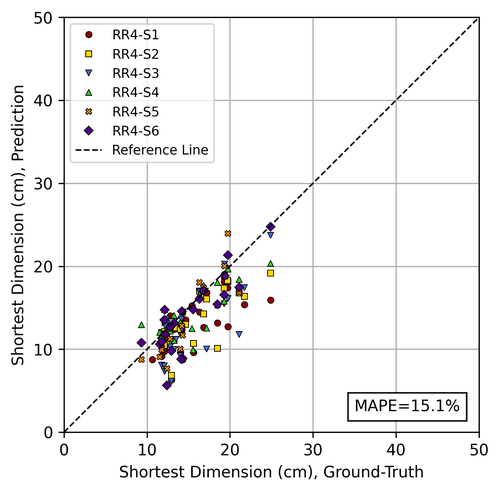}
		\caption{Shortest Dimension}
	\end{subfigure}
	\newline 
	\hfill
	\begin{subfigure}[b]{0.45\textwidth}
		\centering
		\includegraphics[width=\textwidth]{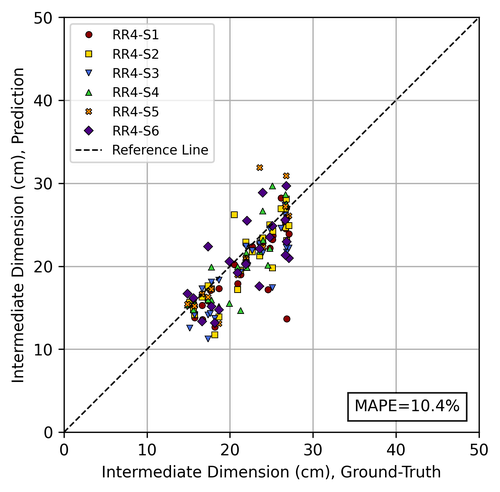}
		\caption{Intermediate Dimension}
	\end{subfigure}
	\hfill
	\begin{subfigure}[b]{0.45\textwidth}
		\centering
		\includegraphics[width=\textwidth]{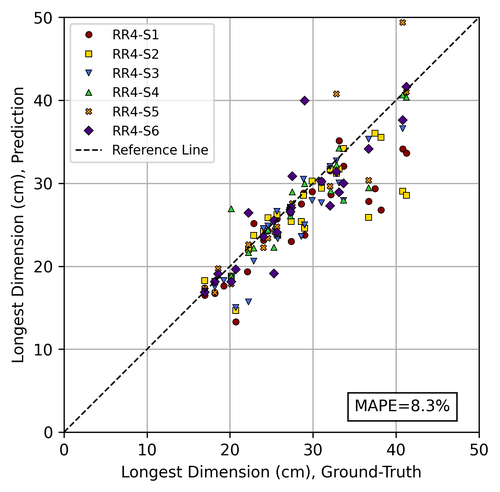}
		\caption{Longest Dimension}
	\end{subfigure}
	\caption{Comparisons of morphological properties between the completed aggregates and ground-truth aggregates for all RR4 stockpiles}
	\label{fig: rr4-all}
\end{figure}

\begin{figure}[t] \ContinuedFloat
	\captionsetup{list=off,format=continued}
	\caption{}
	\begin{subfigure}[b]{0.45\textwidth}
		\centering
		\includegraphics[width=\textwidth]{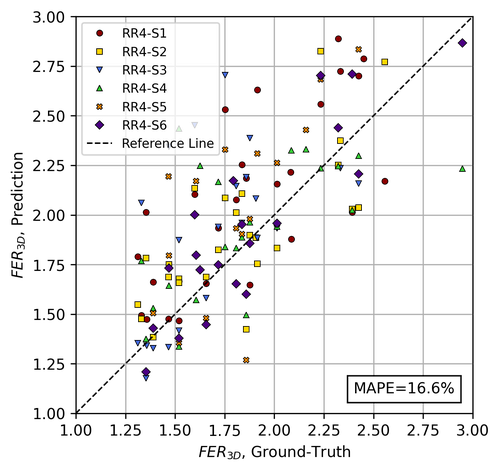}
		\caption{3D FER}
	\end{subfigure}
	\hfill
	\begin{subfigure}[b]{0.45\textwidth}
		\centering
		\includegraphics[width=\textwidth]{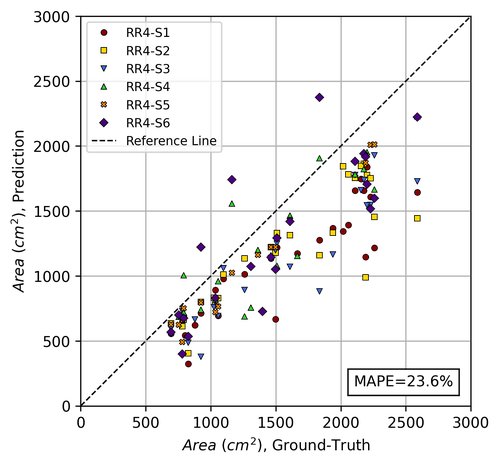}
		\caption{Surface Area}
	\end{subfigure}
	\newline 
	\begin{subfigure}[b]{0.45\textwidth}
		\centering
		\includegraphics[width=\textwidth]{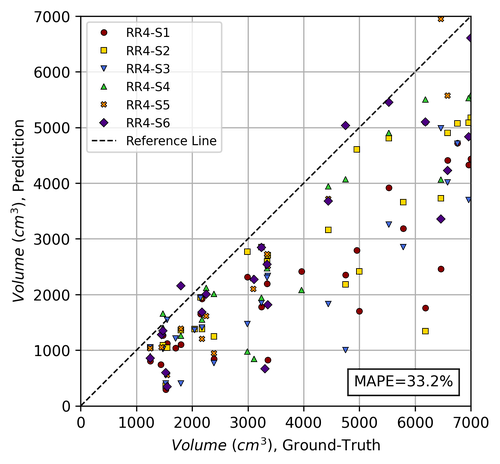}
		\caption{Volume}
	\end{subfigure}
\end{figure}
\clearpage

\subsection{Field Stockpile Validation with Ground-Truth Weight Measurement}

\begin{figure}[!htb]
	\begin{subfigure}[b]{0.45\textwidth}
		\centering
		\includegraphics[width=\textwidth]{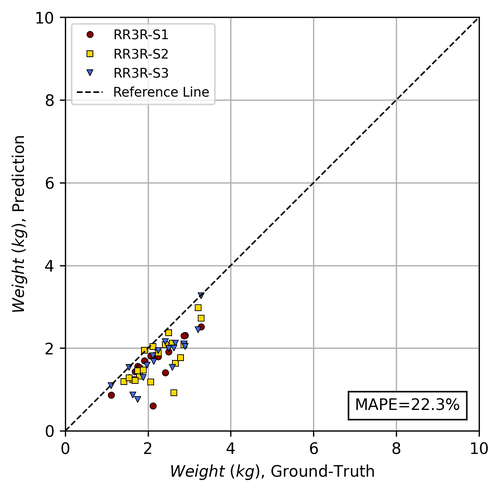}
		\caption{RR3R Weight}
	\end{subfigure}
	\hfill
	\begin{subfigure}[b]{0.45\textwidth}
		\centering
		\includegraphics[width=\textwidth]{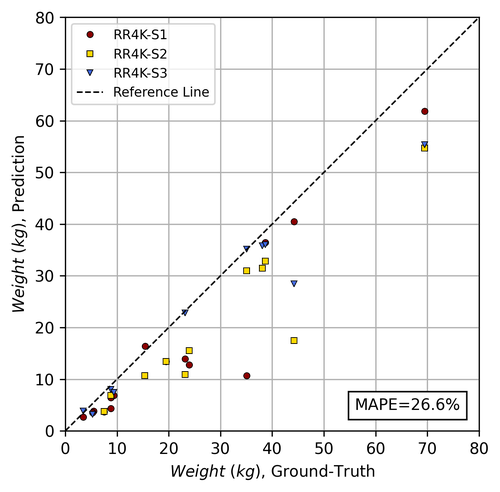}
		\caption{RR4K Weight}
	\end{subfigure}
	\newline
	\begin{subfigure}[b]{0.45\textwidth}
		\centering
		\includegraphics[width=\textwidth]{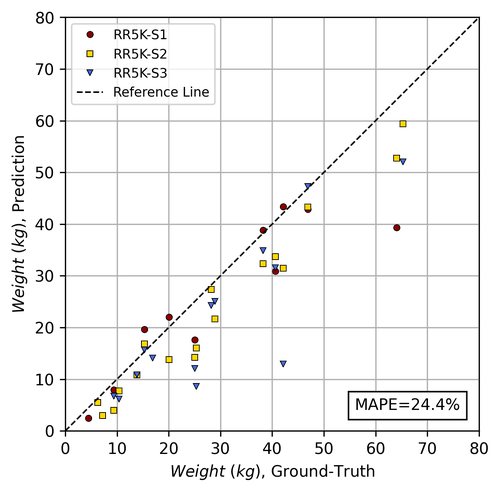}
		\caption{RR5K Weight}
	\end{subfigure}
	\caption{Weight comparisons between the completed aggregates and ground-truth measurements for all RR3R, RR4K and RR5K stockpiles}
	\label{fig: rrk-all}
\end{figure}

For the RR3R, RR4K and RR5K field stockpiles, comparisons were made between the predicted weights and the measured weights. A specific gravity of $G_s=2.65$ was assumed to convert the volume predictions into weight values. As presented in \autoref{fig: rrk-all}, the volume comparisons indicate good and even slightly better results than the RR3 and RR4 results. This may be explained by the fact that the RR3R, RR4K and RR5K are relatively larger rocks (it was noted during the field visit that this batch of RR4K material was larger than normal RR4) thus the degree of occlusion/overlapping in the stockpile is lower than smaller size categories. Note that larger rocks typically have larger void spaces in a stockpile setting, which may lead to better separation during instance segmentation. Also, by comparing the re-engineered stockpiles (\autoref{fig: photo-reengineer}) and field stockpiles (\autoref{fig: photo-field}), it can be observed that the field stockpiles were close to a flat-layered setting, where a larger portion of aggregate surface is visible from multi-view inspection.

The validation results on the field stockpiles further resolve the concern on the generalization ability of the networks used in the integrated framework. The networks have proven to achieve equally well results on stockpiles with known aggregates as well as field stockpiles with unknown properties.

\subsection{Systematic Volume Underestimation} \label{sec: volume-correction}

As shown in \autoref{fig: rr3-all}g, \autoref{fig: rr4-all}g, and \autoref{fig: rrk-all}, a consistent under-estimation of volume predictions was clearly observed. Based on the evidence from all different categories of stockpiles, the author attributes this observation to a systematic deviation due to the following reasons:
\begin{itemize}
	\item The nature of volume prediction in aggregate stockpile analysis. Aggregates in the stockpile form are challenging for volume characterization because volume is a sensitive, high-dimensional physical property. With the principal length dimensions of a shape being the base unit, volume and area are cubic and quadratic quantities of the length dimensions, where the error in estimating the base unit could propagate exponentially to the high-dimensional properties. For example, if the radius of a sphere increases by $20\%$, its volume will change by a much larger percentage of $72.8\%$. Therefore, it should be first acknowledged that volume predictions are likely to carry larger errors than other properties during stockpile analysis. With this fundamental understanding, the sources of error can be analyzed from the observation process and the prediction process.
	
	\item Reconstruction stage (observation error). In the 3D reconstruction stage, the photogrammetry-based multi-view stereo technique follows an optimization mechanism. The reconstruction result is a solution that reaches the best consensus with multi-view observations, which comes with its own error statistics although the reconstruction error is usually relatively small. More importantly, it was observed from experiments that the discrete point cloud does not include certain featureless regions between aggregates. The reconstruction mechanism is based on feature matching. Therefore, those highly occluded or shadowed regions are missing from the observation. In addition, the marker system developed in the framework requires distance measurement. The error in the measurement affects the overall scale of the stockpile and every instance in it.
	
	\item Segmentation stage (observation error). The instance segmentation process tends to generate instance proposals with high confidence, therefore, points that are close to adjacent instances are usually the ones with the highest uncertainty. Hence, the segmentation results are conservative and often a slightly shrunk version of the visible parts.
	
	\item Completion stage (prediction error). As previously discussed, the shape completion process is a probabilistic approach. If the segmentation results are conservative in the first place, the shape completion results are expected to maintain the conservatism during the prediction.
\end{itemize}

\begin{figure}[!htb]
	\begin{subfigure}[b]{0.45\textwidth}
		\centering
		\includegraphics[height=7cm]{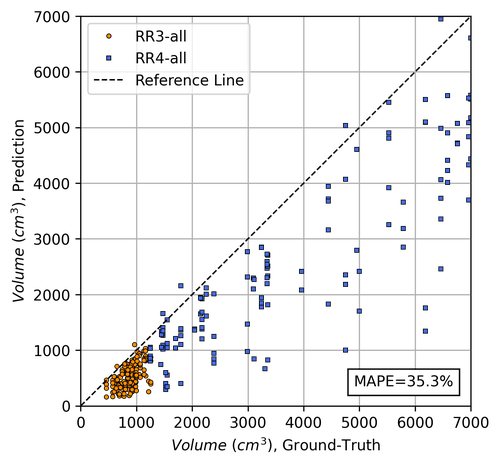}
		\caption{RR3 and RR4 Stockpiles}
	\end{subfigure}
	\hfill
	\begin{subfigure}[b]{0.45\textwidth}
		\centering
		\includegraphics[height=7cm]{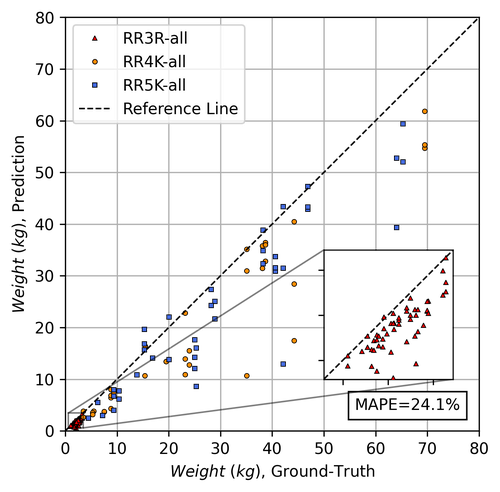}
		\caption{RR3R, RR4K and RR5K Stockpiles}
	\end{subfigure}
	\caption{Systematic underestimation in volume for (a) RR3 and RR4 stockpiles and (b) RR3R, RR4K and RR5K stockpiles}
	\label{fig: correction-stockpile}
\end{figure}

Based on the above discussion, a potential systematic deviation can be further investigated by quantifying the underestimation effect. The volume comparison of the RR3/RR4 stockpiles and RR3R/RR4K/RR5K stockpiles were graphed as two independent groups considering that their stockpile settings are slightly different. The results are presented in \autoref{fig: correction-stockpile}, with enlarged regions for field stockpiles because these aggregates span a large weight range. Regression analysis was conducted on data points within each group and the regression results indicate the systematic underestimation is around $70\%$ for the re-engineered stockpiles and $80\%$ for the field stockpiles. This generally indicates the true volumes are most likely to be $40\%$ to $30\%$ greater than the predicted volumes from the reconstruction-segmentation-completion approach, respectively. 

The two types of stockpile settings (i.e., re-engineered stockpiles in densely-stacked form and field stockpiles in flat-layered form) both demonstrate a systematic underestimation in volume prediction yet to different degrees. For engineering use, however, it is not very practical to clearly distinguish the stockpile forms and determine a case-specific correction factor for the volume estimation. Therefore, it is deemed necessary to further investigate the essential causes of such differences and improve the results by conducting more advanced morphological analyses, as described in the next section.

\subsection{Improvement of Morphological Analysis Results Using Shape Percentage Thresholding}
By further investigating the aggregate shape characteristics in a stockpile, it was observed that the particle aggregate shapes usually exhibit very different visibility levels from a stockpile observation. For example, particles in a flat-layered stockpile setting tend to have better visibility due to larger void space and less occlusion between adjacent aggregates; meanwhile particles are likely to have lower visibility in a densely-stacked stockpile setting. 

In this regard, a quantitative method of characterizing the field ``visibility'' of aggregate shapes was developed. The intuition behind this visibility concept comes from the observation that partial and complete aggregate shapes demonstrate great differences in their spatial occupation patterns. A complete shape is a watertight surface such that a ray originated from the centroid will hit the enclosed surface in any arbitrary direction. However, for partial shapes, the rays originated from the centroid will either hit (for existing regions) or miss (for missing regions). 

The field visibility of aggregate shapes is calculated using a modified ray casting scheme similarly to the one previously described in \cref{chapter-9}. By calculating the ratios between the number of hit rays and the total number of cast rays, a visibility indicator named Shape Percentage (SP) was developed to quantify the partial shape observation, as described below:
\begin{itemize}
	\item Step 1: Initializing a directional sphere at the centroid of the aggregate. Note that the centroid is approximated as the centroid of the partial shape, which may not be exactly at the centroid of the true shape but is the best-possible prediction based on partial observations. Then, a directional sphere with $N=1,000$ equally distributed surface vertices is created at the centroid.
	
	\item Step 2: Raycasting for shape surface intersection. For each vertex on the sphere surface, the directional vector from the centroid to the vertex coordinates forms a ray direction. The ray-surface intersection is then conducted to indicate if the current direction contains a valid shape surface. If the ray hits the surface, the number of ray hits increments; otherwise, this direction represents a missing region.
	
	\item Step 3: Calculating the SP value. After completing ray casting for all $N=1,000$ directional vectors, the SP value is calculated by the ratio between the number of ray hits and the total number of rays.
\end{itemize}

\begin{figure}[!htb]
	\centering
	\includegraphics[height=9cm]{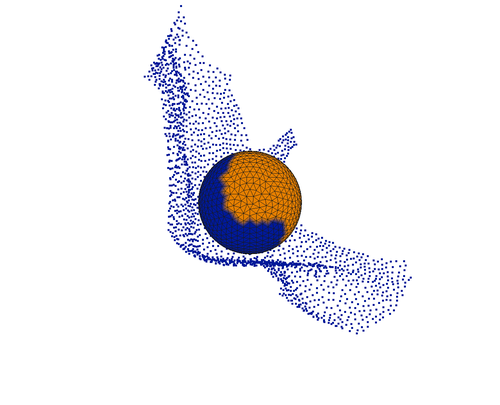}
	\caption{Shape Percentage (SP) concept of a partial shape}
	\label{fig: shape-percentage-concept}
\end{figure}

The demonstration of the SP concept is presented in \autoref{fig: shape-percentage-concept}. The blue region on the directional sphere represents the directions that have ray hits with the partial surface, while the orange region illustrates the missing space from the partial observation.

\begin{figure}[!htb]
	\centering
	\includegraphics[height=9cm]{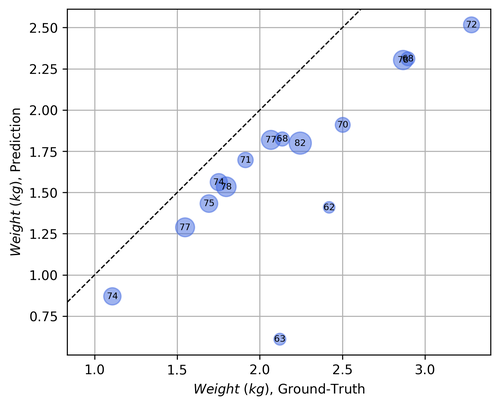}
	\caption{Shape Percentage (SP, in percentage) analysis of RR3R-S1 stockpile results}
	\label{fig: shape-percentage-rr3r}
\end{figure}

Based on the SP indicator, the segmentation and completion results can be interpreted from a more effective perspective for all segmented shapes in the stockpile. An example analysis on RR3R-S1 stockpile is shown in \autoref{fig: shape-percentage-rr3r}. The size of the data points is proportional to the SP value of each partial shape after segmentation. More intuitively, the SP values (in percentage) are directly labeled on each data point. 

As can be observed in \autoref{fig: shape-percentage-rr3r}, most of the outliers with high deviation from the ground-truth have relatively low SP values (e.g., below $70\%$). As previously discussed in \cref{chapter-9}, the shape completion process is probabilistic and learning-based. Therefore, the higher the shape visibility (i.e., the more portion of a shape that can be observed), the better the reliability and robustness of the shape completion results. When interpreting the results for practical use, it is important to screen and select the effective data, i.e., reliable results with high confidence level in the analysis.

\begin{figure}[!htb]
	\centering
	\begin{subfigure}[b]{0.4\textwidth}
		\centering
		\includegraphics[width=\textwidth]{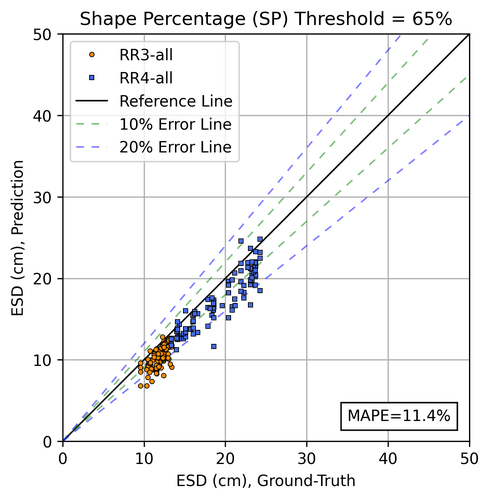}
		\caption{ESD, SP=65\%}
	\end{subfigure}
	\hfill
	\begin{subfigure}[b]{0.42\textwidth}
		\centering
		\includegraphics[width=\textwidth]{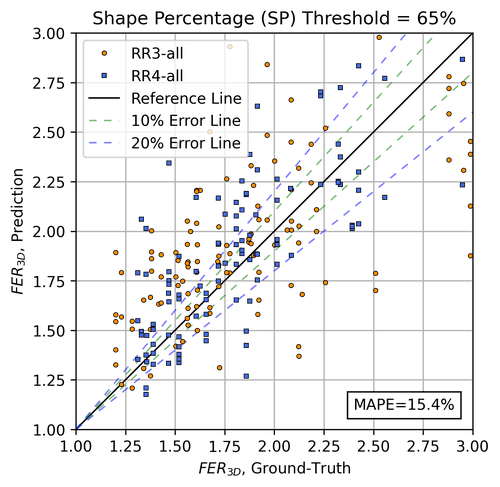}
		\caption{3D FER, SP=65\%}
	\end{subfigure}
	\newline 
	\begin{subfigure}[b]{0.4\textwidth}
		\centering
		\includegraphics[width=\textwidth]{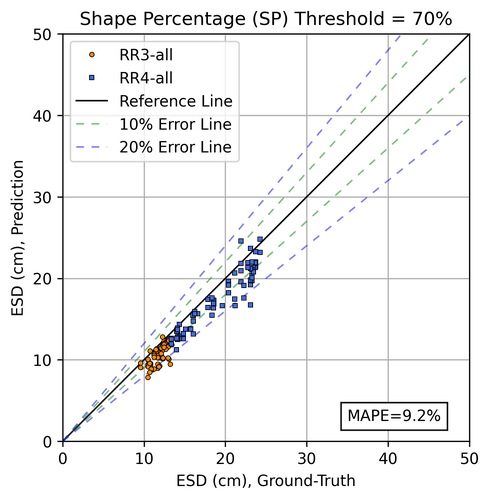}
		\caption{ESD, SP=70\%}
	\end{subfigure}
	\hfill
	\begin{subfigure}[b]{0.42\textwidth}
		\centering
		\includegraphics[width=\textwidth]{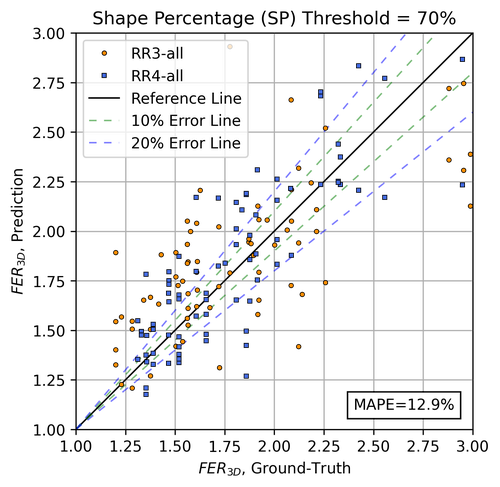}
		\caption{3D FER, SP=70\%}
	\end{subfigure}
	\newline 
	\begin{subfigure}[b]{0.4\textwidth}
		\centering
		\includegraphics[width=\textwidth]{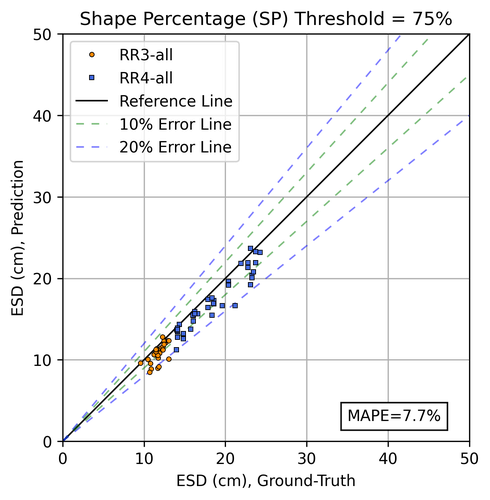}
		\caption{ESD, SP=75\%}
	\end{subfigure}
	\hfill
	\begin{subfigure}[b]{0.42\textwidth}
		\centering
		\includegraphics[width=\textwidth]{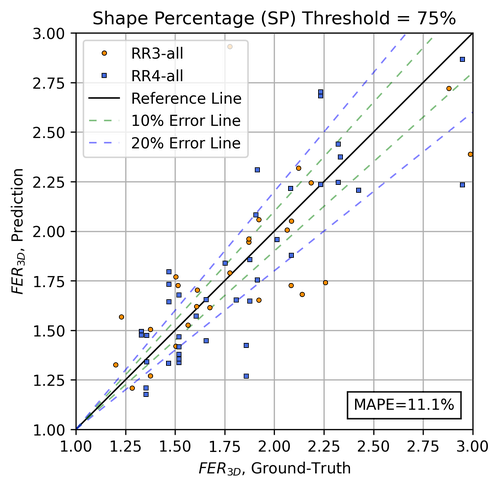}
		\caption{3D FER, SP=75\%}
	\end{subfigure}
	\caption{Effect of SP thresholding on aggregate dimension and shape metrics (ESD and 3D FER) for re-engineered stockpile data}
	\label{fig:sp-size}
\end{figure}

\begin{figure}[!htb] \ContinuedFloat
	\captionsetup{list=off,format=continued}
	\caption{}
	\begin{subfigure}[b]{0.4\textwidth}
		\centering
		\includegraphics[width=\textwidth]{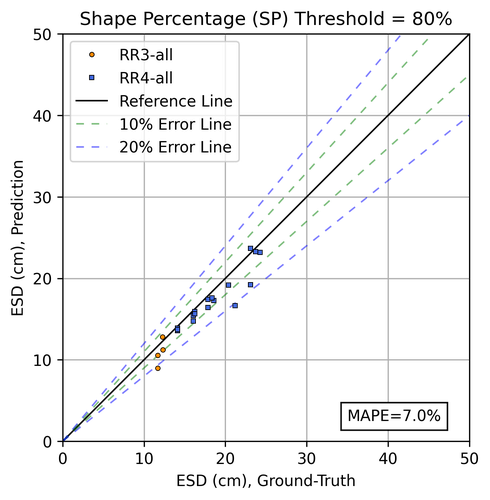}
		\caption{ESD, SP=80\%}
	\end{subfigure}
	\hfill
	\begin{subfigure}[b]{0.42\textwidth}
		\centering
		\includegraphics[width=\textwidth]{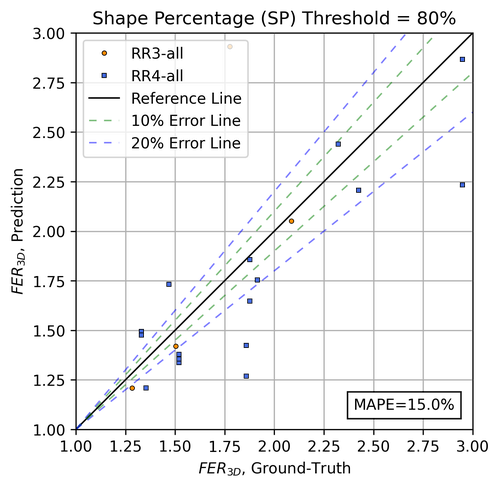}
		\caption{3D FER, SP=80\%}
	\end{subfigure}
\end{figure}

\begin{figure}[!htb]
	\centering
	\begin{subfigure}[b]{0.4\textwidth}
		\centering
		\includegraphics[width=\textwidth]{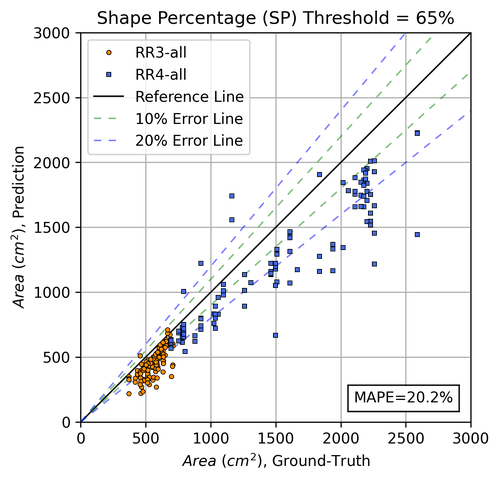}
		\caption{Area, SP=65\%}
	\end{subfigure}
	\hfill
	\begin{subfigure}[b]{0.4\textwidth}
		\centering
		\includegraphics[width=\textwidth]{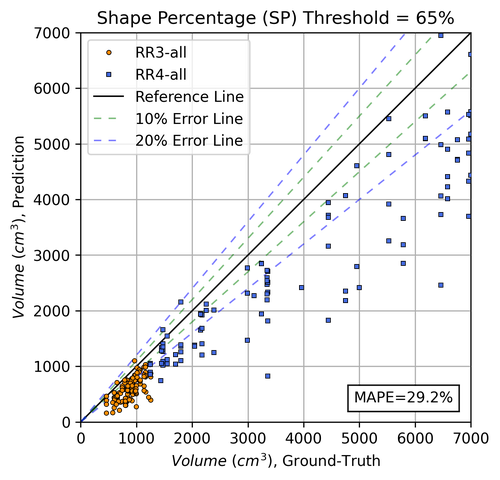}
		\caption{Volume, SP=65\%}
	\end{subfigure}
	\newline 
	\begin{subfigure}[b]{0.4\textwidth}
		\centering
		\includegraphics[width=\textwidth]{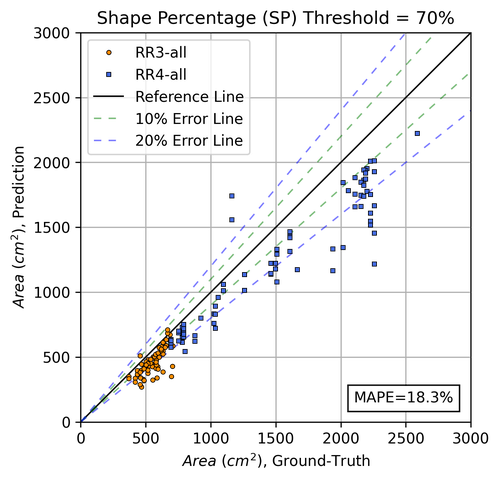}
		\caption{Area, SP=70\%}
	\end{subfigure}
	\hfill
	\begin{subfigure}[b]{0.4\textwidth}
		\centering
		\includegraphics[width=\textwidth]{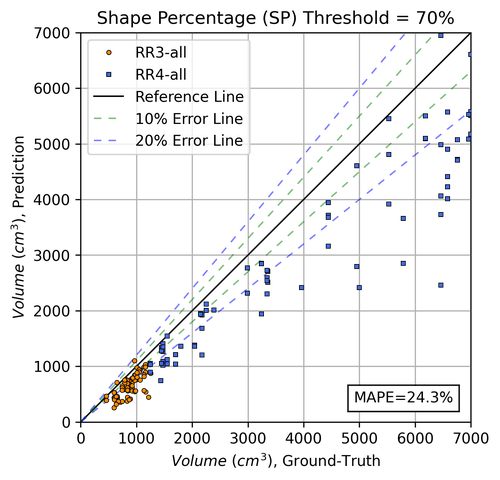}
		\caption{Volume, SP=70\%}
	\end{subfigure}
	\newline 
	\begin{subfigure}[b]{0.4\textwidth}
		\centering
		\includegraphics[width=\textwidth]{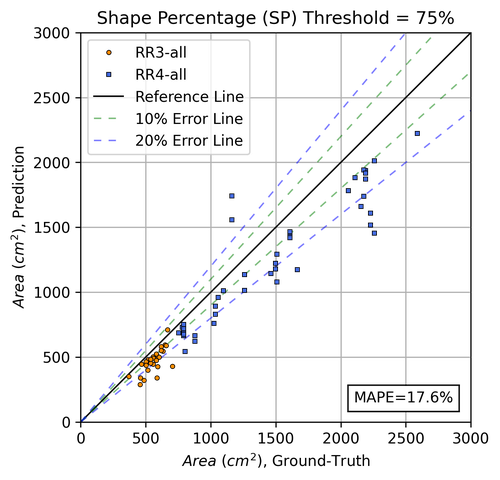}
		\caption{Area, SP=75\%}
	\end{subfigure}
	\hfill
	\begin{subfigure}[b]{0.4\textwidth}
		\centering
		\includegraphics[width=\textwidth]{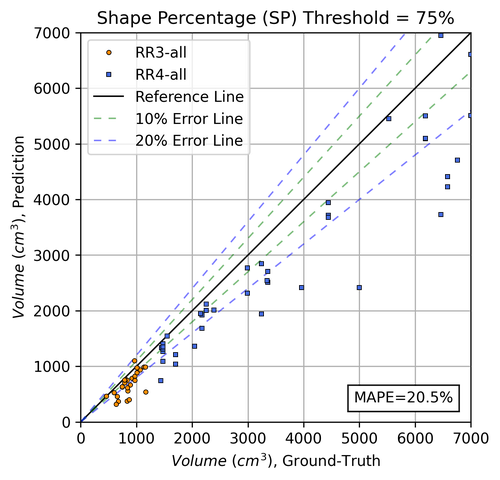}
		\caption{Volume, SP=75\%}
	\end{subfigure}
	\caption{Effect of SP thresholding on high-dimensional metrics (surface area and volume) for re-engineered stockpile data}
	\label{fig:sp-volume}
\end{figure}

\begin{figure}[!htb] \ContinuedFloat
	\captionsetup{list=off,format=continued}
	\caption{}
	\begin{subfigure}[b]{0.4\textwidth}
		\centering
		\includegraphics[width=\textwidth]{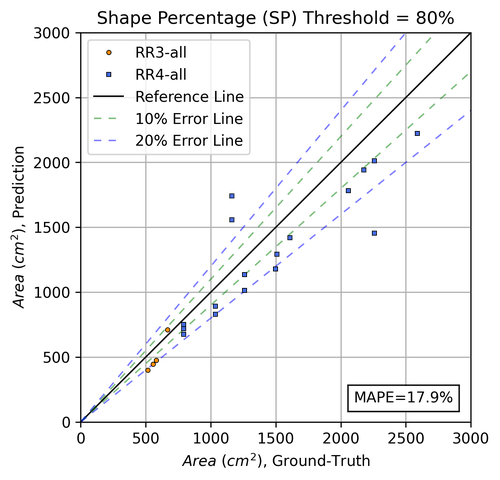}
		\caption{Area, SP=80\%}
	\end{subfigure}
	\hfill
	\begin{subfigure}[b]{0.4\textwidth}
		\centering
		\includegraphics[width=\textwidth]{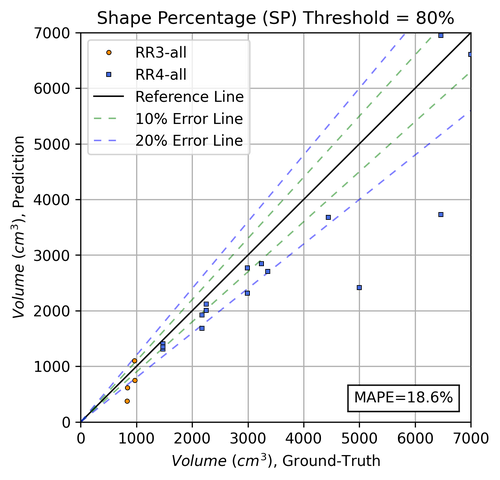}
		\caption{Volume, SP=80\%}
	\end{subfigure}
\end{figure}

\begin{figure}[!htb]
	\centering
	\begin{subfigure}[b]{0.45\textwidth}
		\centering
		\includegraphics[width=\textwidth]{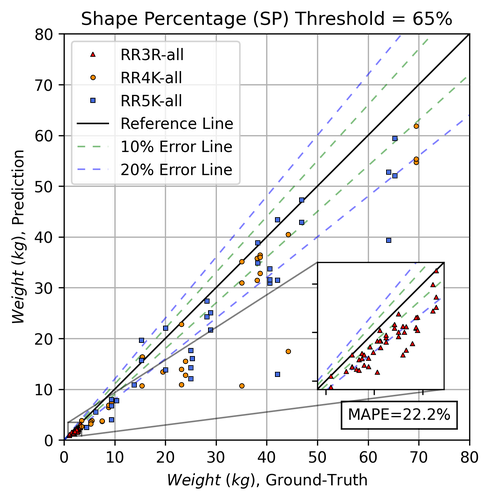}
		\caption{Weight, SP=65\%}
	\end{subfigure}
	\hfill
	\begin{subfigure}[b]{0.45\textwidth}
		\centering
		\includegraphics[width=\textwidth]{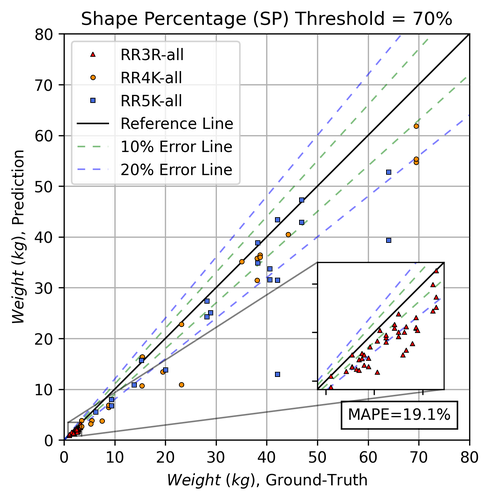}
		\caption{Weight, SP=70\%}
	\end{subfigure}
	\newline 
	\begin{subfigure}[b]{0.45\textwidth}
		\centering
		\includegraphics[width=\textwidth]{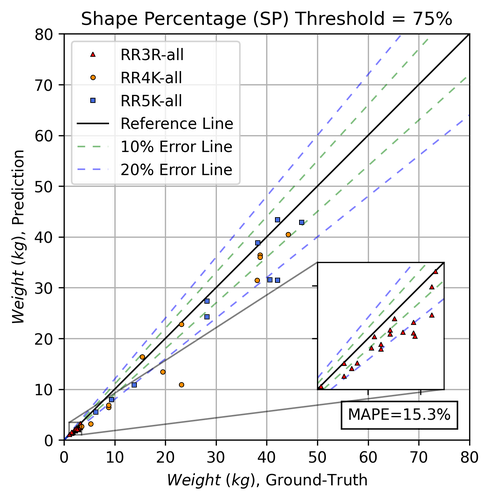}
		\caption{Weight, SP=75\%}
	\end{subfigure}
	\hfill
	\begin{subfigure}[b]{0.45\textwidth}
		\centering
		\includegraphics[width=\textwidth]{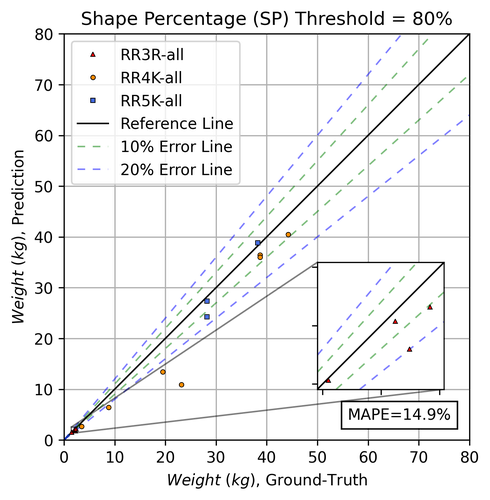}
		\caption{Weight, SP=80\%}
	\end{subfigure}
	\caption{Effect of SP thresholding on high-dimensional metric (weight) for field stockpile data}
	\label{fig:sp-weight}
\end{figure}

Therefore, an SP thresholding process was developed to improve the morphological analyses procedure. Based on the segmented particle shapes from the re-engineered and field stockpile analysis, the most common range of SP values is $60\%$ to $85\%$. Accordingly, a SP threshold series of $\{65\%, 70\%, 75\%, 80\%\}$ was used to investigate the effect of SP thresholding. The morphological analysis results at various SP levels are presented in \autoref{fig:sp-size} and \autoref{fig:sp-volume} for re-engineered stockpile data and in \autoref{fig:sp-weight} for field stockpile data.

\autoref{fig:sp-size} demonstrates the effect of SP thresholding of size dimension and shape metrics (ESD and 3D FER) on re-engineered stockpile data. By comparing the ESD at various SP levels with the raw ESD results in \autoref{fig: rr3-all}a and \autoref{fig: rr4-all}a, it can be observed the MAPE error decreases from around 15\% to less than 8\% (at SP levels of 75\% and 80\%), and the MAPE error of 3D FER (in \autoref{fig: rr3-all}e and \autoref{fig: rr4-all}e) decreases from 19\% to around 11\% (at SP level 75\%). For high-dimensional metrics (surface area and volume), similar improvements over the raw results are noticed by comparing \autoref{fig:sp-volume} to \autoref{fig: rr3-all}f-g and \autoref{fig: rr4-all}f-g. The MAPE error of surface area drops from 23\% to around 18\% (at SP level 75\% and 80\%) and the volume MAPE error is improved significantly from around 35\% to around 20\% (at SP level 75\% and 80\%). For field stockpile data with only weight metric, the comparison between \autoref{fig:sp-weight} and \autoref{fig: correction-stockpile}b shows the MAPE error is decreased from 24.1\% to around 15\% (at SP level 75\% and 80\%). 
\clearpage 

The MAPE statistics clearly indicate that the SP thresholding process effectively improves the results of the RSC-3D framework, in all morphological properties (dimension and shape metrics and high-dimensional metrics). In addition to the MAPE error evaluation (which only gives an average error estimate of all the data points), error bound analysis was also conducted to better reveal the improvements. As shown in \autoref{fig:sp-size}, \autoref{fig:sp-volume} and \autoref{fig:sp-weight}, $\pm 10\%$ and $\pm 20\%$ error lines were added to indicate the range of deviation with respect to ground-truth. It can be observed that the SP thresholding process plays a crucial role in screening most of the less reliable predictions (i.e., those with low SP and limited partial shape observation) and improving the overall confidence level of the morphological analysis. This effect is most remarkable in the 3D FER analysis shown in the second column of \autoref{fig:sp-size}. As compared to \autoref{fig: rr3-all}e and \autoref{fig: rr4-all}e where the raw predicted 3D FERs exhibit a high deviation from the ground-truth, the 3D FER results with SP thresholding dramatically decrease the deviation, which indicates the completed shapes after screening are now much closer to the ground-truth aggregate shapes. This observation coincides perfectly with the intuition of the shape percentage concept. Namely, when an aggregate shape is observed with limited visibility (e.g., SP below 60\%), the shape completion result can only represent one of the best-effort guesses based on the partial observation; conversely, as the partial shape approaches towards a relatively complete observation (e.g., SP over 75\%), the shape completion results may better capture the true aggregate shape with increasing confidence. 

Furthermore, the influence of the SP threshold on the results is two-fold. \autoref{fig:sp-size}, \autoref{fig:sp-volume} and \autoref{fig:sp-weight} show that as the SP threshold increases, usually more outliers with low confidence are screened and the prediction error is decreased. However, it should also be noted that there is a balance between the number of effective data points and the confidence of the results. If the SP value is set relatively high (such as 80\% or higher), the efficiency of the stockpile analysis is likely to be negatively influenced with only few results obtained in a stockpile. By analyzing the results at various SP levels in \autoref{fig:sp-size}, \autoref{fig:sp-volume} and \autoref{fig:sp-weight}, a SP threshold of 75\% is recommended for volume estimation to ensure both a sufficient number of results and improved reliability of RSC-3D framework.

\begin{figure}[!htb]
	\centering
	\begin{subfigure}[b]{0.45\textwidth}
		\centering
		\includegraphics[width=\textwidth]{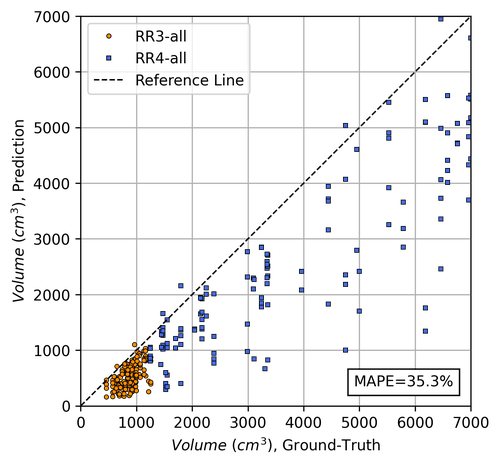}
		\caption{Volume (re-engineered stockpiles), without SP thresholding}
	\end{subfigure}
	\hfill
	\begin{subfigure}[b]{0.45\textwidth}
		\centering
		\includegraphics[width=\textwidth]{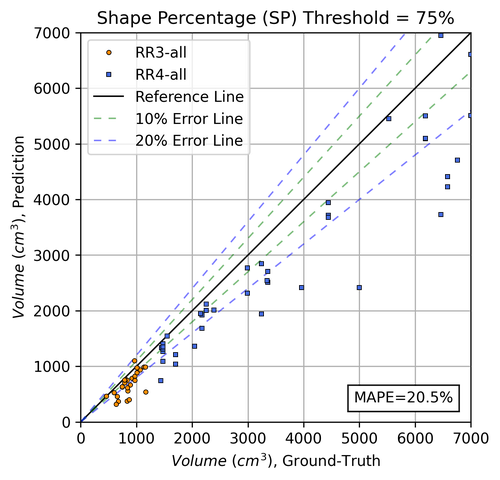}
		\caption{Volume (re-engineered stockpiles), with SP thresholding at 75\%}
	\end{subfigure}
	\newline 
	\begin{subfigure}[b]{0.42\textwidth}
		\centering
		\includegraphics[width=\textwidth]{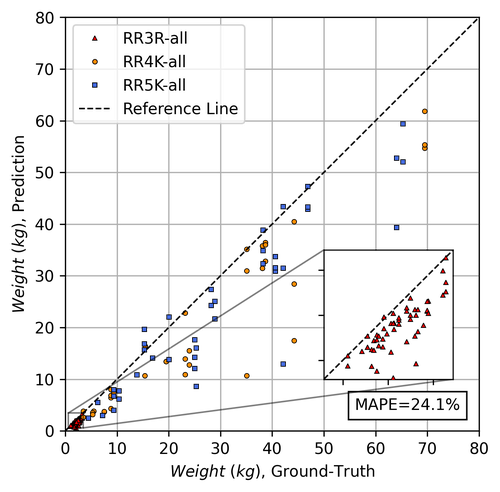}
		\caption{Weight (field stockpiles), without SP thresholding}
	\end{subfigure}
	\hfill
	\begin{subfigure}[b]{0.42\textwidth}
		\centering
		\includegraphics[width=\textwidth]{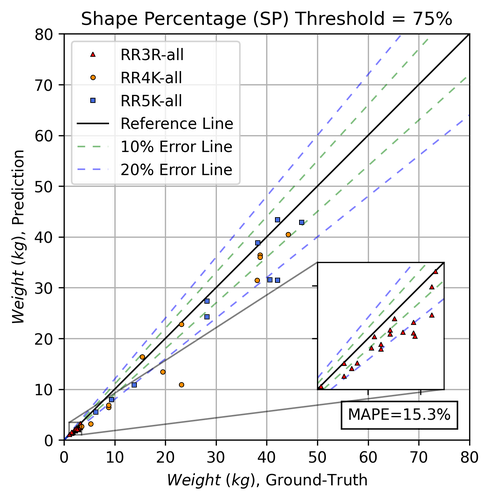}
		\caption{Weight (field stockpiles), with SP thresholding at 75\%}
	\end{subfigure}
	\caption{Systematic volume/weight underestimation for both re-engineered and field stockpiles (with and without SP thresholding)}
	\label{fig:stockpile-form}
\end{figure}

Finally, the SP thresholding process is able to help address the aforementioned issue of variable stockpile forms (at the end of Section \ref{sec: volume-correction}). \autoref{fig:stockpile-form} shows the degree of systematic volume/weight underestimation for both re-engineered and field stockpiles with and without SP thresholding. In addition to the improvement in both re-engineered and field stockpiles, it can also be observed that the degree of systematic volume underestimation after the SP thresholding process (with SP=75\%) becomes less distinct between the densely-stacked form (re-engineered stockpiles) and flat-layered form (field stockpiles). This can be explained by the fact that, although densely-stacked stockpile and flat-layered stockpile are conceptually two different types of macroscopic stockpile forms, the difference in terms of per-aggregate partial shape may not be distinguishable. In other words, even though flat-layered form can have overall higher visibility for aggregate shapes than densely-stacked form, the reliability of shape completion results for the subset of partial shapes with high visibility (e.g., SP over 75\%) may be very similar in the two cases. For example, the raw volume/weight results in \autoref{fig:stockpile-form} show deviations with MAPE=35.3\% and MAPE=24.1\% for re-engineered and field stockpiles, respectively, but after SP thresholding the results are improved to deviations with MAPE=20.3\% and MAPE=15.3\%, respectively. Also, after the SP thresholding, the distributions of the effective data points in both cases lie more consistently near the $-10\%$ and $-20\%$ error bounds. This indicates that, for practical use, it is likely to utilize the quantitative SP thresholding to establish a uniform correction process rather than using a case-specific (i.e., based on the stockpile form) volume correction factor. 

\begin{figure}[!htb]
	\centering
	\begin{subfigure}[b]{0.45\textwidth}
		\centering
		\includegraphics[width=\textwidth]{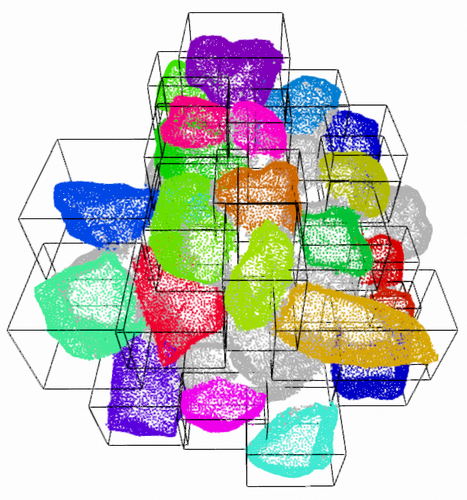}
		\caption{SP thresholding at 65\%}
	\end{subfigure}
	\hfill
	\begin{subfigure}[b]{0.45\textwidth}
		\centering
		\includegraphics[width=\textwidth]{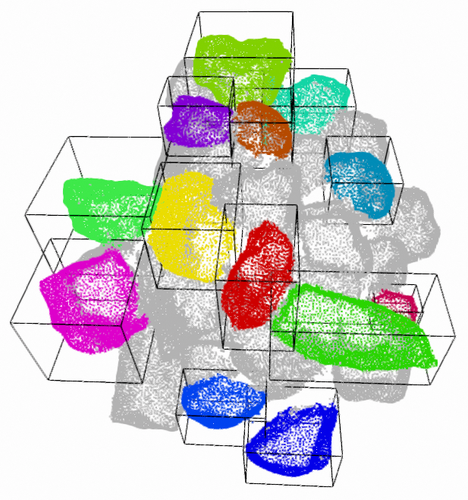}
		\caption{SP thresholding at 70\%}
	\end{subfigure}
	\newline 
	\begin{subfigure}[b]{0.45\textwidth}
		\centering
		\includegraphics[width=\textwidth]{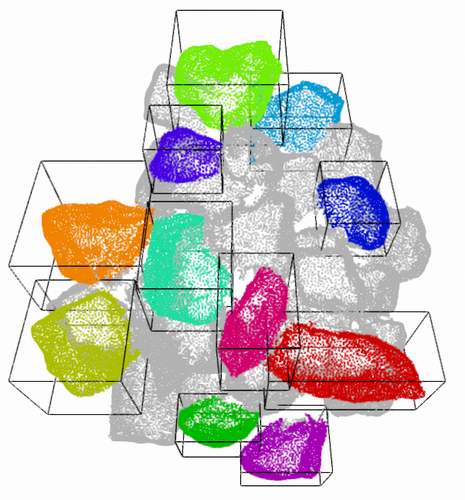}
		\caption{SP thresholding at 75\%}
	\end{subfigure}
	\hfill
	\begin{subfigure}[b]{0.45\textwidth}
		\centering
		\includegraphics[width=\textwidth]{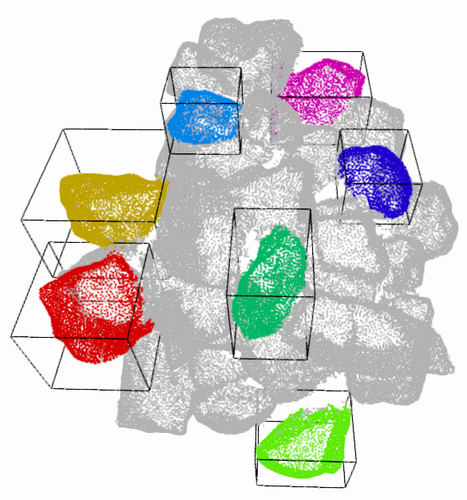}
		\caption{SP thresholding at 80\%}
	\end{subfigure}
	\caption{Effect of SP thresholding on segmented aggregate shapes in a stockpile (RR4-S6)}
	\label{fig:stockpile-sp}
\end{figure}

\autoref{fig:stockpile-sp} illustrates the effect of SP thresholding process on the stockpile analysis results. As the SP threshold increases, the number of effective aggregates decreases, with the remaining shapes at locations of less occupancy and larger open space. Aggregates at these protruding positions typically have better visibility with a large portion of the shape accessible from multi-view observation. Partial shapes segmented from a stockpile, in either densely-stacked or flat-layered forms, can exhibit different SP values, i.e. visibility levels. Generally, flat-layered stockpile form gives higher SP values or better shape visibility than the densely-stacked form. This is because the aggregates in a flat-layered form usually have fewer occlusions from the stacking of particles. Therefore, when a certain SP threshold is used to screen the segmentation and completion results, it is expected that flat-layered stockpiles can have more effective aggregates (i.e., aggregates with SP greater than the threshold) than densely-stacked stockpiles, given the same number of total aggregates in the stockpile. 

Moreover, the balance between the analysis efficiency (i.e., number of effective aggregates from the analysis) and the analysis quality (i.e., confidence or reliability level of the predicted results) is important. This may suggest the following practical guidance on the field implementation and application of the RSC-3D framework for practitioners. First, practitioners can choose the appropriate SP threshold according to the specific field application. For example, if the key metrics during the evaluation are the length dimension metrics that usually have high accuracy (as shown in \autoref{fig:sp-size}), a lower threshold could be used to efficiently capture a sufficient amount of aggregates during the analysis. On the other hand, if the key metrics are the sensitive high-dimensional metrics (volume, area, etc., as shown in \autoref{fig:sp-volume}), the practitioners could set a higher threshold to filter and select fewer yet more reliable shapes. 

Furthermore, to obtain an overall better visibility of aggregate shapes, practitioners at the quarry sites are recommended to form stockpiles in the flat-layered form, such that aggregates can be arranged to have relatively high percentages of surface to be visible (e.g., visually over 60\% or 70\%), as illustrated in \autoref{fig: photo-field}. Based on the field activities undertaken in this study, flat-layered is found to be a very common form of stockpiles at the quarry sites (as shown in the Wolman method in \autoref{fig: wolman}), where the practitioners may conveniently incorporate the RSC-3D framework into their QA/QC activities.

\clearpage
\section{Summary}

This chapter presented an integrated framework based on 3D reconstruction, 3D instance segmentation, and 3D shape completion. The framework follows a reconstruction-segmentation-completion approach to conduct 3D aggregate stockpile analysis. Field application of the framework was demonstrated, and the performance of the framework was validated on 12 re-engineered stockpiles and 9 field stockpiles. The results of instance segmentation and shape completion were evaluated qualitatively, while 3D morphological analysis was conducted to quantitatively validate the approach against ground-truth. Regression analyses were conducted to reveal the systematic volume underestimation, and a shape percentage prediction threshold was further developed towards practical interpretation of the morphological analysis results. Overall, the framework demonstrated good robustness and reliability in characterizing morphological properties of aggregates in stockpiles. 

%% file: chapter11.tex
\chapter{Conclusions and Recommendations for Future Research} \label{chapter-11}

\section{Summary of Findings}

The primary objective of this doctoral research study is to develop a convenient and efficient field imaging framework for aggregates based on computer vision techniques. The framework is supposed to provide an analysis platform for the field collected aggregate data to determine the size, shape, volume/weight, and gradation properties of the large-sized aggregates inspected. Review of the current practice and literature showed that characterizing the morphological properties of riprap and large-sized aggregates in a quantitative manner has not been an area that benefited from the technology advancements in image processing. The state-of-the-practice methods by engineers and practitioners mainly rely on visual inspection and manual measurements. By incorporating the state-of-the-art technology in computer vision and computer graphics, the developed framework in this study enables the characterization of aggregates at different sophistication levels: (i) individual and isolated aggregates for volumetric estimation, (ii) in-place aggregates in a stockpile for 2D image analyses, as well as (iii) in-place aggregates in a stockpile or constructed layer for 3D point cloud analyses. The major research findings of this doctoral study are summarized and highlighted as follows.

\subsection{Summary of Findings from the Individual-Aggregate Study}
The following findings can be summarized related to the individual-aggregate field imaging system for the characterization of size and weight information of individual riprap rocks and large-sized aggregates:
\begin{itemize}
	
	\item A field imaging system was designed and built as a portable and versatile toolkit for the convenience of efficient and reliable image acquisition needs. Image segmentation and volumetric reconstruction algorithms were developed for individual aggregate particles or rocks with the capabilities of extracting them under uncontrolled field lighting conditions and reconstructing them volumetrically with necessary calibration and correction. 

	\item The robustness and accuracy of the developed algorithms were studied on 85 riprap aggregate particles collected from two quarry sites. The Mean Average Percentage Error (MAPE) between the ground-truth volume/weight measurements and the image-based volumetric reconstruction results was $3.6\%$ and $7.9\%$ for different material sources after applying rotate-repetitions. For all studied particles, the volumetric reconstruction results show that most data points lie within $\pm 20\%$ error band from the ground-truth reference, and more than half of the results locate within the $\pm 10\%$ band.
	
	\item Comparisons were made between the image-based volumetric reconstruction results and the state-of-the-practice manual measurements. Significant improvements were achieved using the developed field imaging system, from MAPE=$68.3\%$ for manual measurement results to MAPE=$8.2\%$ imaging-based results. 
\end{itemize}

\subsection{Summary of Findings from the 2D Aggregate Stockpile Study}
Based on the 2D aggregate stockpile imaging study, the following findings can be summarized:
\begin{itemize}

	\item This study adopted and successfully implemented a neural network to accomplish the stockpile aggregate image segmentation task. By establishing an image dataset of 164 aggregate stockpile images with 11,795 labeled aggregates, a segmentation kernel was trained to learn the instance segmentation task on aggregate stockpile images.
	
	\item The trained segmentation kernel achieved an average completeness of $88\%$ and an average IoU precision of $87\%$, with a standard deviation of $7.1\%$ and $1.5\%$ respectively. The developed approach allows extracting individual aggregate particles in an automated manner, thus greatly enhances the efficiency of morphological analysis. 
	
	\item Morphological analyses were conducted on the segmented aggregate particles to generate size and shape distribution curves. Analysis results were verified with ground-truth labeling to measure the robustness and accuracy of the segmentation approach.

\end{itemize}

\subsection{Summary of Findings from the 3D Aggregate Stockpile Study}
Based on the 3D stockpile study, the following findings can be summarized:
\begin{itemize}	
	\item A marker-based 3D reconstruction approach was developed as a cost-effective and flexible procedure to allow full 3D reconstruction of aggregates. A 3D aggregate particle library of 46 RR3 and 36 RR4 aggregate samples collected from field studies was established using this approach. The resolution of the 3D reconstruction results was around 1 $point/mm^2$, and the Mean Percentage Error (MPE) of the reconstructed volumes is around $+2\%$ with respect to the ground-truth volume measurements.
	
	\item A comparative analysis was conducted on the 3D particle library regarding the statistical differences between the 2D and 3D morphology. The comparison indicated a potential intrinsic relationship between the true 3D morphological index and its 2D morphological equivalence, which was validated across different aggregate shapes from various aggregate size categories. It was found the 2D indicators obtained from single-view analysis are likely to capture the intermediate dimension ratios rather than the longest-shortest dimension ratios.
	
	\item Based on the 3D particle library, high-quality datasets were prepared for multiple purposes of the deep learning tasks. First, a synthetic dataset of 300 aggregate stockpiles and 105,054 total aggregates was prepared with ground-truth labels leveraging the developed raycasting techniques. The dataset was used during the training of the 3D segmentation network. A dataset of 9,184 partial-complete shape pairs were also generated from the particle library based on the developed varying-visibility and varying-view raycasting schemes, at seven visibility levels and 16 model orientations for each of the 82 models in the particle library. 
	
	\item 3D instance segmentation and 3D shape completion networks were implemented, trained, and tested. The 3D instance segmentation network achieved an average completeness of $78\%$ and an average Intersection over Union (IoU) precision of $82\%$, with a standard deviation of $6.3\%$ and $4.8\%$, respectively. The segmentation network effectively learns the per-point offset vector to shift the original point cloud into an optimized clustered coordinate space, from which the instance proposals are generated. The 3D shape completion network achieved 0.00019 in. (0.00483 mm) and 0.00022 in. (0.00559 mm) Chamfer Distance (CD) on the validation and test sets, respectively. The completion network effectively learns the global and local shape context of the partial input point cloud and predicts the missing regions with fine-grained details. Both components demonstrated very good performance in the stockpile segmentation and shape completion tasks. 
	
	\item Based on the development of neural networks, an integrated 3D Reconstruction-Segmentation-Completion (RSC-3D) framework was proposed. The robustness and reliability of the framework were validated against 12 re-engineered stockpiles and 9 field stockpiles. The size dimension metrics demonstrated MAPE error around $8\%$ to $18\%$ against the ground-truth, while a higher systematic deviation around $25\%$ to $35\%$ was observed in terms of the high-dimensional measures. Further, Shape Percentage (SP) thresholding study was conducted to analyze and address the systematic deviation for morphological analysis results. An SP threshold of 75\% was recommended based on statistically analysis. The SP thresholding quantitatively characterized the partial observation process and reduced the systematic volume underestimation to approximately $15\%$ to $20\%$, which allows a uniform correction for different stockpile forms.
\end{itemize}

\section{Conclusions and Major Contributions}

This research effort presents major contributions to the practical characterization methods for determining morphological properties (size, shape, volume, etc.) of aggregates as well as the computer vision techniques developed by establishing a multi-scenario solution for field imaging of aggregates. The developed framework encompasses three major approaches that characterize various forms and representations of field aggregates with increasing analysis complexity: (i) a volumetric reconstruction approach for individual and non-overlapping aggregates; (ii) a 2D instance segmentation and morphological analysis approach for aggregates in stockpiles based on 2D image analysis; and (iii) a 3D reconstruction-segmentation-completion approach for aggregates in stockpiles based on 3D point cloud analysis. The framework also has a focus on relatively large-sized aggregates, for which effective and efficient field characterization methods are extremely lacking. 

To state-of-the-practice methods, the developed framework extends the set of feasible tools and techniques for practitioners and engineers. First, the volumetric reconstruction approach and/or the marker-based 3D reconstruction approach can be used for the assessment of individual aggregates, which can relieve the labor-intensive weighing process and has proven to be much more accurate than the manual dimension measurement practice. Second, the 2D and 3D stockpile segmentation approaches can provide quantitative characterization of aggregate morphology, which greatly improves the rough and less informative stockpile visual inspection practice. Lastly, with the size dimension measure from the 3D integrated approach, it is likely to refine and improve the current specifications and standards of large-sized aggregates by imposing dimensional requirements in addition to the sole particle weight requirement. 

To state-of-the-art aggregate imaging methods, the developed framework expands the domain of available algorithms and applications by a great margin. First, this framework fills the gap between laboratory-oriented aggregate imaging systems and the challenging field conditions. All approaches developed in this framework are naturally designed, tested, and validated under field conditions. Second, this framework extends the aggregate size range limit of existing aggregate imaging systems to large-sized aggregates. The flexible designs in the volumetric reconstruction approach, 2D segmentation approach, and 3D segmentation approach are all versatile in that aggregate particles are allowed to span from regular-sized to large-sized. Third, this framework addresses the challenging stockpile segmentation problem by proposing both 2D and 3D segmentation approaches. Lastly, this framework completes the realistic 3D field imaging research with stockpile analysis of aggregates. To the author's knowledge, the research effort in this study advances over the state-of-the-art serving as: (i) the first comprehensive aggregate imaging study that considers large-sized aggregates, (ii) the first field volumetric reconstruction approach, (iii) the first deep learning based 2D aggregate stockpile/assembly segmentation approach, (iv) the first 3D imaging and analysis approach for aggregate stockpiles in field conditions, and (v) the first 3D aggregate shape study that involves partial observations and recreation of particle shape.

\section{Recommendations for Future Research}

Based on the good performance of the framework developed in this study, more advancements are envisioned and recommended for future research related to the similar topics. Major promising future directions are discussed as follows, including a few preliminary and/or proof-of-concept studies as part of the author's ongoing research efforts. Before that, several limitations of the current framework are noteworthy for discussing the future directions:
\begin{itemize}
	\item The current framework requires calibration objects to determine the scale of the image (for 2D approach) or the scene (for 3D approach).
	\item The Structure-from-Motion (SfM) based 3D reconstruction approach is not real-time and is very computationally intensive. Also, the vision-based reconstruction is more sensitive to shadowing condition as compared to physics-based reconstruction methods such as laser and/or Light Detection and Ranging (LiDAR) approaches.
	\item The statistics of the stockpile only represent the aggregates that are on the stockpile surface. The interior of aggregate stockpiles cannot be and is not expected to be inspected naturally by the definition of the problem.
\end{itemize}

\subsection{Progressive Improvement of the Framework}

Due to the data-driven nature of the deep learning networks in this framework, the performance is expected to gain progressive and scalable improvement with increased dataset size. The potential improvements on datasets include: 

\begin{itemize}
	\item Enriching 2D stockpile image dataset by collecting and labeling more stockpile images from diverse geological origins and rock types (limestone, granite, sandstone, trap rock, etc.). Aggregate images containing various backgrounds such as the in-situ background of aggregates in constructed layers can also be included in the database.
	
	\item Extending the 3D particle library by collecting more aggregates from different origins and size groups (e.g., ballast, gravel, etc.). The size and quality of the 3D particle library will directly reflect on the quality of the 3D synthetic stockpile dataset and the partial-complete shape dataset.
	
	\item Exploring and improving the robustness of 3D instance segmentation and 3D shape completion network. As previously discussed in \cref{chapter-7}, the surface normals, point colors and other features of the point cloud may be beneficial for a more robust segmentation and completion process. By addressing the challenges of estimating normals from field stockpile data and high color variation of aggregates, these additional per-point features could help to improve the performance of the networks.
	
	\item With convenient tools for 3D instance labeling, the manual labeling process may still be adapted to generate a small quantity of labeled points clouds of real stockpile data to serve as the training/test set for the 3D segmentation network.
\end{itemize}

\subsection{Integration with Intelligent Sensing Technologies}

The current framework adopts the traditional SfM techniques for obtaining the 3D point clouds of aggregate stockpiles. Note that the three major components of the framework are standalone. Namely, as long as the input to the 3D segmentation and shape completion networks follows the point cloud format, the framework does not need to be bonded to certain 3D reconstruction techniques. With the rapid development in 3D visualization and augmented reality, it is expected that more advanced technologies for 3D sensing will be readily available in the future. For example, potential methods for the 3D reconstruction step can be further developed as follows: (i) LiDAR devices that directly capture the point cloud and (ii) Dense Simultaneous Localization and Mapping (SLAM) techniques that leverage RGB-D sensors and optical flow methods.

\begin{figure}[!htb]
	\centering
	\includegraphics[width=\textwidth]{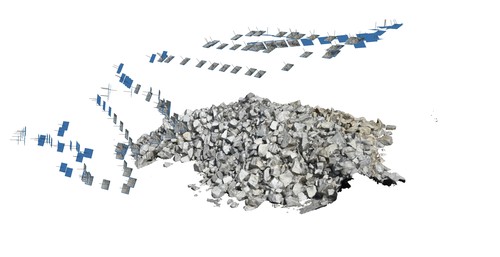}
	\caption{Full 3D reconstruction of a larger field stockpile from manually controlled UAV images}
	\label{fig: drone}
\end{figure}

On the other hand, the data acquisition devices are not limited to handheld sensors. For example, to embed the developed framework deeply into the aggregate production line, it is best to attach sensors to the conveyor system, which allows better statistical coverage of most aggregates before they become the stockpiles. Further, intelligent methods of acquiring stockpile aggregate images can be integrated with advanced aerial photography techniques. For example, Unmanned Aerial Vehicle (UAV) can greatly help with the image acquisition step for multi-spot or all-around inspection of a large stockpile, especially when intelligent route planning techniques are used. An example of preliminary study of 3D reconstruction from UAV images is illustrated in \autoref{fig: drone}. The preliminary study proves that UAV images can give high-quality reconstruction of a full stockpile. By dividing the large stockpile into chunks, the entire stockpile can be analyzed by the framework incrementally.

Also note the great potential of using UAV for calibration-free reconstruction. Commercial or industry level UAVs usually have open-source Software Developer ToolKit (SDK) that allows reading of the internal Inertial Measurement Unit (IMU) data. With such flight route data integrated into the 3D reconstruction step, it is likely to achieve a completely calibration-free reconstruction of the stockpile.

\subsection{Generalized Applications to All Aggregate Sizes and Categories}

\begin{figure}[!htb]
	\centering
	\begin{subfigure}[b]{0.45\textwidth}
		\centering
		\includegraphics[height=5cm]{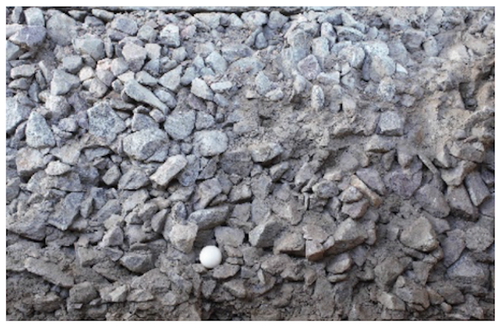}
		\caption{}
	\end{subfigure}
	\hfill
	\begin{subfigure}[b]{0.45\textwidth}
		\centering
		\includegraphics[height=5cm]{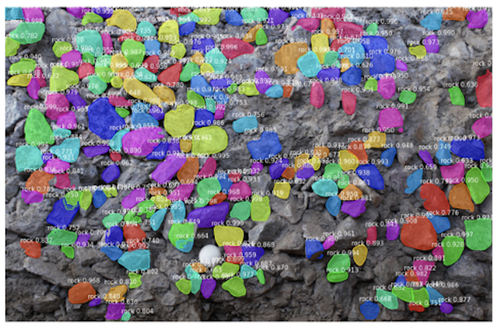}
		\caption{}
	\end{subfigure}
	\caption{(a) Cross-sectional ballast image of a trench cut and (b) segmentation results}
	\label{fig: ballast-1}
\end{figure}

Lastly, based on the methodology and the deep learning nature of the developed framework, its performance is very likely to generalize well onto broader size categories of aggregates (e.g., coarse aggregates used in pavement construction, ballast in railway engineering, etc.). The author believes the main difference between the relatively large-sized aggregates and regular-sized aggregates is the scale when taking the images. By moving closer to the surface when inspecting smaller aggregates, the per-instance resolution or point density can be maintained at a similar level to the large-sized case. Therefore, the framework is expected to have generalized performance since the essence of the tasks for different types of aggregates is almost identical once they are at the same scale. An example of the performance of the 2D instance segmentation network on ballast stockpile images is illustrated in \autoref{fig: ballast-1}. The network was fine-tuned based on the developed large-sized aggregates model with very few labeled ballast images. It can be observed the network generalizes reasonably well onto a different type of aggregates. 
\clearpage 

\begin{figure}[!htb]
	\centering
	\includegraphics[trim=0 0 200 0, clip, width=0.8\textwidth]{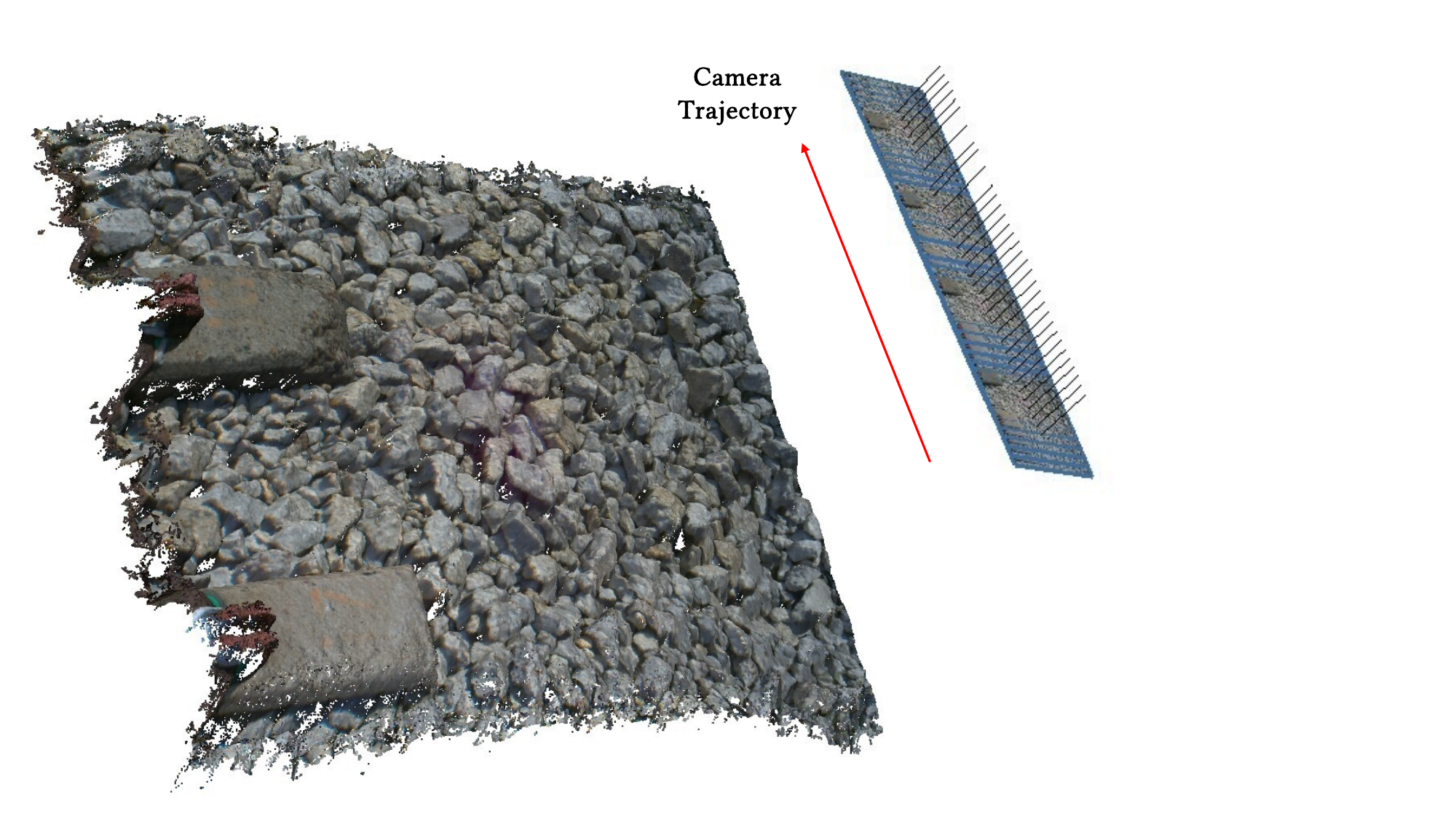}
	\caption{3D reconstruction of the shoulder of a ballasted track}
	\label{fig: ballast}
\end{figure}

Similarly, a 3D reconstruction step was tested on a railway testing facility by moving the camera along the track. It can be seen in \autoref{fig: ballast} that the 3D point cloud of the ballast can be obtained with high-resolution. Then, the developed 3D reconstruction-segmentation-completion framework is very likely to perform equally well on the ballast assemblies which mostly resemble the form of an aggregate stockpile.

\clearpage